%% file: neurips_2019.tex
\documentclass{article}

\PassOptionsToPackage{numbers, compress}{natbib}

\usepackage[preprint]{neurips_2019}

\input{macros}

\usepackage[utf8]{inputenc} 
\usepackage[T1]{fontenc}    
\usepackage{hyperref}       
\usepackage{url}            
\usepackage{booktabs}       
\usepackage{amsfonts}       
\usepackage{nicefrac}       
\usepackage{graphicx}
\usepackage{array}
\usepackage{microtype}      
\usepackage{amsmath}

\title{Time-Series Analysis via Low-Rank Matrix Factorization Applied to Infant-Sleep Data}

\author{
    Sheng Liu$^1$ \qquad Mark Cheng$^2$ \qquad Hayley Brooks$^{3,4}$ \qquad Wayne Mackey$^4$ \\ \textbf{David J. Heeger$^{4,5}$ \qquad Esteban G. Tabak$^{6}$ \qquad Carlos Fernandez-Granda$^{1,6}$}\\
    $^1$Center for Data Science, NYU \qquad 
    $^2$Technische Universit\"at Berlin\\
    $^3$Department of Psychology, University of Denver \qquad
    $^4$Center for Neural Science, NYU\\
    $^5$Department of Psychology, NYU \\
    $^6$Courant Institute of Mathematical Sciences, NYU
}
\begin{document}

\maketitle
\vspace{-7mm}
\begin{abstract}
  We propose a nonparametric model for time series with missing data based on low-rank matrix factorization. The model expresses each instance in a set of time series as a linear combination of a small number of shared basis functions. Constraining the functions and the corresponding coefficients to be nonnegative yields an interpretable low-dimensional representation of the data. A time-smoothing regularization term ensures that the model captures meaningful trends in the data, instead of overfitting short-term fluctuations. The low-dimensional representation makes it possible to detect outliers and cluster the time series according to the interpretable features extracted by the model, and also to perform forecasting via kernel regression. We apply our methodology to a large real-world dataset of infant-sleep data gathered by caregivers with a mobile-phone app. Our analysis automatically extracts daily-sleep patterns consistent with the existing literature. This allows us to compute sleep-development trends for the cohort, which characterize the emergence of circadian sleep and different napping habits. 
\end{abstract}
\vspace{-5mm}
\section{Introduction}

Analysis of time series is a fundamental challenge in modern healthcare applications. The data are often very noisy, incomplete, and high dimensional, which precludes the direct application of standard machine-learning tools. We propose a method to extract meaningful features from such data, which are associated to regular intervals, such as hours, days, weeks, etc. This yields an interpretable representation of the data in terms of a superposition of daily, hourly, or weekly patterns that evolve over time. The representation can be used to perform several tasks, including trend extraction, outlier detection, clustering, and forecasting. We apply the proposed methodology to study infant sleep data gathered by caregivers via a mobile app. Understanding infant sleeping behavior is a problem of great importance to researchers, clinicians and, of course, parents. Traditional studies tend to be limited either to small number of infants from relatively homogeneous backgrounds observed intermittently (e.g. for a few hours, one day per month) or to data obtained from questionnaires, instead of direct observations \cite{henderson2011consolidation}. Large-scale datasets logged directly by users present an unprecedented opportunity for data-driven analysis of sleep development across large populations.






\textbf{Technical significance.} Our main methodological contribution is a novel approach to analyze a set of multiple time series with missing entries. The goal of the analysis is to extract meaningful patterns associated to regular intervals that partition the time series, such as days, hours, months, etc. To achieve this, we combine ideas from low-rank matrix completion~\cite{candes2009exact}, singular spectrum analysis~\cite{vautard1992singular}, and nonnegative matrix factorization~\cite{lee1999learning}. 
We represent the data in terms of a small number of basis functions that capture common patterns associated to the intervals of interest. The coefficients of the representation encode the time evolution of these patterns in each of the time series. Constraining the functions and coefficients to be nonnegative avoids cancellations in the model and yields more interpretable basis functions. A smoothing regularization term on the coefficients ensures that they capture meaningful trends in the data, instead of overfitting short-term fluctuations. The proposed model is validated on a real-world dataset, where we verify that the learned representation can be used to detect outliers and cluster the time series. 


\textbf{Clinical relevance.} Recent works suggest that sleep affects memory and language development~\cite{seehagen2015timely,dionne2011associations}, and may be used to predict attention and behavior problems~\cite{sadeh2015infant}. 
The analysis of large-scale infant-sleep data gathered with hand-held devices therefore opens up the door to many promising clinical applications. In this work, we showcase several of them on a real-world dataset. Even though we do not make any prior assumptions about sleeping behavior (beyond nonnegativity and temporal smoothness), our analysis automatically extracts daily-sleep patterns-- such as night sleep combined with one or two daily naps-- that are consistent with the existing literature. 

\textbf{Related work.} There is no previous analysis of infant-sleep data applying low-rank modeling (to the best of our knowledge). Ref. \cite{mindell2016development} studies sleep development from mobile-phone data using more traditional techniques, such as averaging in predetermined time intervals. Ref. \cite{weinraub2012patterns} applies a growth mixture model to analyze patterns of developmental change in infants' sleep. Other relevant studies in infant sleep development include \cite{henderson2011consolidation,henderson2010sleeping,weissbluth1995naps,iglowstein2003sleep}. The ingredients in our methodological approach are inspired by several existing techniques, but to the best of our knowledge their combination is novel. Low-rank matrix completion~\cite{candes2009exact} and nonnegative matrix factorization~\cite{lee1999learning} have been widely used to perform imputation, denoising and factor analysis of high-dimensional data in collaborative filtering, topic modeling, and many other domains. Diverse forms of temporal regularization have been proposed in the application of these models to time series in \cite{chen2005nonnegative,rallapalli2010exploiting,cheung2015decomposing,yu2016temporal,tabak2018conditional}. However, these works consider a single multivariate time series, as opposed to a set of multiple one-dimensional time series. The idea of reshaping a single time series as a matrix dates back to at least the work of \cite{vautard1992singular} on singular spectrum analysis. The reshaping procedure differs from ours because the rows overlap (the matrix is Toeplitz), so the learned patterns are not associated to fixed time intervals as in our case. Our approach is closer to very recent work \cite{agarwal2018time}, where the authors apply low-rank models to perform denoising, imputation and forecasting on a single time series (as opposed to multiple time series, which is our setting of interest). Code and data are available at \url{https://cims.nyu.edu/~sl5924/infant.html}.

\vspace{-2mm}
\section{Methods}

%


\subsection{Low-rank modeling of time-series data}

\begin{figure}[tp]
\begin{tabular}{ >{\centering\arraybackslash}m{0.15\linewidth} >{\centering\arraybackslash}m{0.02\linewidth} >{\centering\arraybackslash}m{0.15\linewidth}  >{\centering\arraybackslash}m{0.02\linewidth} >{\centering\arraybackslash}m{0.15\linewidth}  >{\centering\arraybackslash}m{0.15\linewidth} >{\centering\arraybackslash}m{0.15\linewidth} }
\vspace{-0.3cm}\includegraphics[width=0.5\linewidth]{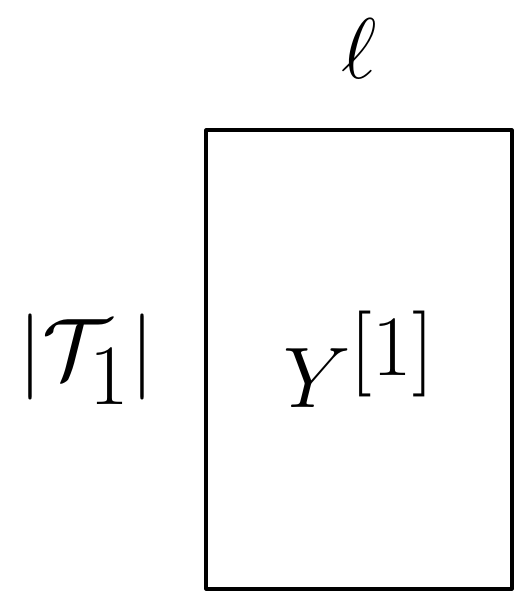} & $ \approx$ & \includegraphics[width=0.45\linewidth]{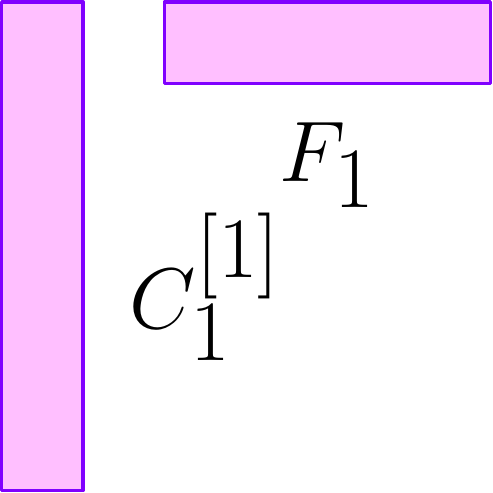} & + & \includegraphics[width=0.45\linewidth]{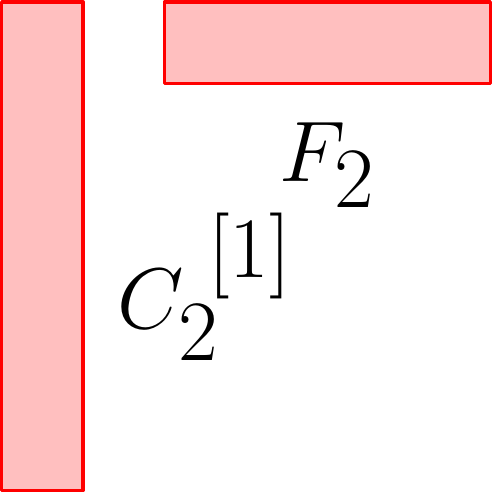} & $+$ $ \quad \cdots \quad$ $+$ & \includegraphics[width=0.45\linewidth]{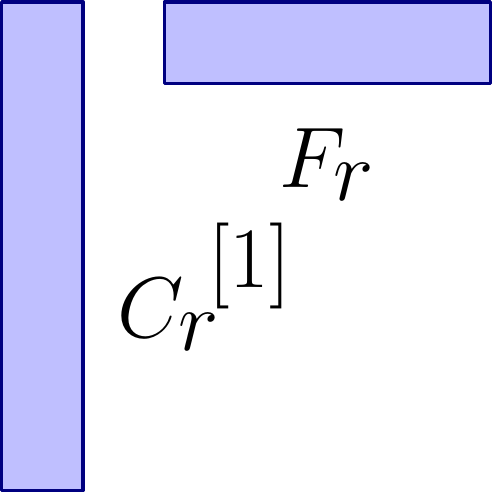} \\
\includegraphics[width=0.5\linewidth]{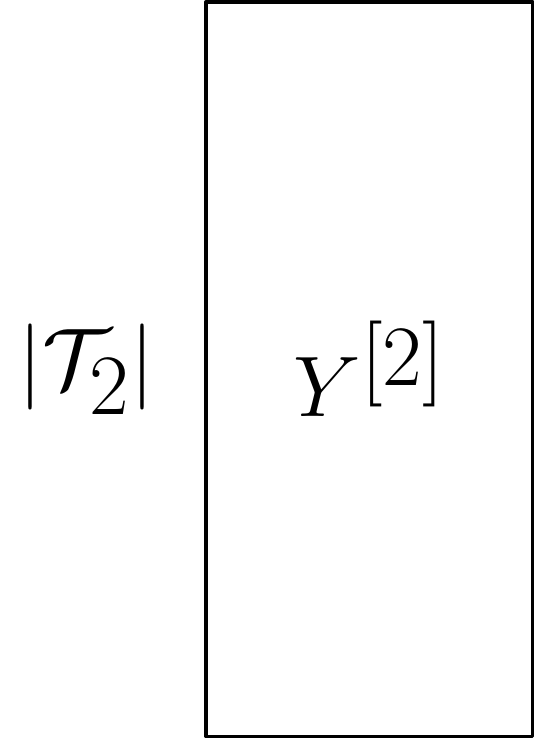} & $ \approx$ & \includegraphics[width=0.45\linewidth]{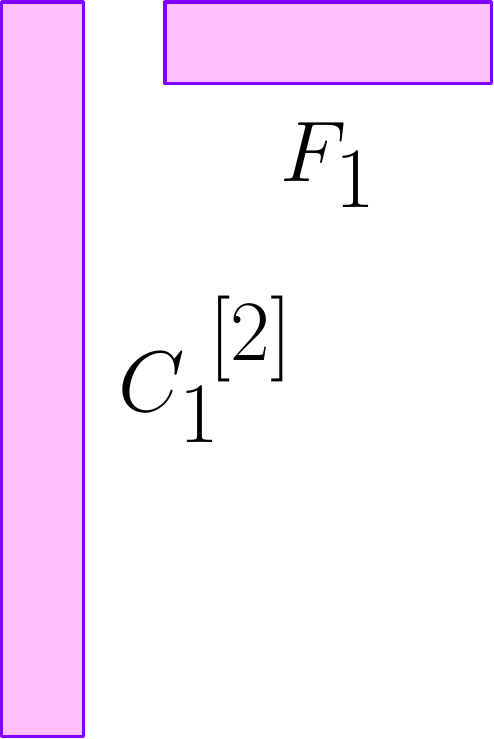} & + & \includegraphics[width=0.45\linewidth]{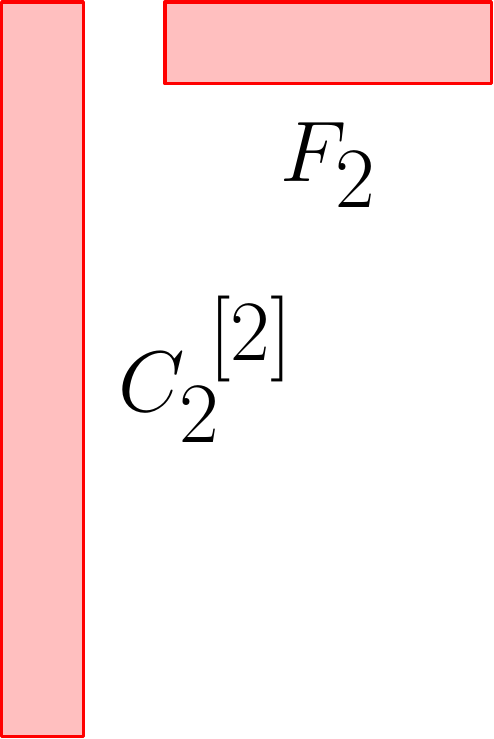} & $+$ $ \quad \cdots \quad$ $+$ & \includegraphics[width=0.45\linewidth]{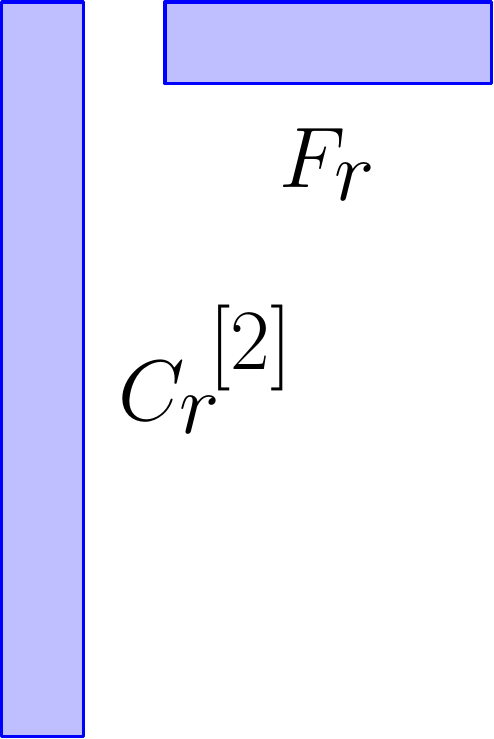} \\
& $\cdots$ & \\
\hspace{-0.1cm}\includegraphics[width=0.5\linewidth]{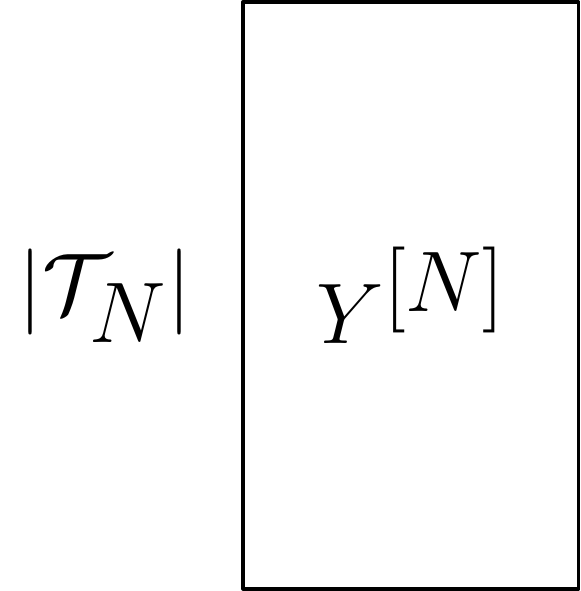} & $ \approx$ & \includegraphics[width=0.45\linewidth]{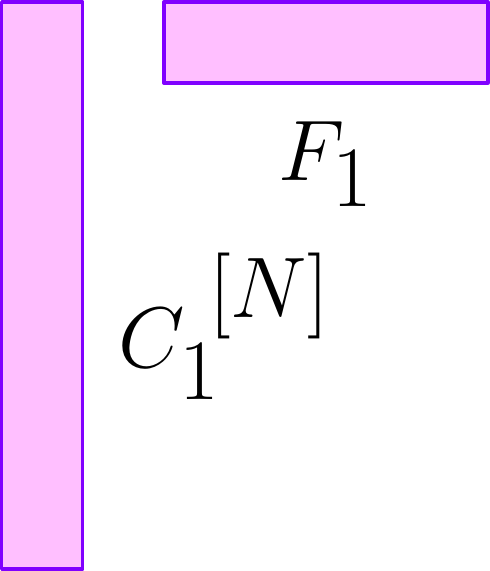} & + & \includegraphics[width=0.45\linewidth]{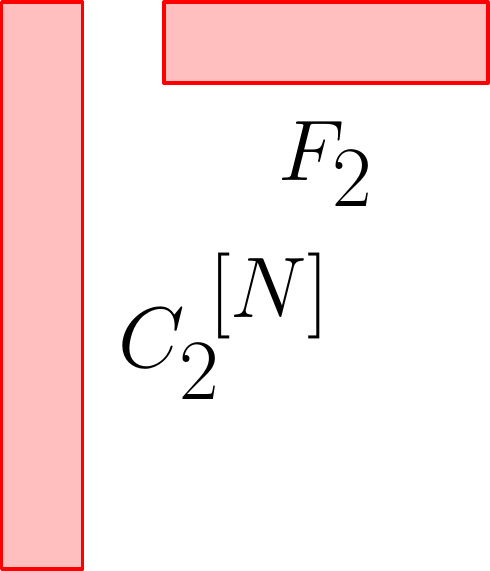} & $+$ $ \quad \cdots \quad$ $+$ & \includegraphics[width=0.45\linewidth]{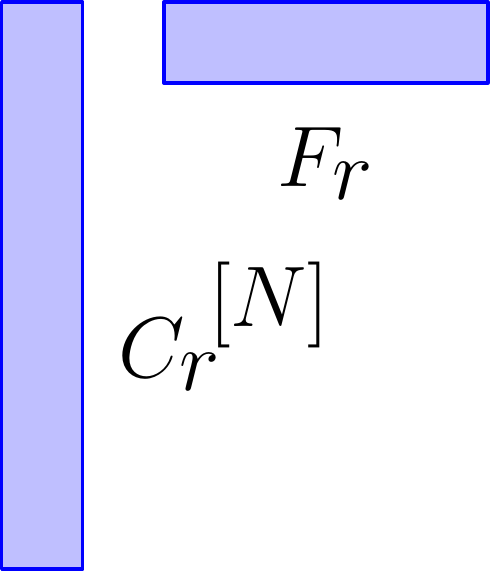}  
 \end{tabular}
 \caption{Proposed low-rank model applied on a set of time series. Each time series, represented by a matrix $Y^{[n]}$, $1\leq n \leq N$, is approximated by a sum of $r$ components corresponding to basis functions $F_{1}$, \ldots, $F_{r}$, shared across the different time series. The coefficients $C^{[n]}_{1}$, \ldots, $C^{[n]}_{r}$ provide a representation of reduced dimensionality for each $Y^{[n]}$. $\ml{T}_n$ is the set of observed rows in $Y^{[n]}$.
 }
 \vspace{-3mm}
 \label{fig:diagram_lowrank}

\end{figure}

We consider a dataset of $N$ discrete time series $y^{[n]}$, sampled at the same rate. The time series may have missing entries. We partition each time series into intervals of a predefined length $\ell$. Our main assumption is that the intervals can be approximated as superpositions of a small number of shared basis functions. In the case of the infant-sleep data the intervals correspond to days, because the aim is to extract daily sleeping patterns. We express each partitioned time series as a matrix $Y^{[n]}$ with $\ell$ columns, where the rows contain non-overlapping segments of length $\ell$, 
\begingroup
{\small
\renewcommand*{\arraystretch}{1.5}
\begin{align}
Y^{[n]} := \MAT{ y^{[n]}[1] & y^{[n]}[2] & \cdots &  y^{[n]}[\ell] \\ y^{[n]}[\ell+1] & y^{[n]}[\ell+2] & \cdots &  y^{[n]}[2\ell] \\ y^{[n]}[2\ell+1] & y^{[n]}[2\ell+2] & \cdots &  y^{[n]}[3\ell] \\ \cdots & \cdots & \cdots & \cdots }.
\end{align}
}
\endgroup
For simplicity, we assume that each row is present or missing, but our methodology is straightforward to extend to the case where single entries are missing. 
Our goal is to learn a joint low-rank representation for these matrices, such that each $Y^{[n]}$ is well approximated by a rank $r $ decomposition of the form
\begin{align}
\label{eq:lowrank_model}
Y^{[n]}\brac{t,i} \approx \sum_{j=1}^{r} C^{[n]}_{j}\brac{t} F_{j}\brac{i}, \quad 1 \leq i \leq \ell,  
\end{align}
for all observed values of $t$. $F_{1}$, \ldots, $F_{r}$ are basis functions shared by all the time series. The model yields a representation of each $Y^{[n]}$ as $r \ll \ell $ coefficient time series $C^{[n]}_{1}$, \ldots, $C^{[n]}_{r}$. The dimensionality of the data is consequently reduced by a factor of $\ell/r$. Figure~\ref{fig:diagram_lowrank} shows a diagram of the low-rank model. In the application to infant-sleep data, the basis functions $F_{1}$, \ldots, $F_{r}$ represent \emph{daily-sleep patterns} common to all the infants in the population. The coefficients $C^{[n]}_{1}$, \ldots, $C^{[n]}_{r}$ used to combine these basis vectors are unique to each infant. They describe the sleep-development dynamics of the infant in terms of the daily-sleep patterns. 

\subsection{Nonnegative matrix factorization with time smoothing}
\label{sec:rnmf}
In order to extract meaningful patterns, we incorporate additional structure on the low-rank model defined by Eq.~\eqref{eq:lowrank_model}. First, we constrain the basis functions and the coefficients to be nonnegative. In applications such as image processing and topic modeling, nonnegative constraints often yield low-rank models that are more interpretable~\cite{lee1999learning}. 
Second, we favor coefficients that vary smoothly in time. The aim is to capture gradual tendencies in the data, instead of overfitting rapid, local fluctuations. In order to fit a low-rank model with these characteristics we solve the following optimization problem:
 \begin{align}
 \label{eq:rnmf_cost_function}
& \text{minimize} \quad \sum_{n = 1}^{N} \sum_{t\, \in \, \ml{T}_n} \sum_{i=1}^{\ell}  \brac{ Y^{[n]}\brac{t,i}  - \sum_{j=1}^{r} C^{[n]}_{j}\brac{t} F_{j}\brac{i} }^2 +  \lambda \sum_{n = 1}^{N} \sum_{j=1}^{r} \normTwo{ \ml{D} C^{[n]}_{j}}^2 \\
& \text{such that} \quad C^{[n]}_{j}\brac{t} \geq 0, \; F_{j}\brac{i} \geq 0, \; \normTwo{F_{j}} =1,    \;1\leq j \leq r, \; 1 \leq n \leq N, \; t\, \in \, \ml{T}_n, \; 1\leq i \leq \ell, \notag
\end{align}
where $\lambda > 0$ is a parameter that controls the tradeoff between the fit error and the regularization term, and the rest of the notation is consistent with Eq.~\eqref{eq:lowrank_model}. $\ml{D}$ is a finite-difference operator that computes a discrete approximation to the derivative of the coefficients (we use a second-order operator). 
Crucially, the presence of the time-smoothing operator $\ml{D}$ links the data of each individual time series. Otherwise, the low-rank model would be equivalent to performing principal-component analysis (PCA \cite{jolliffe1986principal}) or nonnegative matrix factorization (NMF \cite{lee1999learning}) over the whole time-series dataset, interpreting each vector $Y^{[n]}\brac{\cdot,t}$ as a separate data point. We call our approach nonnegative matrix factorization with time smoothing (NMF-TS). For the infant-sleep data, NMF-TS yields smooth coefficients as shown in Figure~\ref{fig:age_development_patterns}. In contrast, applying NMF or PCA results in very irregular coefficients, which overfit short-term variations in the data. The resulting low-rank approximations learned by PCA, NMF and NMF-TS for a specific infant are shown in Figure~\ref{fig:low_rank_approximation}. Additional examples of NMF-TS coefficients are shown in Figure~\ref{fig:random_samples}.

\section{Results}
\label{sec:results}

In this section we report the results of applying the proposed low-rank model to analyze the infant-sleep dataset described in Section~\ref{sec:dataset} of the appendix. 

\begin{figure}[t]
\hspace{-.5cm} 
\begin{tabular}{ >{\centering\arraybackslash}m{0.05\linewidth} >{\centering\arraybackslash}m{0.35\linewidth} >{\centering\arraybackslash}m{0.35\linewidth}   }
j \vspace{0.35cm} & Daily-sleep pattern $F_j$ \vspace{0.35cm} &  Development statistics \vspace{0.35cm}\\
1& \includegraphics[scale=0.15]{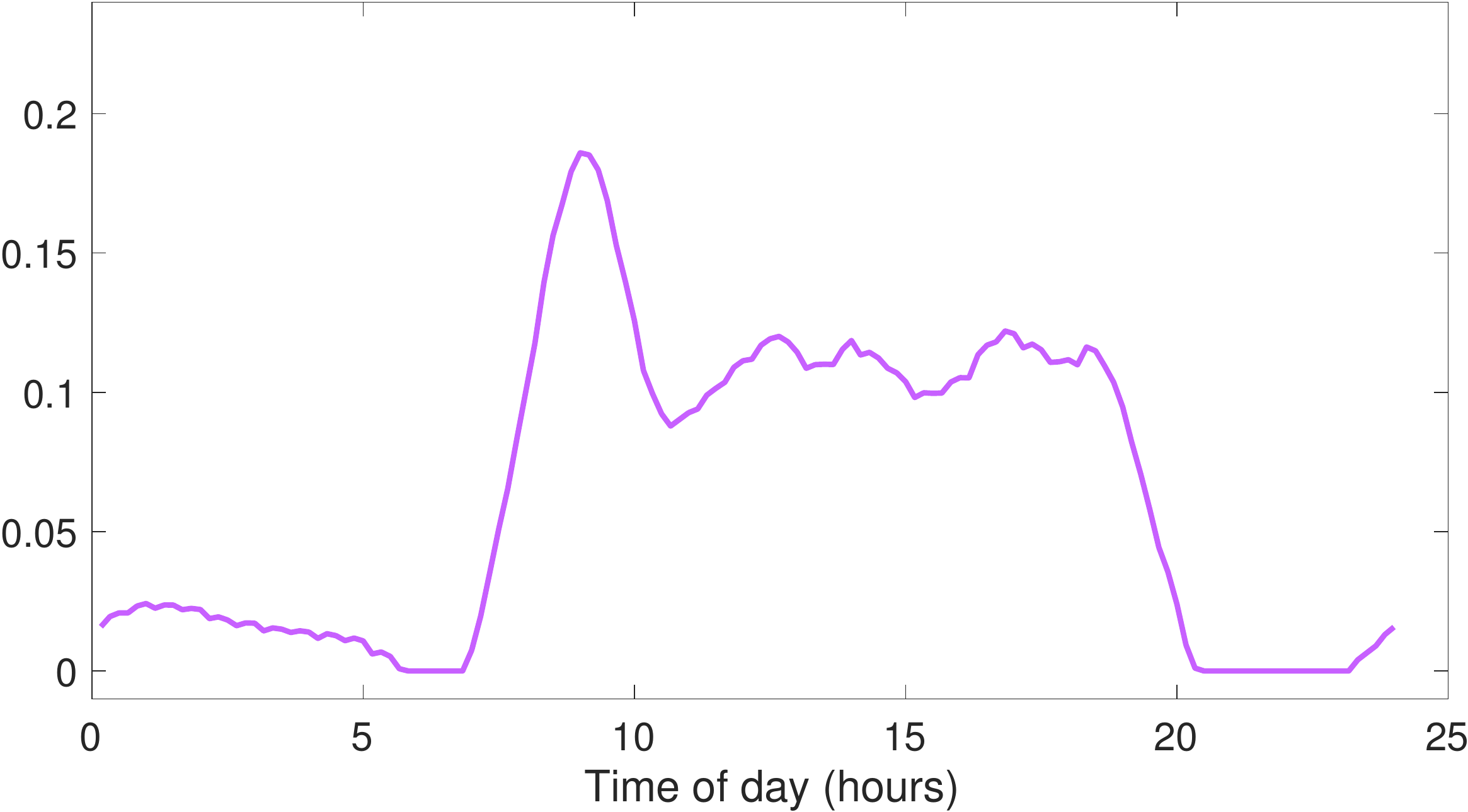} &  \includegraphics[scale=0.15]{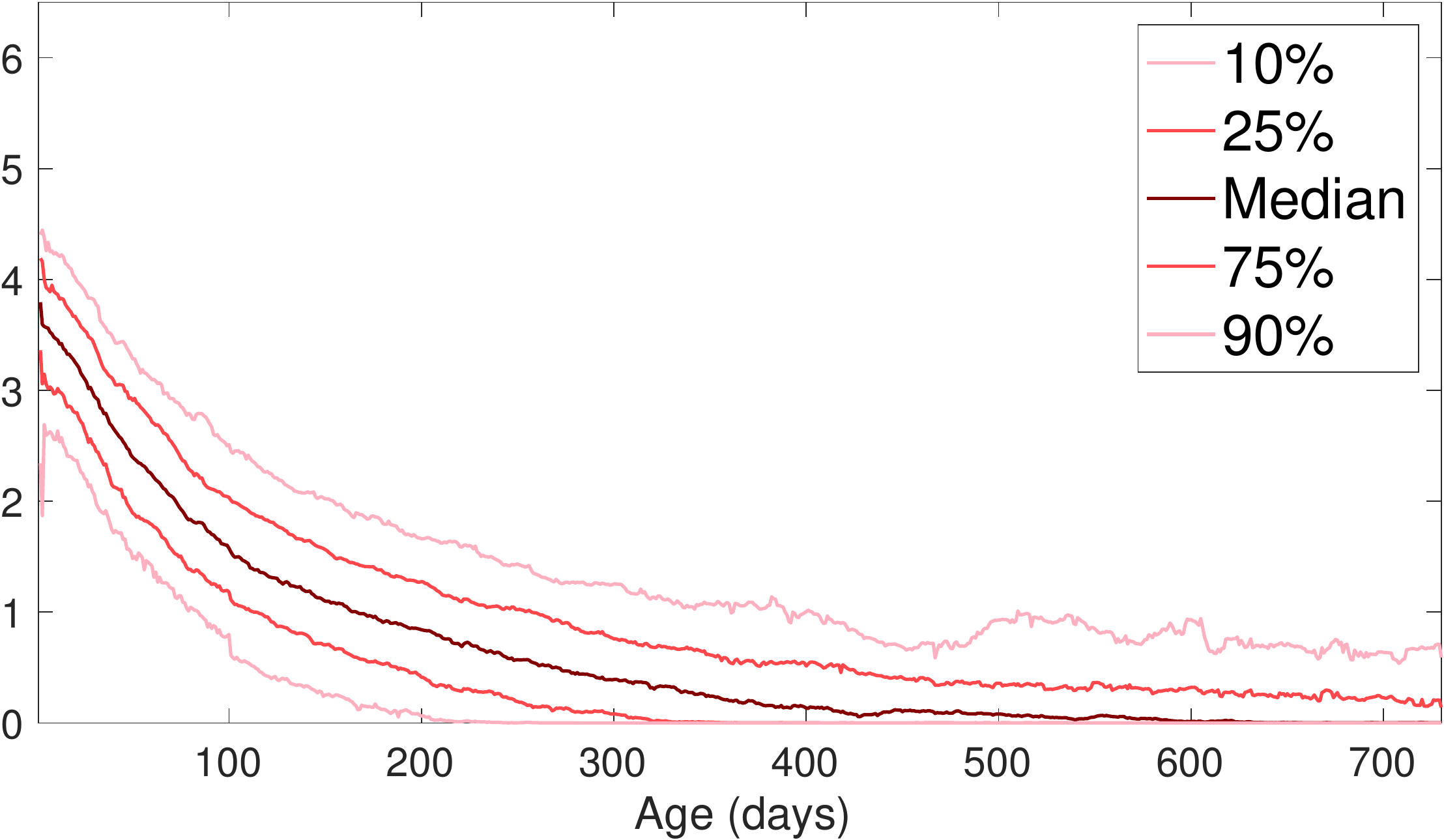} \\
2& \includegraphics[scale=0.15]{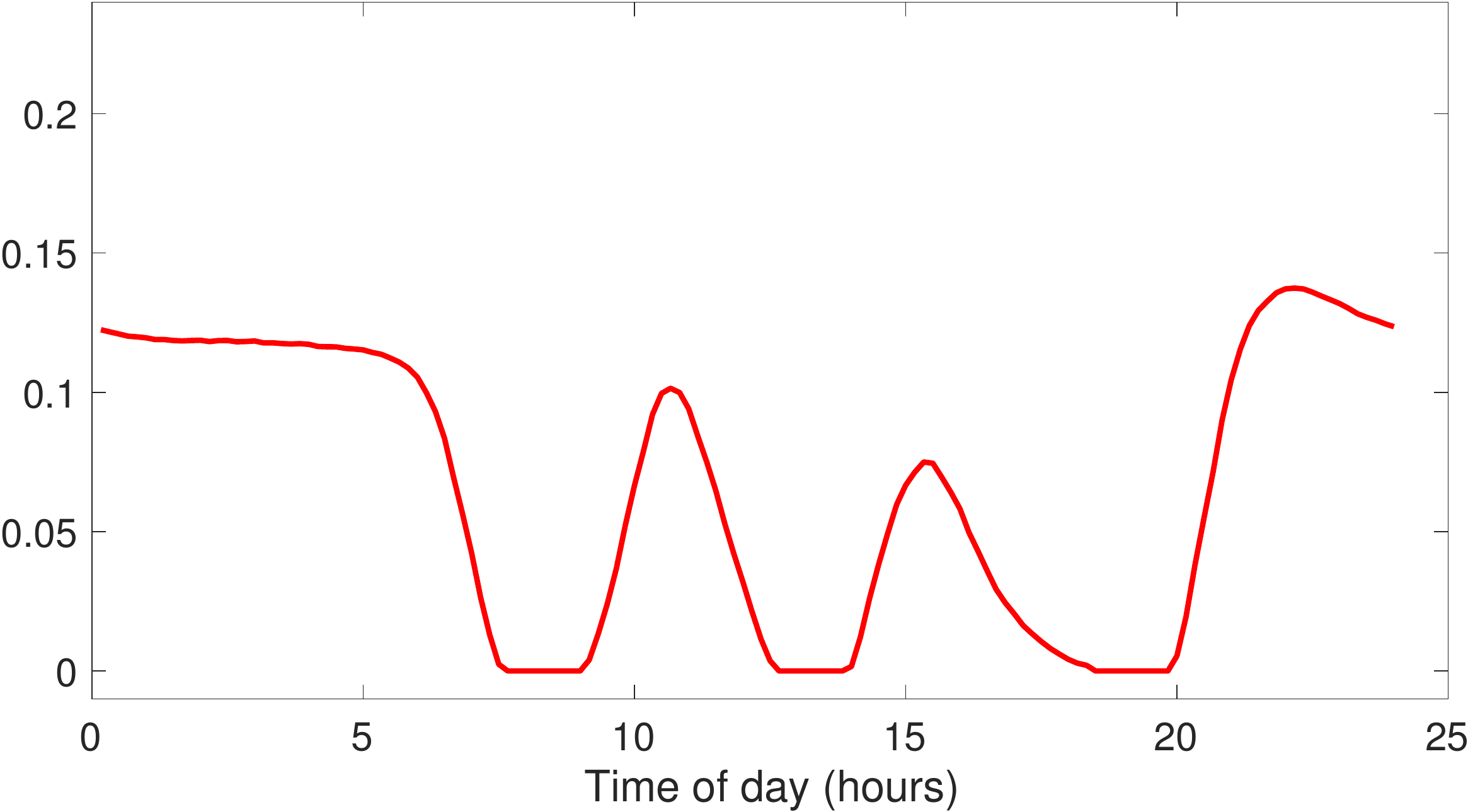} &  \includegraphics[scale=0.15]{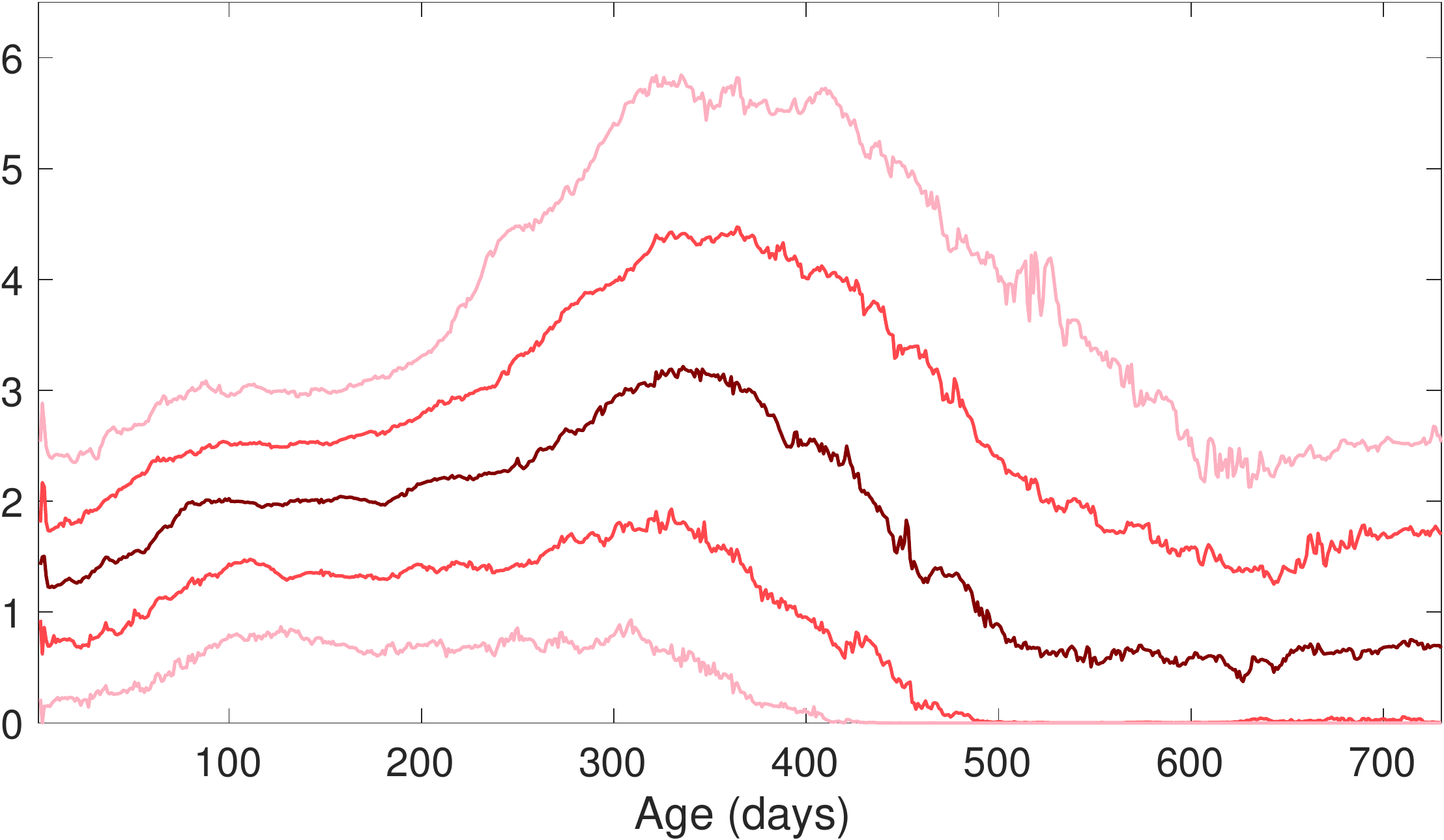} \\
3& \includegraphics[scale=0.15]{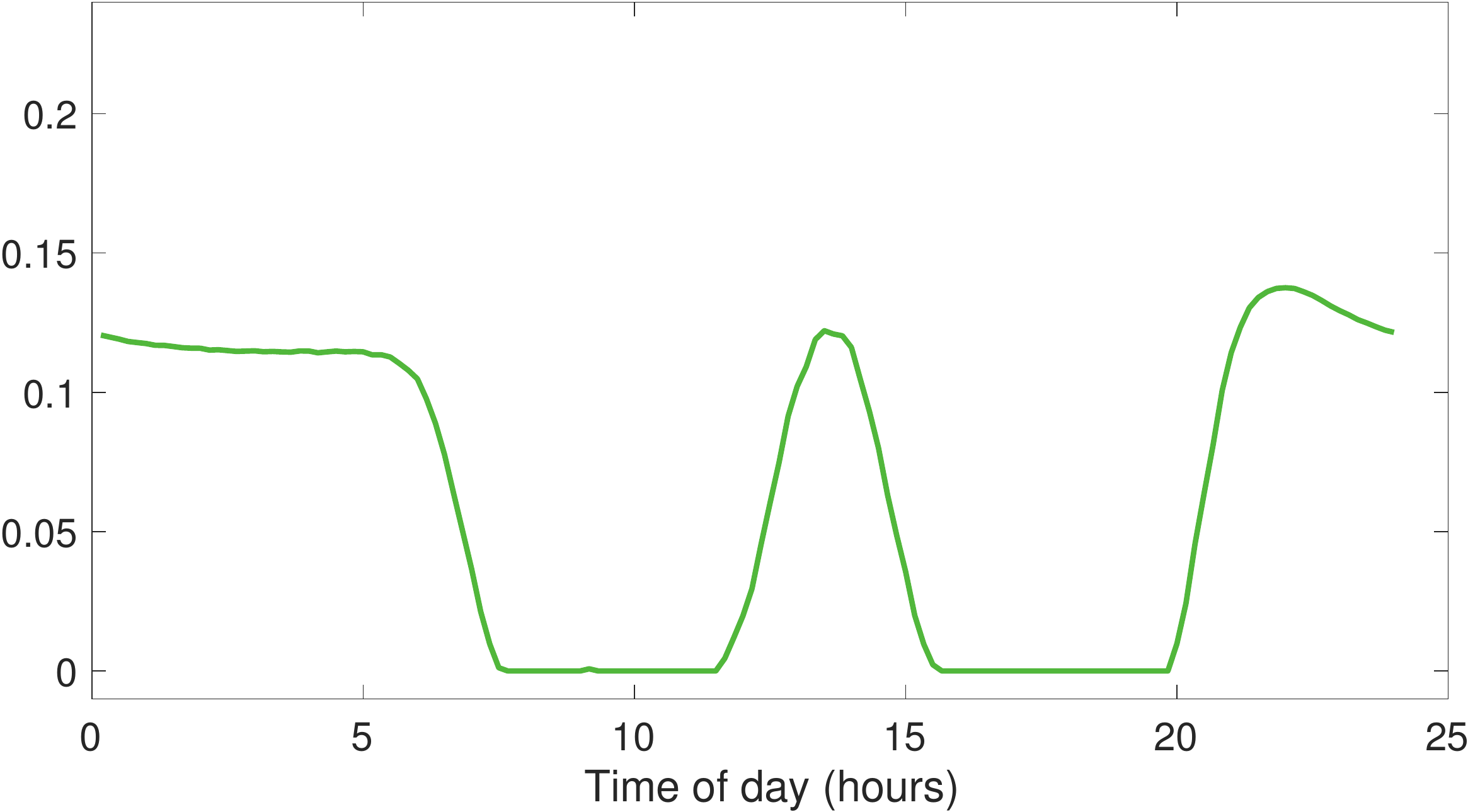} &  \includegraphics[scale=0.15]{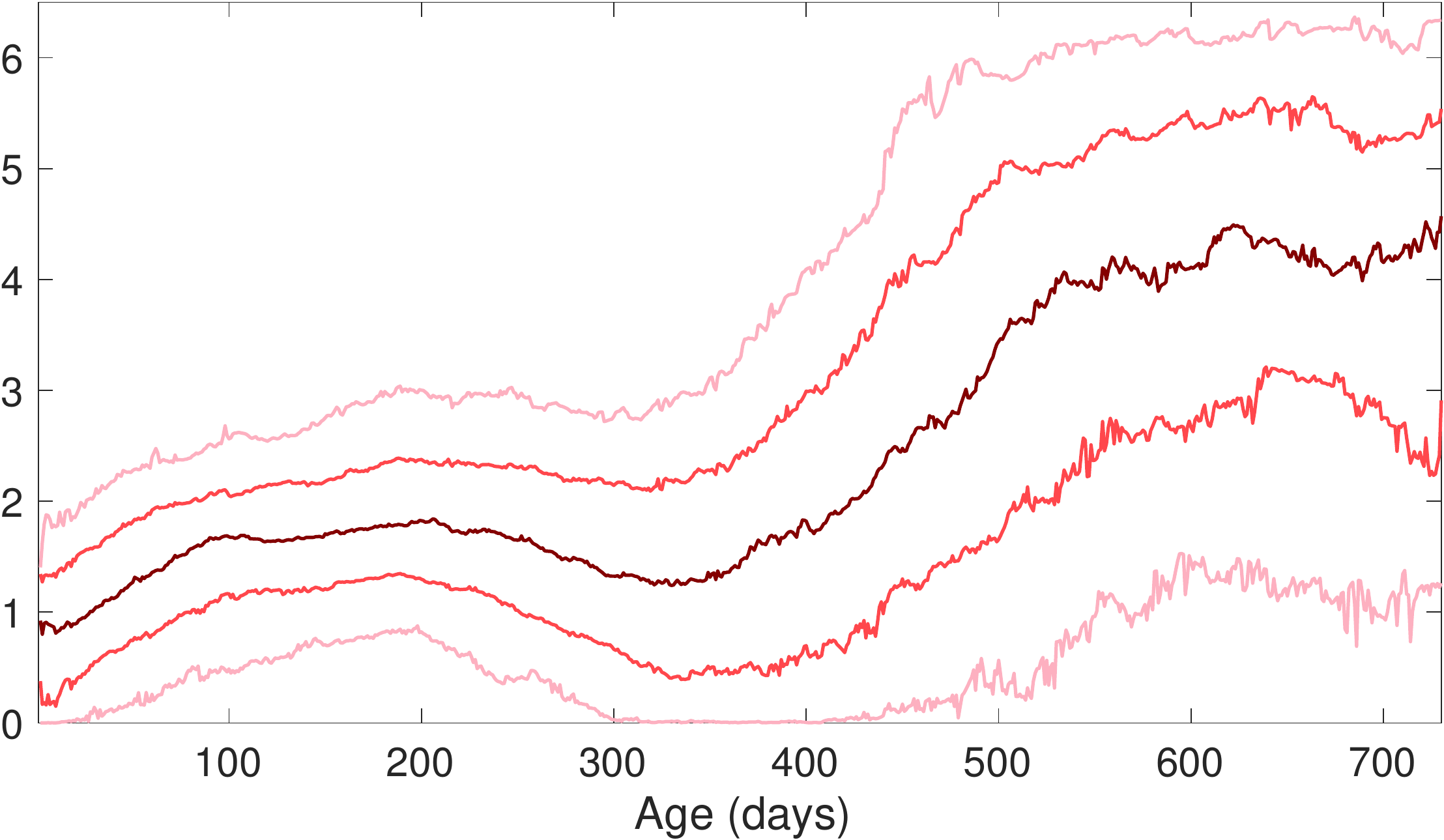} \\
4& \includegraphics[scale=0.15]{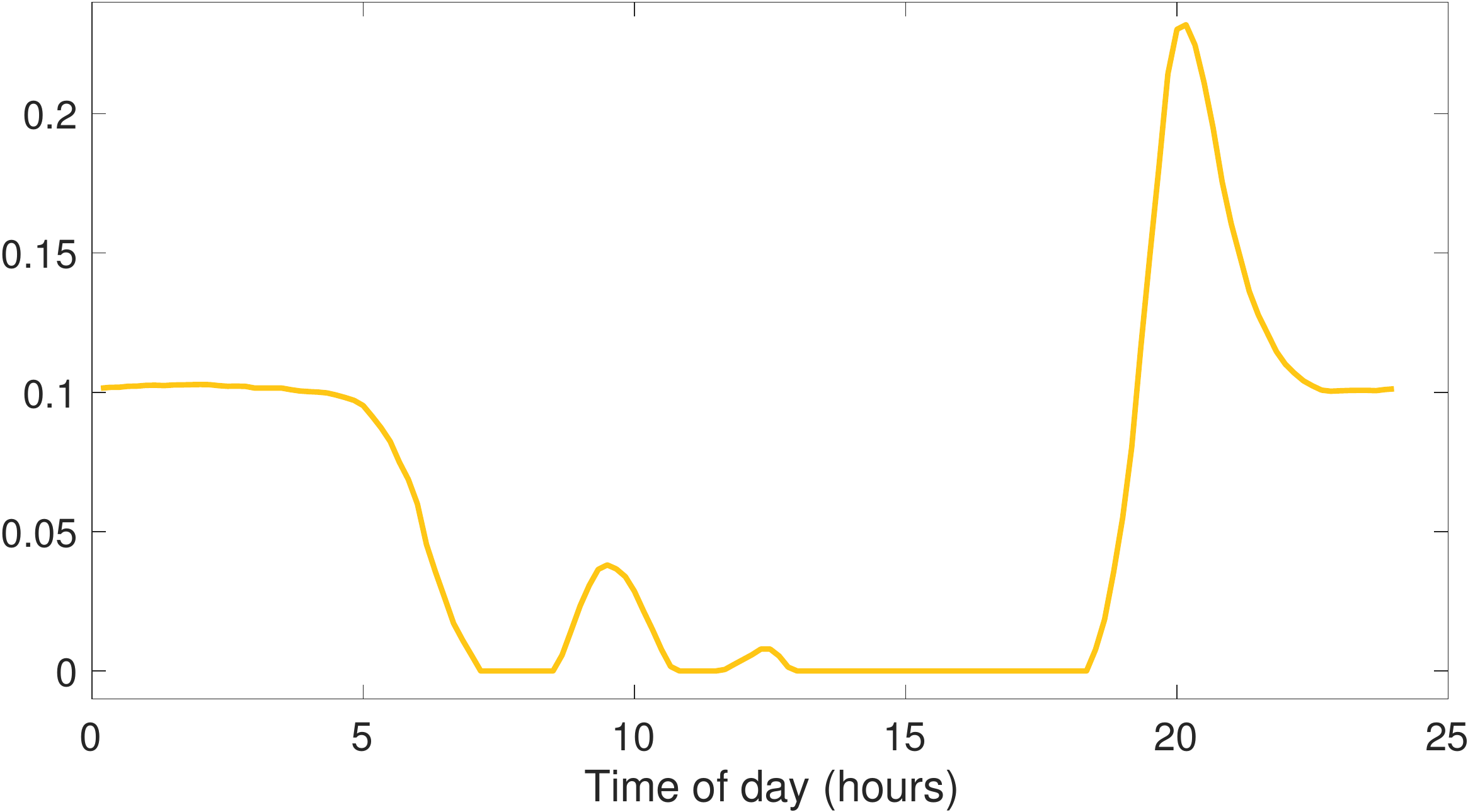} &  \includegraphics[scale=0.15]{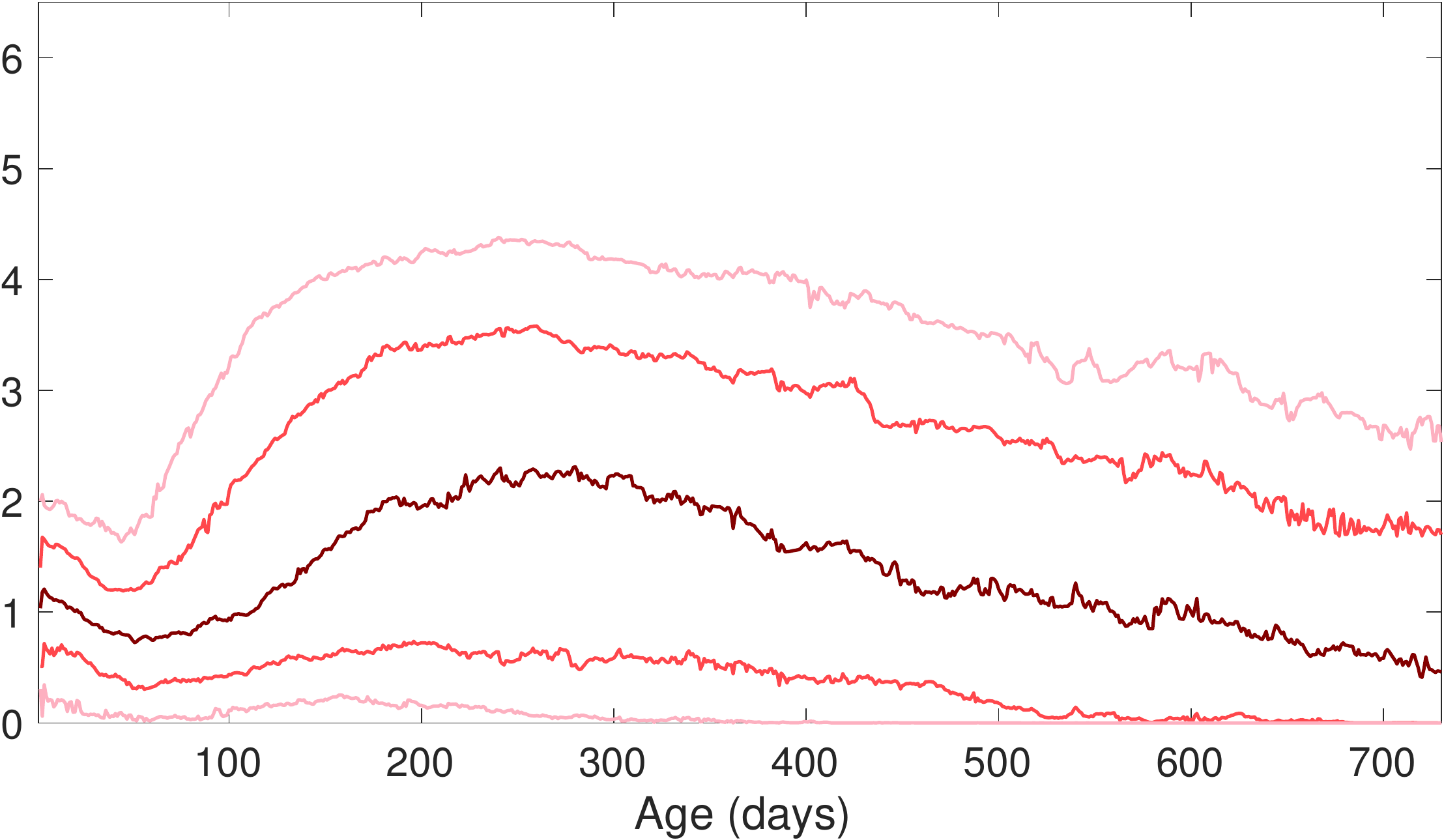} \\
5& \includegraphics[scale=0.15]{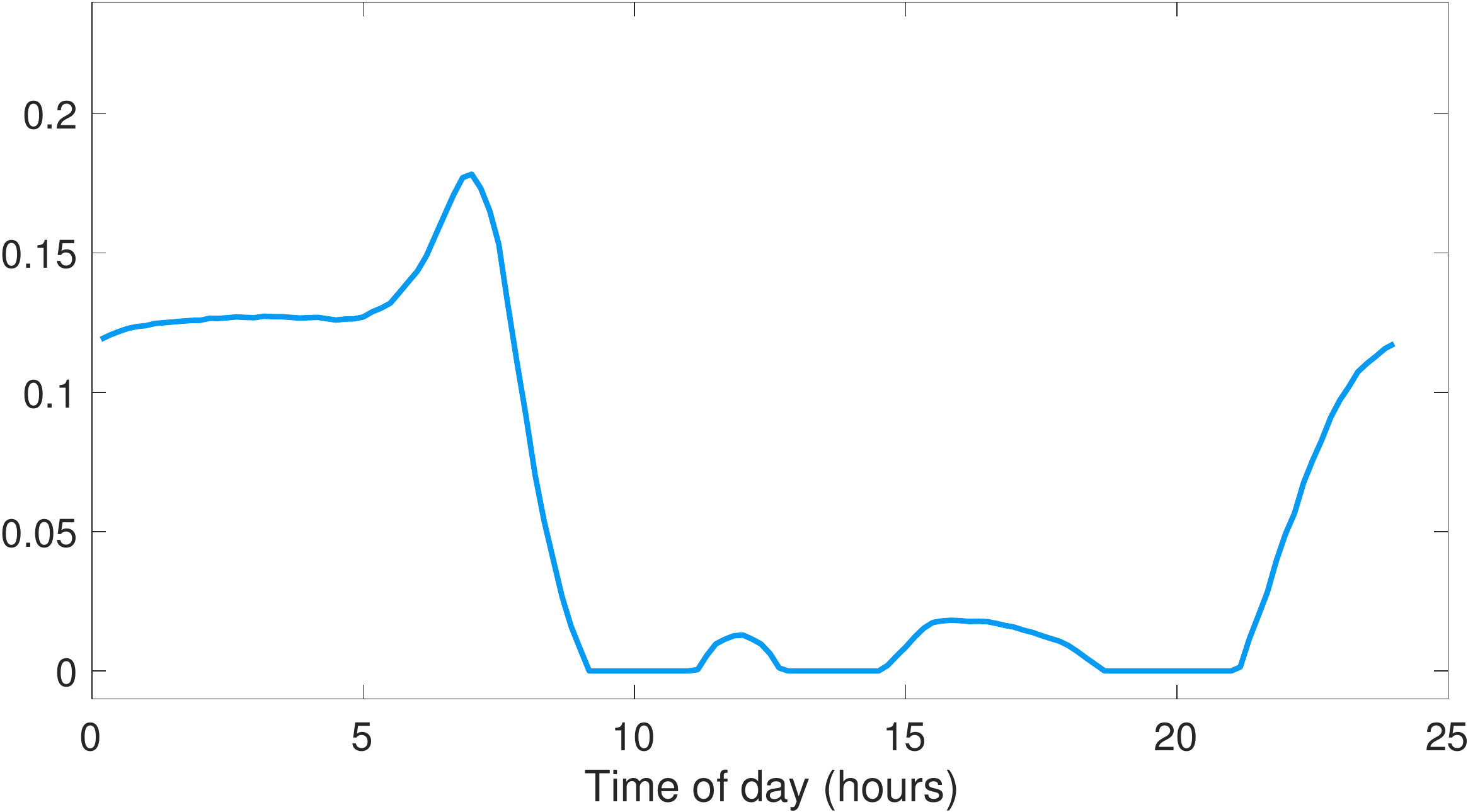} &  \includegraphics[scale=0.15]{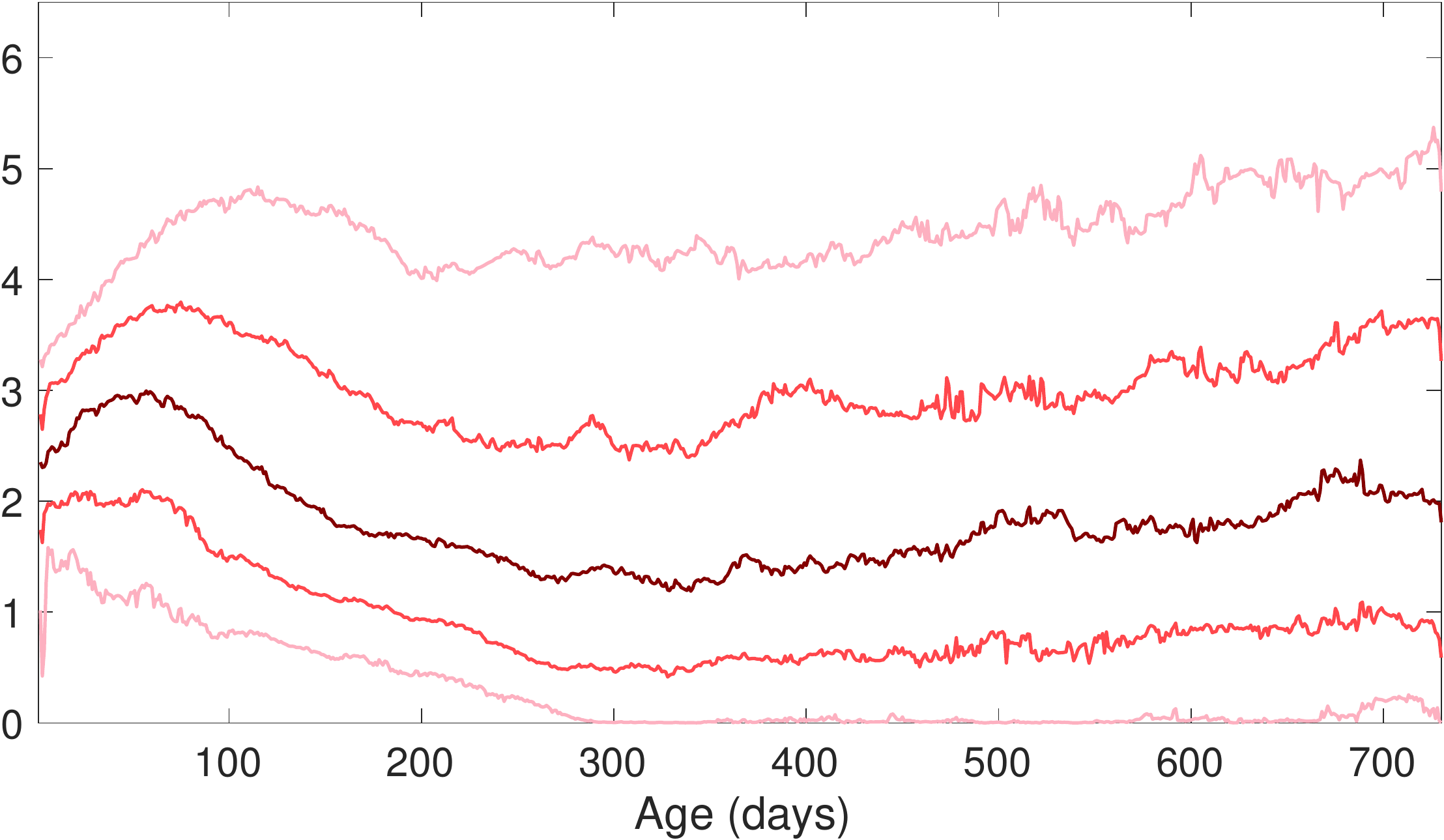} 
 \end{tabular}
 \caption{The left column shows the daily-sleep patterns learned by the proposed method. The right column shows the order statistics of the corresponding coefficients in the low-rank model.}
  \vspace{-3mm}
 \label{fig:pattern_development}
\end{figure}

The basis functions obtained by applying the proposed low-rank decomposition to the infant sleep data are directly interpretable in terms of sleeping behavior: (1) daytime sleep, (2) nighttime sleep with two naps, (3) nightime sleep with one nap, (4) nighttime sleep starting and ending early, (5) nighttime sleep starting and ending late. The patterns are shown in the left column of Figure~\ref{fig:pattern_development} (and also in the rightmost column of Figure~\ref{fig:daily_sleep_patterns}). The coefficients in the low-rank approximation of each infant indicate the degree to which each of these patterns is present at each day of age. They capture the dynamics of sleep development in terms of the learned daily-sleep patterns. We can therefore visualize and quantify these dynamics using the order statistics of the coefficients $C^{[n]}_1$, \ldots, $C^{[n]}_5$ over the whole dataset. Figure~\ref{fig:pattern_development} shows the median, the first and third quartiles and the 10th and 90th percentiles at each day of age.

Pattern 1 (daytime sleep) is prominent mainly during the first days of age. It decreases rapidly over the first year, signaling the emergence of circadian sleep. Pattern 2 (nighttime sleep with two naps) is particularly significant between 200 and 500 days of age. Pattern 3 (nighttime sleep with a single nap) emerges in the second year. The evolution of these three factors suggest that there are three stages in early-sleep development: irregular sleep, nighttime sleep with two naps, and nighttime sleep with a single nap. This is consistent with the literature on infant sleep development (see~\cite{henderson2010sleeping, weissbluth1995naps, iglowstein2003sleep}). Patterns 4 and 5 account for shifts in the nighttime-sleep schedules of the different infants in the population. Their order statistics remain stable once nighttime sleep consolidates.

In Section~\ref{sec:c} of the appendix, we show the results of applying our methodology to detect anomalous individuals, to cluster the cohort into groups with different sleeping tendencies, and to obtain improved predictions of future sleep behavior.

\section{Discussion and Future Directions} 

In this work we propose a nonparametric model for the analysis of multiple time series based on low-rank matrix estimation. The method is shown to extract meaningful patterns from infant-sleep data, which can be used to characterize general trends in the population, detect anomalous sleeping behavior, cluster the infants according to their sleeping habits, and perform forecasting. A limitation of our approach is that it does not account for shifts in the schedule within the individual time series. Modifying low-rank models to enforce invariance of the basis functions to such shifts is an interesting direction of future research. Other directions include using additional features (such as feeding habits) to improve forecasting, incorporating nonlinearities in the low-rank model, and designing parametric models of infant sleep behavior based on the patterns extracted by our method.   


{\small \bibliography{refs}}
\bibliographystyle{ieeetr}

\appendix
\section{Cohort and Data Preprocessing}
\label{sec:dataset}

The data used in this work were shared voluntarily by parents using a mobile-phone app, which allows multiple caregivers to track infant sleep. The raw measurements consist of a list of times when the caregiver reported the beginning or end of a sleeping period during the first two years of the infant's life. 

For each infant, days in the data that seem biologically implausible are discarded, since they are likely due to input errors. The criteria to discard days include sleeping periods longer than 16 hours, awake periods longer than 20 hours, no sleep during the night (9pm to 7am), and being surrounded by 5 days or more without data. 
More than half of the infants are missing at least 50\% of the days. Infants for which more than 90 \% of the days are missing are not included in the dataset. The resulting number of infants in the dataset is 701 (reduced from original 837 infants).

We represent the time series of sleep data for each infant $n$ as a matrix $Y^{[n]}$. Each row corresponds to a day (row 1 corresponds to the day the infant was born). Columns correspond to the time of the day, discretized into 144 10-minute intervals. Each entry $Y^{[n]}(t,i)$ contains the fraction of time the infant was reported to be asleep on day $t$ during the $i$th 10-minute interval (1 indicates that the infant slept during the whole period, 0 that they were awake). Figure~\ref{fig:random_samples} in the appendix shows several examples.

\section{Implementation details and parameter selection}
\label{sec:parameters}

We implement our methodology in Matlab, using the function \emph{nmf} to obtain an initialization for the basis functions based on nonnegative matrix factorization, and the function \emph{lsqlin} to solve the nonnegative least squares subproblems defined in Section~\ref{sec:optimization}. We fix a rank of $r:=5$ in our analysis. As shown in Figure~\ref{fig:svs} of the appendix, the singular values computed over the whole dataset suggest that the data is well approximated by a dimension-5 subspace (there is a jump between the 5th and 6th singular value). Figure~\ref{fig:rank} in the appendix shows basis functions learned using other ranks. The value of the smoothness regularization parameter is fixed to $\lambda := 10^5$. The results are very robust to changes in this parameter, as illustrated in Figure~\ref{fig:smoothness} of the appendix. The basis functions change only slightly when $\lambda$ is large enough, whereas the coefficient vectors become smoother gradually as we increase $\lambda$. 

\subsection{Optimization algorithm}
\label{sec:optimization}

To minimize the cost function~\eqref{eq:rnmf_cost_function} we apply alternating minimization. The coefficients are updated by solving $N$ separate nonnegative least-squares problem, one for each time series,
\begin{align}
\widehat{C}^{[n]}_{1}, \ldots,\widehat{C}^{[n]}_{r} := & \arg \min_{\substack{C_1,\ldots,C_r \in \R^{\abs{\ml{T}_n}} \\ C_{j}\brac{t} \geq 0, \forall j,t} }  \sum_{t\, \in \, \ml{T}_n} \sum_{i=1}^{\ell}  \brac{ Y^{[n]}\brac{t,i}  - \sum_{j=1}^{r} C_{j}\brac{t} F_{j}\brac{i} }^2 +  \lambda \sum_{j=1}^{r} \normTwo{ \ml{D} C_{j}}^2 \notag
\end{align}
for each $n$. These $N$ subproblems are completely decoupled and can consequently be solved in parallel. The basis functions are updated by solving the nonnegative least-squares problem,
\begin{align}
\widehat{F}_{1},\ldots,\widehat{F}_{r} := & \arg \min_{\substack{F_1,\ldots,F_r \in \R^{\ell} \\ F_{j}\brac{i} \geq 0, \forall j,i} } \sum_{n = 1}^{N} \sum_{t\, \in \, \ml{T}_n} \sum_{i=1}^{\ell}  \brac{ Y^{[n]}\brac{t,i}  - \sum_{j=1}^{r} C^{[n]}_{j}\brac{t} F_{j}\brac{i} }^2,
\end{align}
and then normalizing each function to have unit $\ell_2$ norm.
\section{Other applications\label{sec:c}}
\subsection{Outlier detection}
\label{sec:outlier_detection}


\begin{figure}[hp]
\hspace{-0.9cm} 
\begin{tabular}{ >{\centering\arraybackslash}m{0.33\linewidth} >{\centering\arraybackslash}m{0.33\linewidth} >{\centering\arraybackslash}m{0.33\linewidth}   }
 &  Raw data &  \\
 &\includegraphics[width=\linewidth]{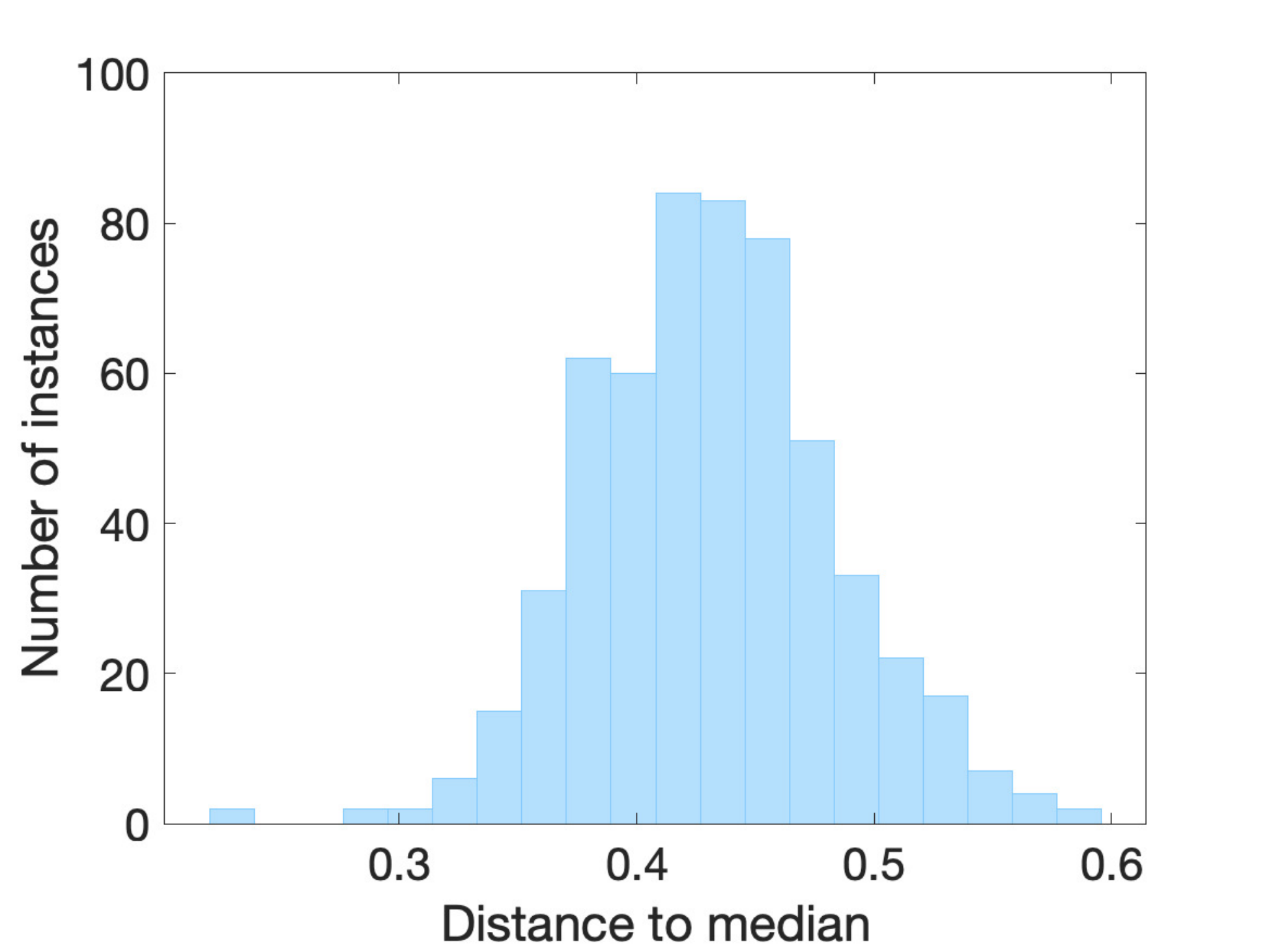}&\\
Daytime sleep &  Two naps & One nap \\
 \includegraphics[width=\linewidth,height=0.8\linewidth]{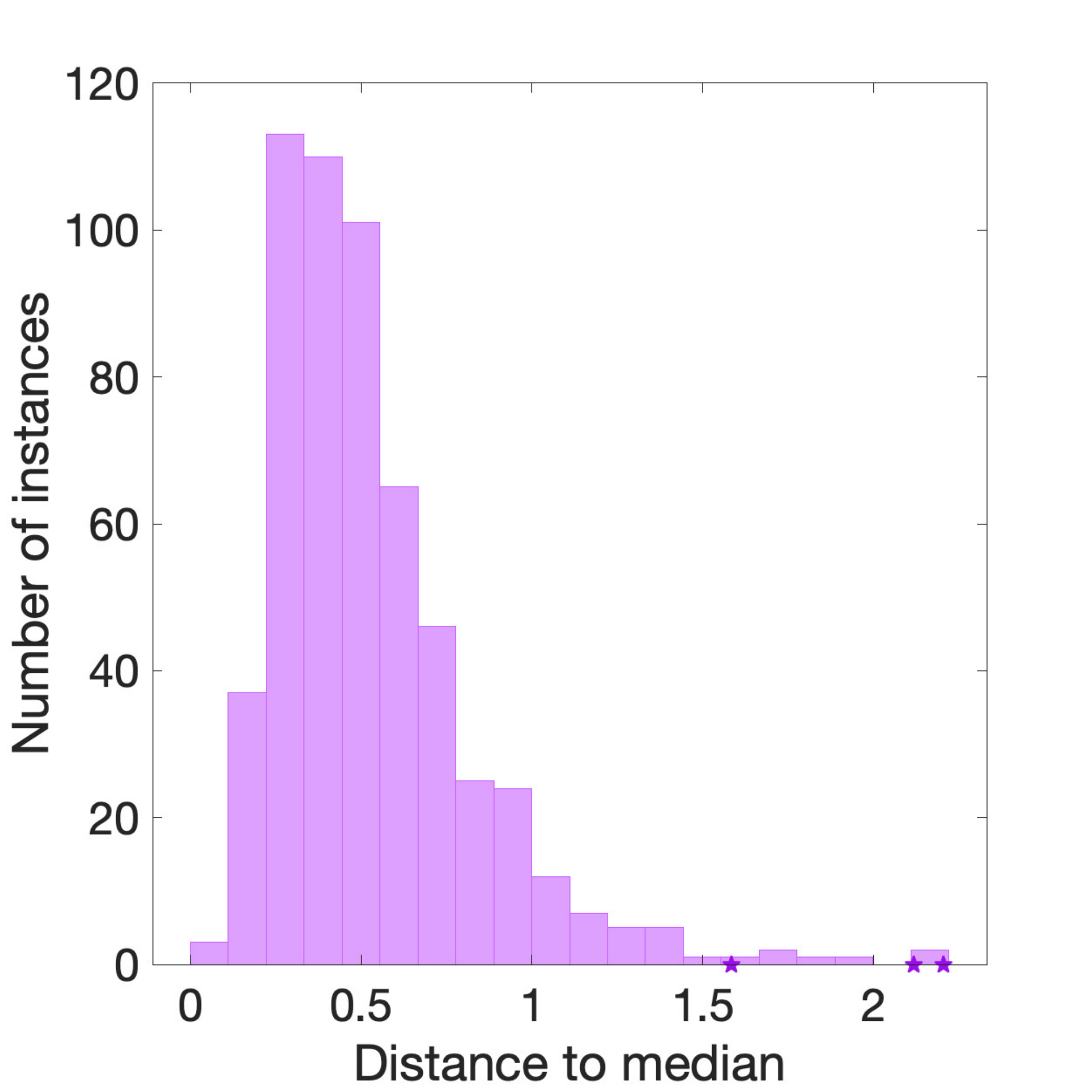}  & \includegraphics[width=\linewidth,height=0.8\linewidth]{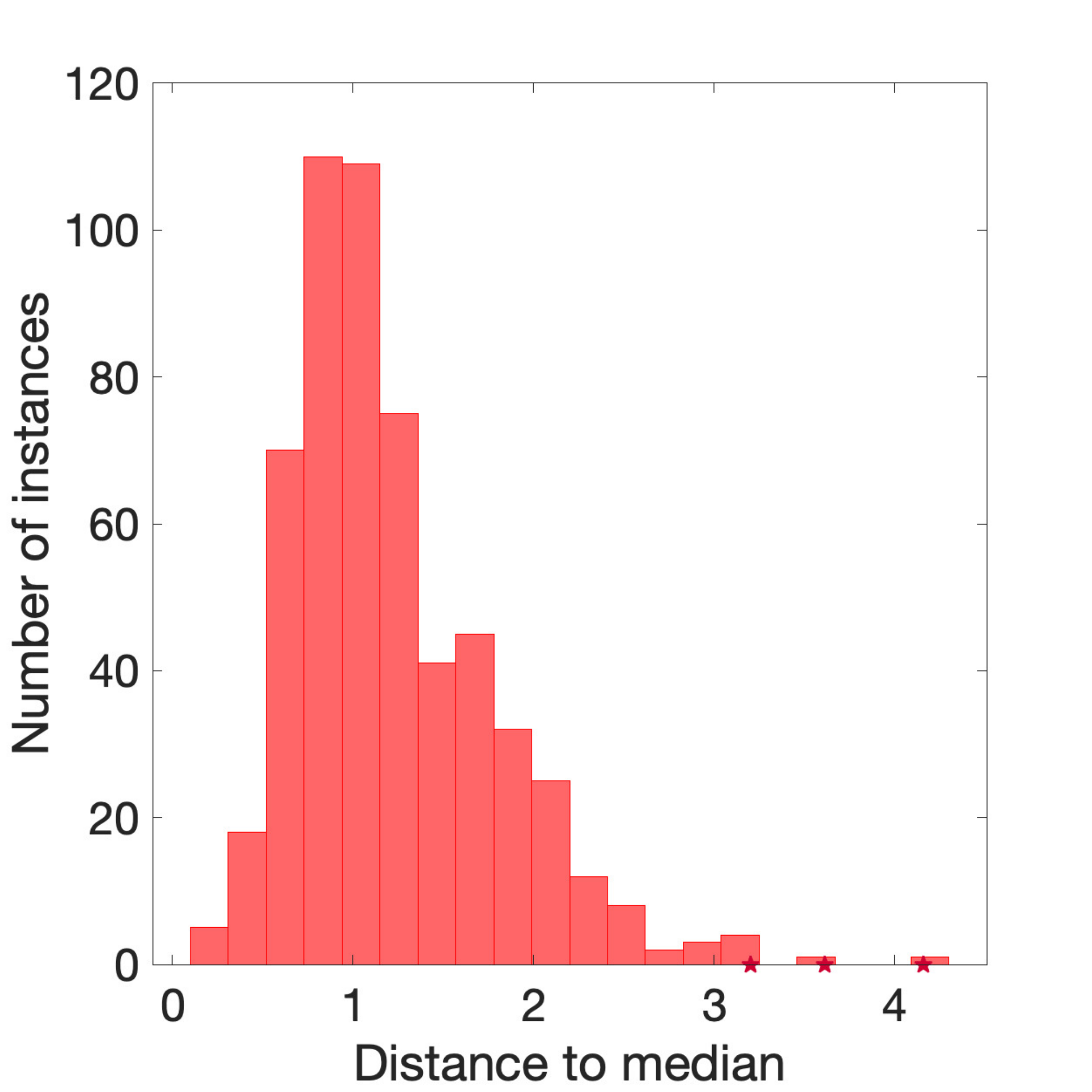} & \includegraphics[width=\linewidth,height=0.8\linewidth]{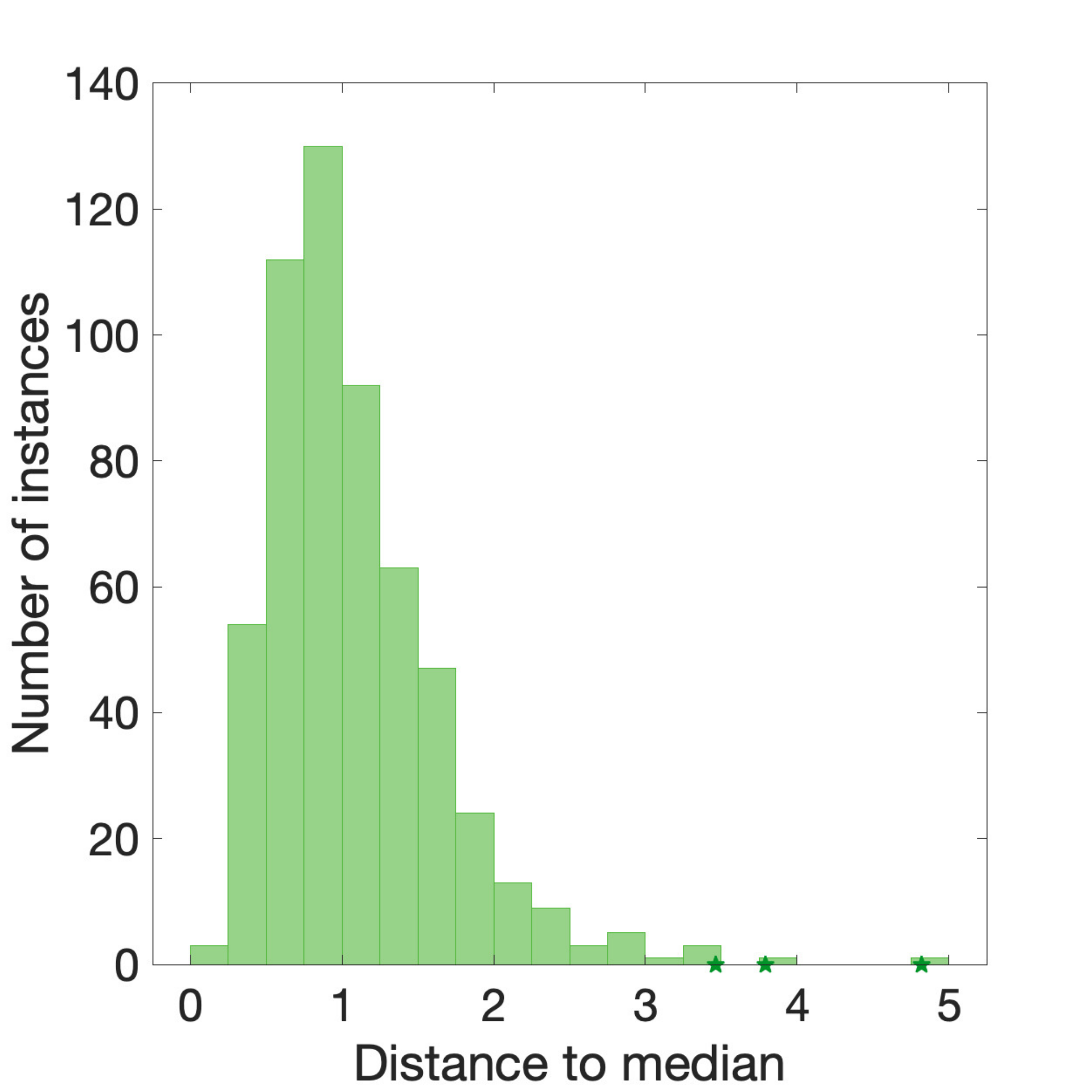} \\
 \includegraphics[width=0.83\linewidth]{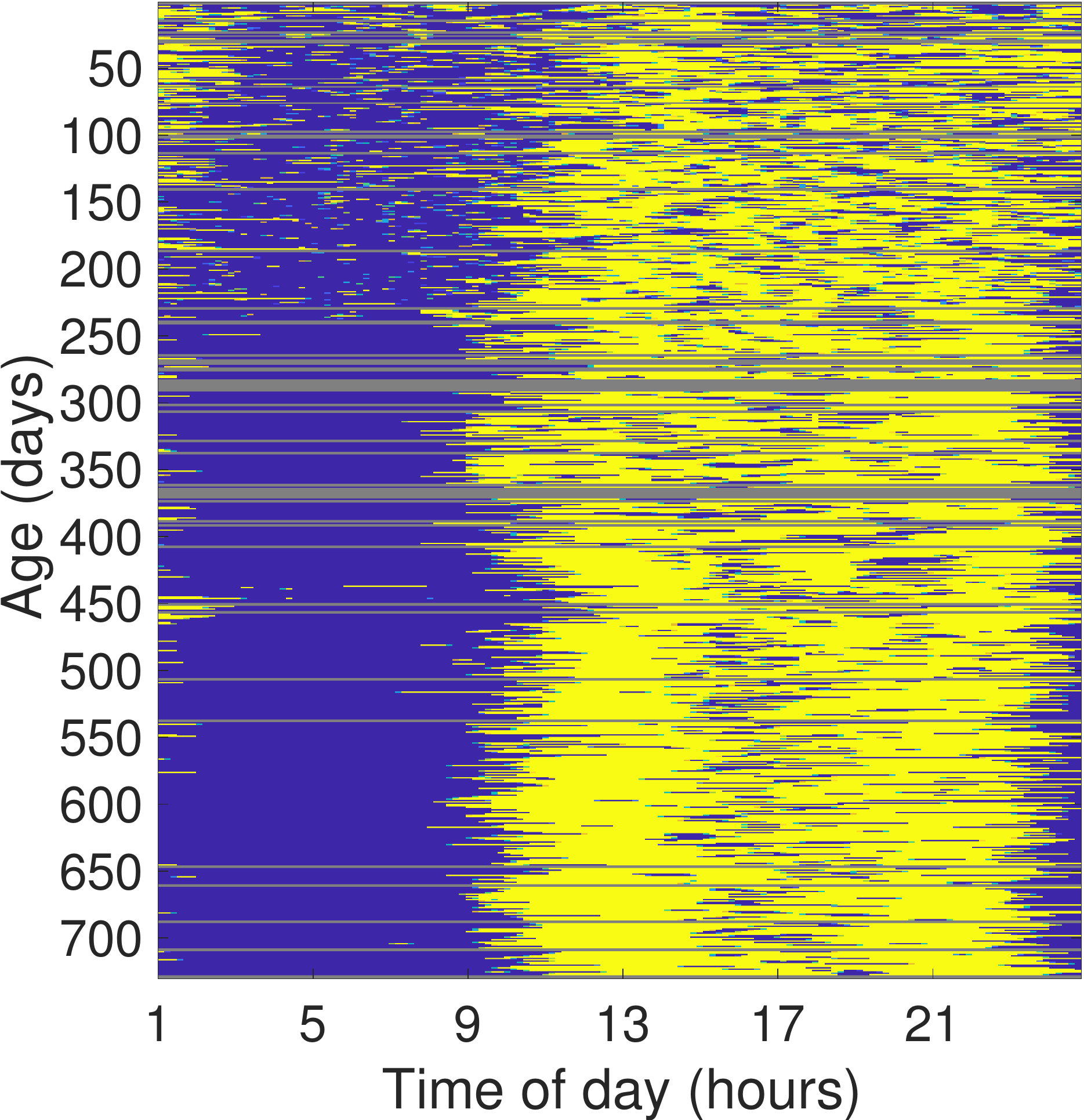} & \includegraphics[width=0.83\linewidth]{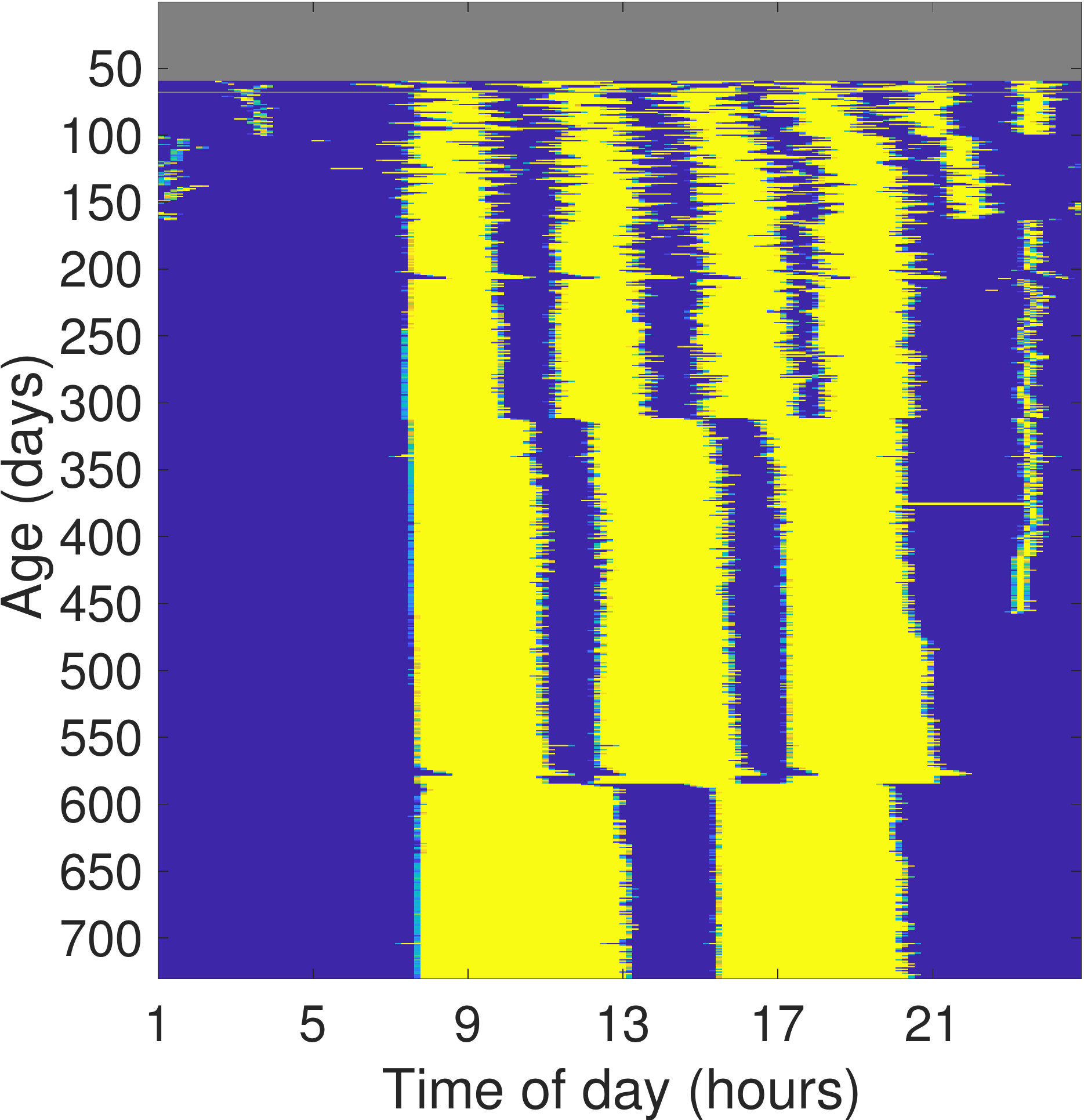} & \includegraphics[width=0.83\linewidth]{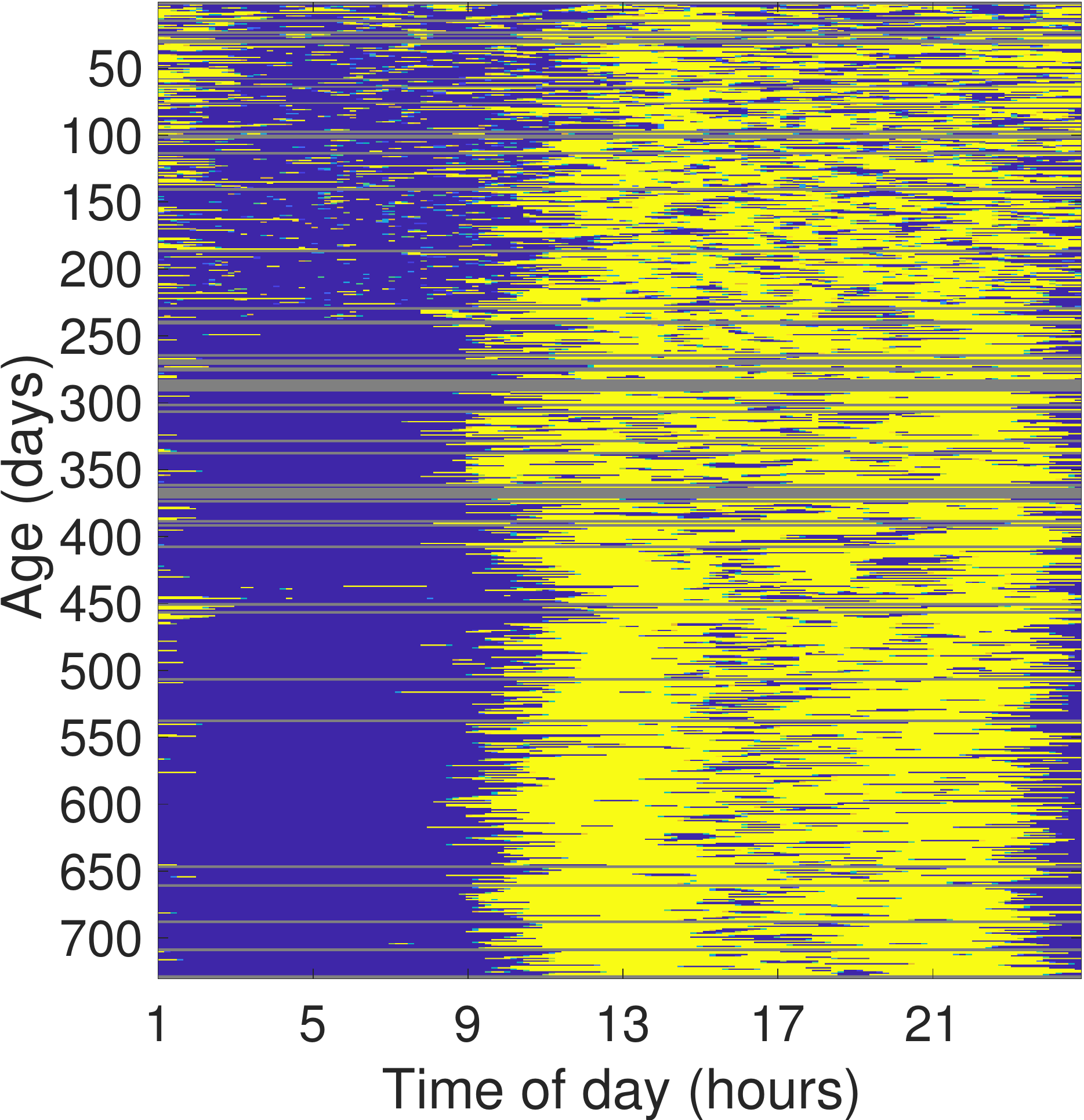} \\
 \includegraphics[width=0.83\linewidth]{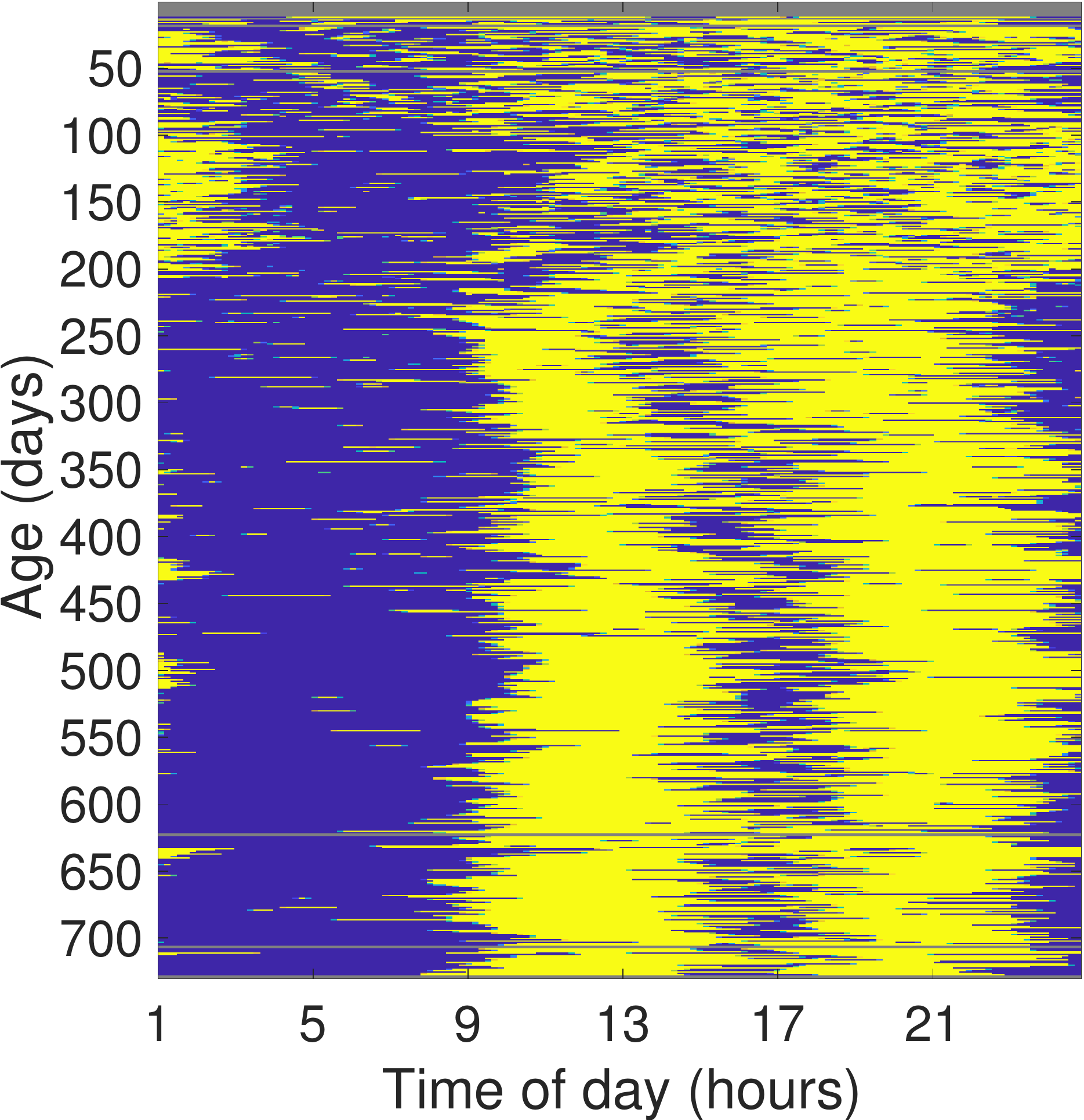} & \includegraphics[width=0.83\linewidth]{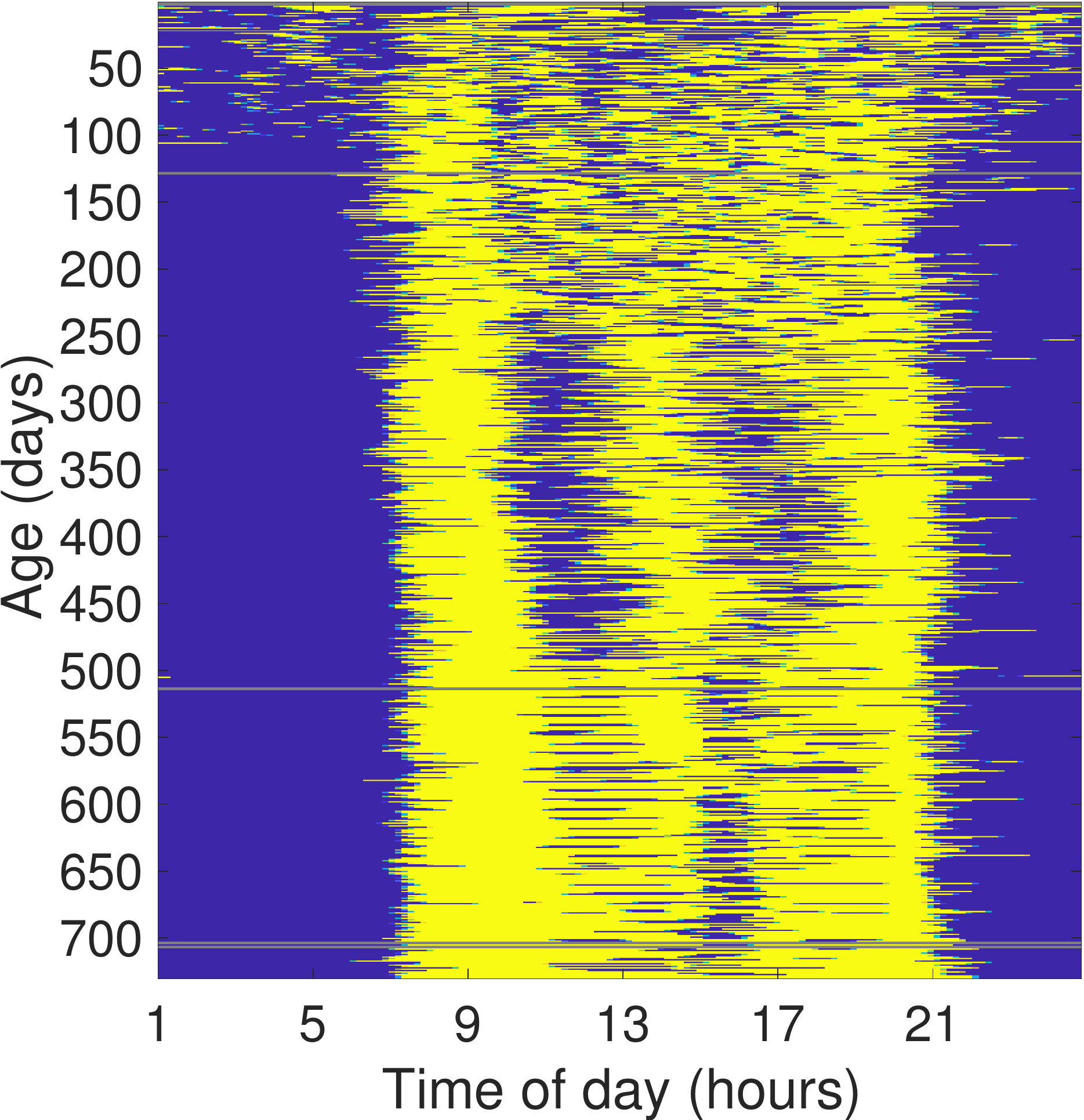} & \includegraphics[width=0.83\linewidth]{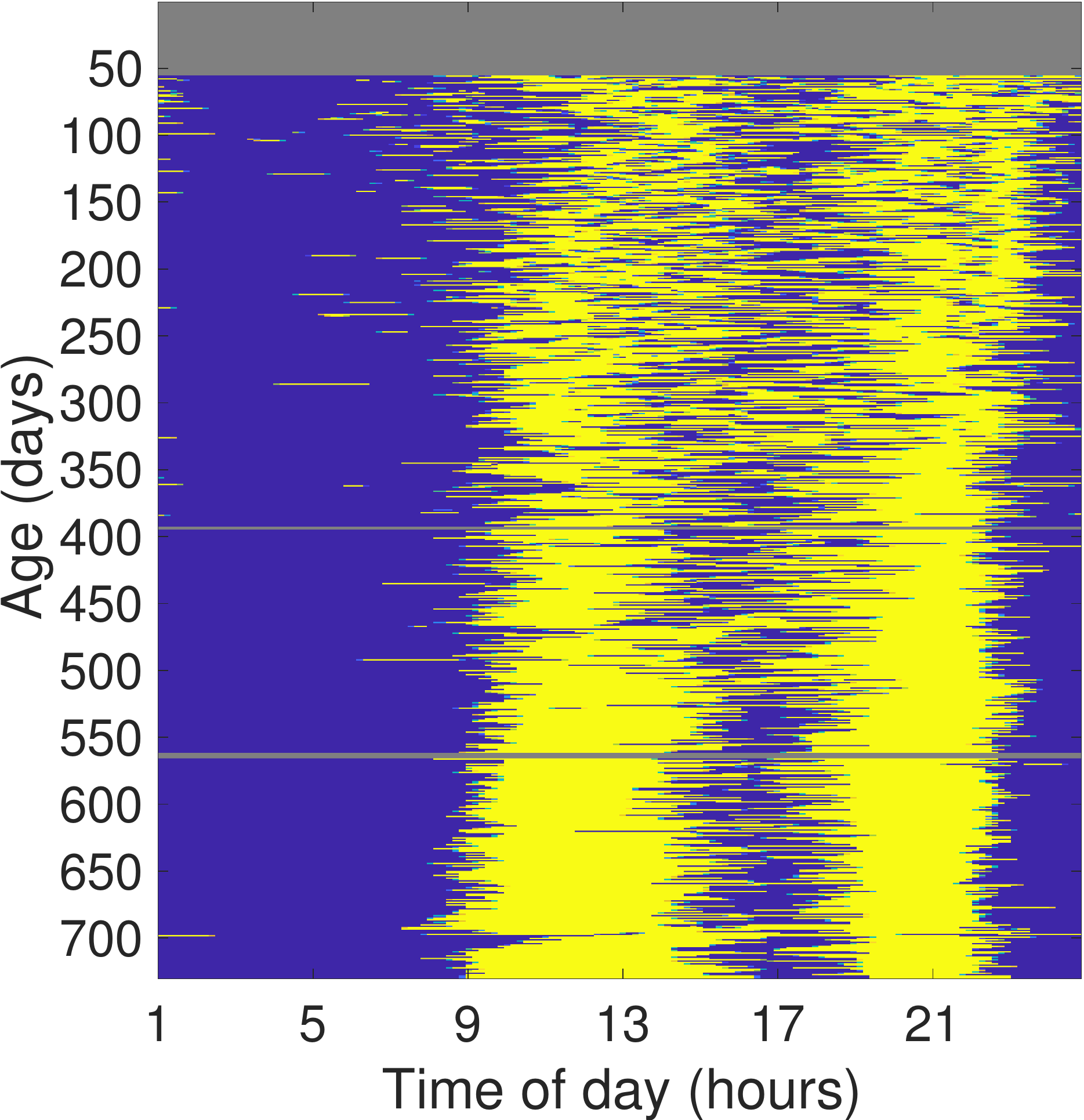} \\
 \includegraphics[width=0.83\linewidth]{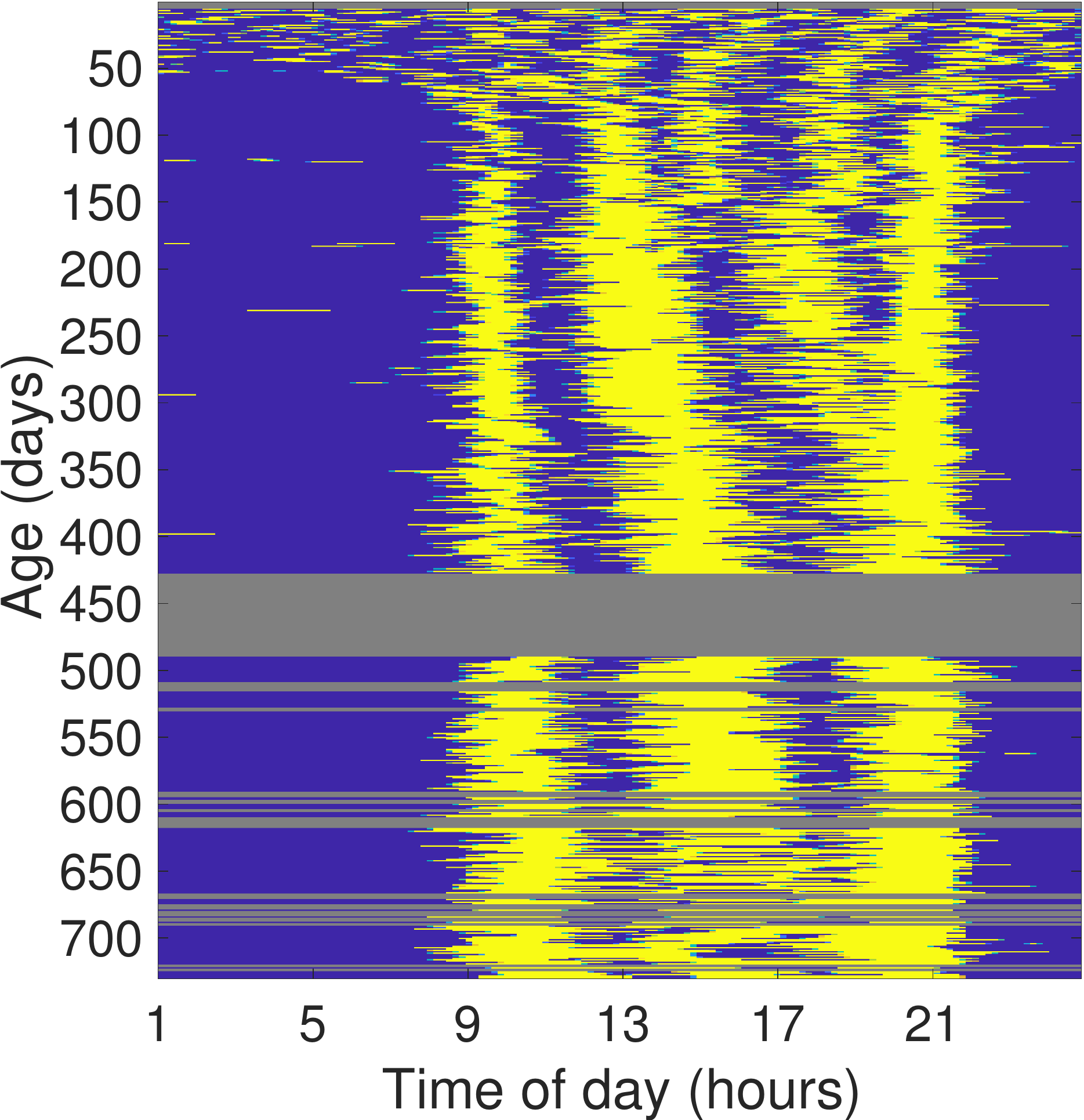} & \includegraphics[width=0.83\linewidth]{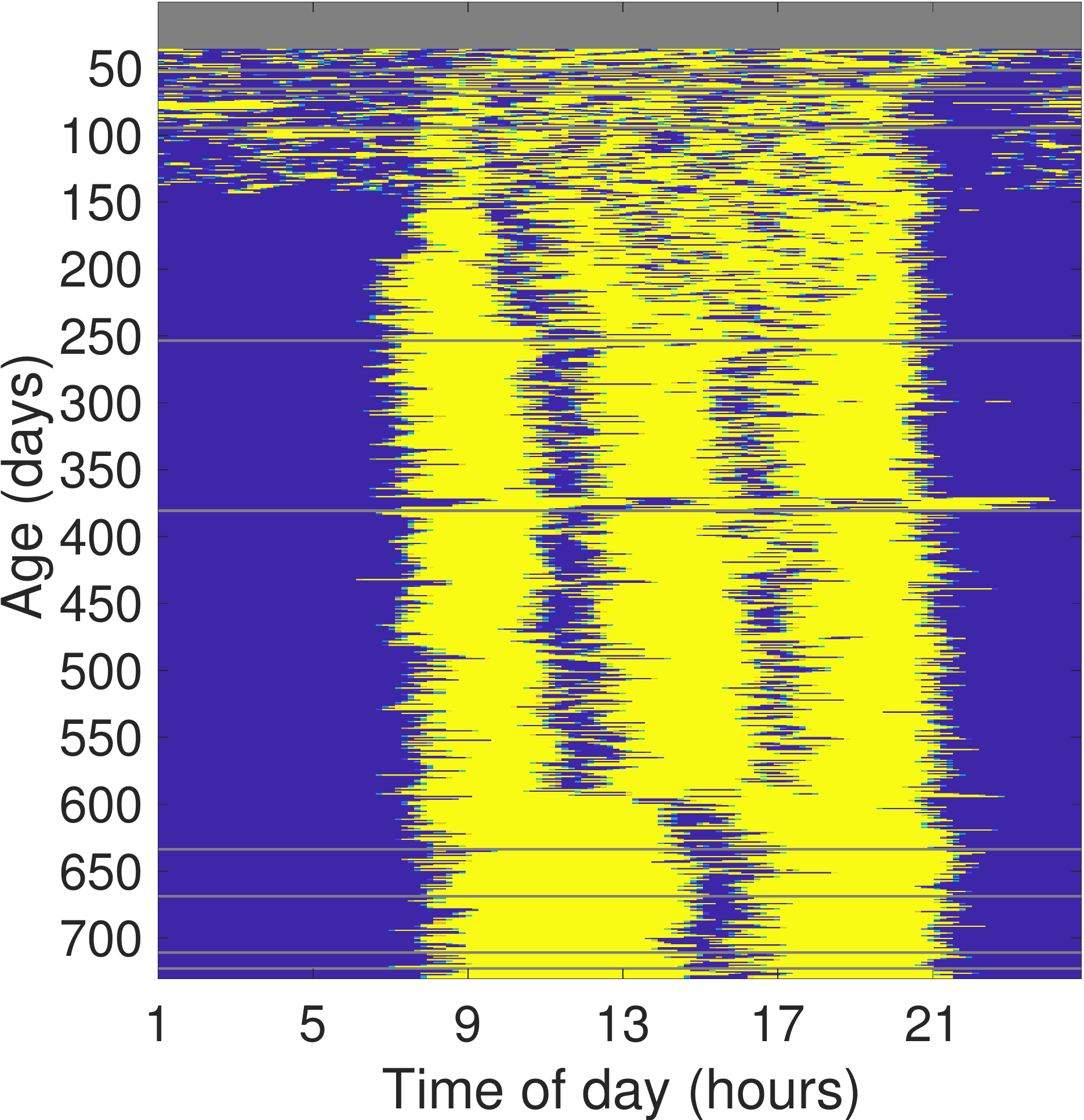} & \includegraphics[width=0.83\linewidth]{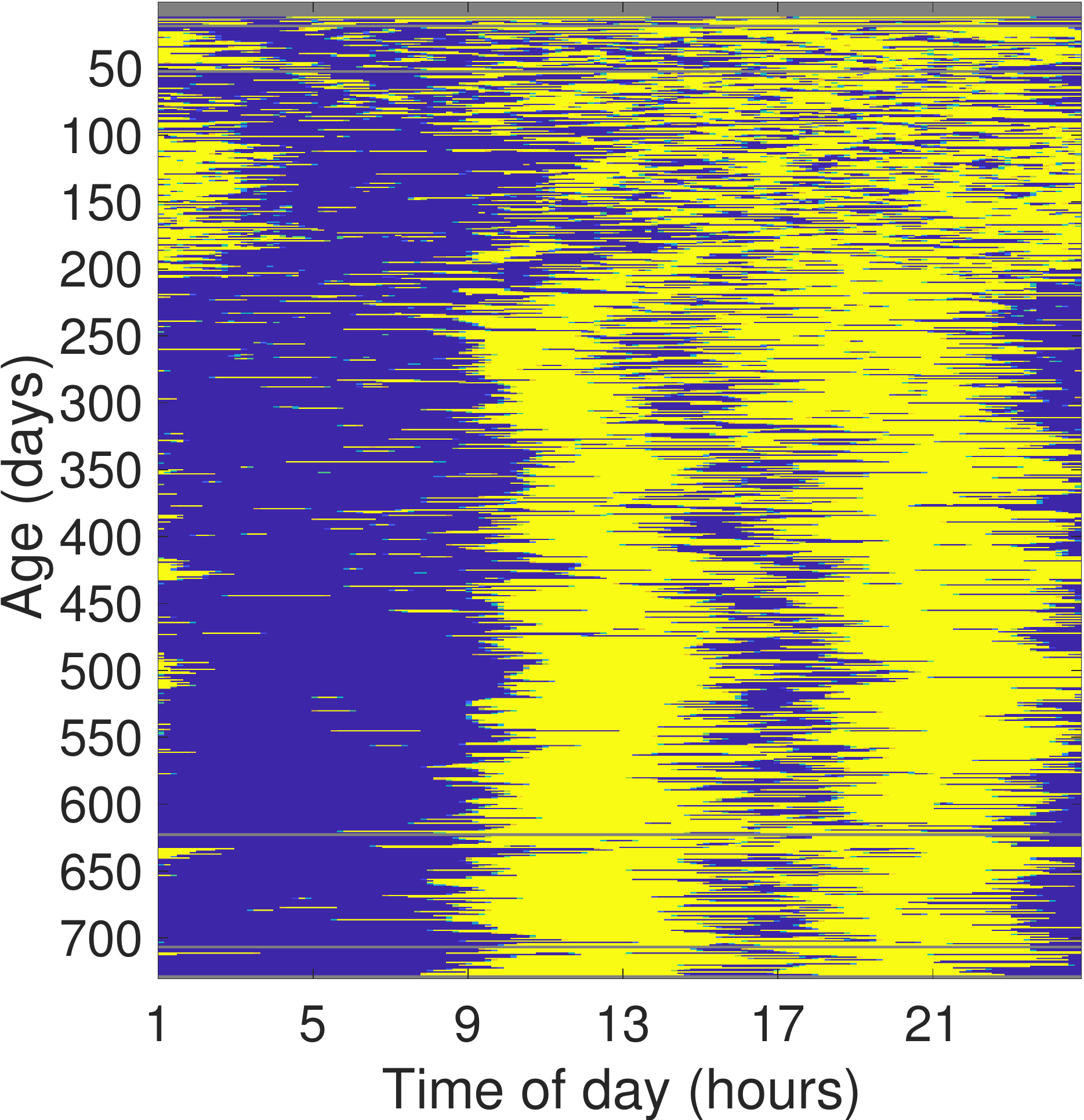} 
 \end{tabular}
 \caption{The top graph shows the histogram of the $\ell_2$-norm distances between the time series in the infant-sleep dataset and their median. The second row shows histograms of distances between the low-rank components extracted by the proposed methodology and the corresponding median (shown in Figure~\ref{fig:pattern_development}). For each component we show three examples of outliers marked with stars on the histogram. Yellow indicates that the infant is awake, and blue that they are asleep. Gray values indicate missing data.}
 \label{fig:outliers}
\end{figure}

Identifying anomalous sleeping patterns in infants has great potential clinical value. Unfortunately, detecting outliers in a set of time series is very challenging because the data are usually very high dimensional. Points in high-dimensional spaces tend to all be at a similar distance from each other, so that comparing these distances is not meaningful, a phenomenon commonly known as the \emph{curse of dimensionality}. The histogram at the top of Figure~\ref{fig:outliers} illustrates this for our infant-sleep data. We tackle this issue by using the coefficients in our low-rank model as a representation of reduced dimensionality. This has the added benefit that the outliers can be interpreted in terms of the sleeping patterns described in Section~\ref{sec:results}. We are particularly interested in the first three patterns (irregular sleep, nighttime sleep with two naps, and nighttime sleep with a single nap), since the remaining two are associated to shifts in waking hours, which are not as meaningful from a clinical point of view. The second row of Figure~\ref{fig:outliers} displays the histogram of distances to the medians of each component (shown in Figure~\ref{fig:pattern_development}). In contrast to the histogram for the raw data, the distribution of distances is heavily skewed towards smaller distances. This makes it possible to characterize outliers as infants that are on the right tail of the histogram. 

Figure~\ref{fig:outliers} also shows several outliers associated to each of our patterns of interest. By comparing to the general trends in Figure~\ref{fig:pattern_development}, we can analyze how they deviate from the normal behavior in the population. For the daytime-sleep pattern, the first and second outliers develop circadian sleep appears later than usual, whereas the third develops structured naps very early. For the two-nap pattern, the first outlier develops three structured naps exceptionally early, the second has a very fuzzy two-nap schedule, and the third one has two naps until an abnormally late age. For the one-nap pattern, the first outlier never develops a one-nap habit, whereas the second and third have a fuzzy one-nap habit that starts unusually early.

\subsection{Clustering}
\label{sec:clustering}
\begin{figure}[t]
\hspace{-0.6cm} 
\begin{tabular}{ >{\centering\arraybackslash}m{0.23\linewidth} >{\centering\arraybackslash}m{0.23\linewidth} >{\centering\arraybackslash}m{0.23\linewidth}
>{\centering\arraybackslash}m{0.23\linewidth}}
\multicolumn{2}{c}{\textbf{Raw Data}} & \multicolumn{2}{c}{\textbf{First 3 NMF-TS coefficients}}\\
Cluster 1 Mean & Cluster 2 Mean & Cluster 1 Mean & Cluster 2 Mean\\
 \includegraphics[width=\linewidth]{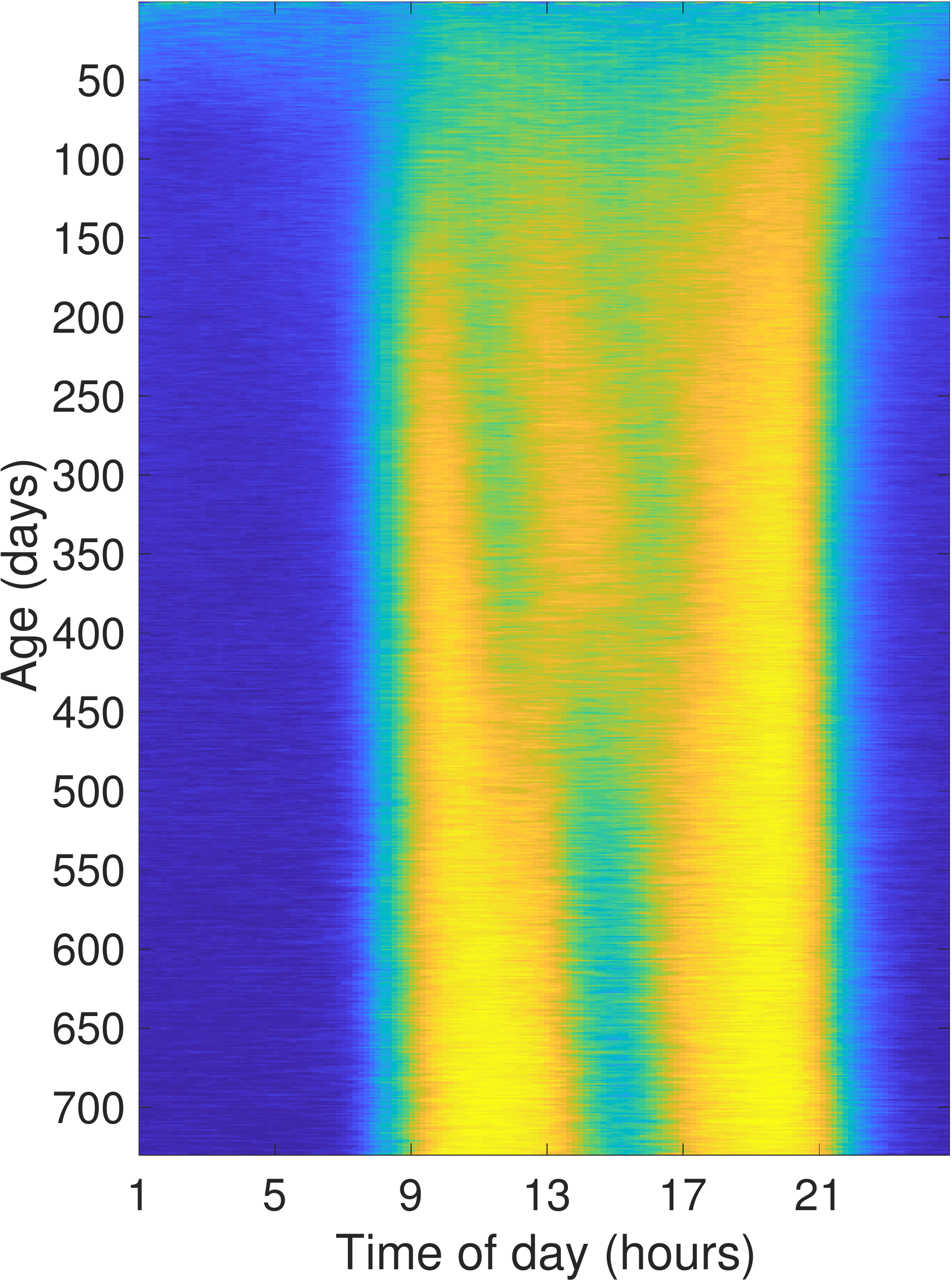}  & \includegraphics[width=\linewidth]{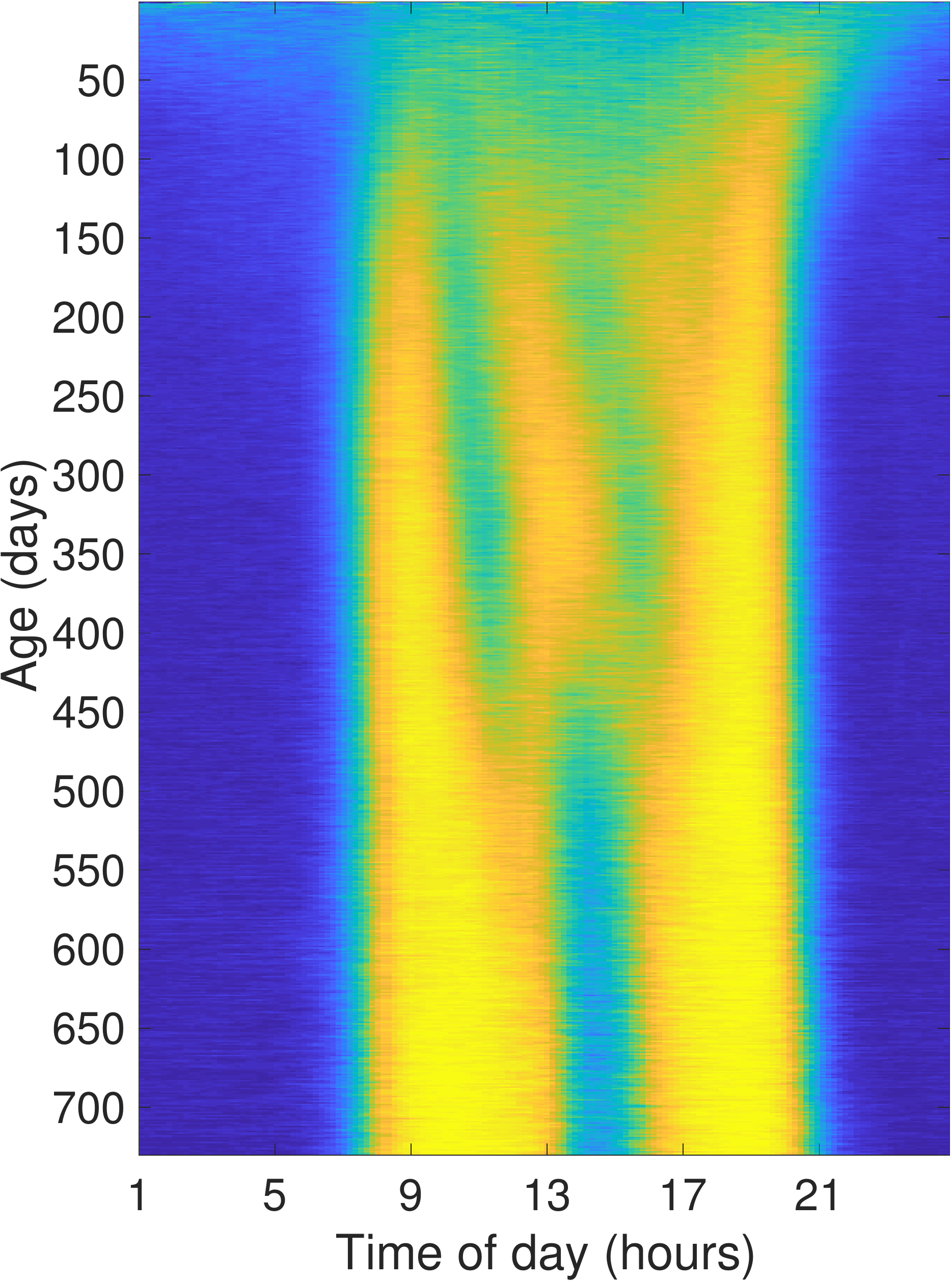} & \includegraphics[width=\linewidth]{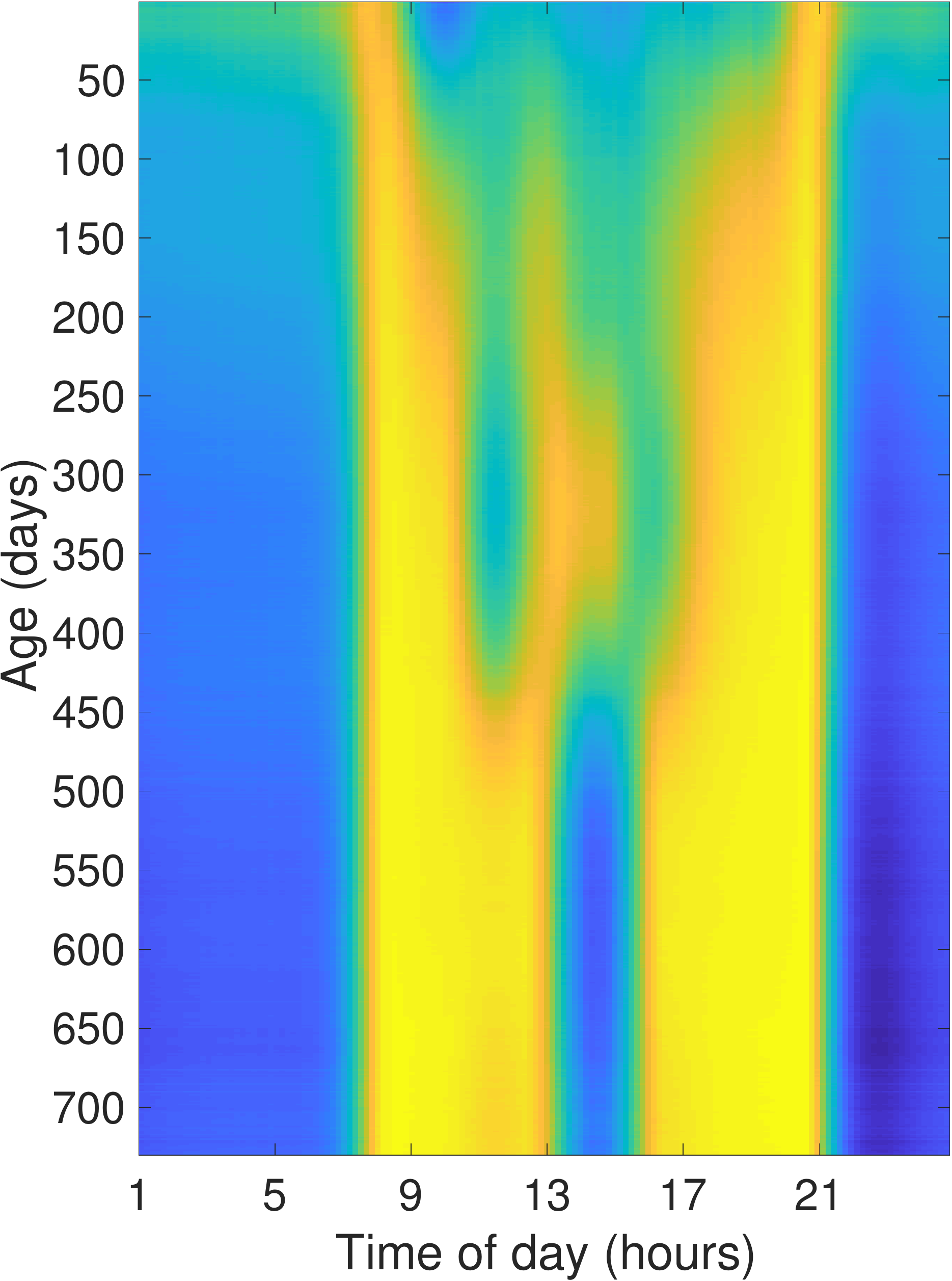}& \includegraphics[width=\linewidth]{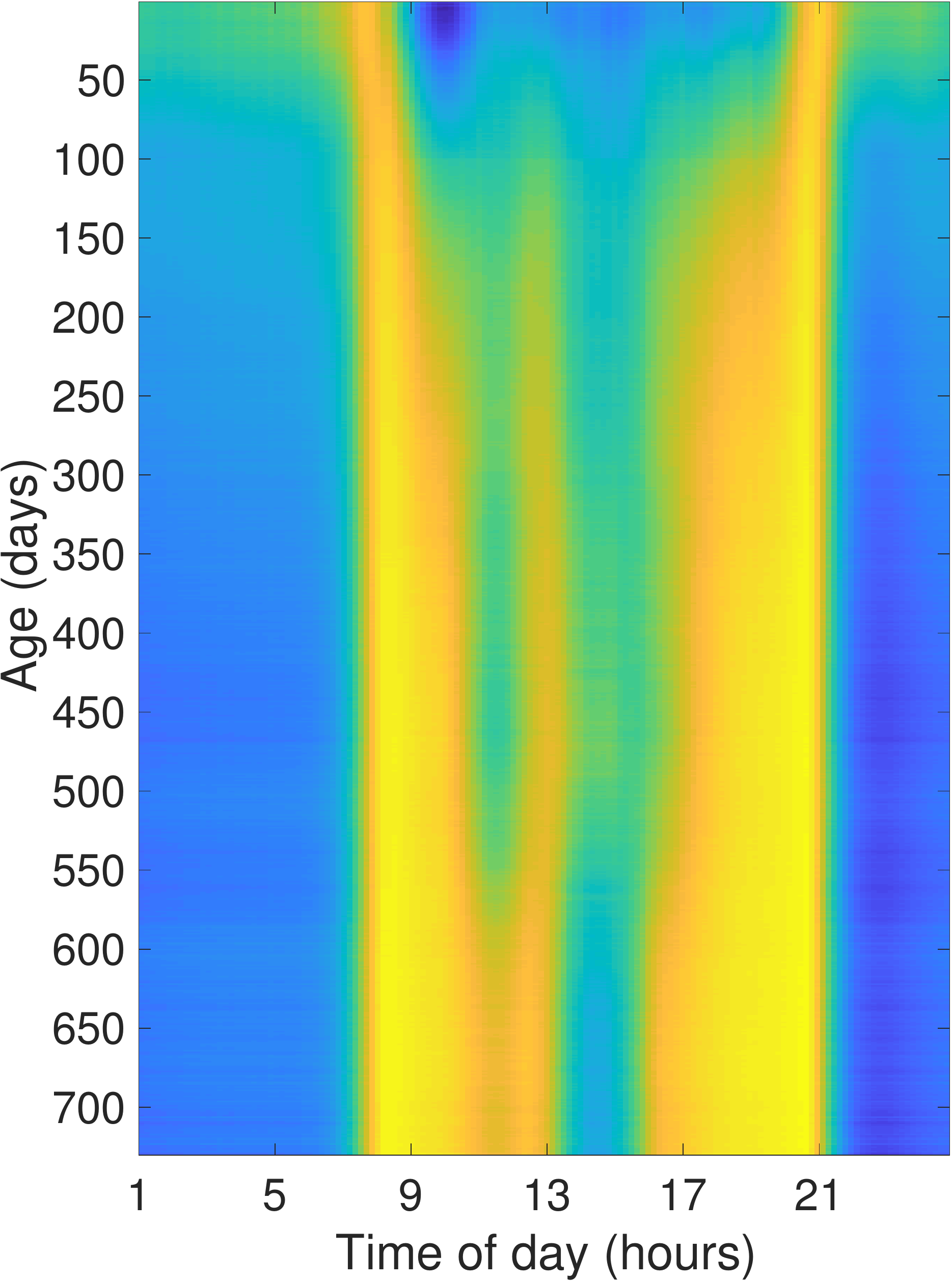} 
 \end{tabular}
 \caption{Cluster centers obtained by applying Lloyd’s algorithm to perform k-means clustering with $k:= 2$ on the raw data (left) and on the first 3 NMF-TS coefficients (right). Clustering the raw data just captures differences in nighttime sleep schedule, whereas clustering the coefficients uncovers differences in nap behavior.}
 \label{fig: cluster_centers}
\end{figure}

In this section we consider the problem of automatically clustering infants according to their sleeping behavior. From a clinical point of view, the goal is to determine whether there are subpopulations of infants with distinct tendencies, as a preliminary step in the study of factors that affect sleep development. Figures~\ref{fig: cluster_centers} and~\ref{fig:rawdata_results} shows the result of separating the data into two clusters using a standard algorithm by~\cite{lloyd1982least} to minimize the k-means cost function \cite{macqueen1967some}. Unfortunately, the clustering procedure just captures differences in nighttime sleep schedules that are not clinically meaningful: cluster 1 contains infants that wake up and go to bed earlier than those in cluster 2. 

\begin{figure}[t]
{\small
\begin{tabular}{ >{\centering\arraybackslash}m{0.08\linewidth} >{\centering\arraybackslash}m{0.27\linewidth} >{\centering\arraybackslash}m{0.27\linewidth} >{\centering\arraybackslash}m{0.27\linewidth}}
Cluster & Daytime sleep & Two naps &  One nap \\
1 & \includegraphics[width=0.9\linewidth]{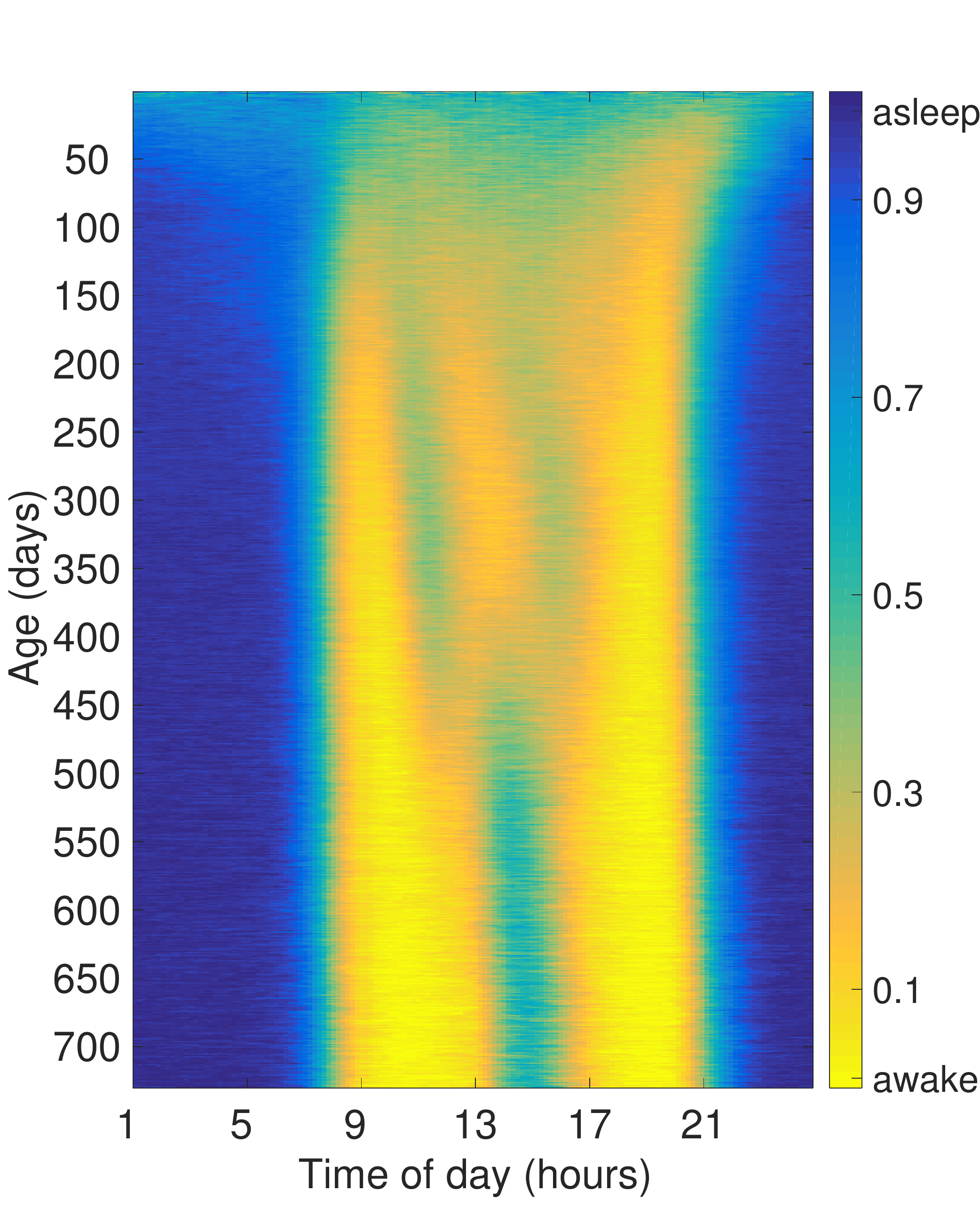} & \includegraphics[width=0.9\linewidth]{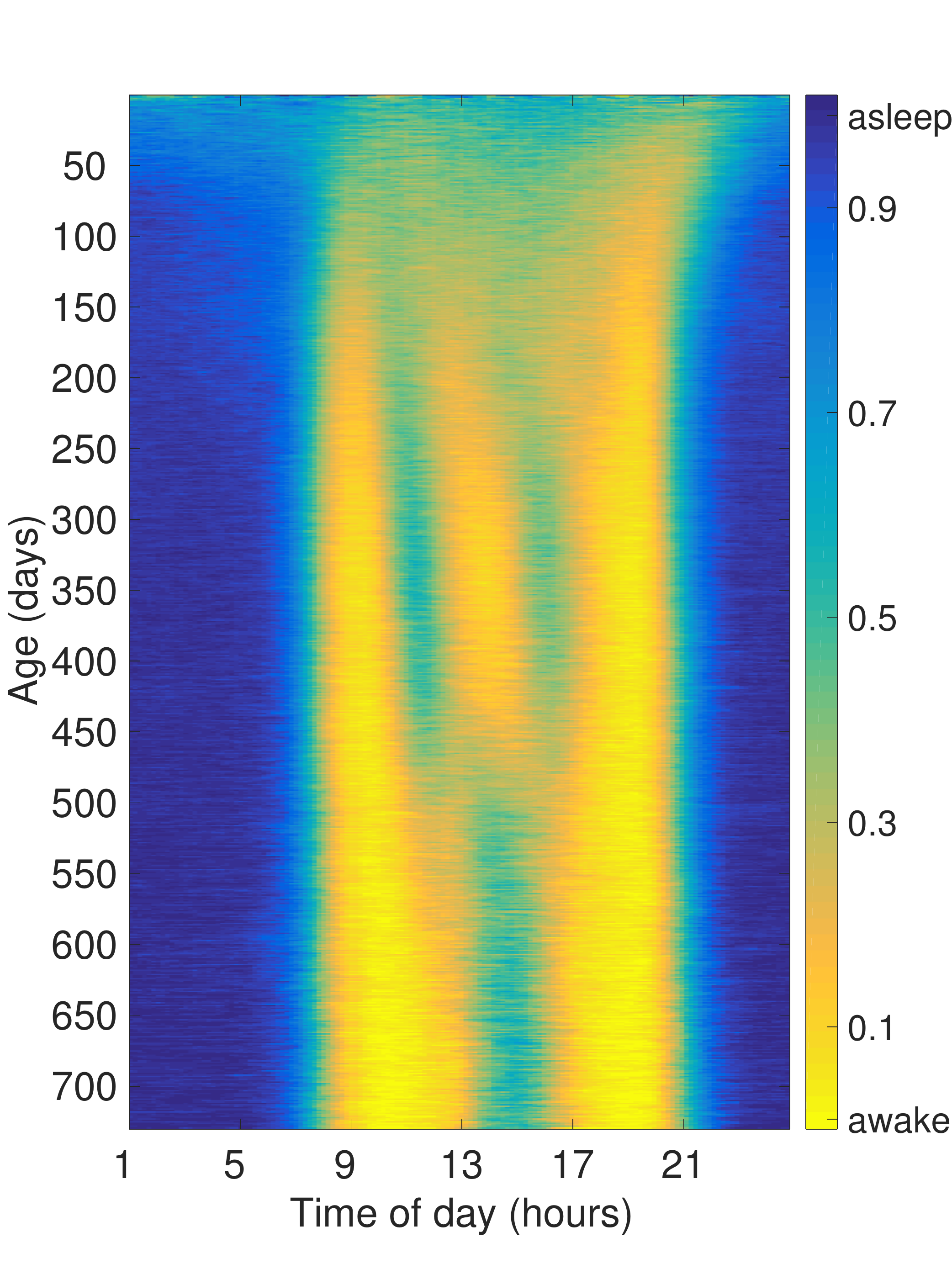} &  \includegraphics[width=0.9\linewidth]{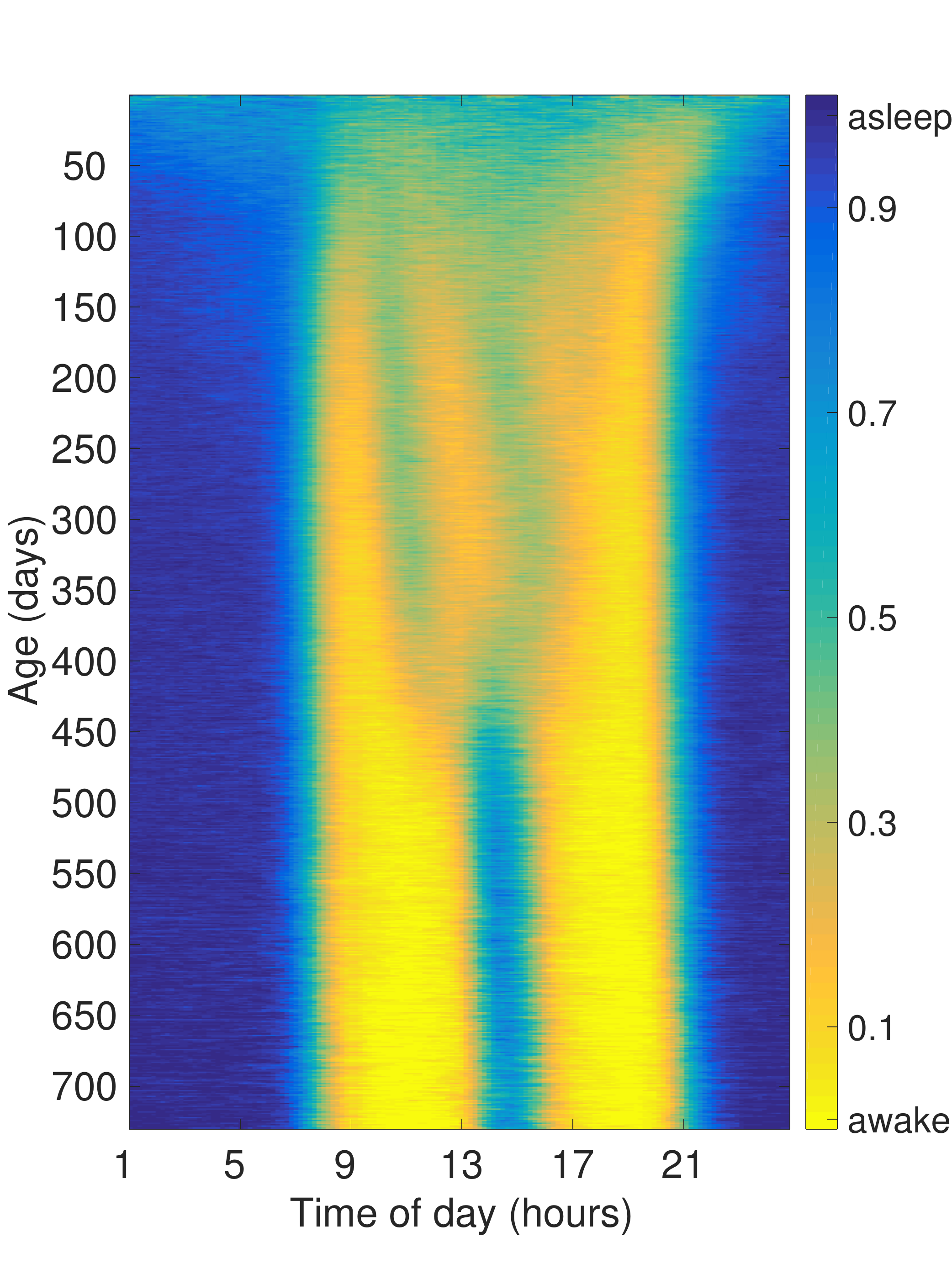}\\
2 & \includegraphics[width=0.9\linewidth]{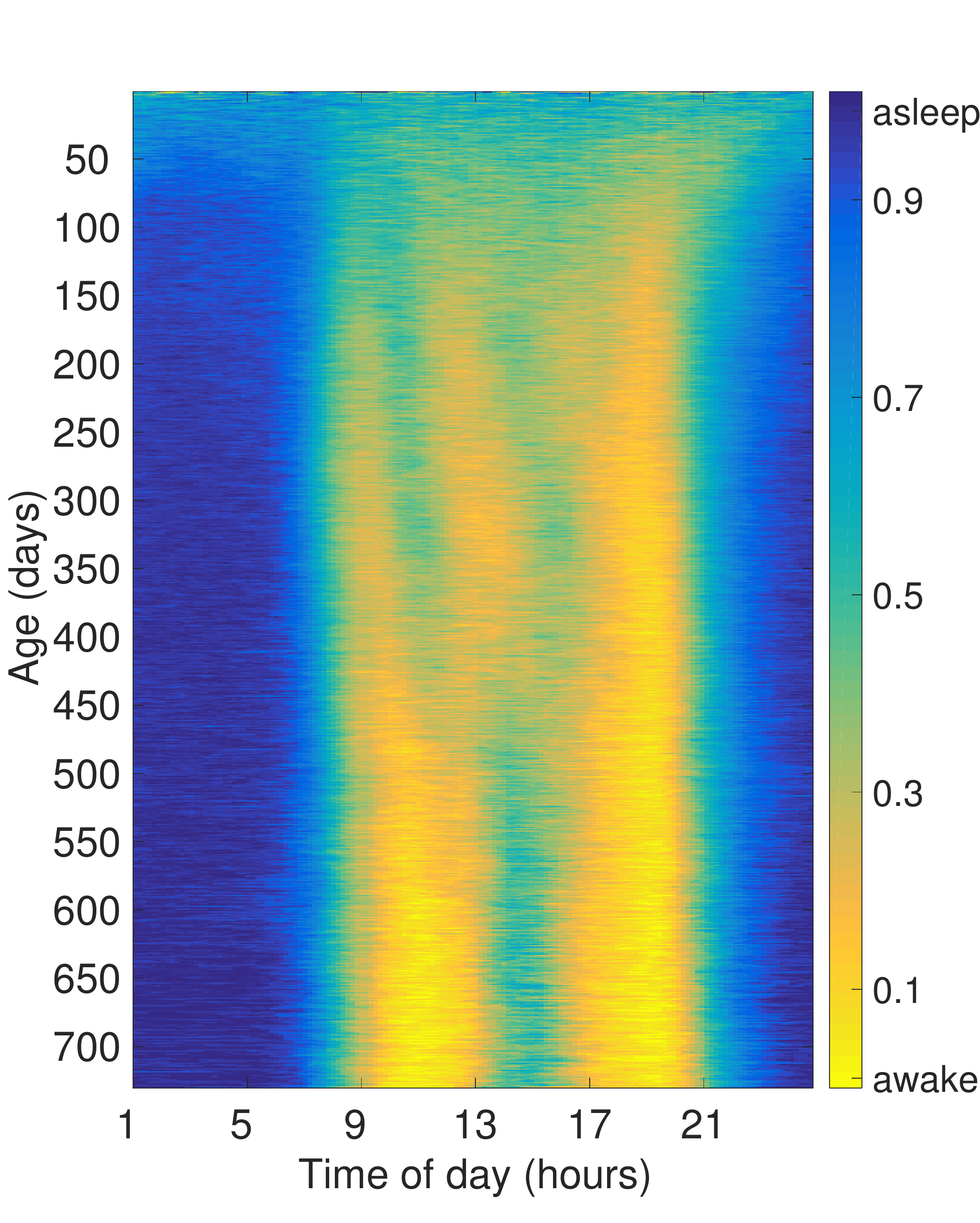} & \includegraphics[width=0.9\linewidth]{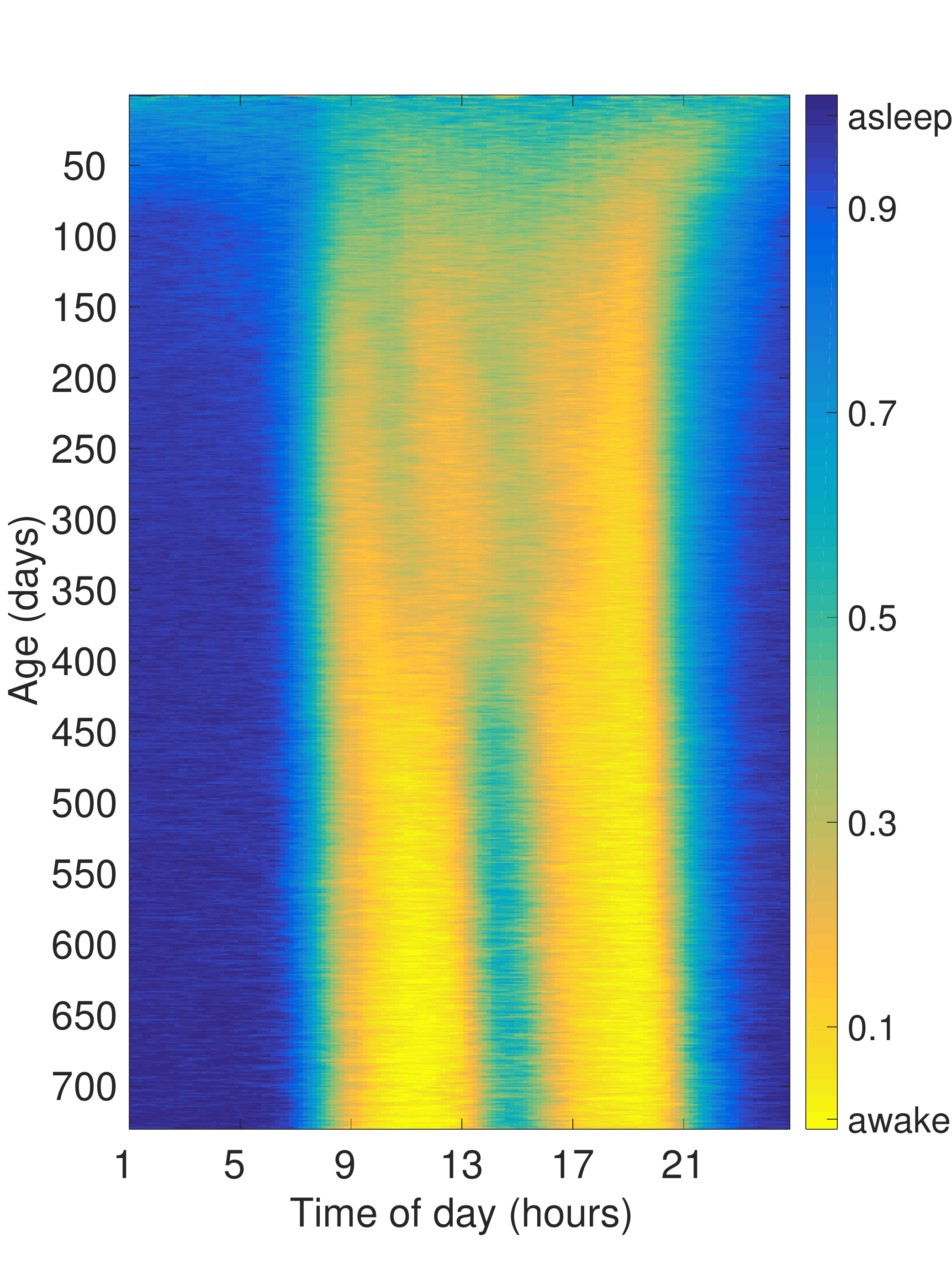} &  \includegraphics[width=0.9\linewidth]{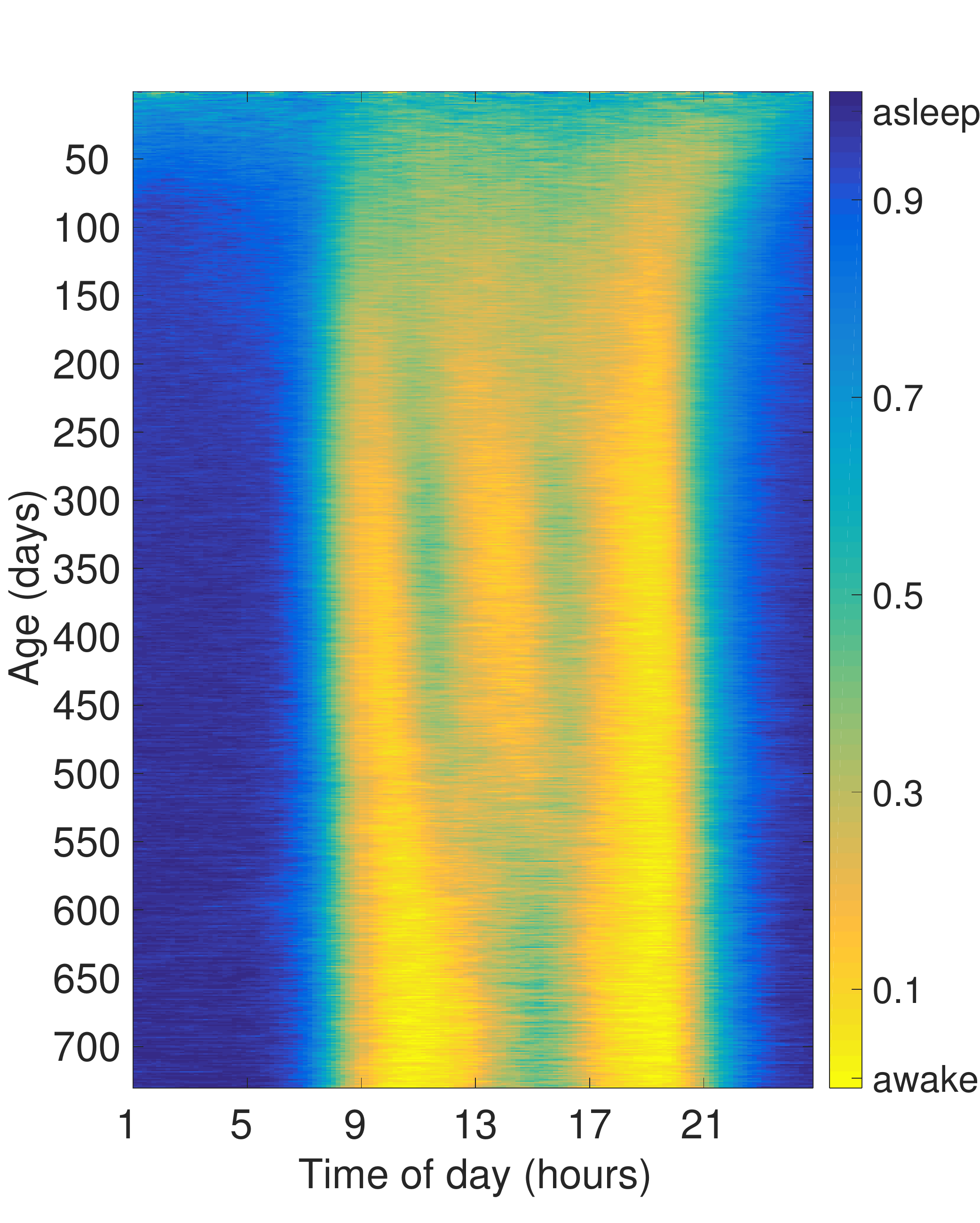}
 \end{tabular}
}
 \caption{Cluster centers obtained by performing k-means clustering with $k:=2$ using a distance associated to each of the first three basis functions in Figure~\ref{fig:pattern_development}. The difference between the centers depends on the corresponding sleeping pattern. In particular, the centers in the top row show a more clearly marked emergence of circadian sleep, two-nap schedule, and one-nap schedule respectively.}
 \label{fig:rnmf_results_factor}
\end{figure}

As described in Section~\ref{sec:results}, three of the daily-sleep patterns extracted by NMF-TS are associated to meaningful sleeping habits, whereas the remaining two just capture shifts in nighttime sleep. We build upon this insight to design a metric that is approximately invariant to shifts. Let $Y^{[n]}$ and $Y^{[n']}$ be the data corresponding to two different infants. The metric is equal to the $\ell_2$ norm of the difference between the coefficients associated to the first three patterns for each of the infants,
\begin{align}
d_{\op{NMF-TS}}\brac{Y^{[n]},Y^{[n']}}:=\sqrt{\frac{ 1}{ \abs{\ml{T}_{n} \cap \ml{T}_{n'}}} \sum_{j =1}^{3} \sum_{t \in \ml{T}_{n} \cap \ml{T}_{n'} } \brac{ C^{[n]}_j(t) - C^{[n']}_j(t) }^2}. \label{eq:metric}
\end{align}
The metric only takes into account entries in $\ml{T}_{n} \cap \ml{T}_{n'}$ (i.e. observations present in the two time series). It is normalized by $\abs{\ml{T}_{n} \cap \ml{T}_{n'}}$ to render it as invariant as possible to the number of missing entries. Figure ~\ref{fig: cluster_centers} and ~\ref{fig:rnmf_results} shows the results of applying k-means clustering based on $d_{\op{NMF-TS}}$. The clusters separate the infants according to sleep behavior: cluster 1 contains infants that have a more regular napping behavior, a more consolidated circadian rhythm, and develop a one-nap schedule earlier compared to the infants in cluster 2. Our methodology also makes it possible to cluster the population according to more specific behaviors. This is achieved by only including the coefficients corresponding to a single basis function in Eq~\eqref{eq:metric}. Figure~\ref{fig:rnmf_results_factor} shows the cluster centers obtained by applying k means with the metric associated to each of the first three basis functions (instead of all three). The clustering clearly captures differences associated to the corresponding sleeping pattern.

\subsection{Forecasting sleeping behavior}
\label{sec:prediction}

In this section we consider the problem of forecasting the sleep tendencies of infants based on their past sleeping behavior using kernel regression, as described in Section~\ref{sec:kernel_regression}. In order to produce forecasts that are robust to noise and spurious short-term fluctuations, we propose to estimate the coefficients in the rank-5 NMF-TS decomposition, instead of the raw data. To evaluate our approach we separate the infant-sleep dataset into a training set with 561 time series, and a test set with 140 examples. The low-rank model is fit to the training data. The test data are only used to compute forecasting error. Table~\ref{table:forecasting} reports the performance of the method over three different 100-day periods using sleeping data from the previous 100 days. For each time period, we only take into account time series that have at least 70\% observed data in that interval. We compare the result of (1) prediction based on the mean of the training data, (2) kernel regression (KR) using the raw data, and (3) kernel regression using the NMF-TS coefficients. We also compare to the ground-truth rank-5 coefficients computed from the true future data (this is the best-case scenario for a rank-5 model). The error is high for all methods, in the range 0.25-0.4 for true values that are between 0 (awake) and 1 (asleep). This is expected: the dimension of the estimated signals is $100 \ell =$ 14,400, an order of magnitude larger than the number of examples! We can only hope to predict general trends in the future sleep behavior, which motivates our approach of predicting smooth coefficients associated to a small number of basis functions. The results show that this provides an improvement over using the raw data for all three time intervals. 

\begin{table}
\begingroup
\renewcommand*{\arraystretch}{1.5}
{\small
\begin{center}
\begin{tabular}{ |c|c|c|c| } 
\hline
 & \multicolumn{3}{c|}{Forecasting period (days of age)}\\
& 100-200 from 1-100 & 300-400 from 200-300 &  600-700 from 500-600 \\
 \hline
Mean & 0.387 ($3.66 \cdot10^{-2}$) & 0.337 ($3.68 \cdot10^{-2}$) &0.292 ($3.78 \cdot10^{-2}$)\\
KR (raw data) & 0.382 ($9.76 \cdot10^{-2}$) & 0.326 ($4.04 \cdot10^{-2}$) & 0.273 ($3.65 \cdot10^{-2}$)\\ 
KR (NMF-TS) & 0.374 ($2.81 \cdot10^{-2}$) & 0.318 ($3.27 \cdot10^{-2}$) & 0.270 ($3.54 \cdot10^{-2}$)\\ 
Rank-5 ground truth & 0.359 ($3.11 \cdot10^{-2}$) & 0.305 ($3.35 \cdot10^{-2}$)& 0.257 ($3.28 \cdot10^{-2}$)\\ 
 \hline
\end{tabular}
\end{center}
}
\endgroup
\caption{Average forecasting error per 10-min interval for the infant-sleep dataset on held-out data. The standard deviation of the error over the test set is given in brackets.}
\label{table:forecasting}
\end{table}
\subsection{Kernel Regression}
\label{sec:kernel_regression}
First, we fit the rank-5 NMF-TS model on the training data to obtain the basis functions $F_1$, \ldots, $F_5$ and the corresponding coefficients. We then extract the coefficients corresponding to the \emph{past} interval and to the \emph{future} interval that we want to predict,   
\begin{align}
\brac{ \keys{C^{[n]}_{1,\op{past}}, \ldots, C^{[n]}_{5,\op{past}}}, \keys{C^{[n]}_{1,\op{future}}, \ldots, C^{[n]}_{5,\op{future}}} }, \quad 1 \leq n \leq N_{\op{train}}.
\end{align} 
Given a test example for which we only have past data, we compute its rank-5 NMF-TS coefficients $\keys{C^{\op{test}}_{1,\op{past}}, \ldots, C^{\op{test}}_{5,\op{past}}}$ using the basis functions obtained from the training set. Then we apply kernel regression, a nonparametric method proposed by \cite{nadaraya1964estimating} and \cite{watson1964smooth}, to estimate its future coefficients: 
\begin{align}
\widehat{C}^{\op{test}}_{i,\op{future}} & := \frac{\sum_{n=1}^{N_{\op{train}}} K (C^{[n]}_{i,\op{past}},C^{\op{test}}_{i,\op{past}}) C^{[n]}_{i,\op{future}}}{\sum_{n=1}^{N_{\op{train}}} K (C^{[n]}_{i,\op{past}},C^{\op{test}}_{i,\op{past}}) }, \quad 1 \leq i \leq r,
\end{align}
The kernel $K$ is Gaussian with standard deviation $\sigma$, which is set via cross-validation on the training set. Intuitively, the estimate is a weighted average of instances that have similar past data to the test example. The estimated coefficients are then combined using the basis functions from the training set, to produce a low-rank estimate of future sleep. 

\section{Additional Figures}
\begin{figure}[tp]
\hspace{-.5cm}
\begin{tabular}{ >{\centering\arraybackslash}m{0.20\linewidth} >{\centering\arraybackslash}m{0.20\linewidth} >{\centering\arraybackslash}m{0.20\linewidth} >{\centering\arraybackslash}m{0.23\linewidth}}
Raw data & PCA & NMF & NMF-TS \\ 
  \includegraphics[scale=0.21]{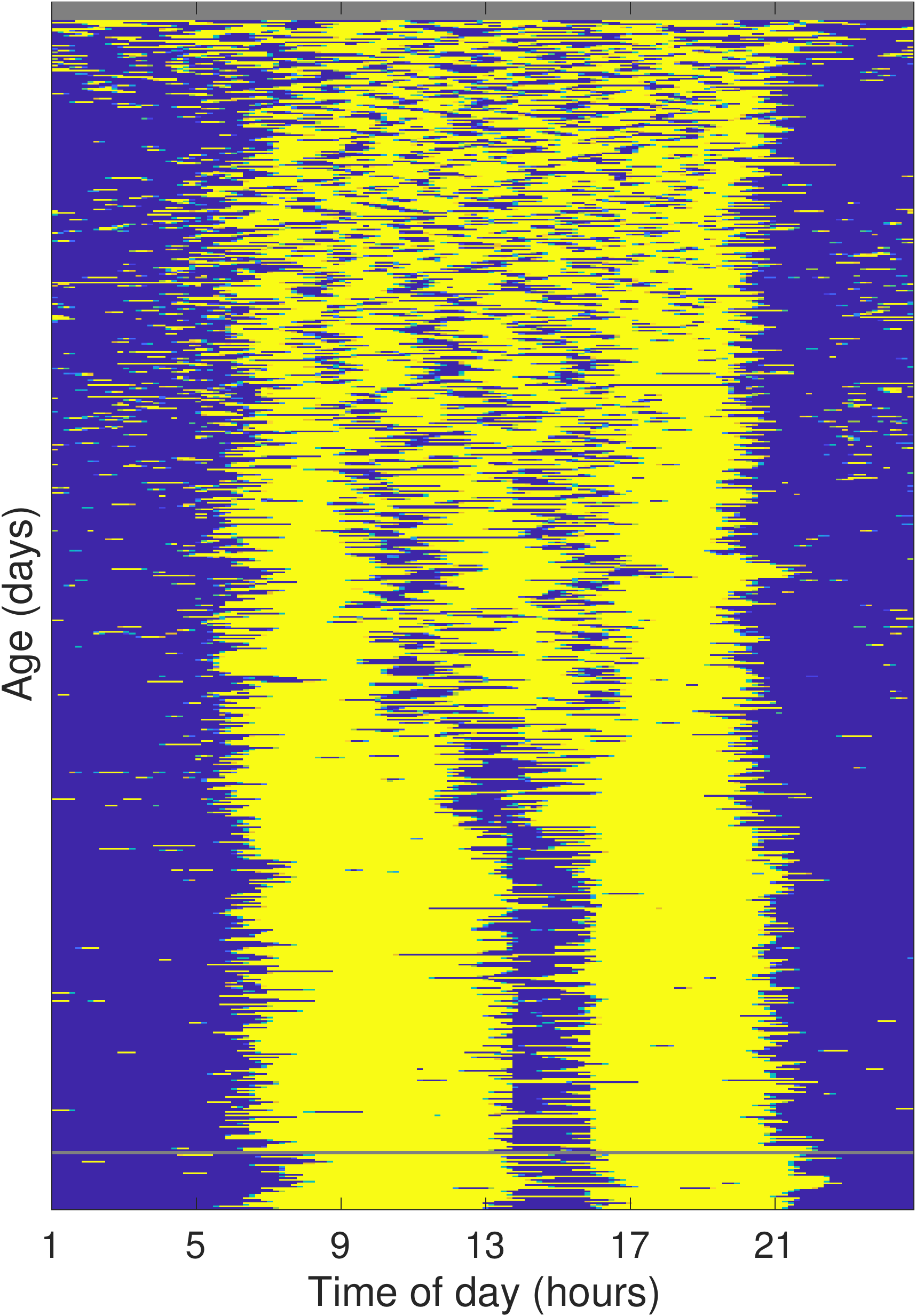} &  
 \includegraphics[scale=0.21]{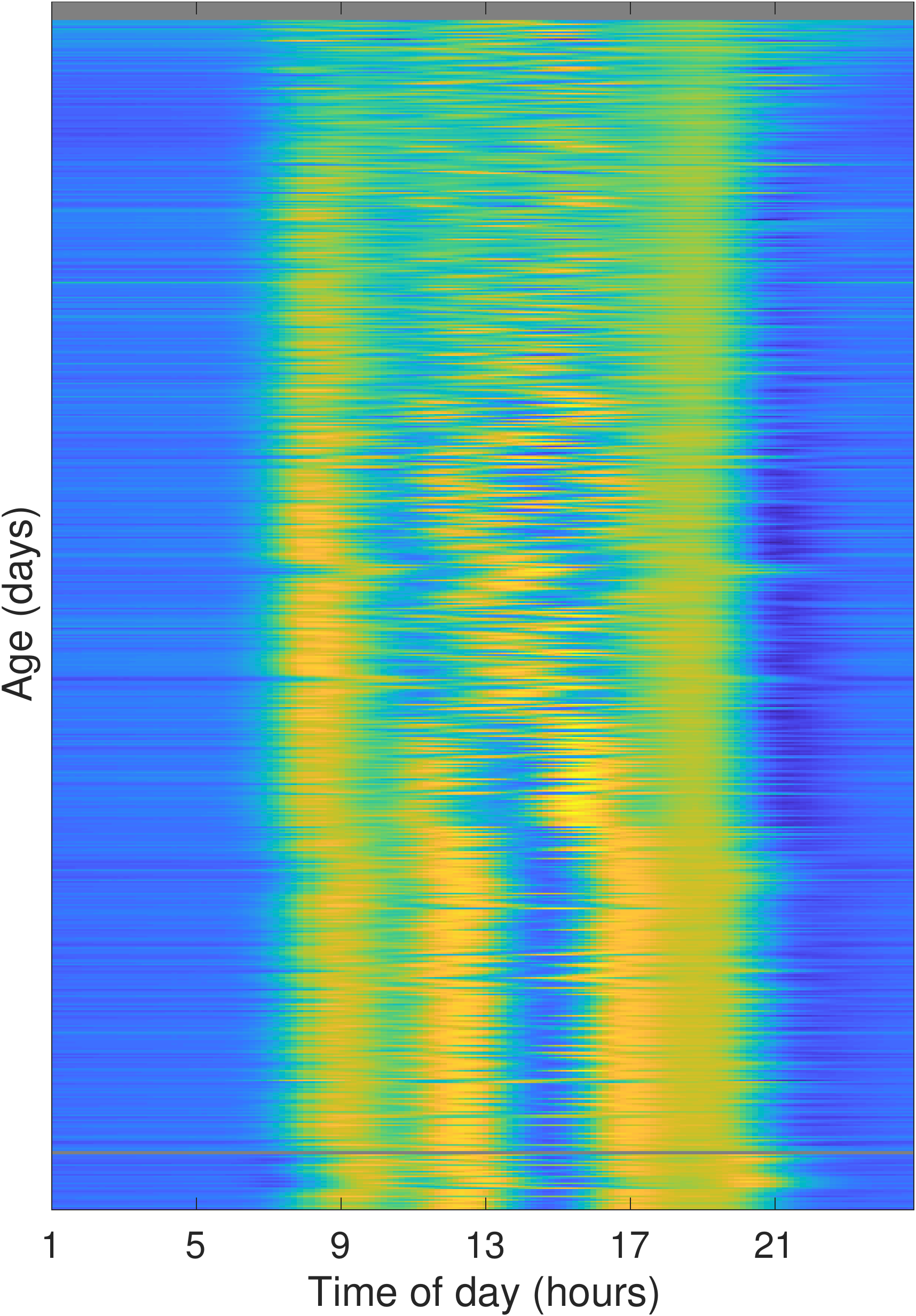} 
&  \includegraphics[scale=0.21]{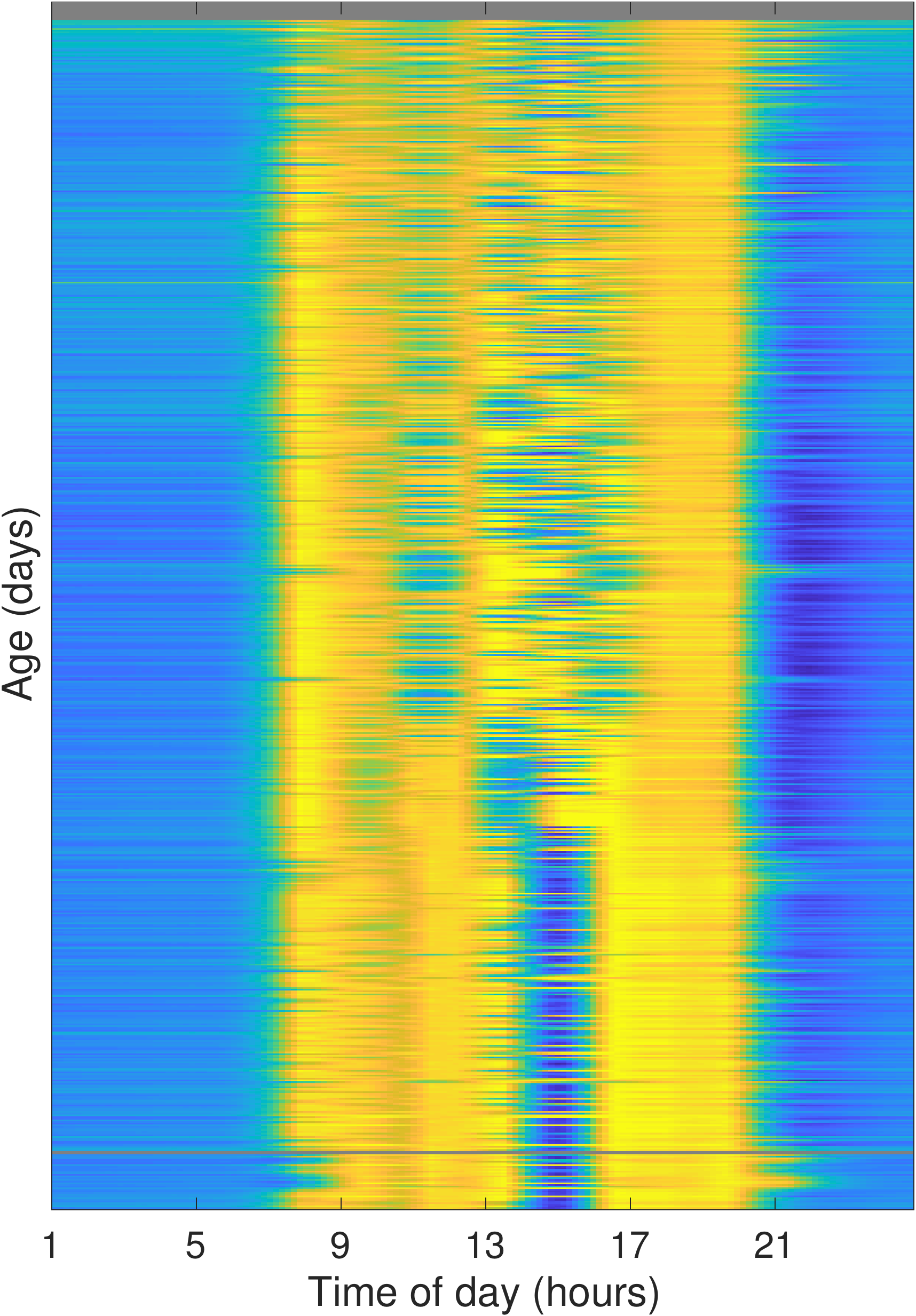}
 &  \includegraphics[scale=0.225]{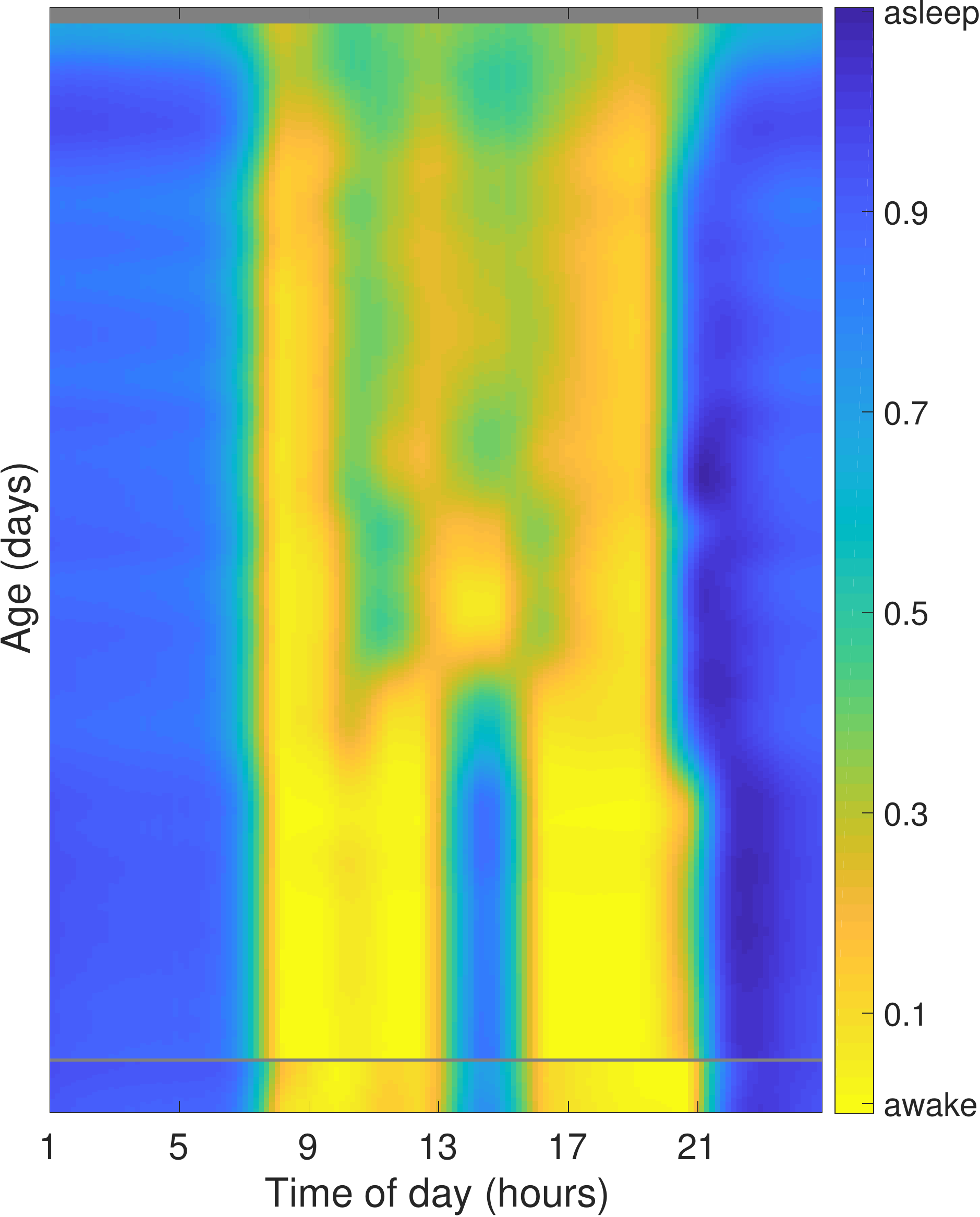}  
 \end{tabular}
 \caption{Low-rank approximation (given by equation~\eqref{eq:lowrank_model}) of the data corresponding to a specific infant obtained combining the daily-sleep patterns and the age development patterns extracted by PCA (2nd column), NMF (3rd column) and the proposed method (4th column). The raw data is shown in the first column. Yellow indicates that the infant is awake, whereas dark blue indicates that they are asleep. The average fitting error per entry for this infant is equal to 0.091 for PCA, 0.104 for NMF and 0.116 for NMF-TS (as expected, since additional constraints reduce overfitting). Gray values indicate missing data.}
 \label{fig:low_rank_approximation}
\end{figure}

\begin{figure}[pt]
\hspace{-0.4cm} 
\begin{tabular}{ >{\centering\arraybackslash}m{0.02\linewidth} >{\centering\arraybackslash}m{0.3\linewidth} >{\centering\arraybackslash}m{0.3\linewidth} >{\centering\arraybackslash}m{0.3\linewidth}   }
$j$ & PCA \vspace{0.25cm} & NMF \vspace{0.25cm}  & NMF-TS \vspace{0.25cm}  \\
1 & \includegraphics[scale=0.23]{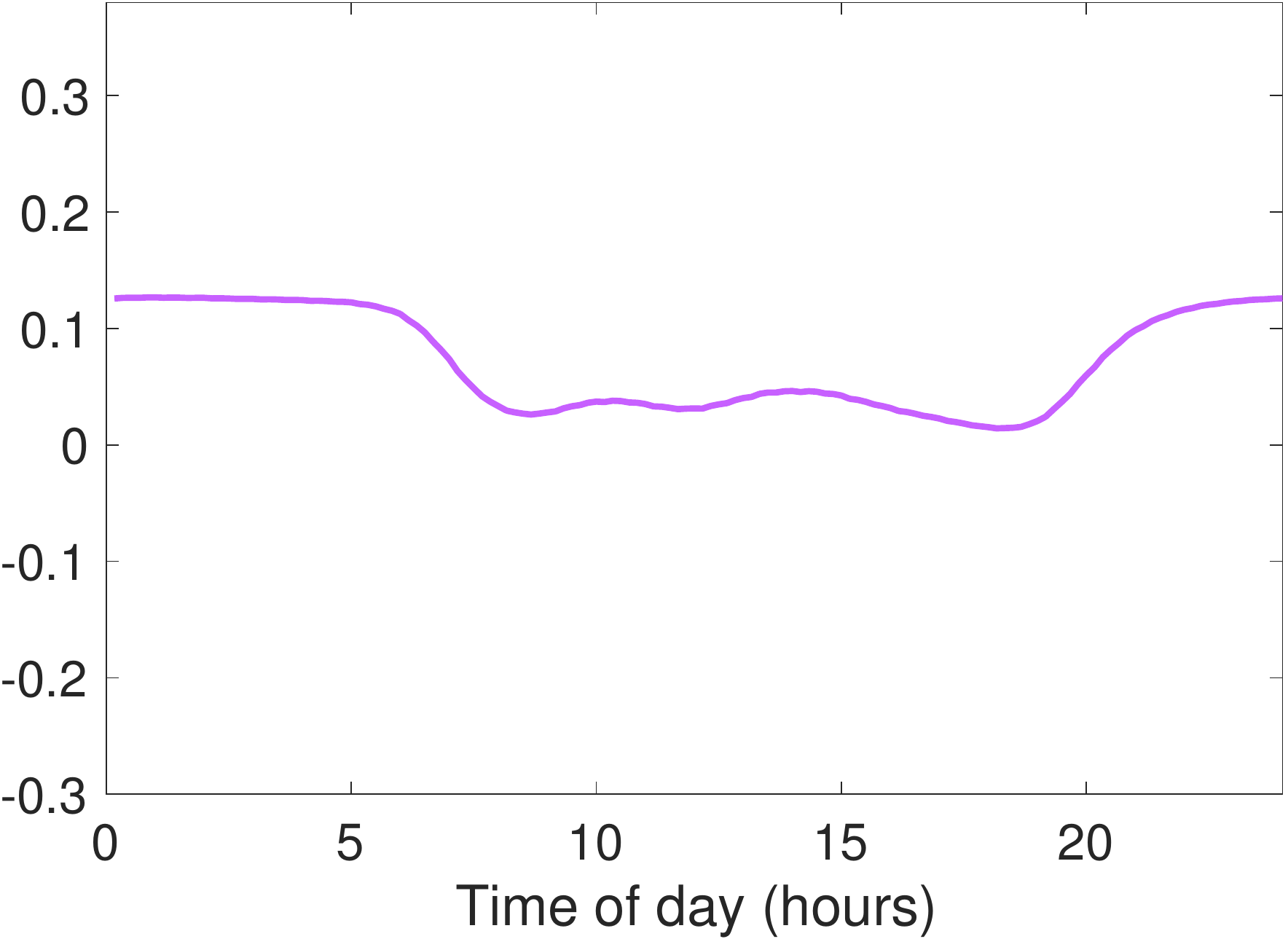} &  \includegraphics[scale=0.25]{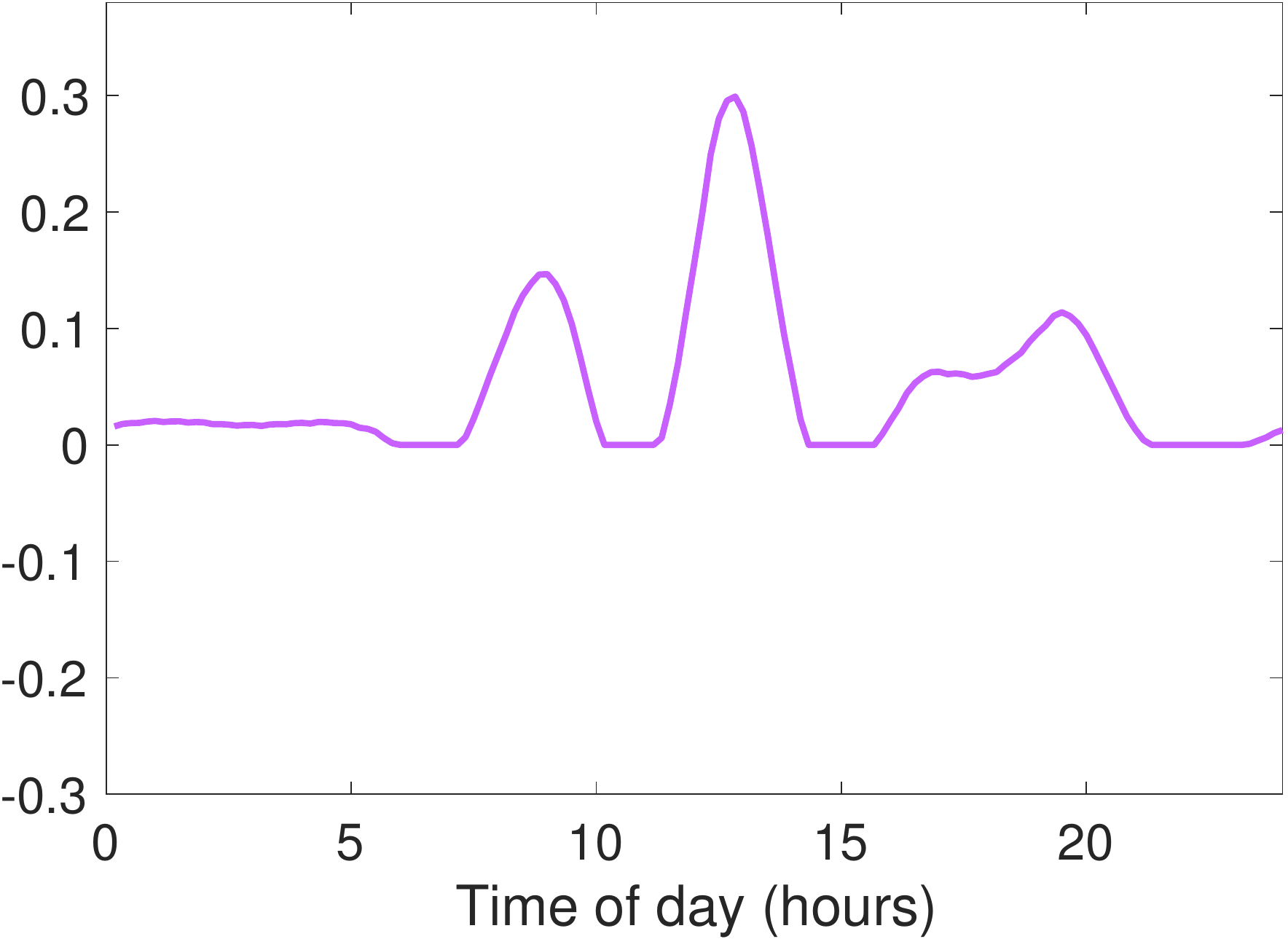} & \includegraphics[scale=0.23]{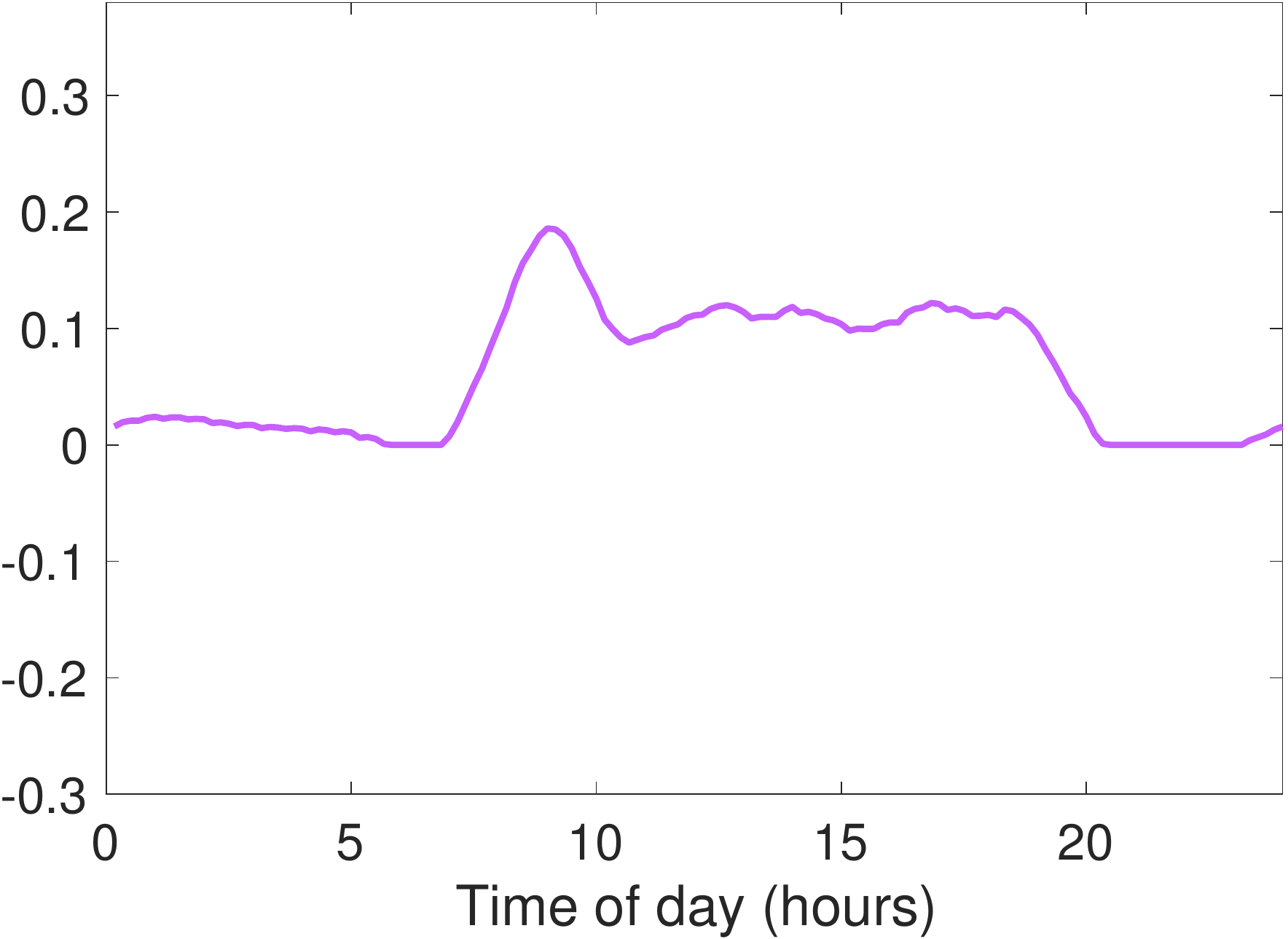} \\
2 & \includegraphics[scale=0.23]{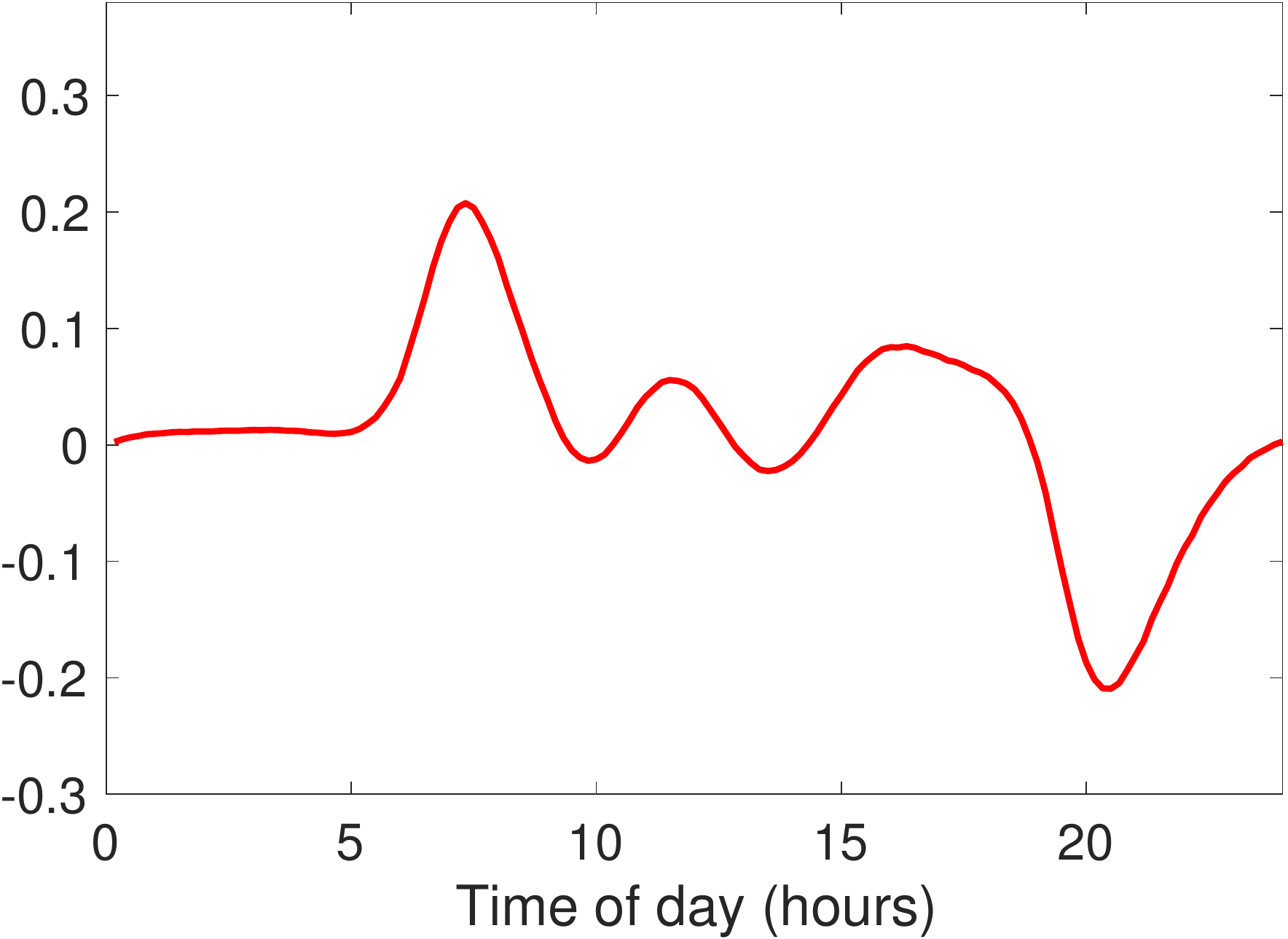} &  \includegraphics[scale=0.23]{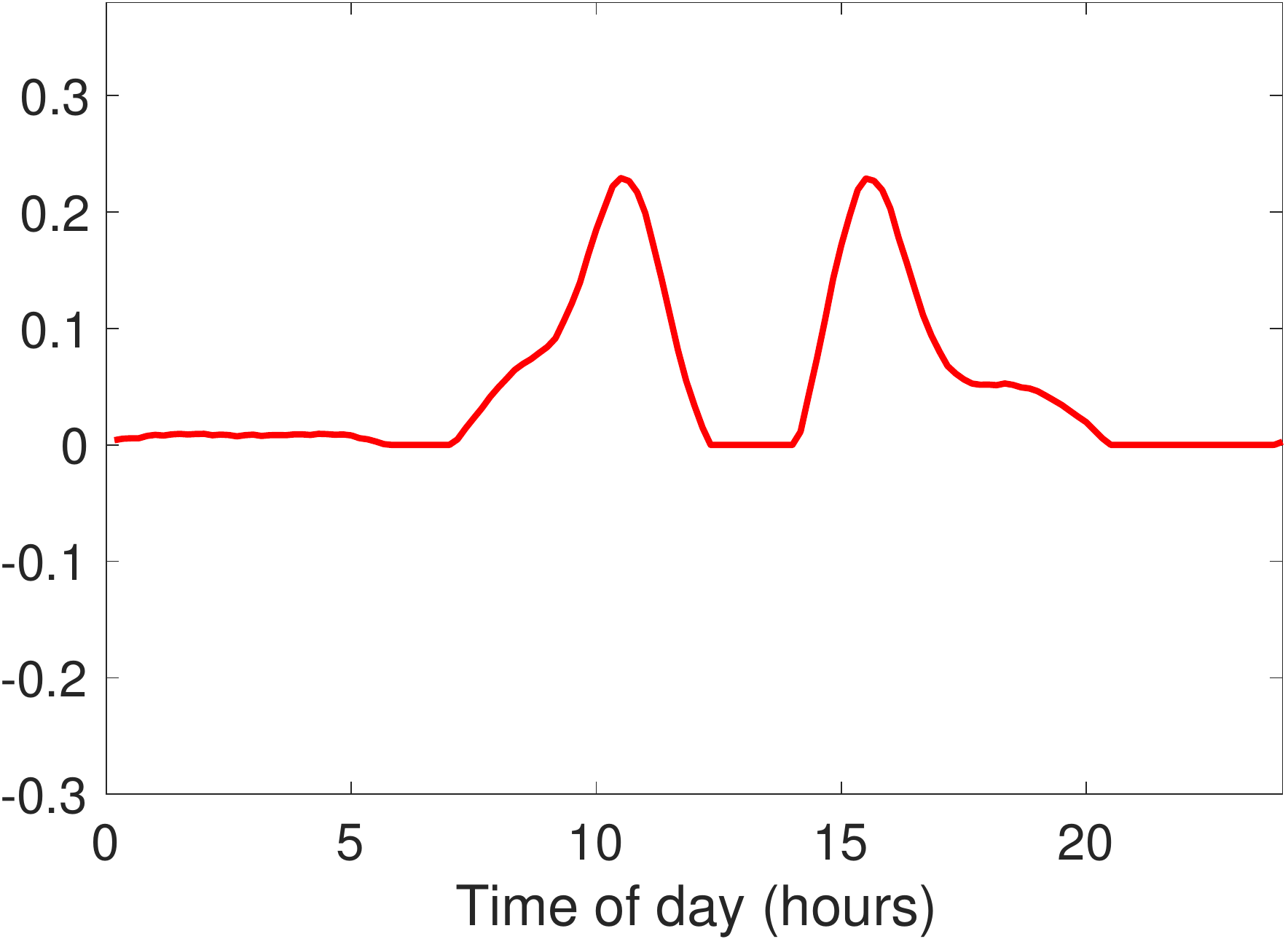} & \includegraphics[scale=0.23]{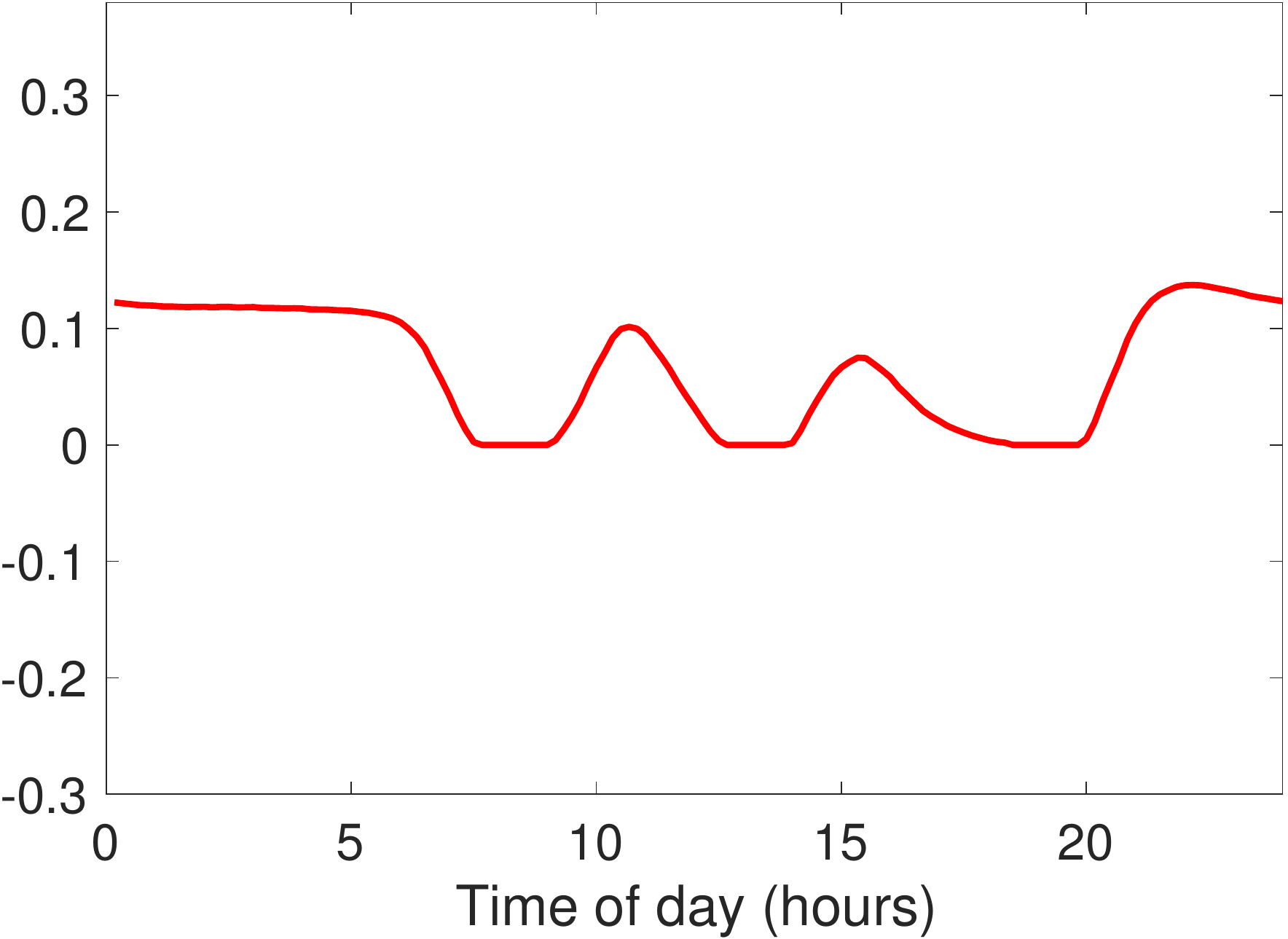} \\
3 & \includegraphics[scale=0.23]{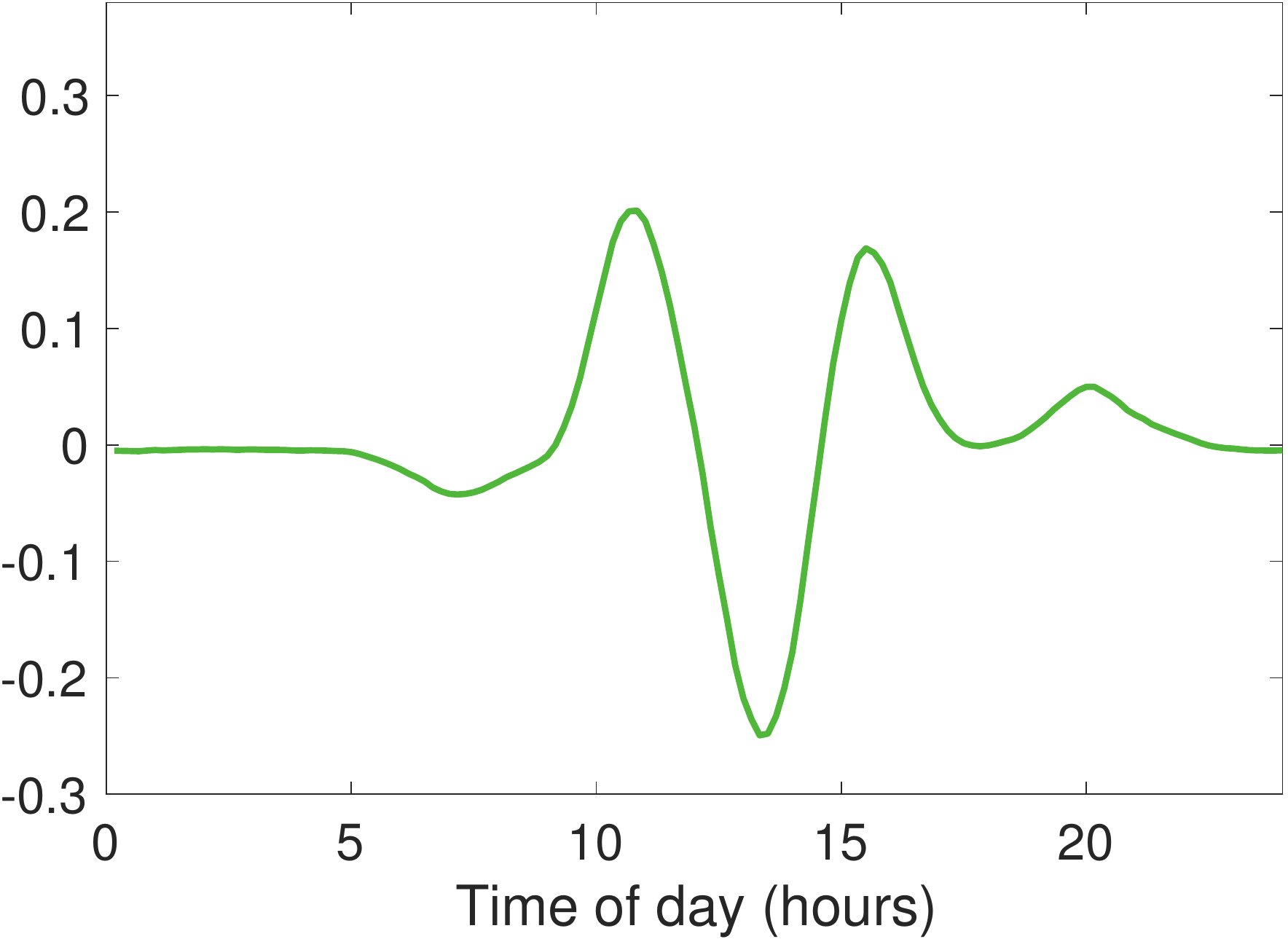} &  \includegraphics[scale=0.23]{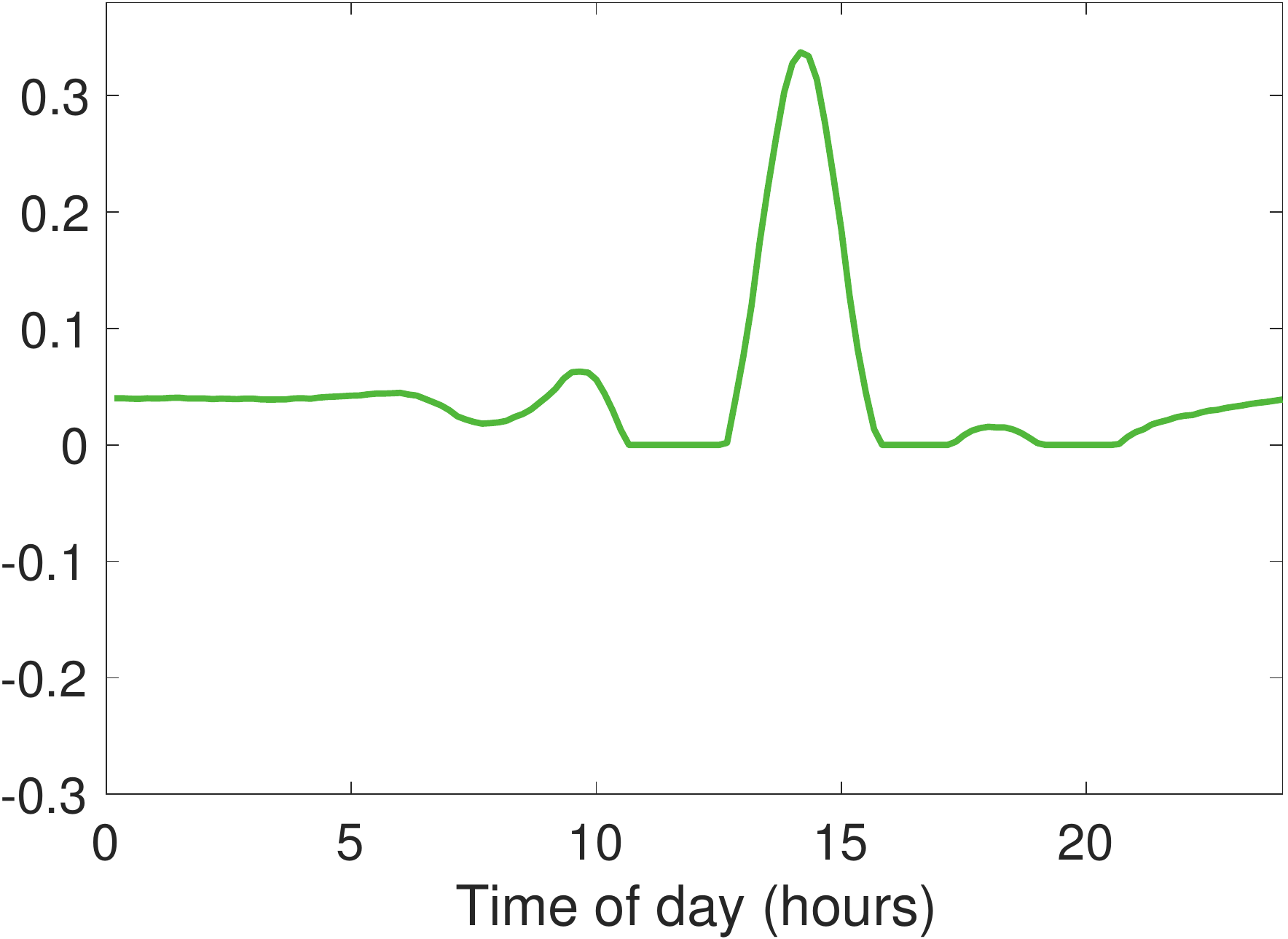} & \includegraphics[scale=0.23]{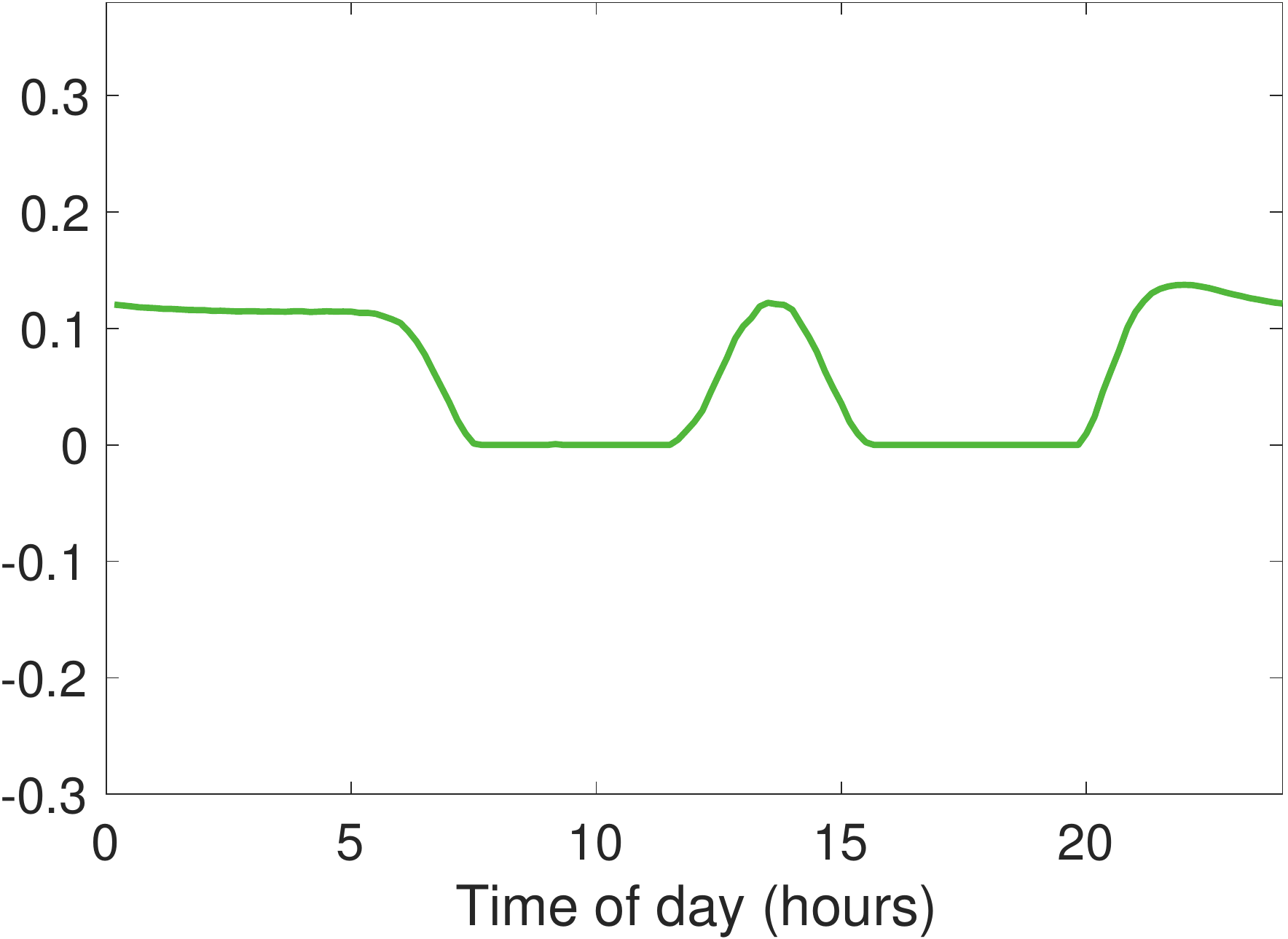} \\
4 & \includegraphics[scale=0.23]{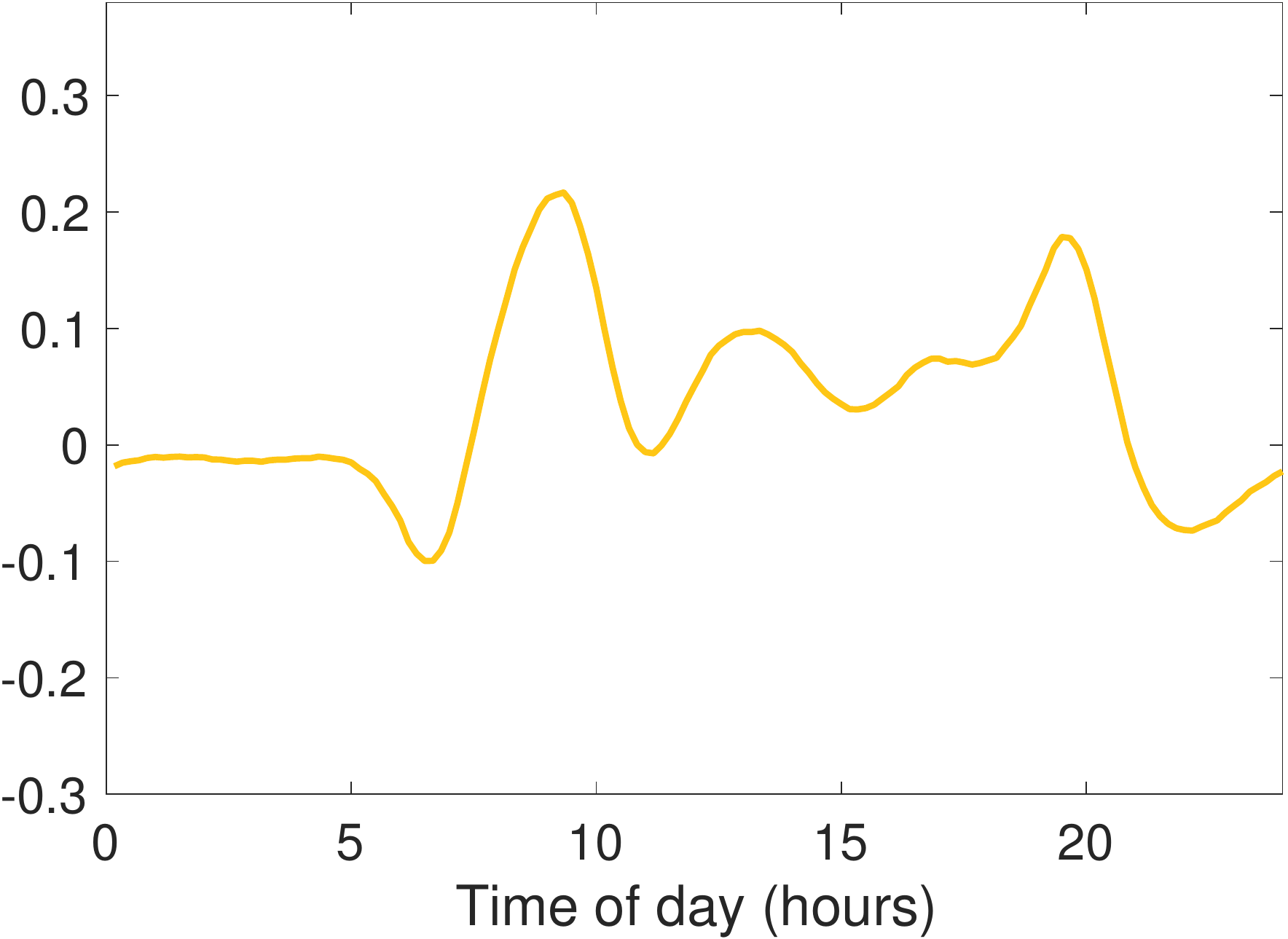} &  \includegraphics[scale=0.23]{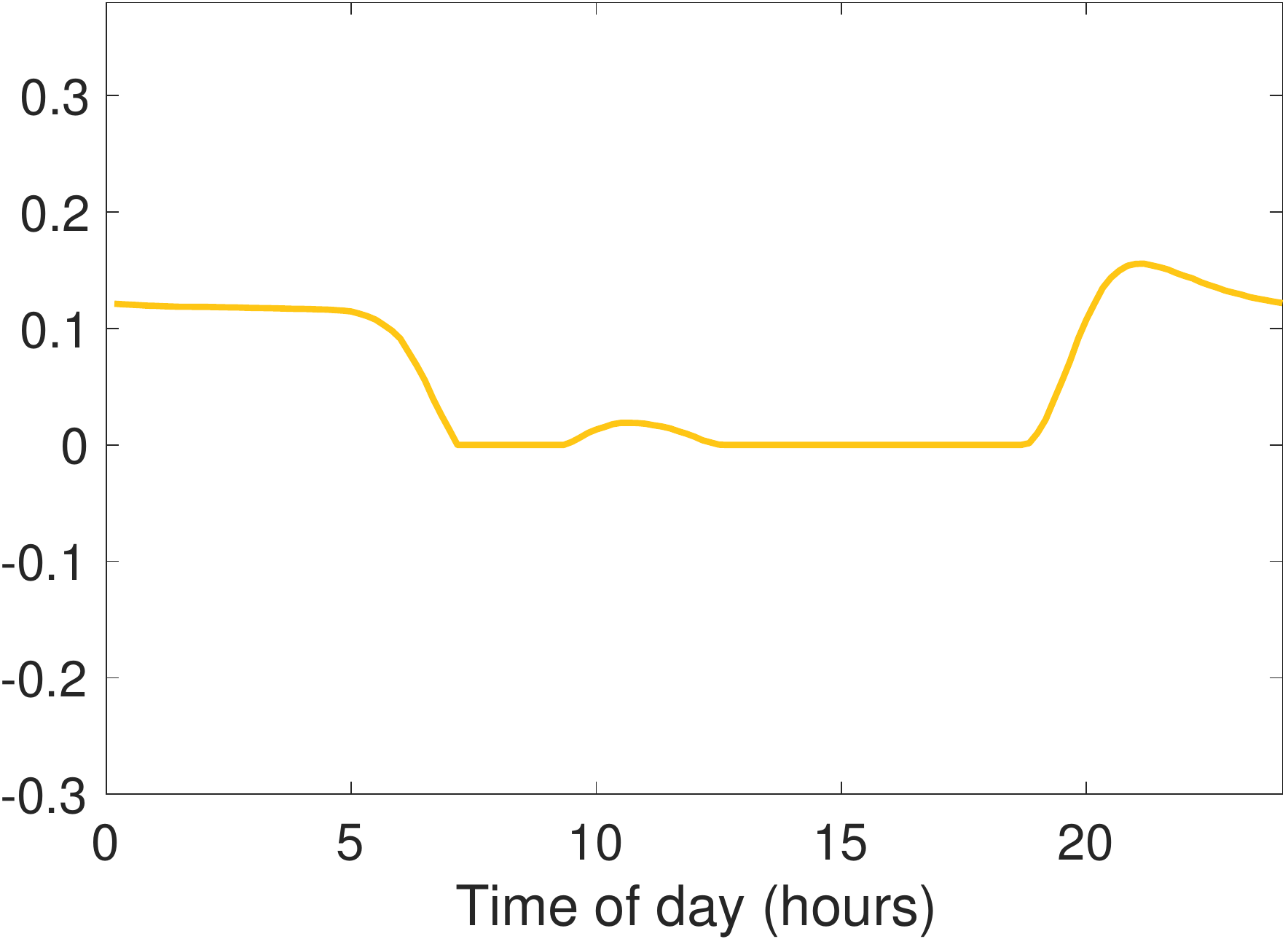} & \includegraphics[scale=0.23]{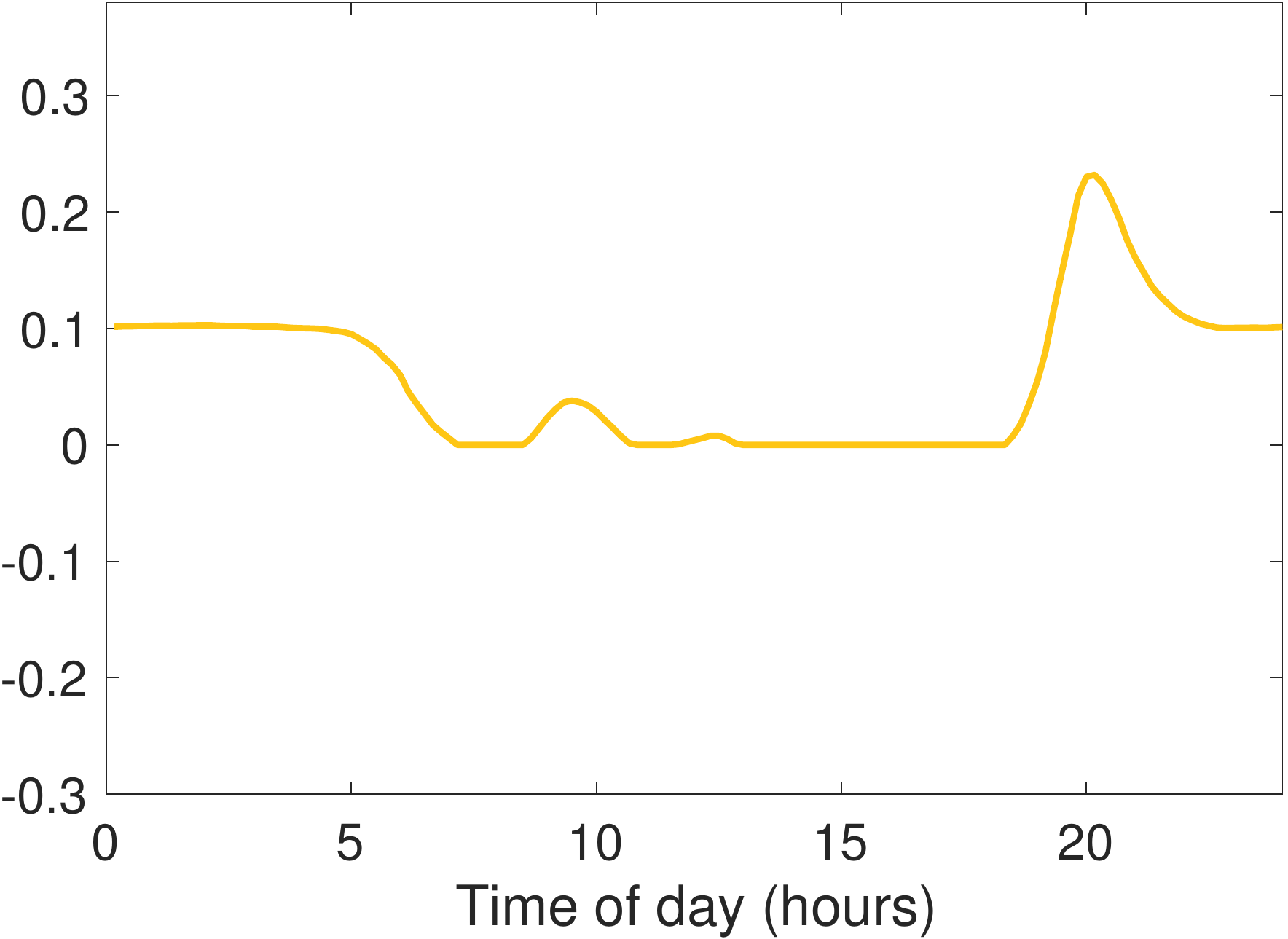} \\
5 & \includegraphics[scale=0.23]{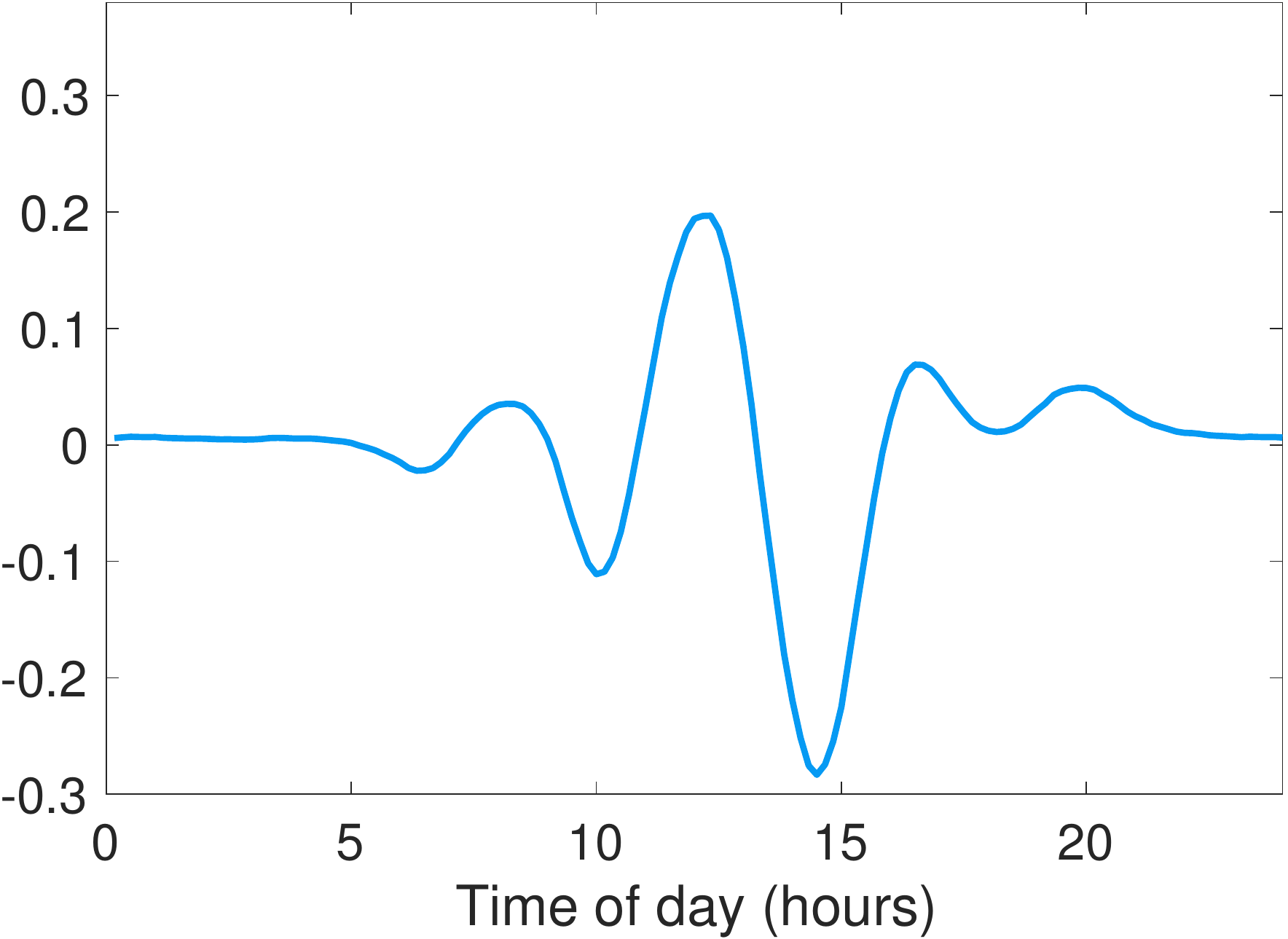} &  \includegraphics[scale=0.23]{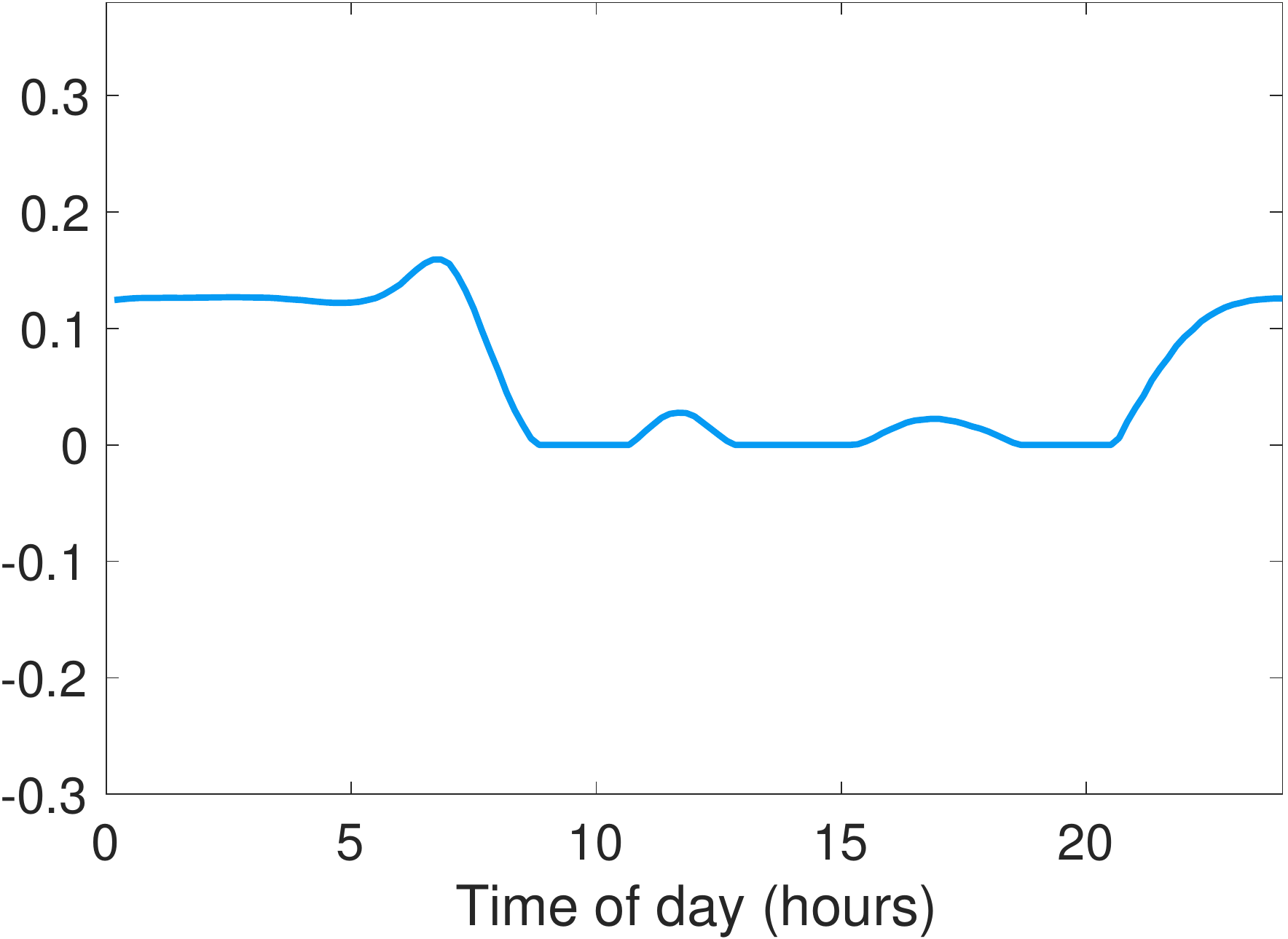} & \includegraphics[scale=0.23]{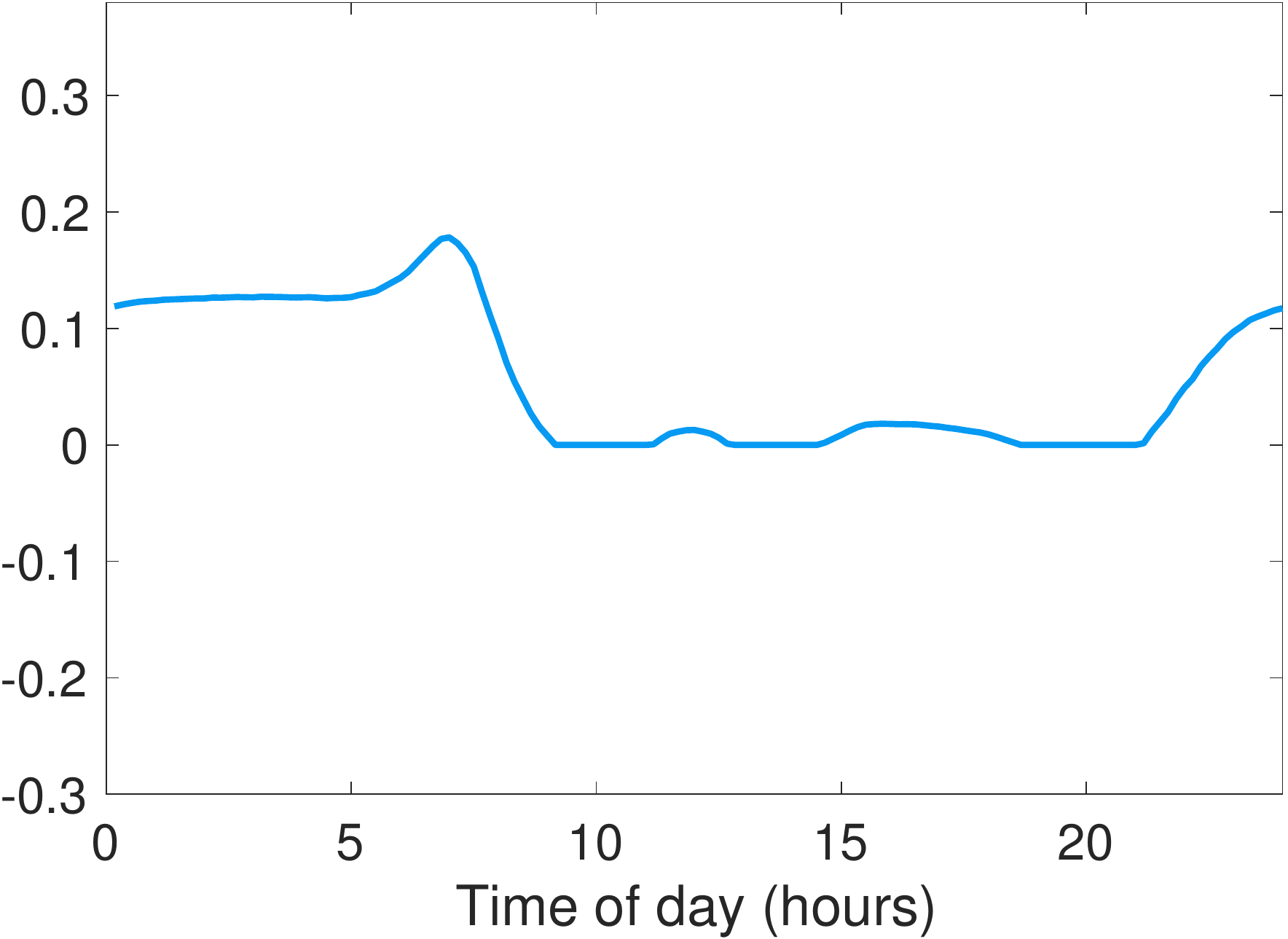} \\
 \end{tabular}
 \caption{Basis functions $F_1$, \ldots, $F_r$ extracted from the infant-sleep dataset using PCA (left), NMF (center) and NMF-TS (right). The rank of the representation is set to $r:=5$. Both NMF and NMF-TS recover patterns that are nonnegative and therefore more interpretable in terms of sleeping behavior. Large values are associated to sleep, whereas values close to zero indicate wakefulness.}
 \label{fig:daily_sleep_patterns}
\end{figure}

\begin{figure}[pt]
\hspace{-0.9cm} 
\begin{tabular}{ >{\centering\arraybackslash}m{0.02\linewidth} >{\centering\arraybackslash}m{0.31\linewidth} >{\centering\arraybackslash}m{0.31\linewidth} >{\centering\arraybackslash}m{0.3\linewidth}   }
$j$ & PCA \vspace{0.25cm} & NMF \vspace{0.25cm}  & NMF-TS \vspace{0.25cm}  \\
1 & \includegraphics[scale=0.18]{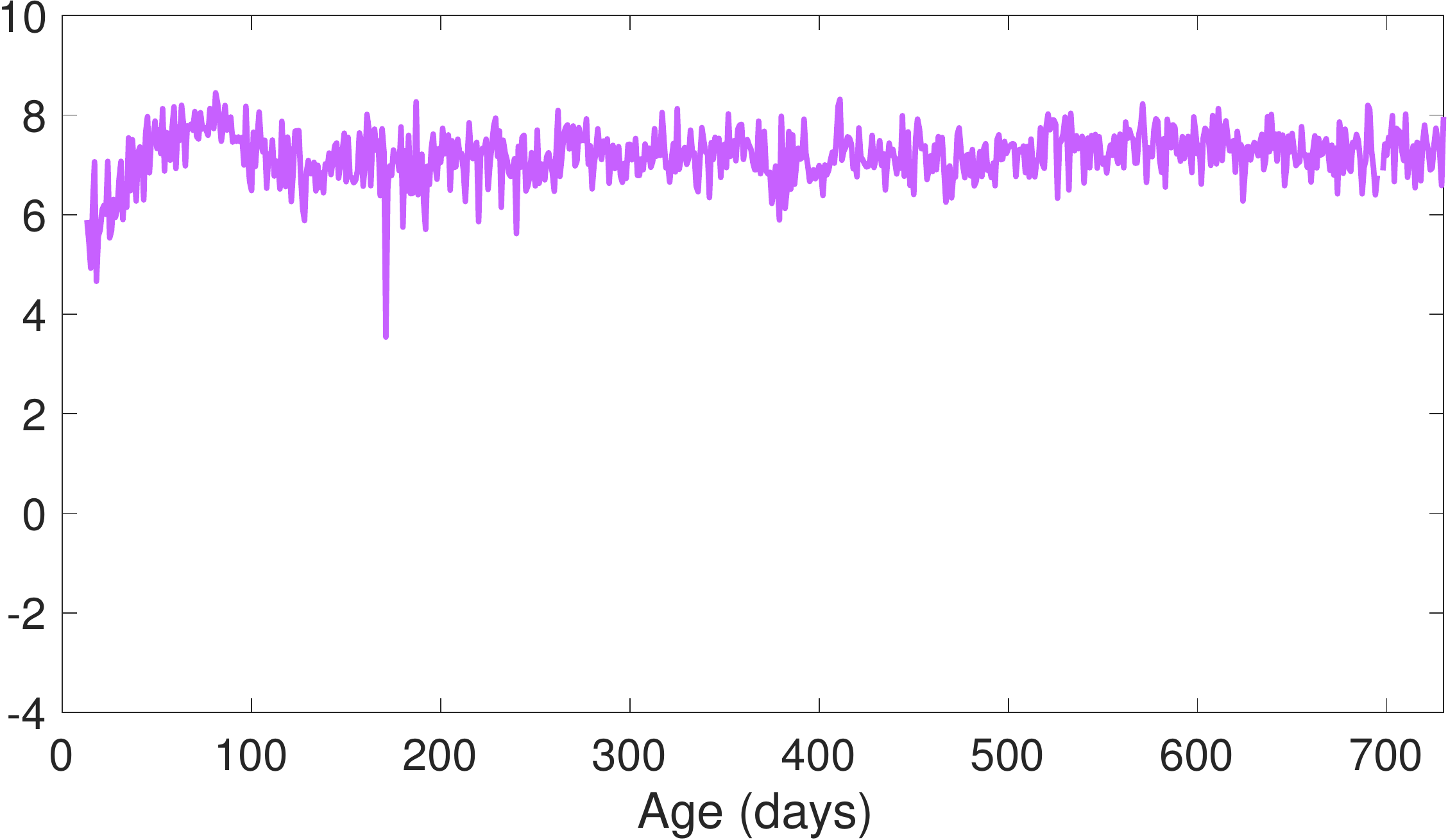} &  \includegraphics[scale=0.18]{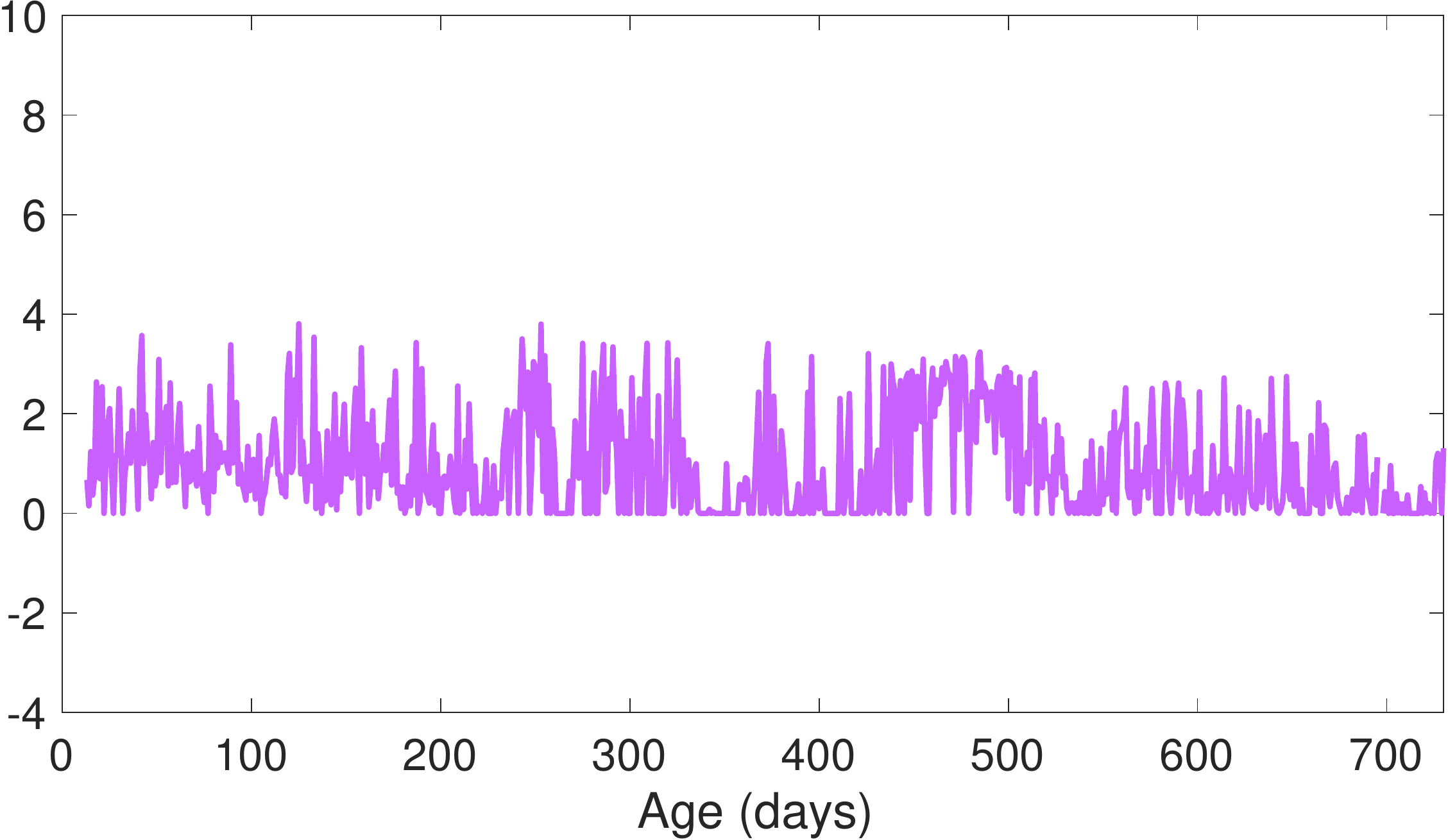} & \includegraphics[scale=0.18]{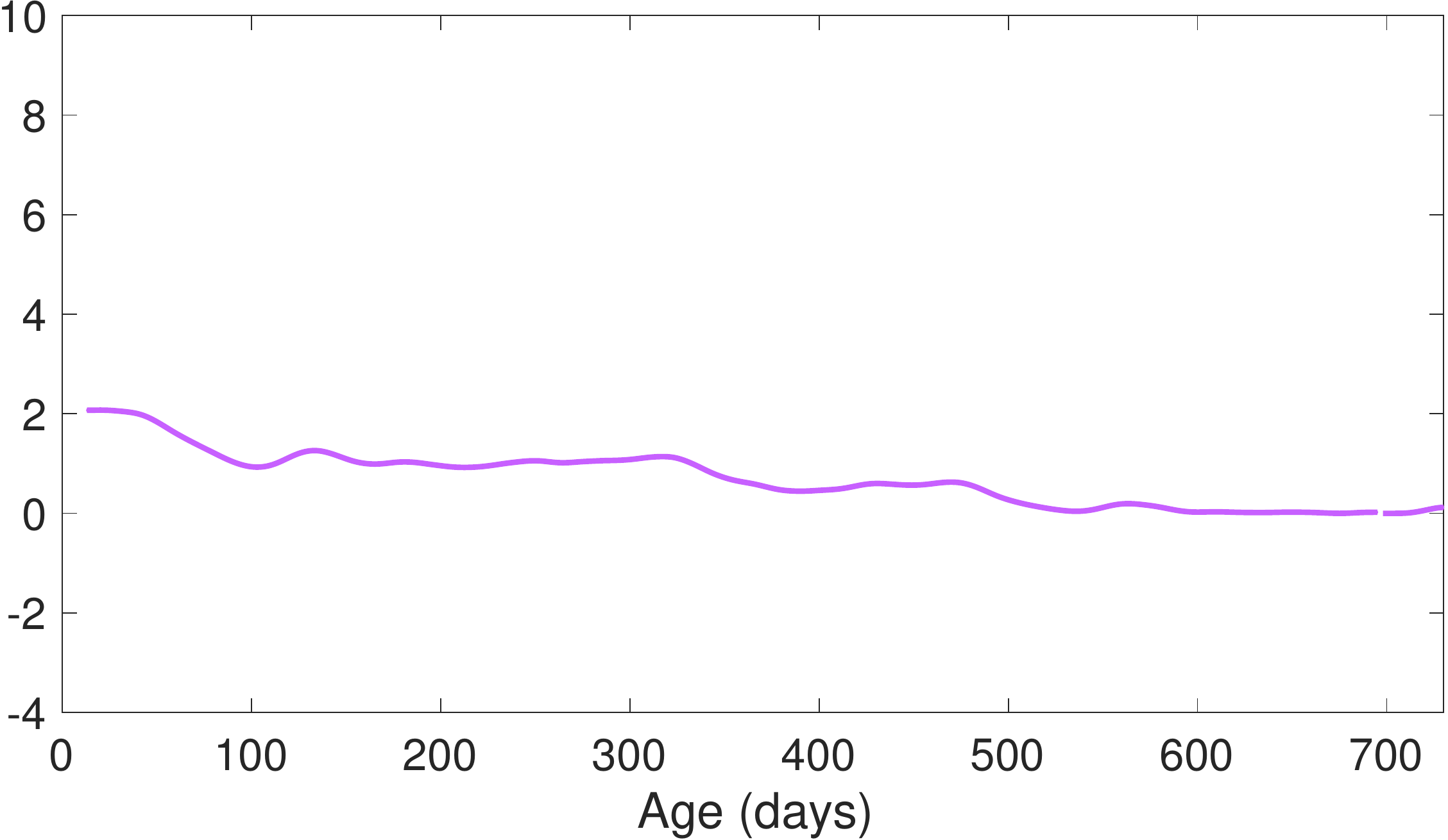} \\
2 & \includegraphics[scale=0.18]{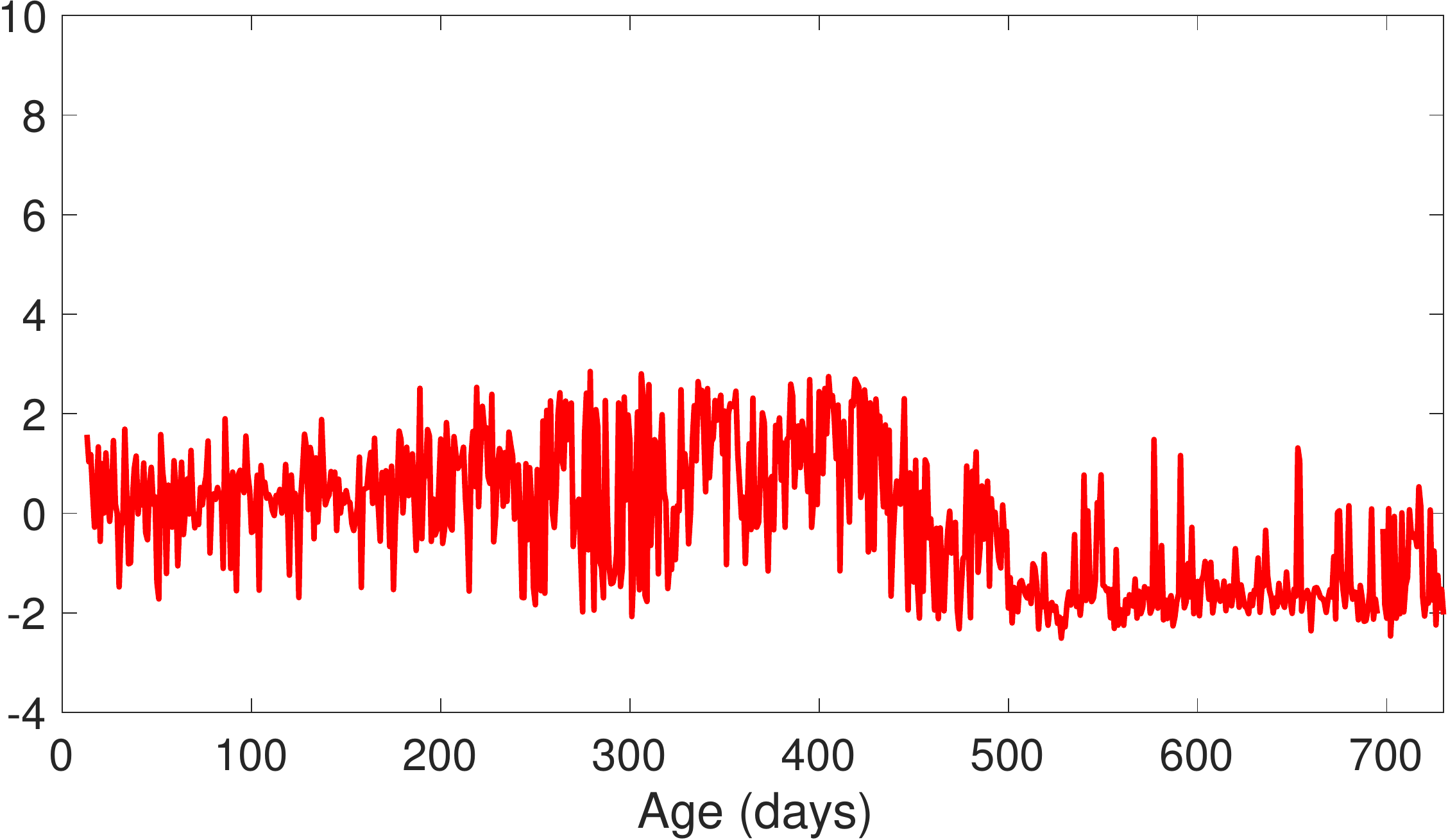} &  \includegraphics[scale=0.18]{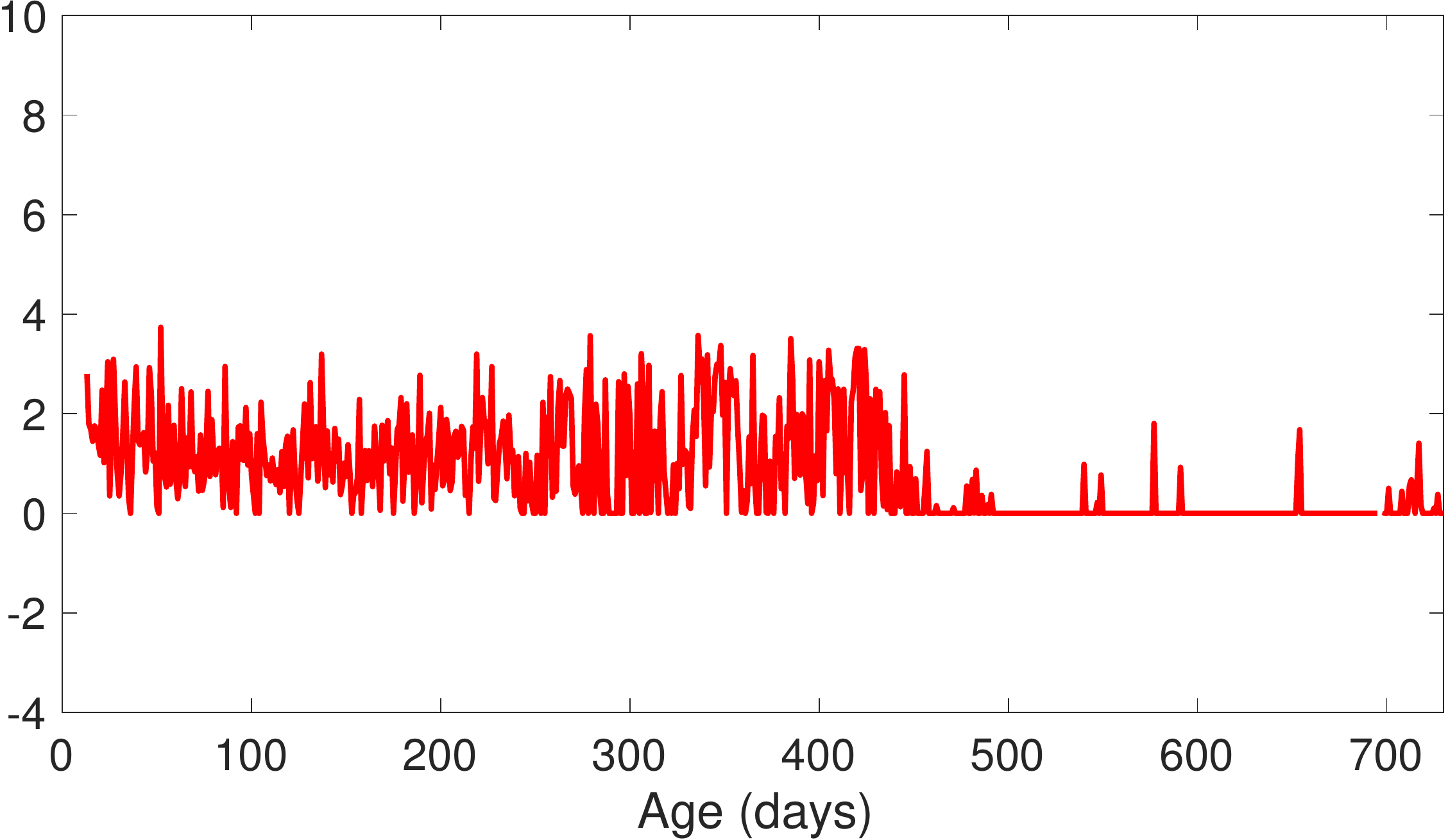} & \includegraphics[scale=0.18]{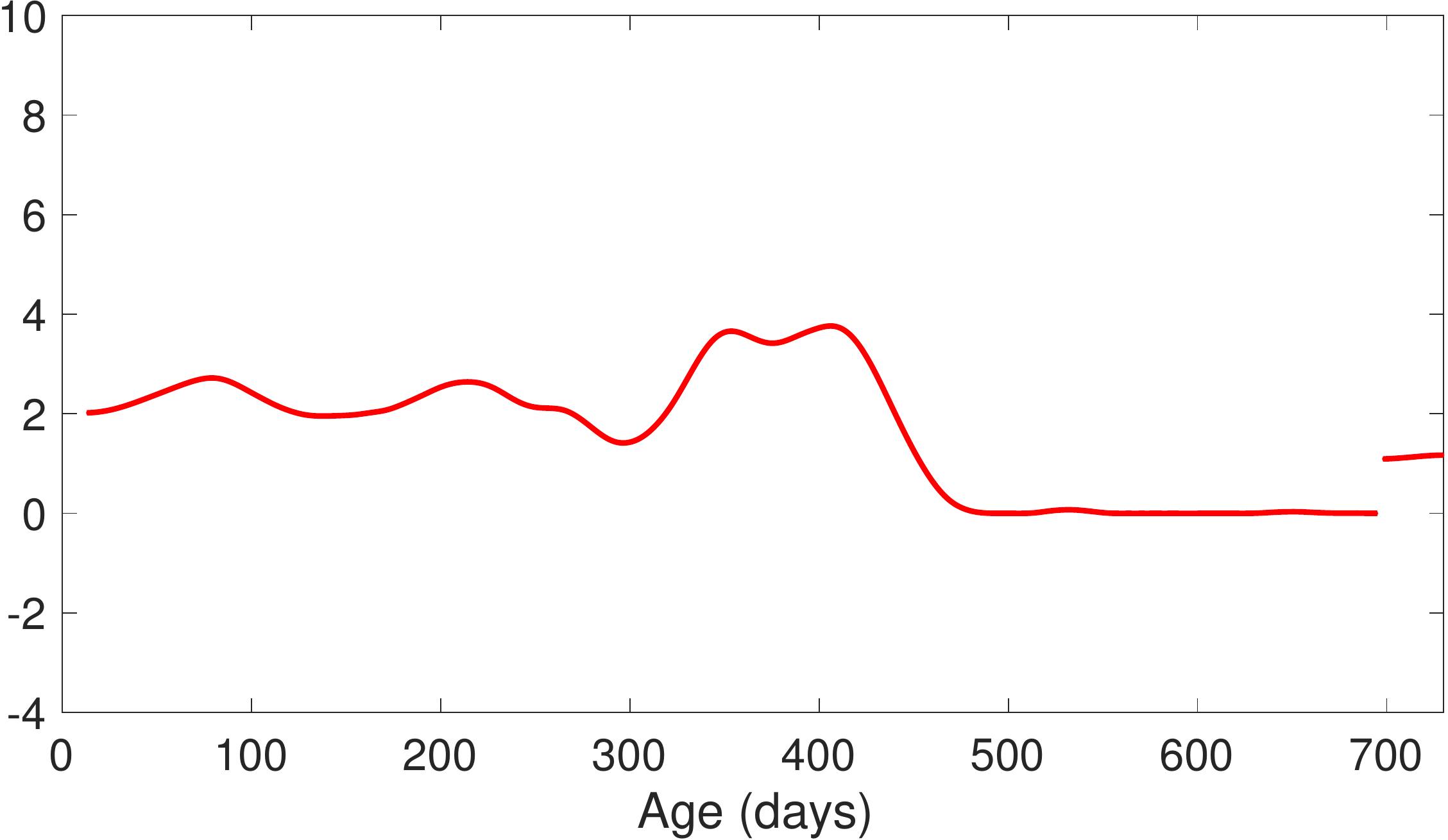} \\
3 & \includegraphics[scale=0.18]{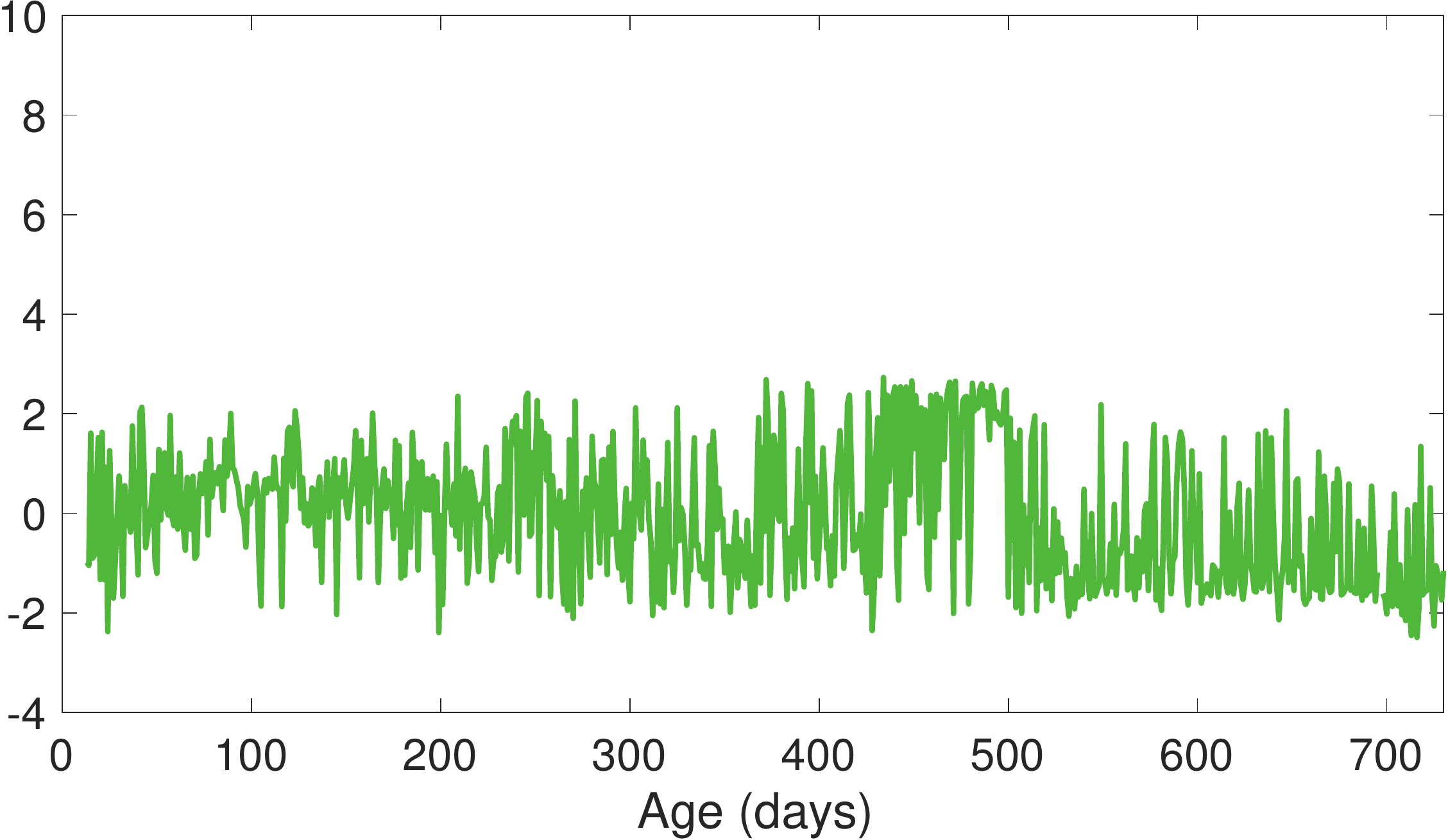} &  \includegraphics[scale=0.18]{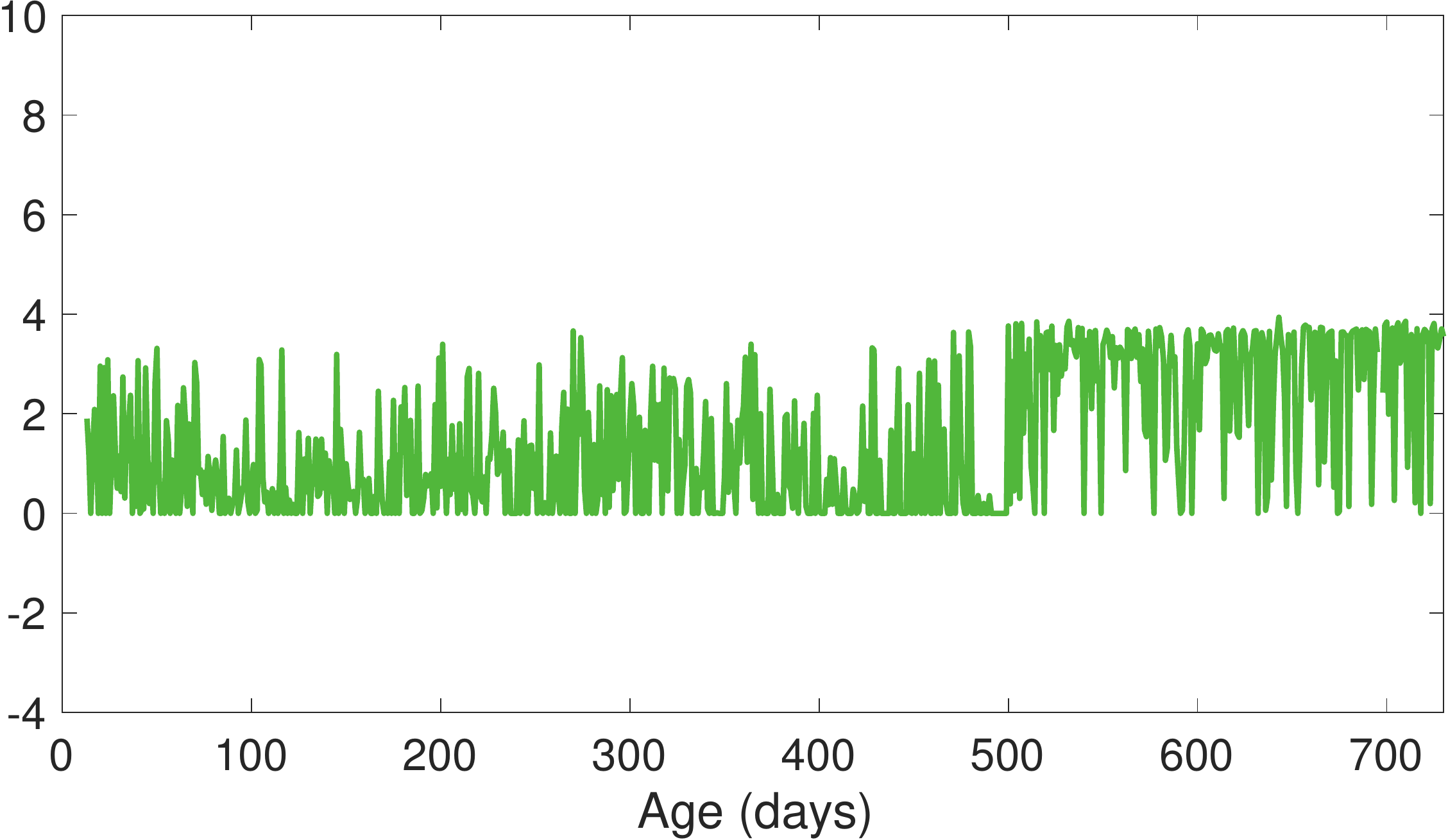} & \includegraphics[scale=0.18]{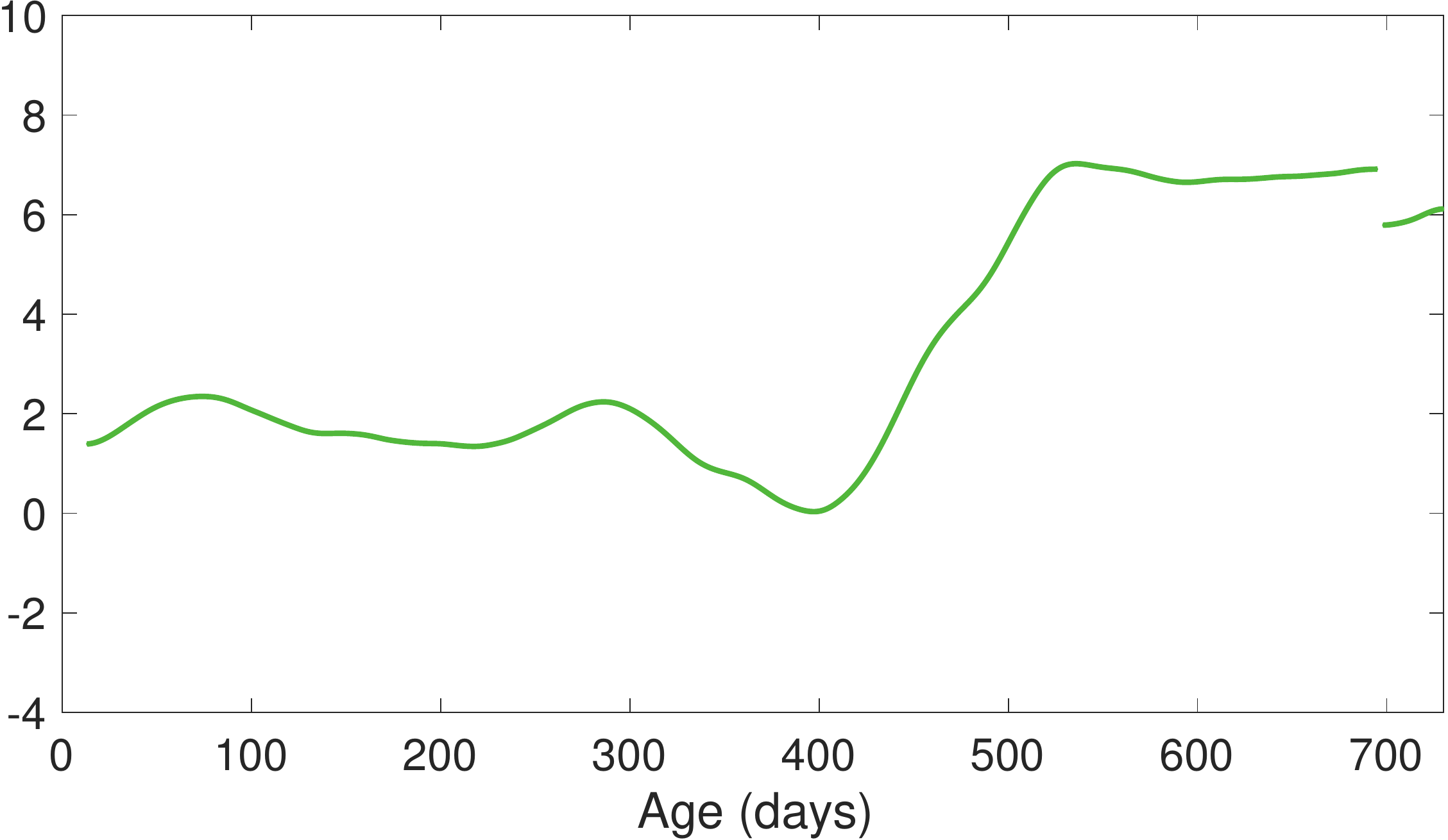} \\
4 & \includegraphics[scale=0.18]{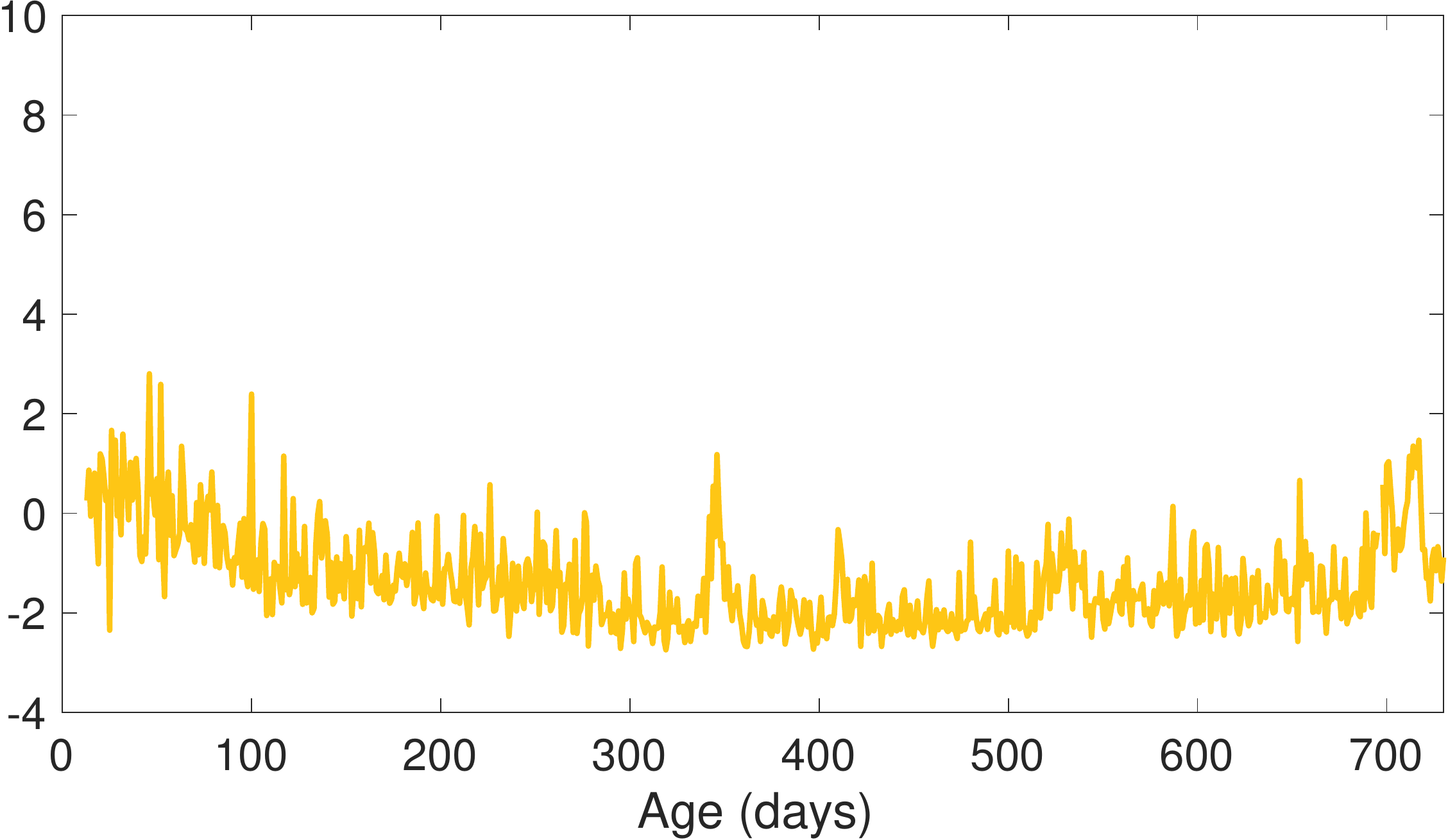} &  \includegraphics[scale=0.18]{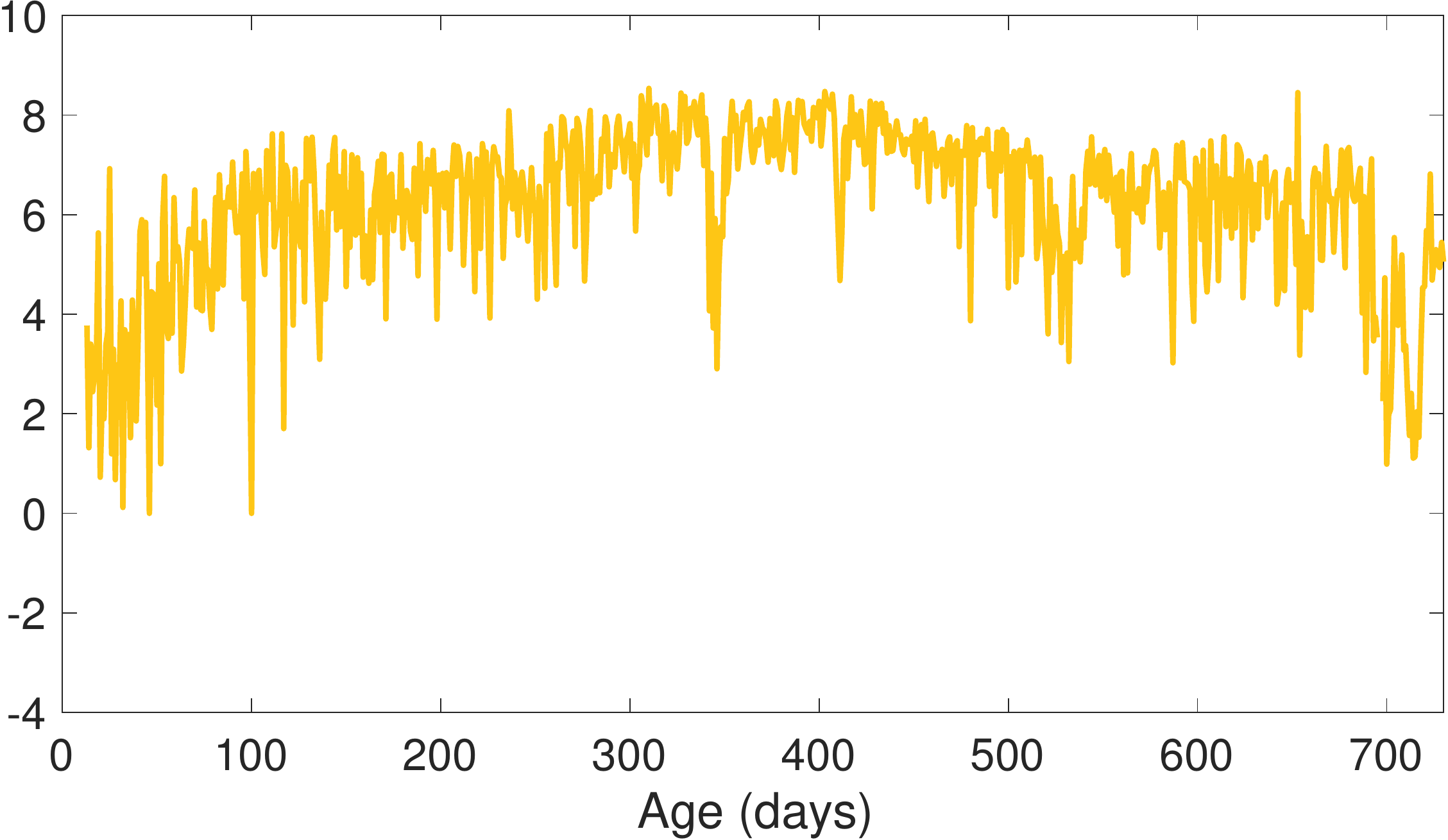} & \includegraphics[scale=0.18]{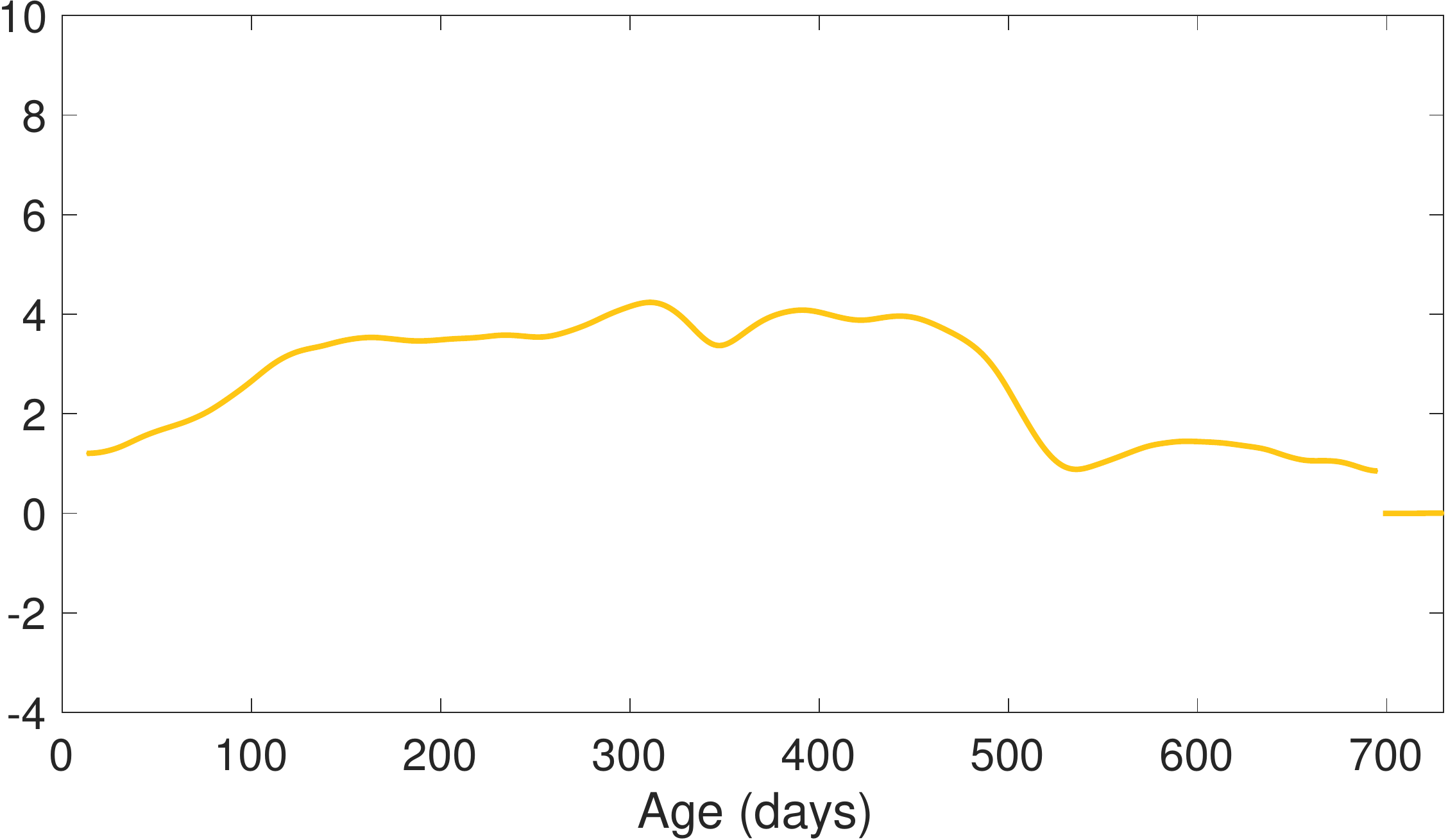} \\
5 & \includegraphics[scale=0.18]{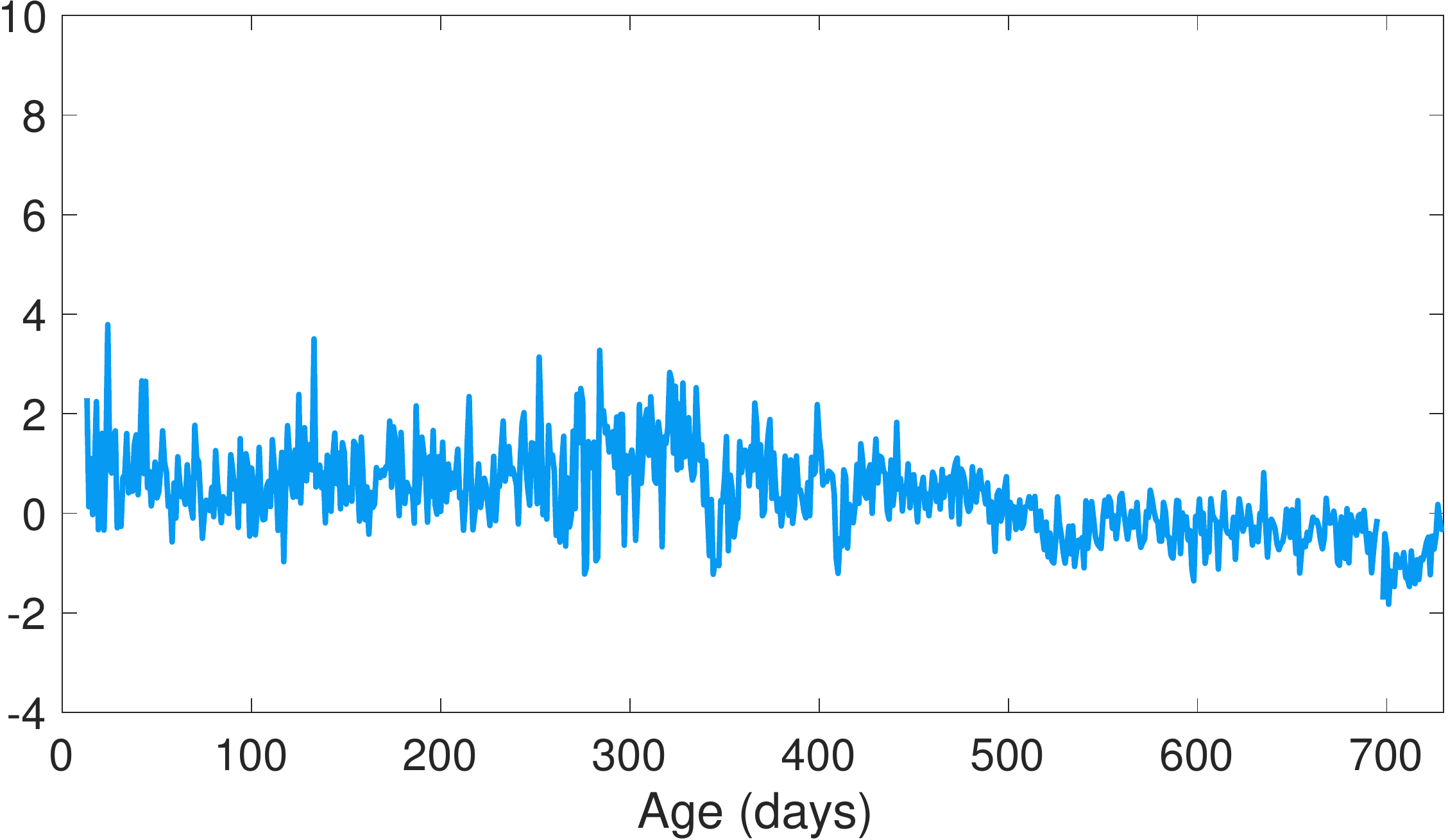} &  \includegraphics[scale=0.18]{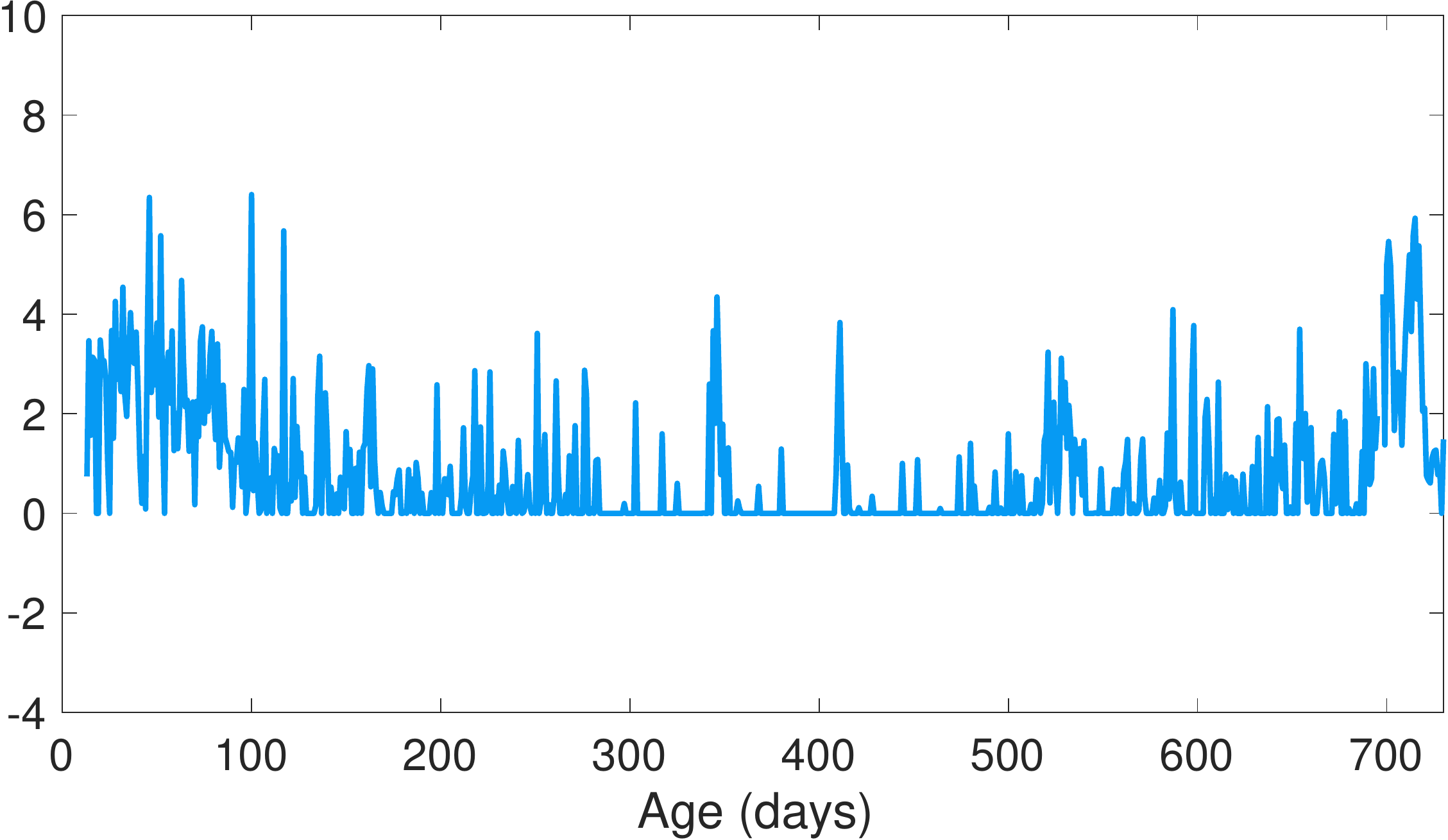} & \includegraphics[scale=0.18]{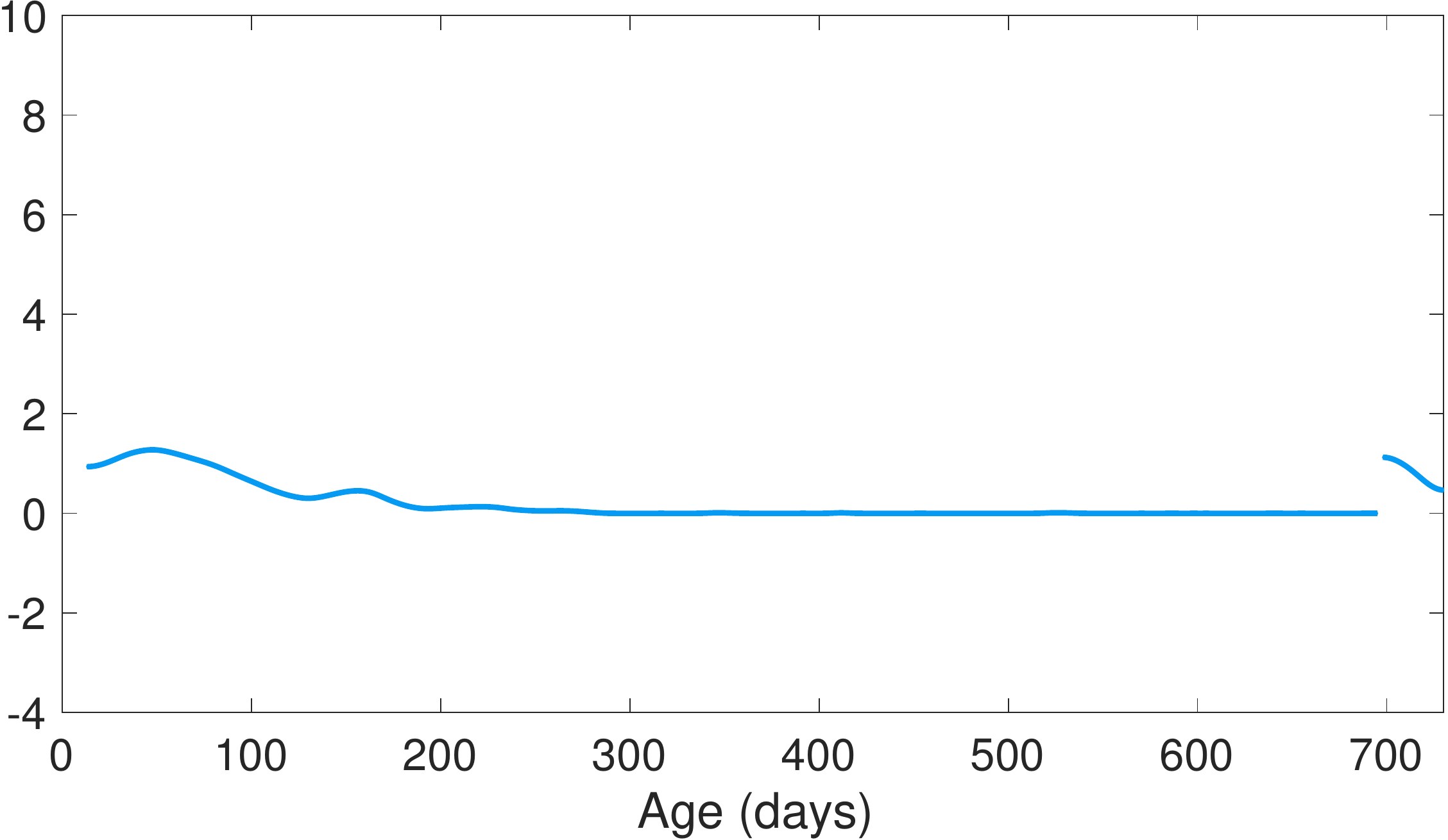} \\
 \end{tabular}
  \caption{Coefficients $C^{[n]}_{1}$, \ldots, $C^{[n]}_{r}$ corresponding to the basis functions in Figure~\ref{fig:daily_sleep_patterns} for the infant whose data is shown in Figure~\ref{fig:low_rank_approximation}. The patterns obtained from NMF are more interpretable than those obtained from PCA, but are also extremely noisy. In contrast, the patterns recovered by NMF-TS are smooth and can be interpreted in terms of sleeping behavior that are consistent with the infant's raw data in Figure~\ref{fig:low_rank_approximation}. }
 \label{fig:age_development_patterns}
\end{figure}

\begin{figure}[h]
\begin{tabular}{ >{\centering\arraybackslash}m{0.17\linewidth} >{\centering\arraybackslash}m{0.17\linewidth} >{\centering\arraybackslash}m{0.17\linewidth} >{\centering\arraybackslash}m{0.17\linewidth} >{\centering\arraybackslash}m{0.17\linewidth} >{\centering\arraybackslash}m{0.17\linewidth}}
\includegraphics[scale=0.12]{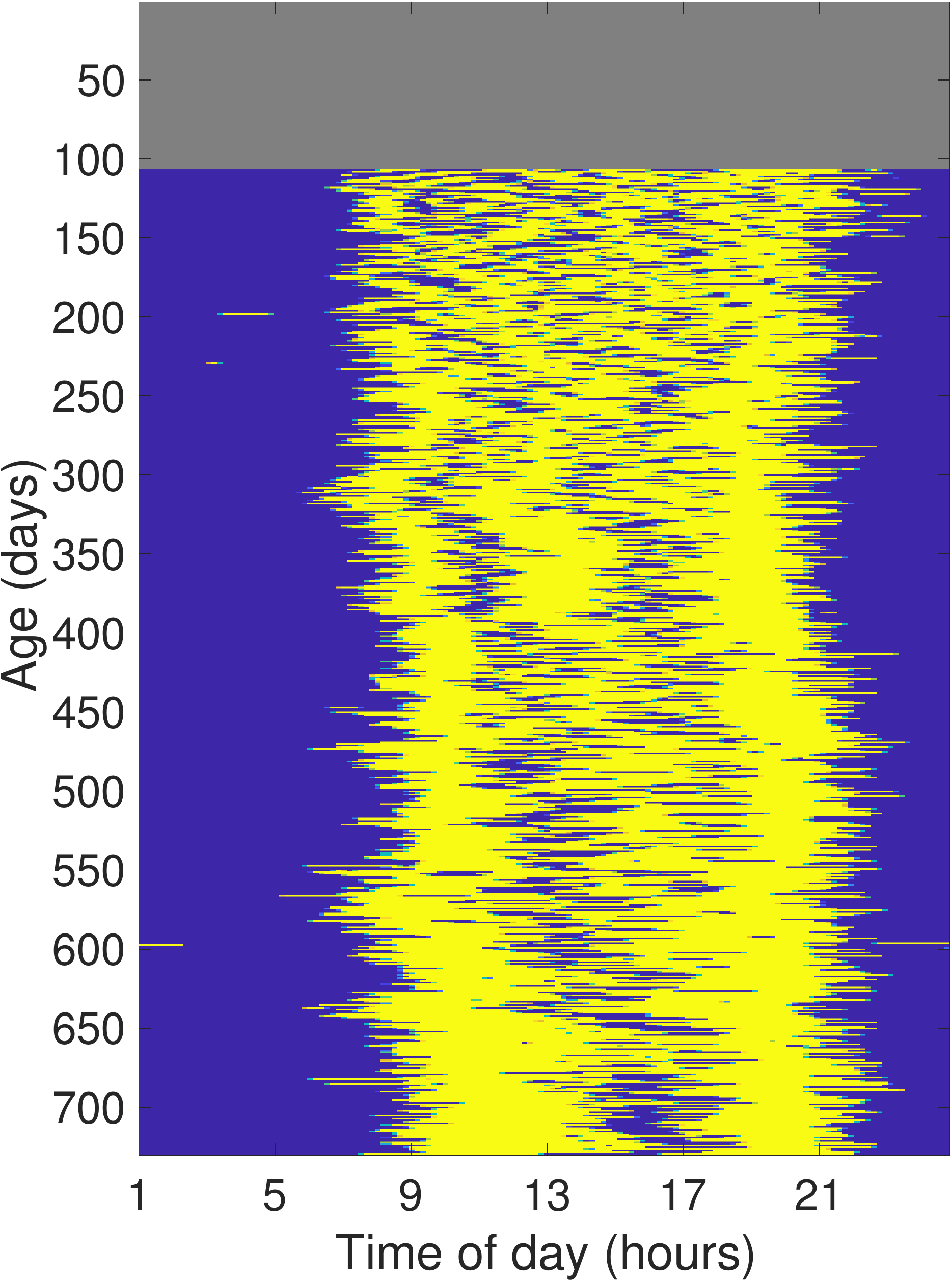} & \includegraphics[scale=0.12]{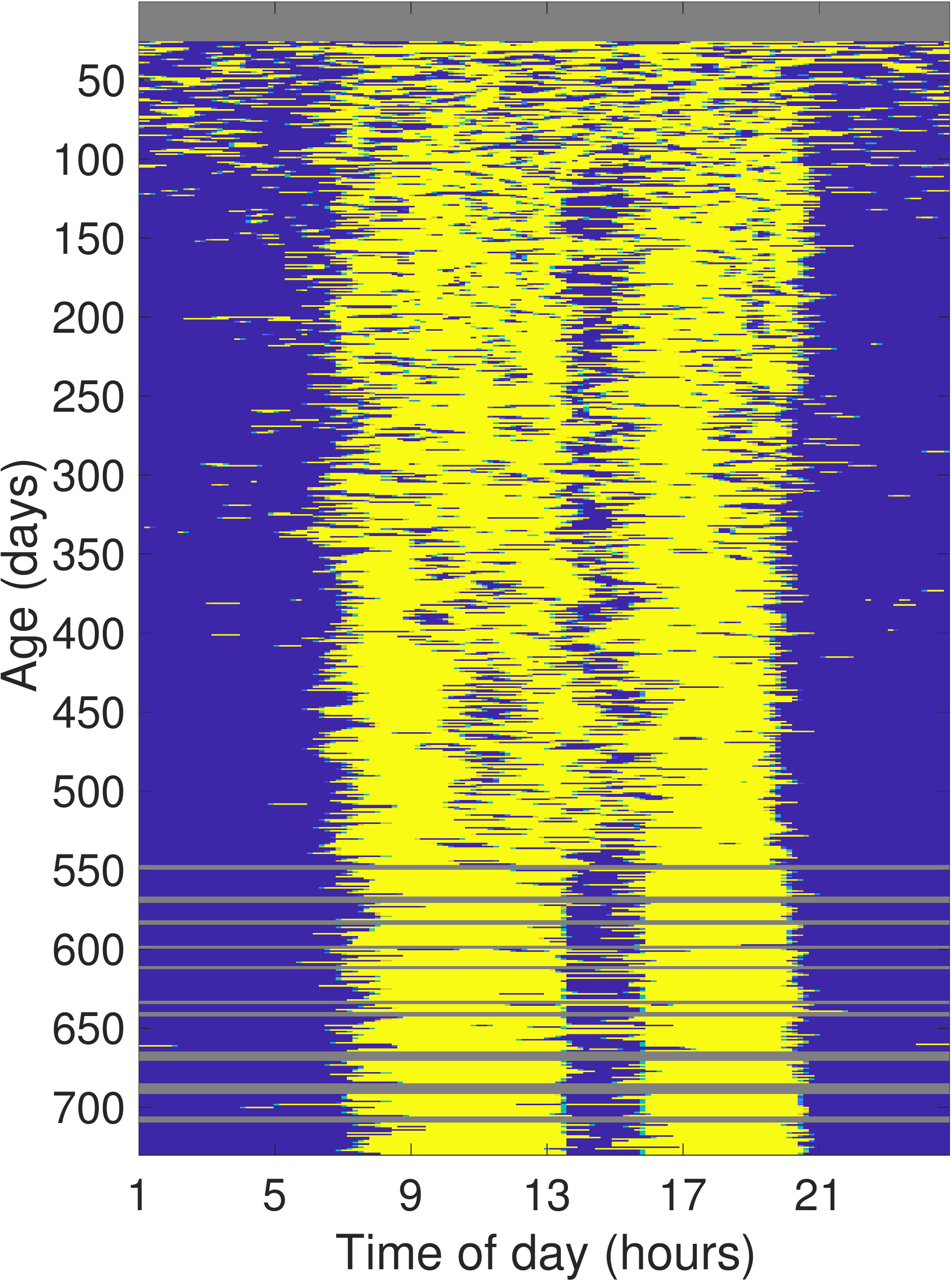} &  \includegraphics[scale=0.12]{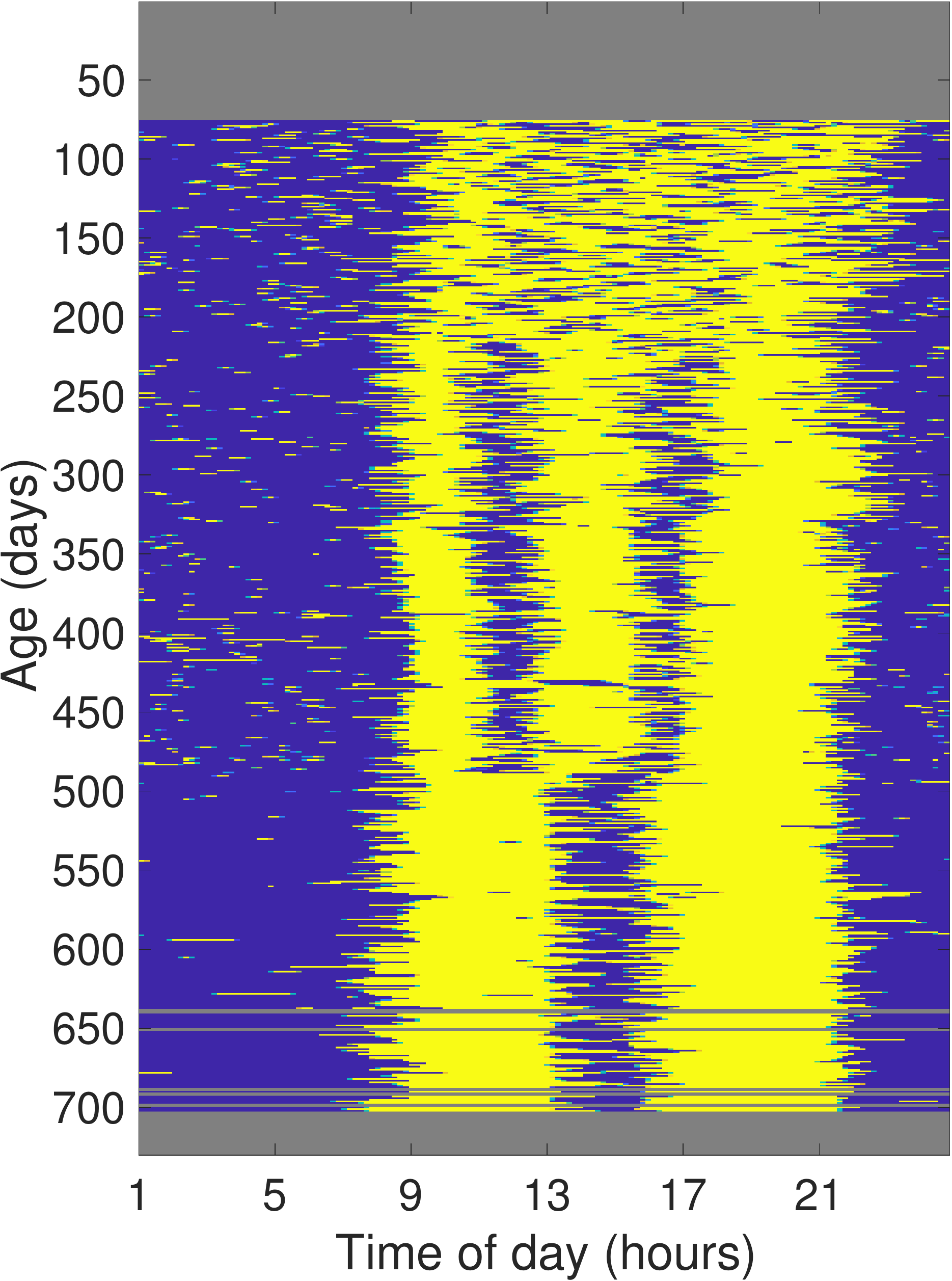} &  \includegraphics[scale=0.12]{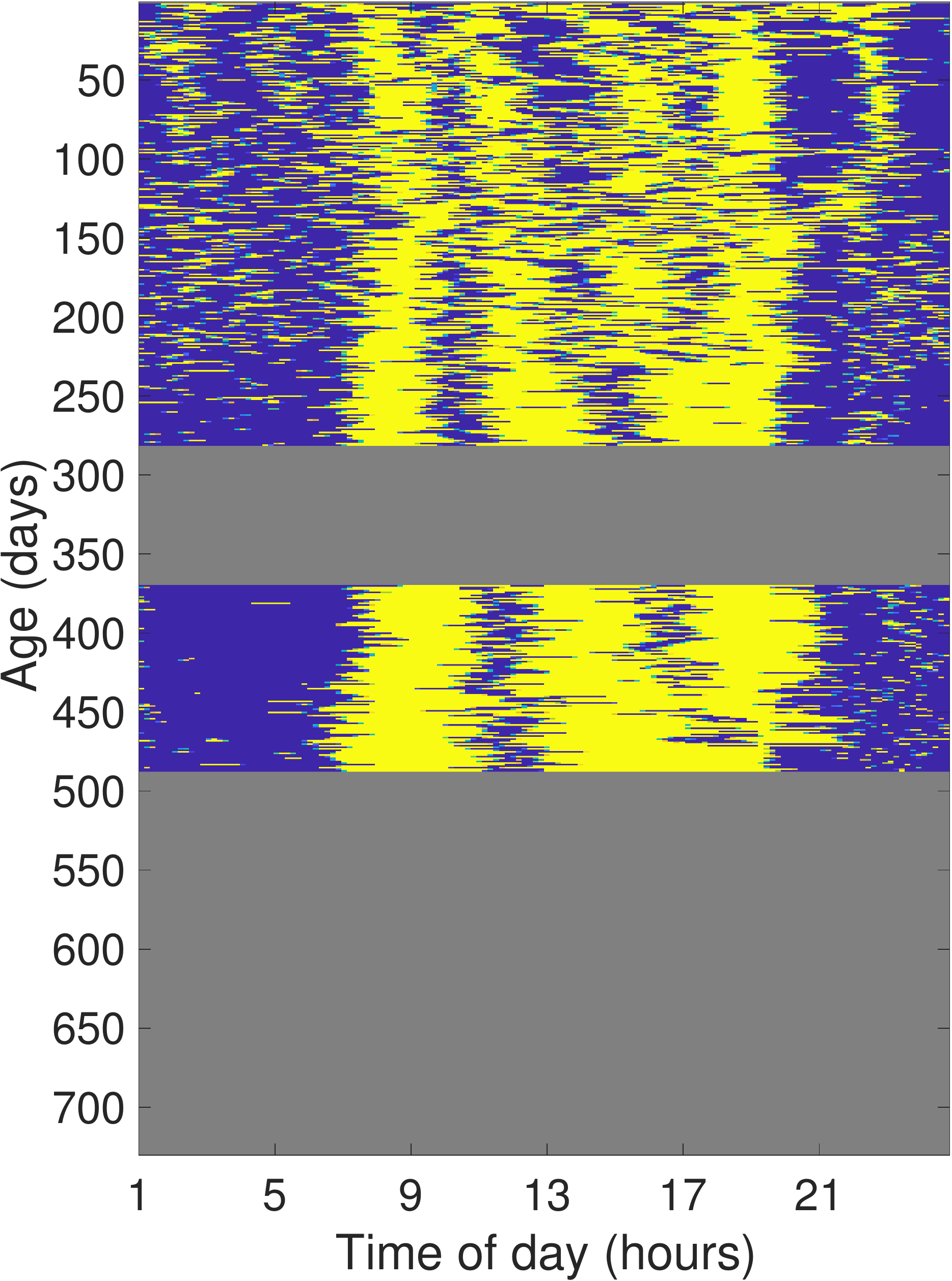}&  \includegraphics[scale=0.12]{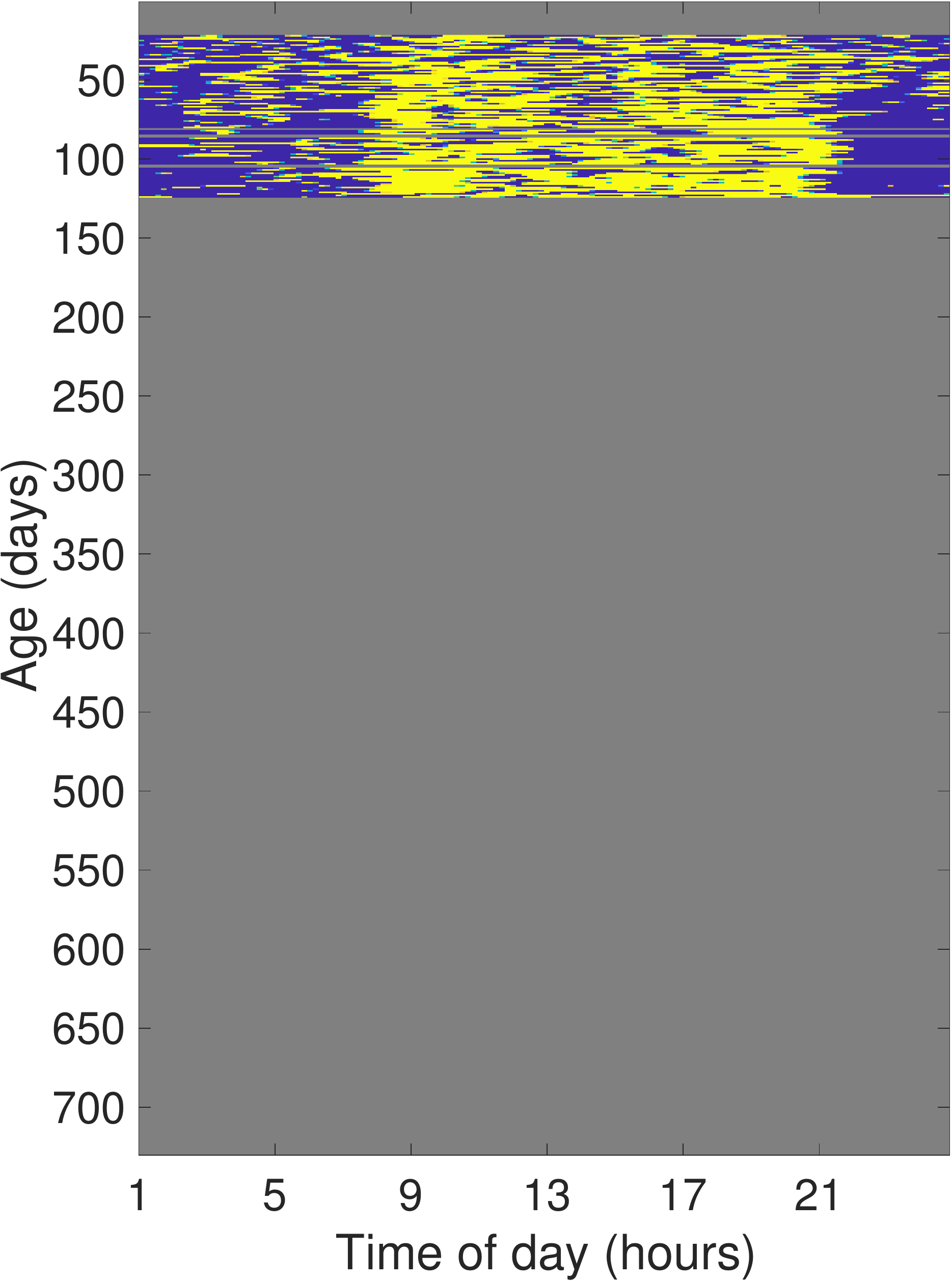} \\
\includegraphics[scale=0.125]{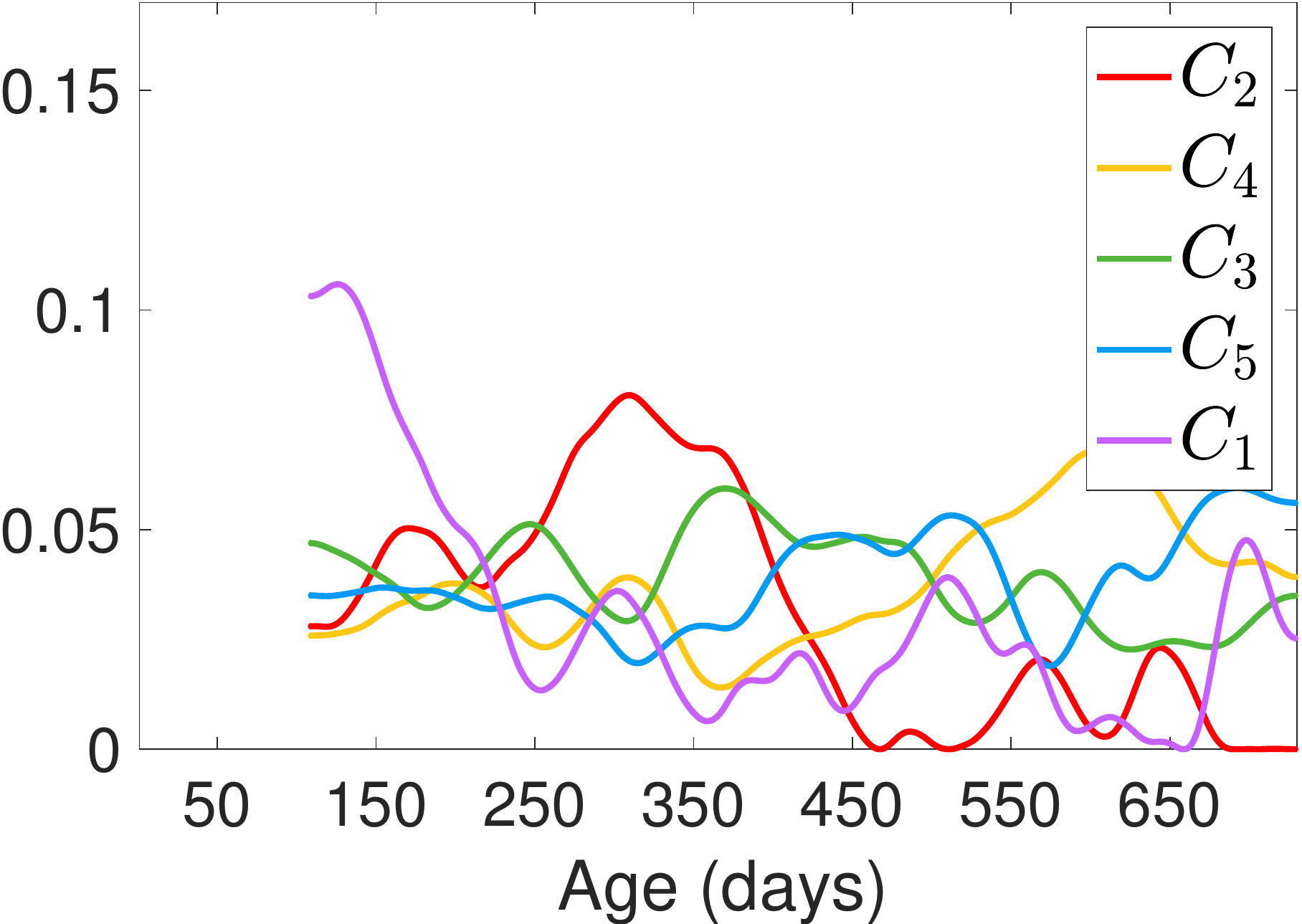} & \includegraphics[scale=0.125]{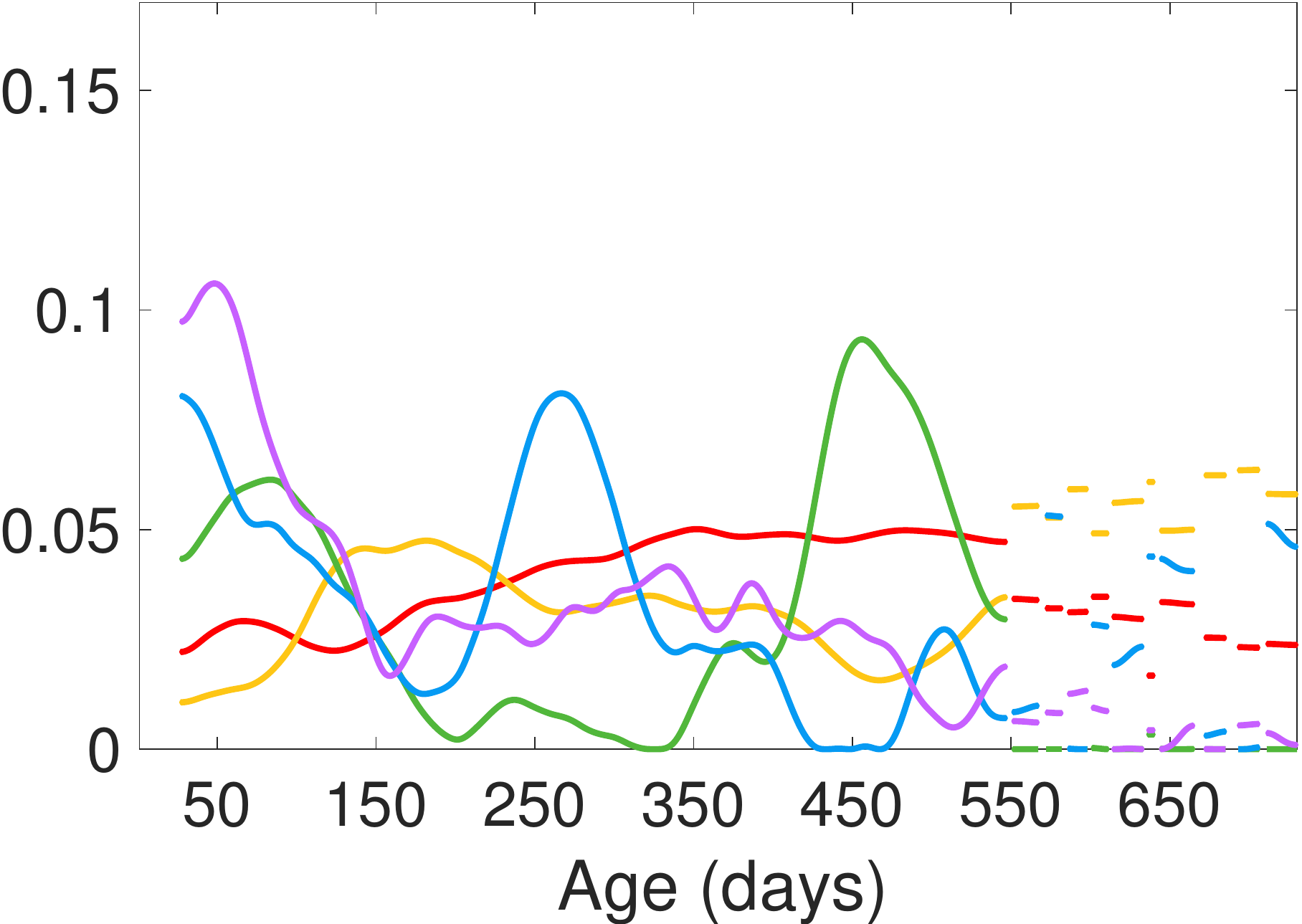} &  \includegraphics[scale=0.125]{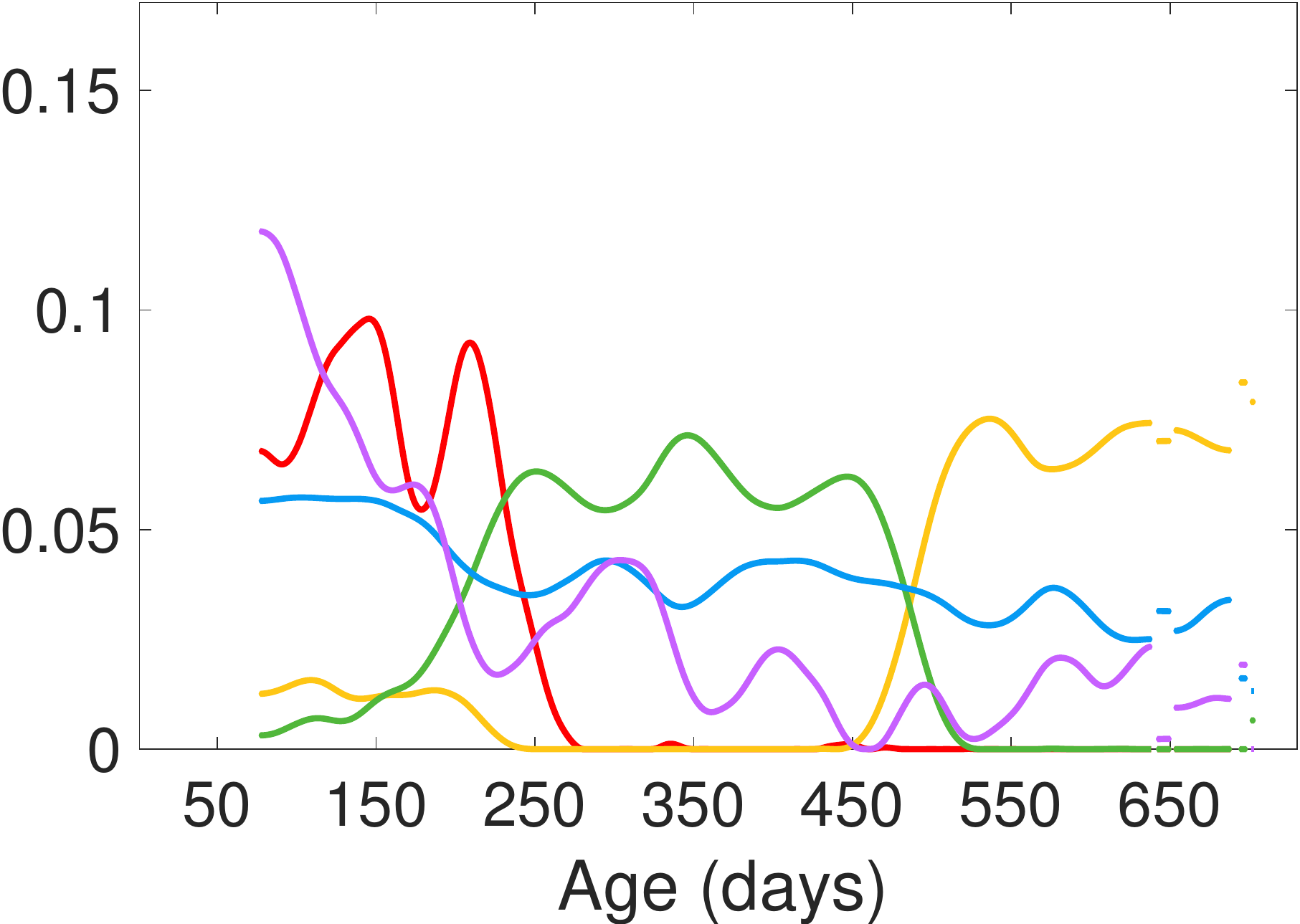} &  \includegraphics[scale=0.125]{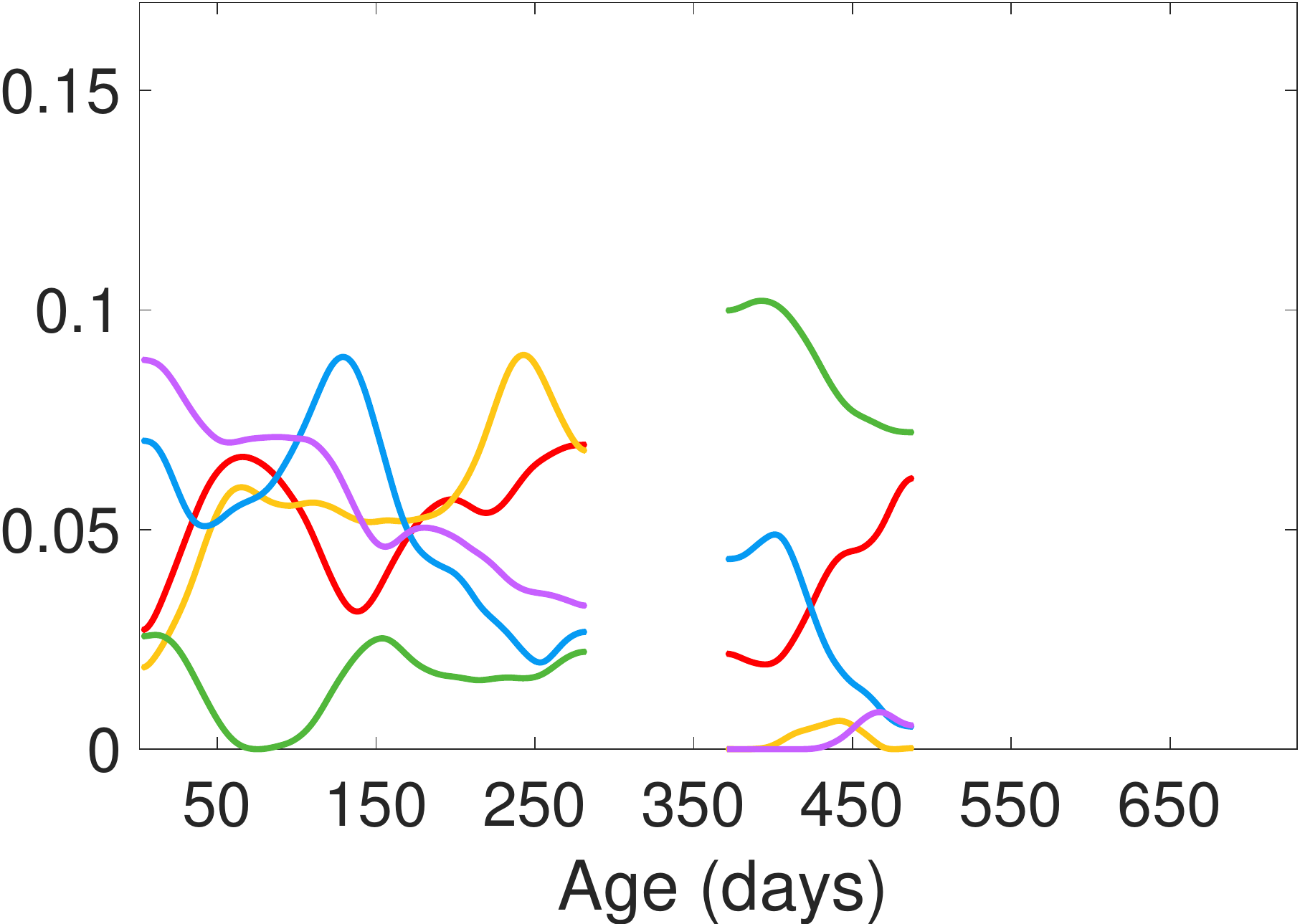}&  \includegraphics[scale=0.125]{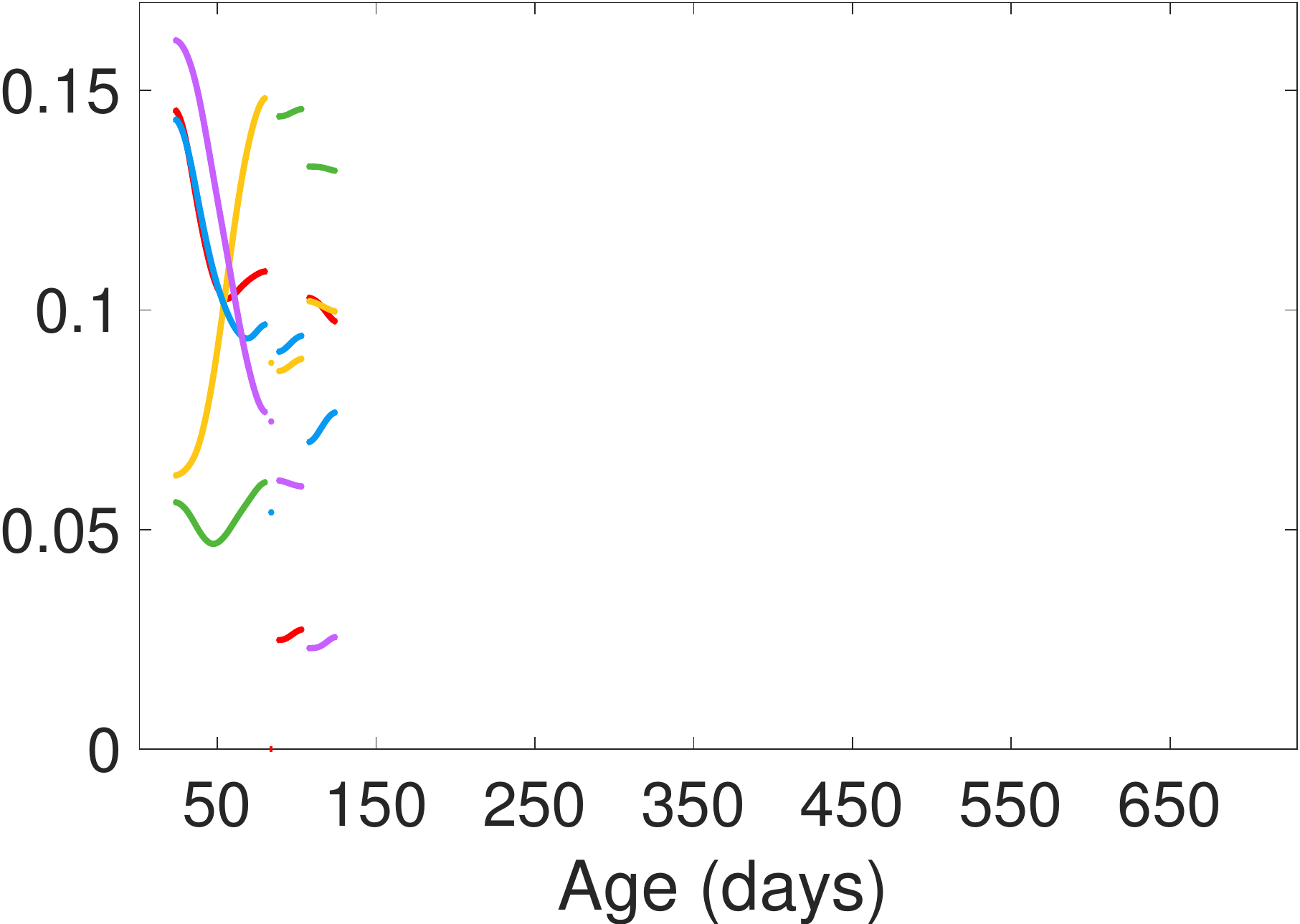} \\
\includegraphics[scale=0.12]{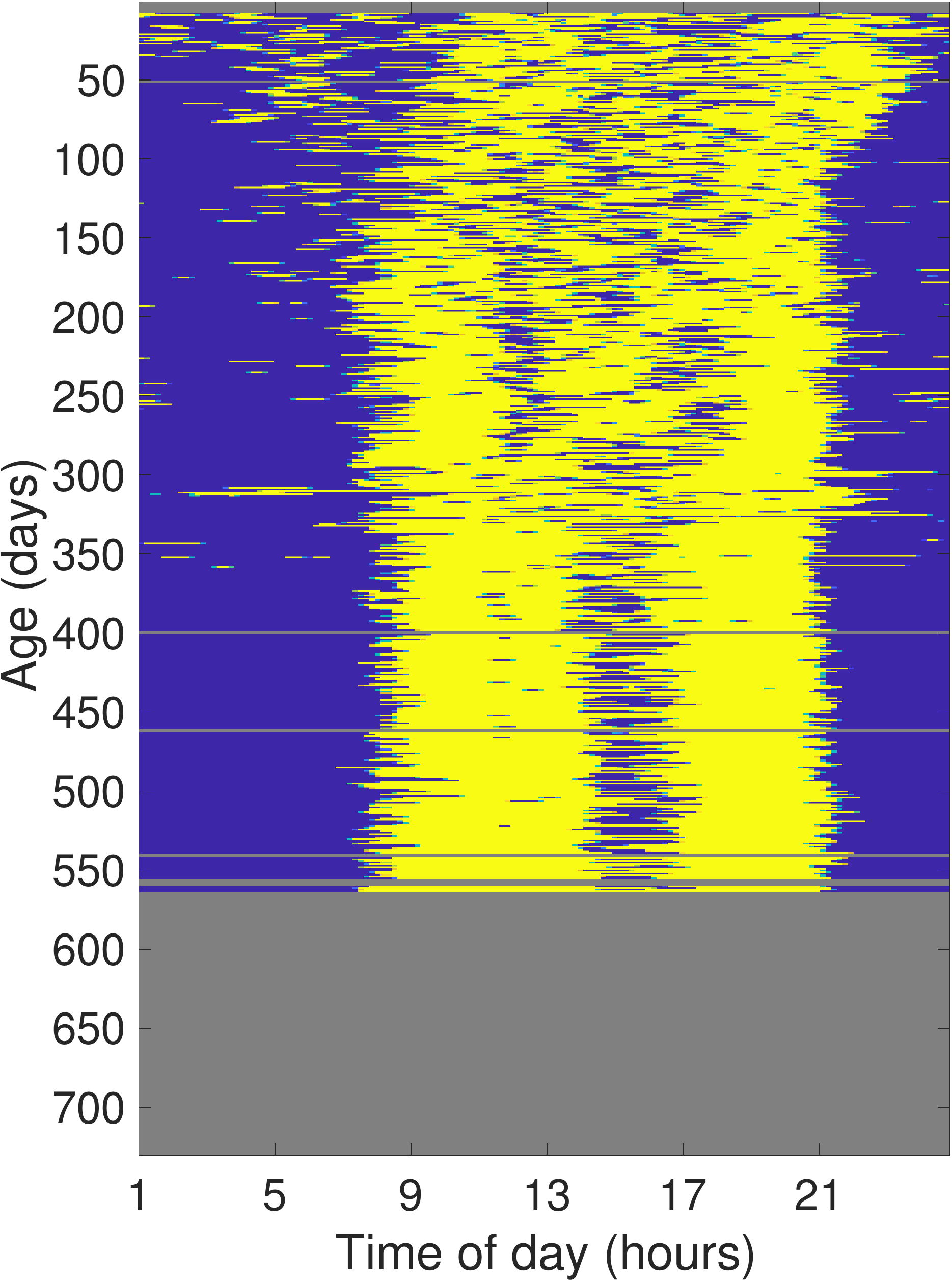} & \includegraphics[scale=0.12]{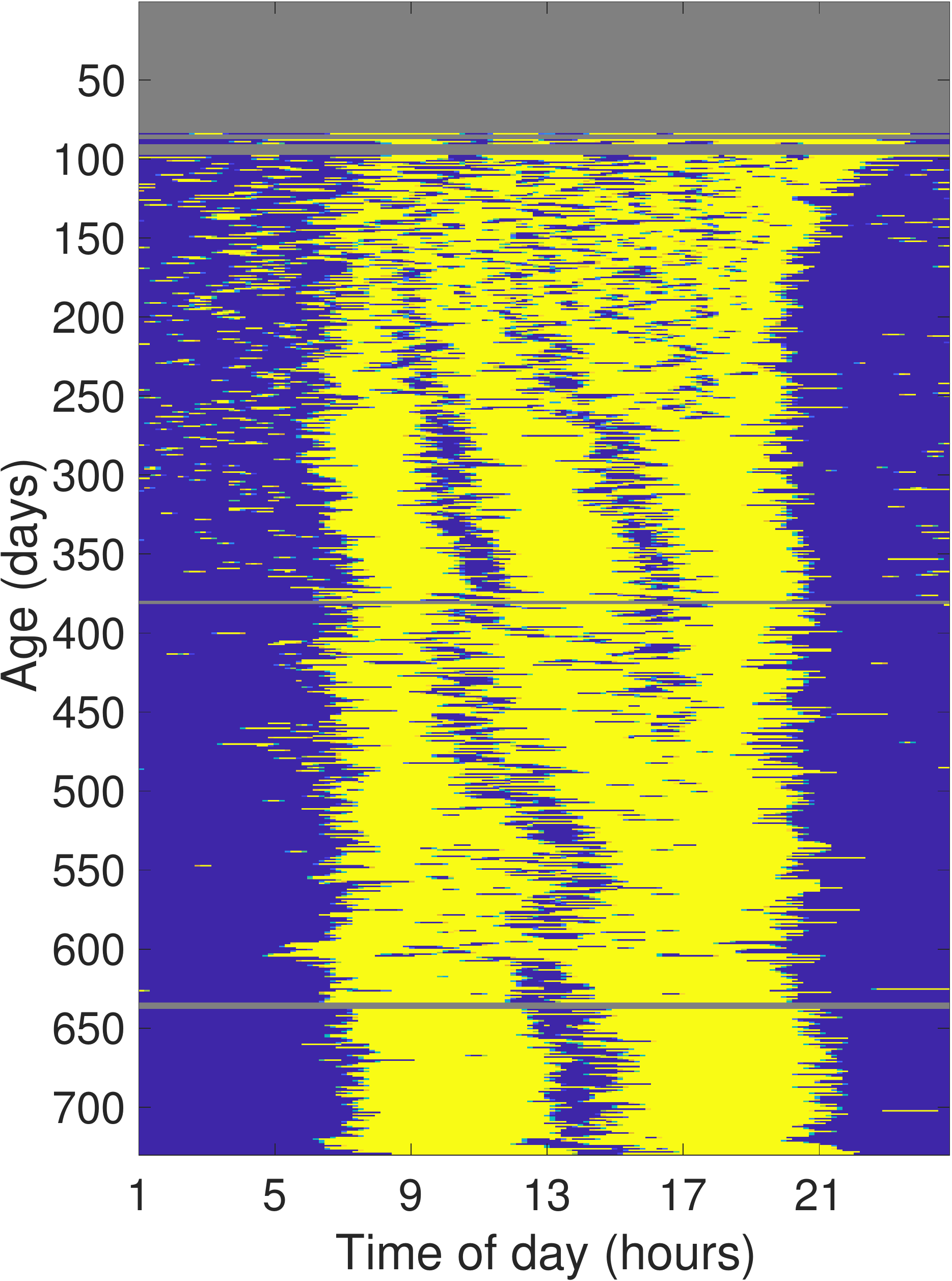} &  \includegraphics[scale=0.12]{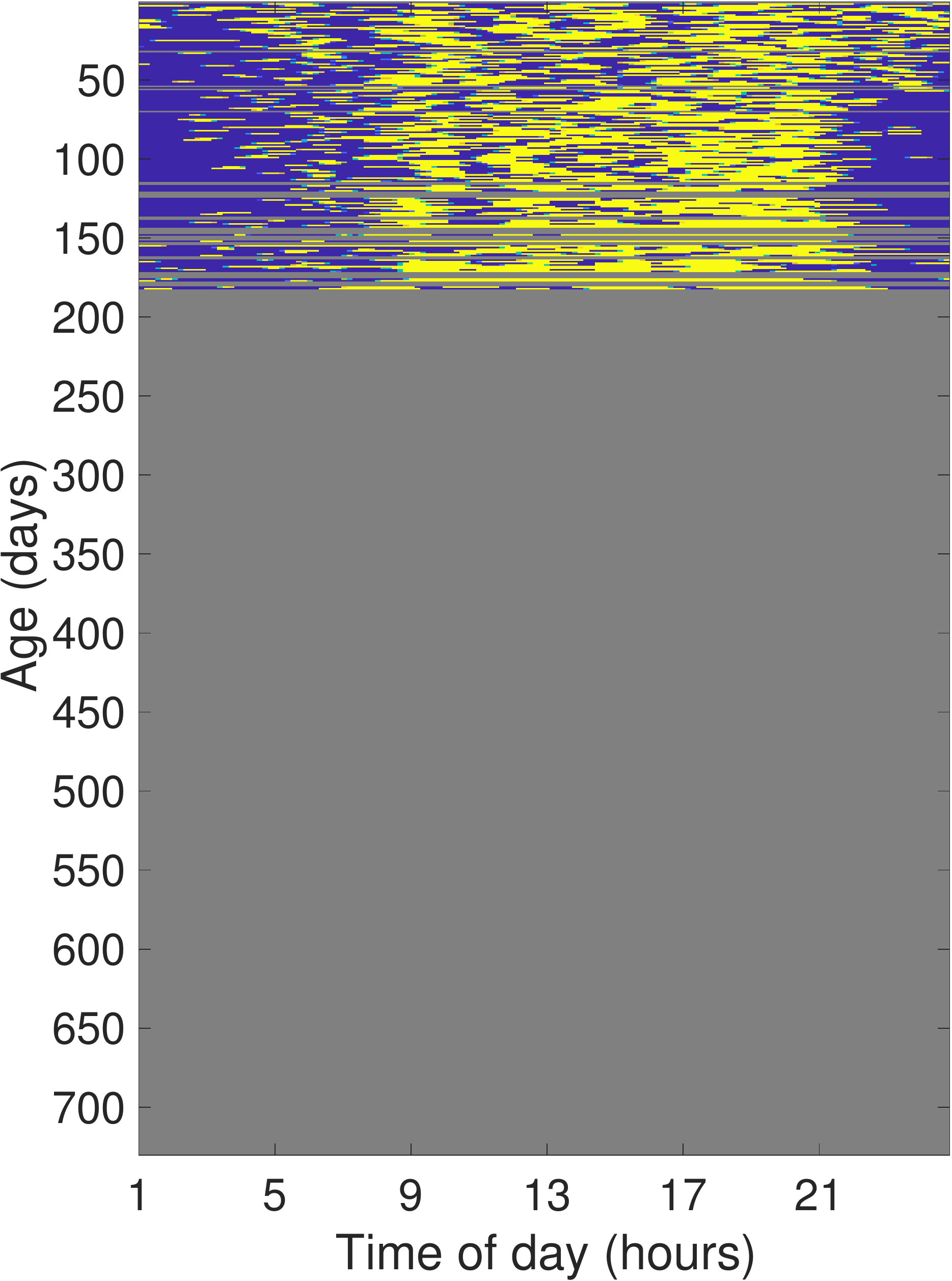} &  \includegraphics[scale=0.12]{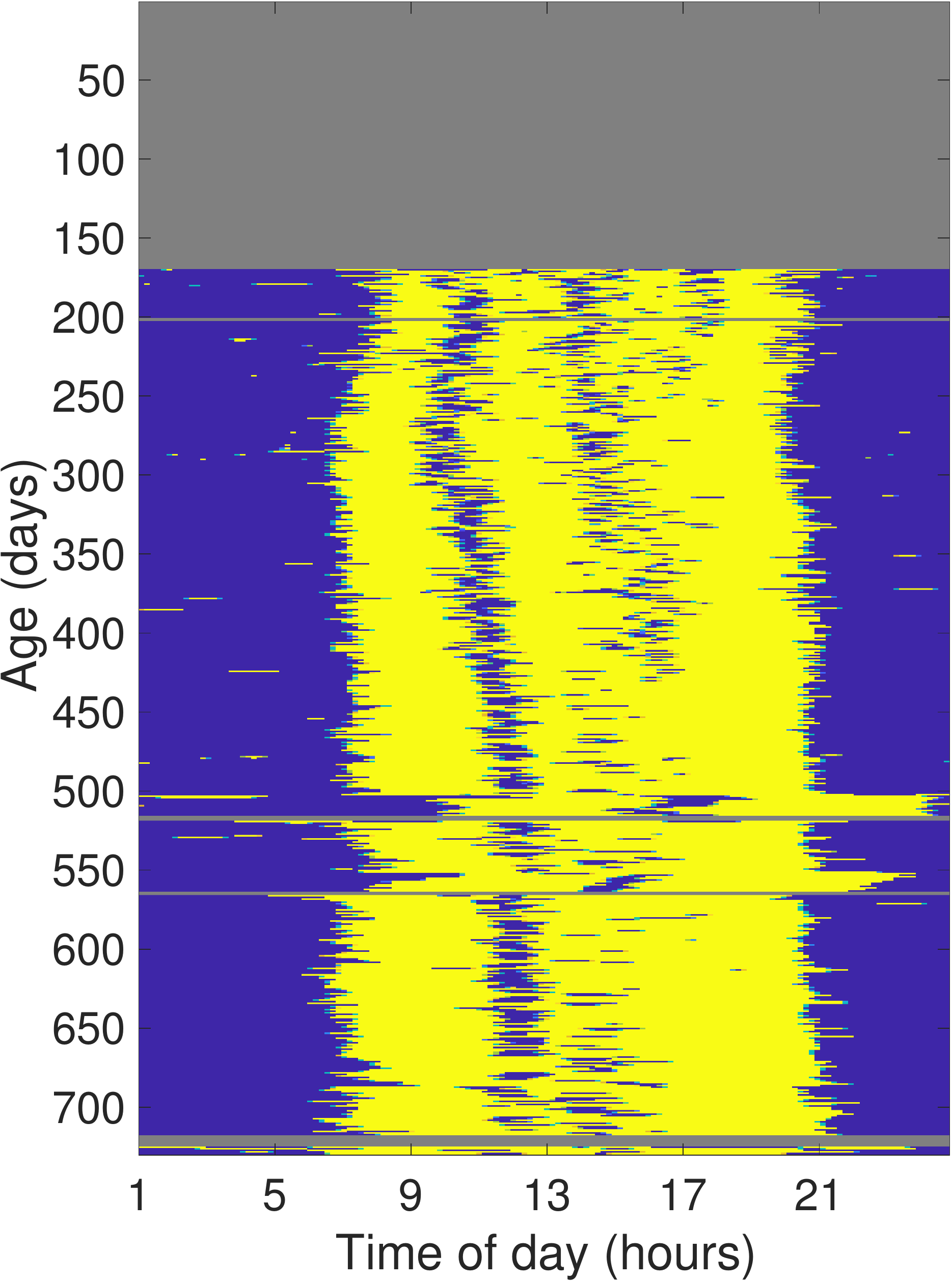}&  \includegraphics[scale=0.12]{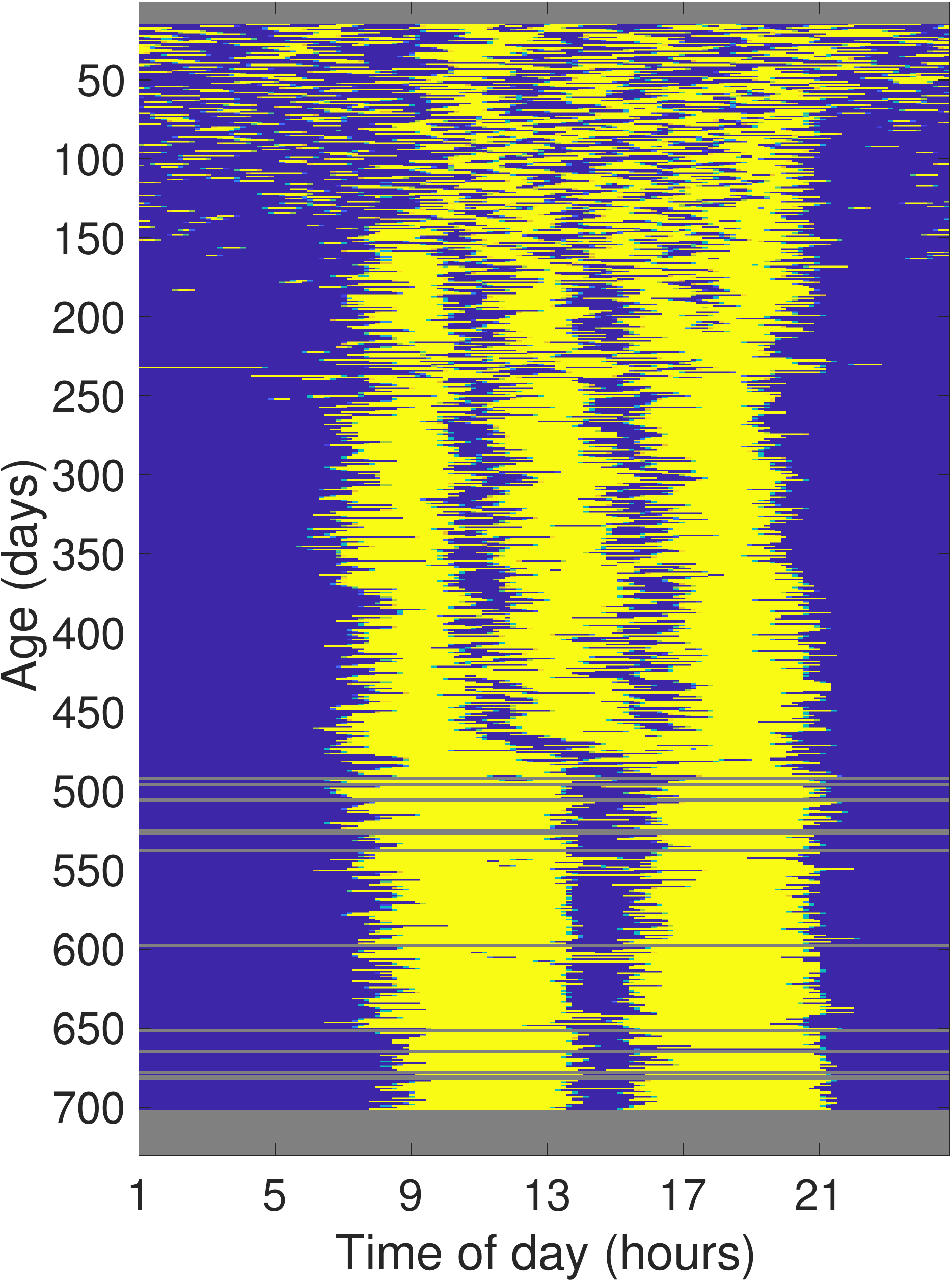} \\
\includegraphics[scale=0.125]{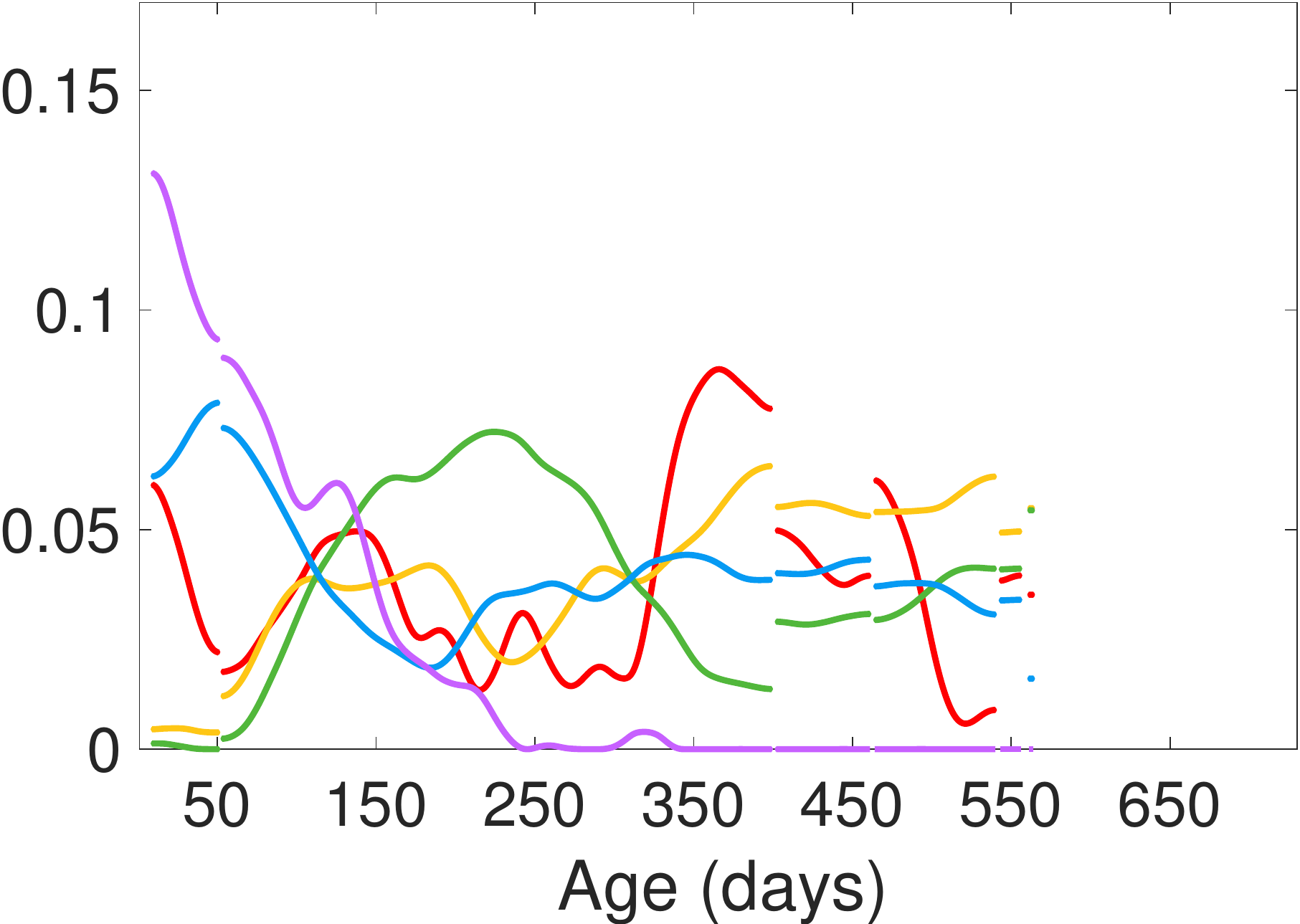} & \includegraphics[scale=0.125]{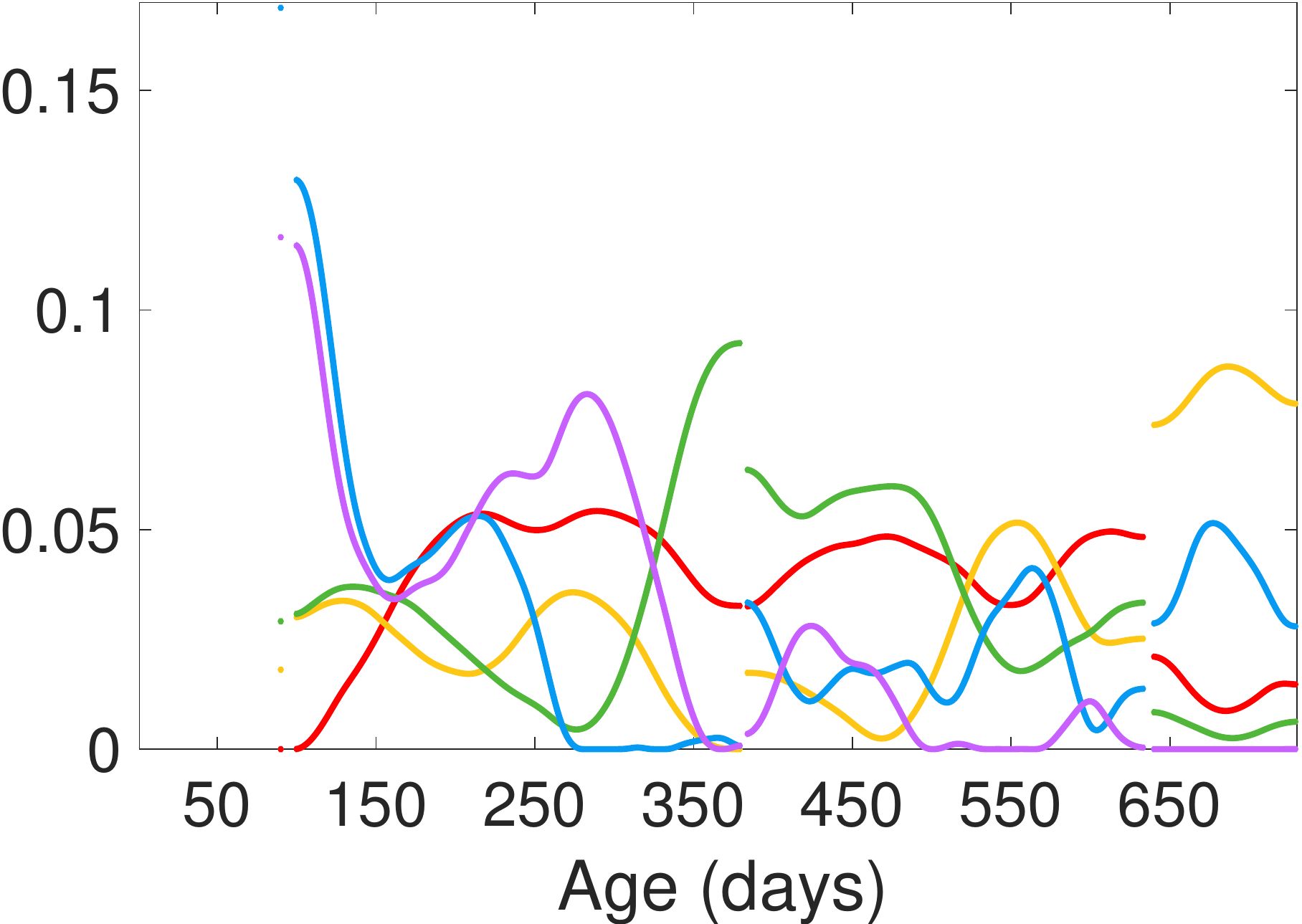} &  \includegraphics[scale=0.125]{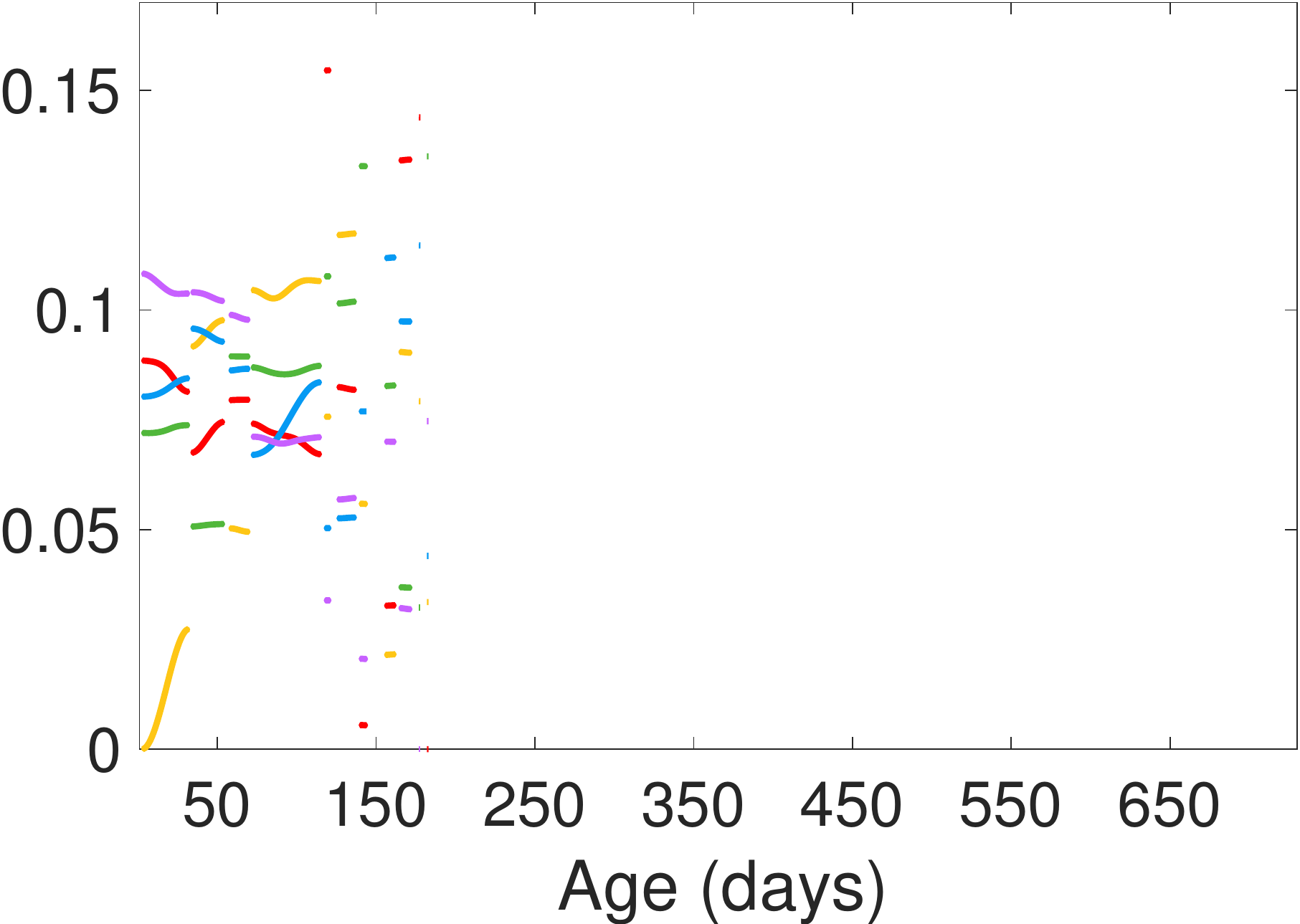} &  \includegraphics[scale=0.125]{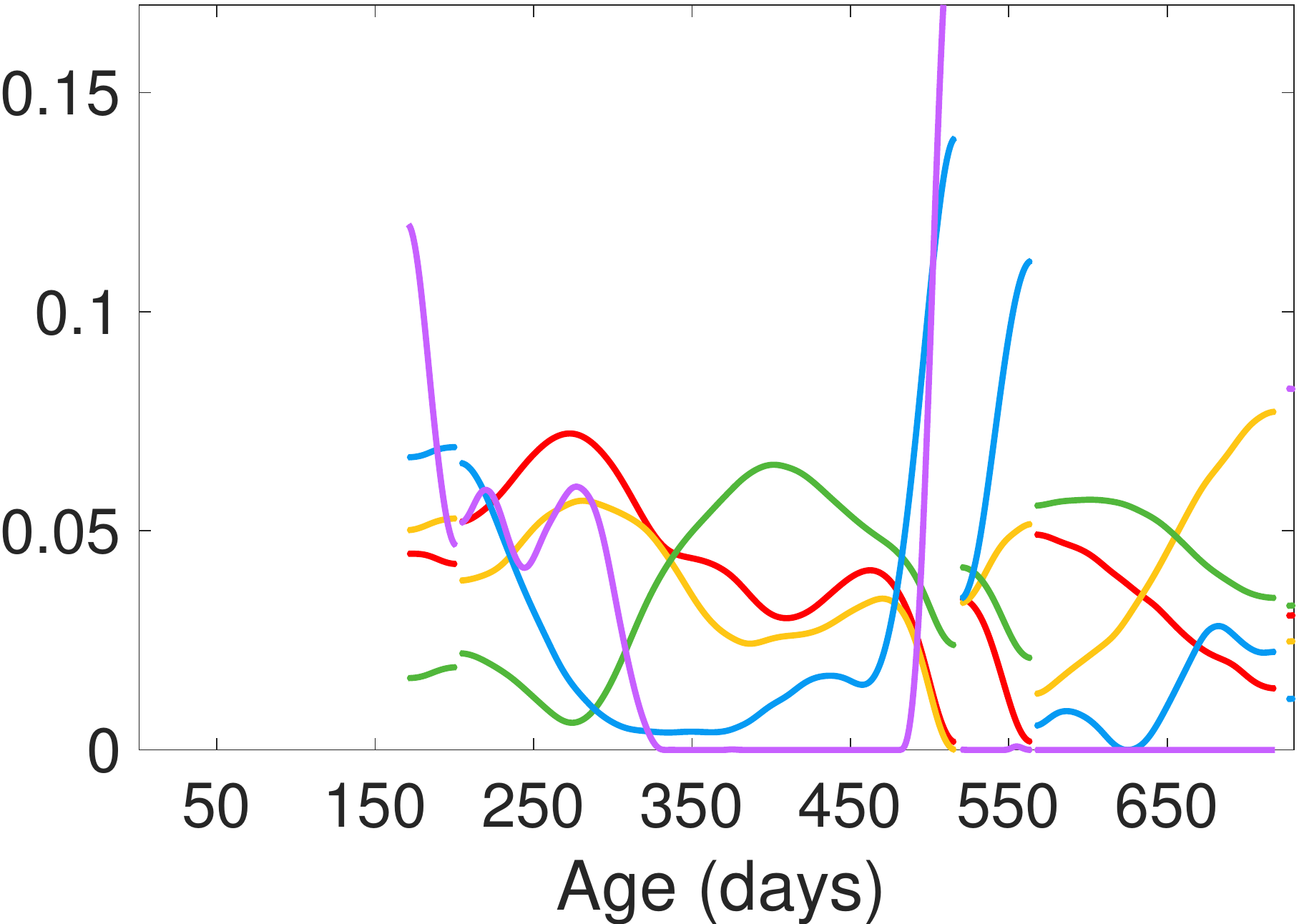}&  \includegraphics[scale=0.125]{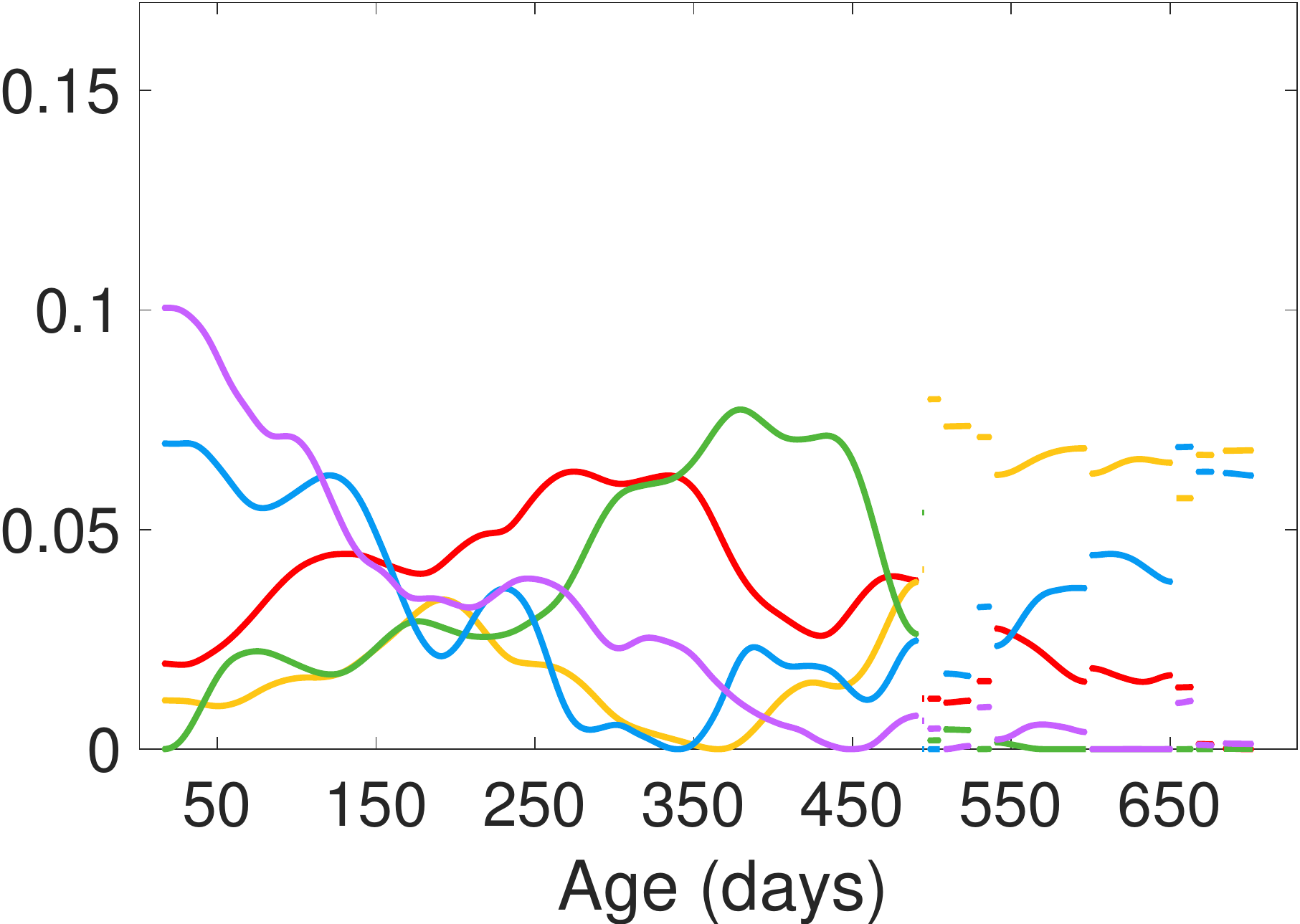} \\
\includegraphics[scale=0.12]{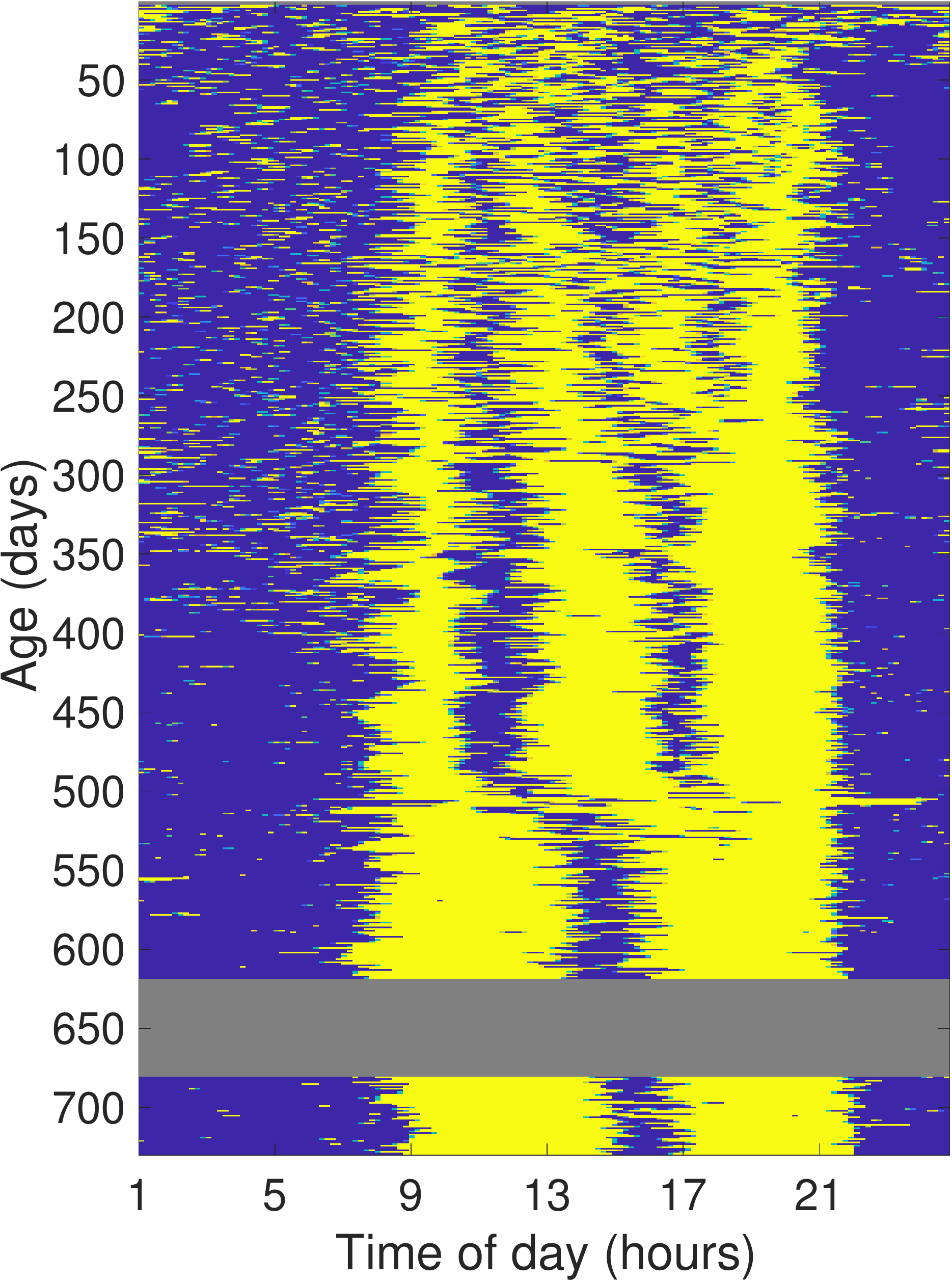} & \includegraphics[scale=0.12]{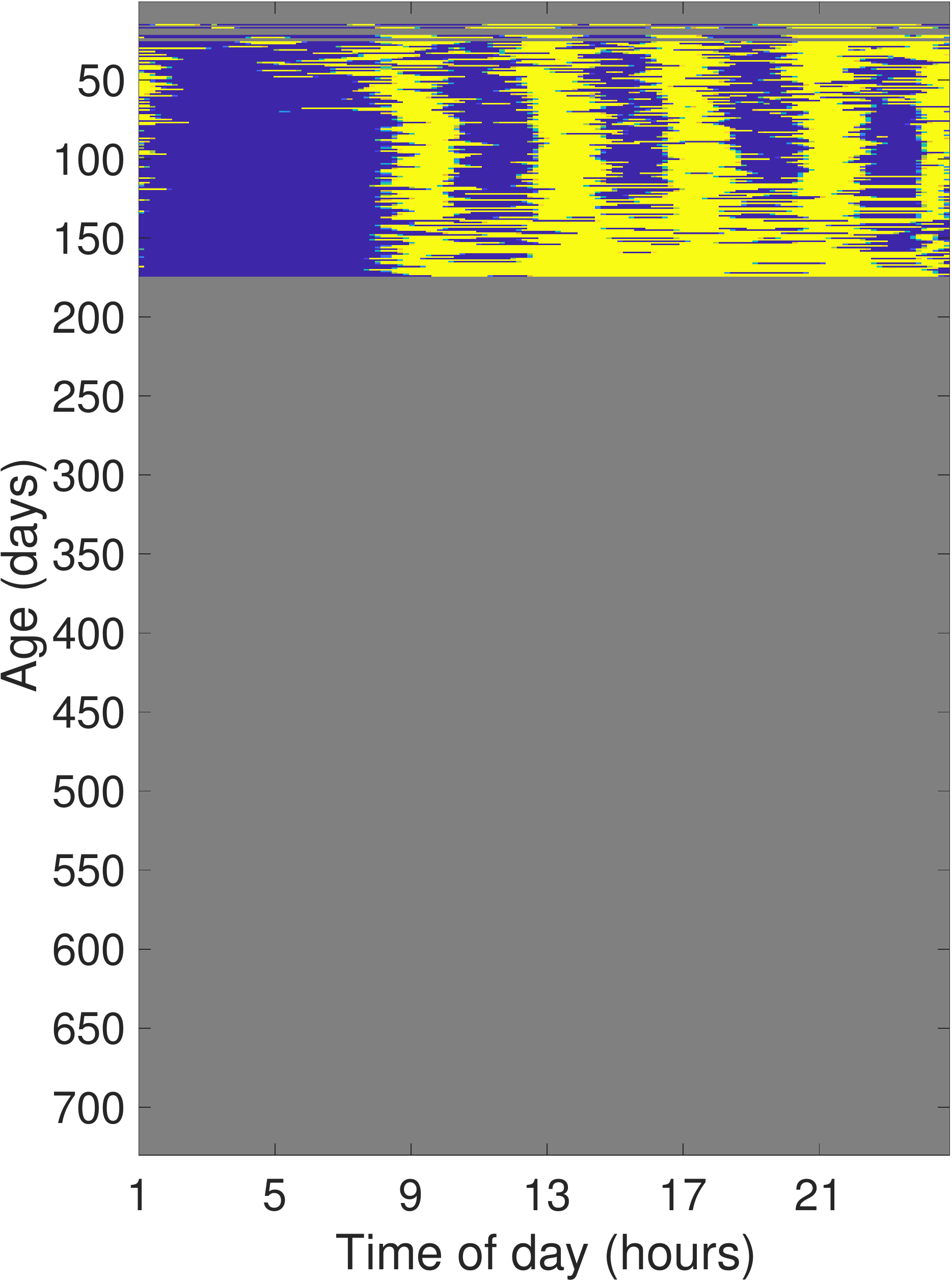} &  \includegraphics[scale=0.12]{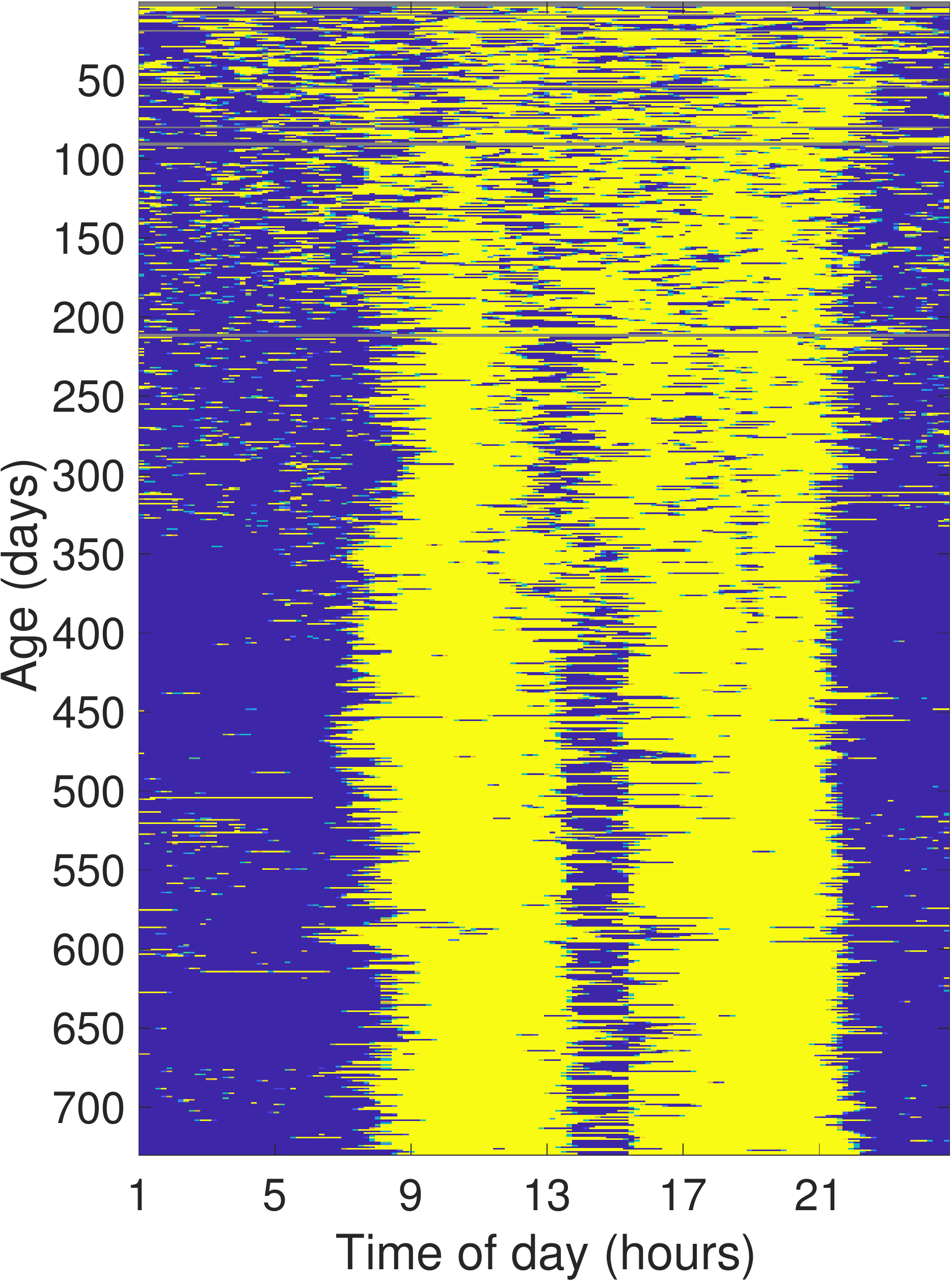} &  \includegraphics[scale=0.12]{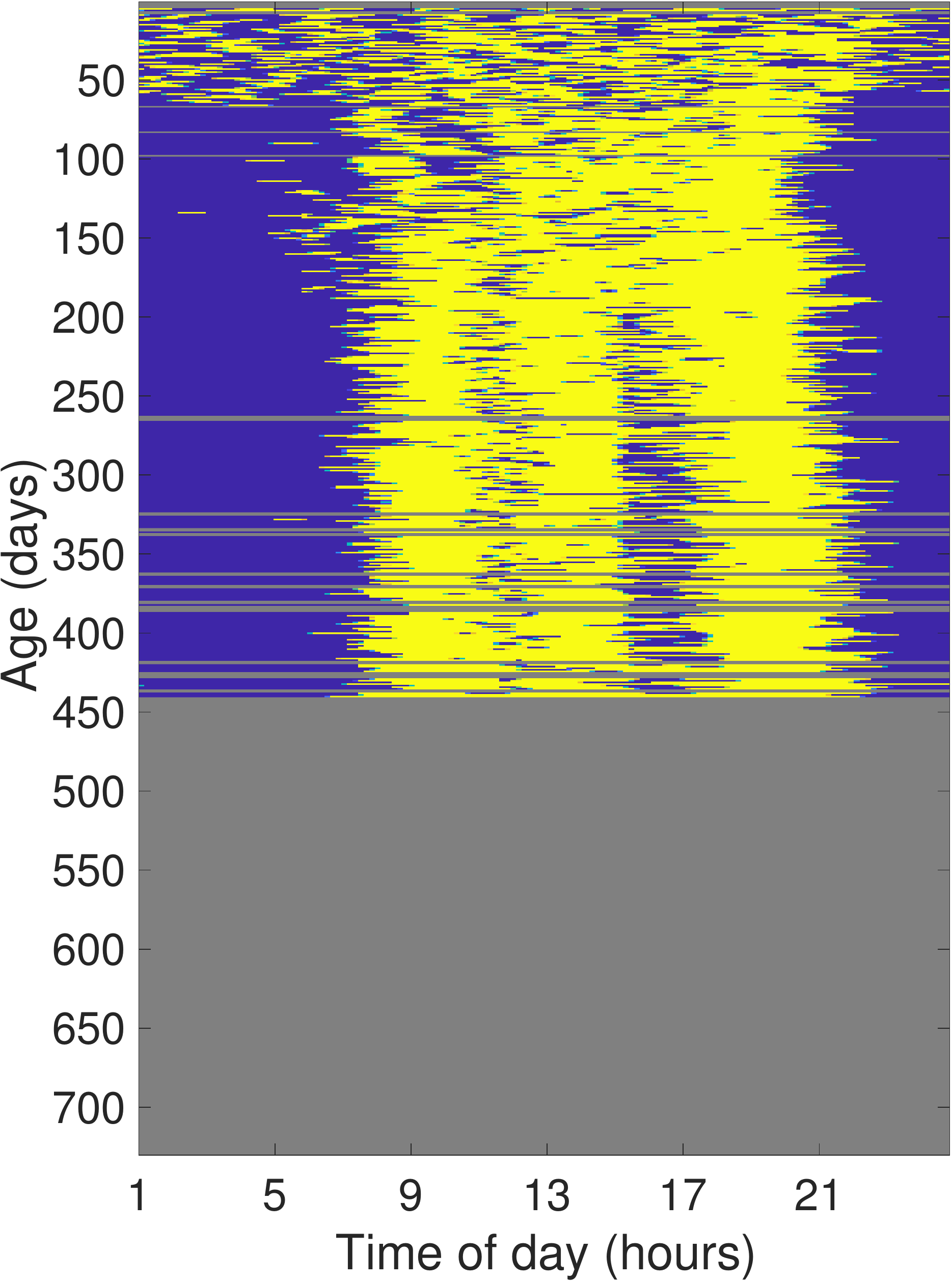}&  \includegraphics[scale=0.12]{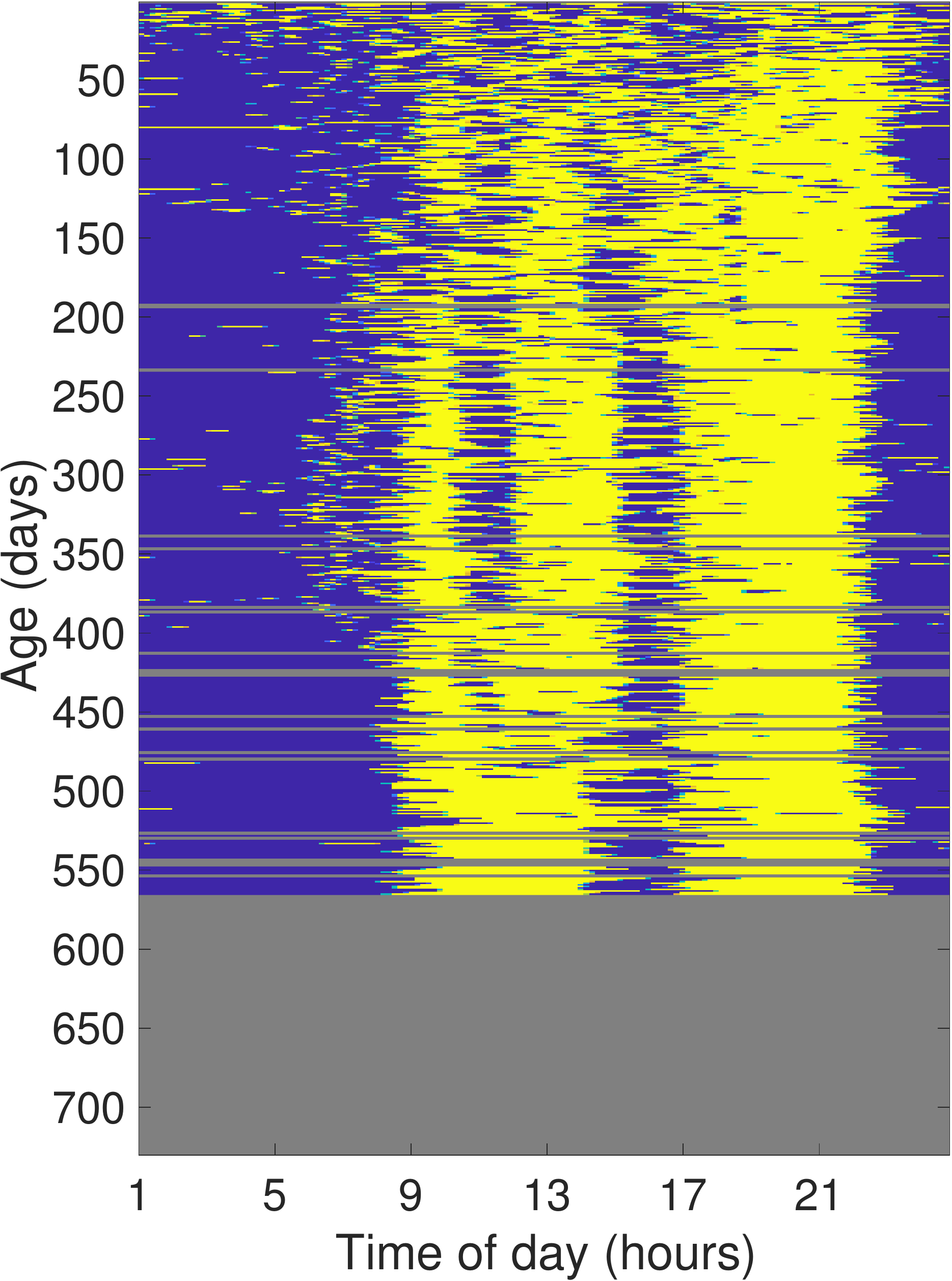} \\
\includegraphics[scale=0.125]{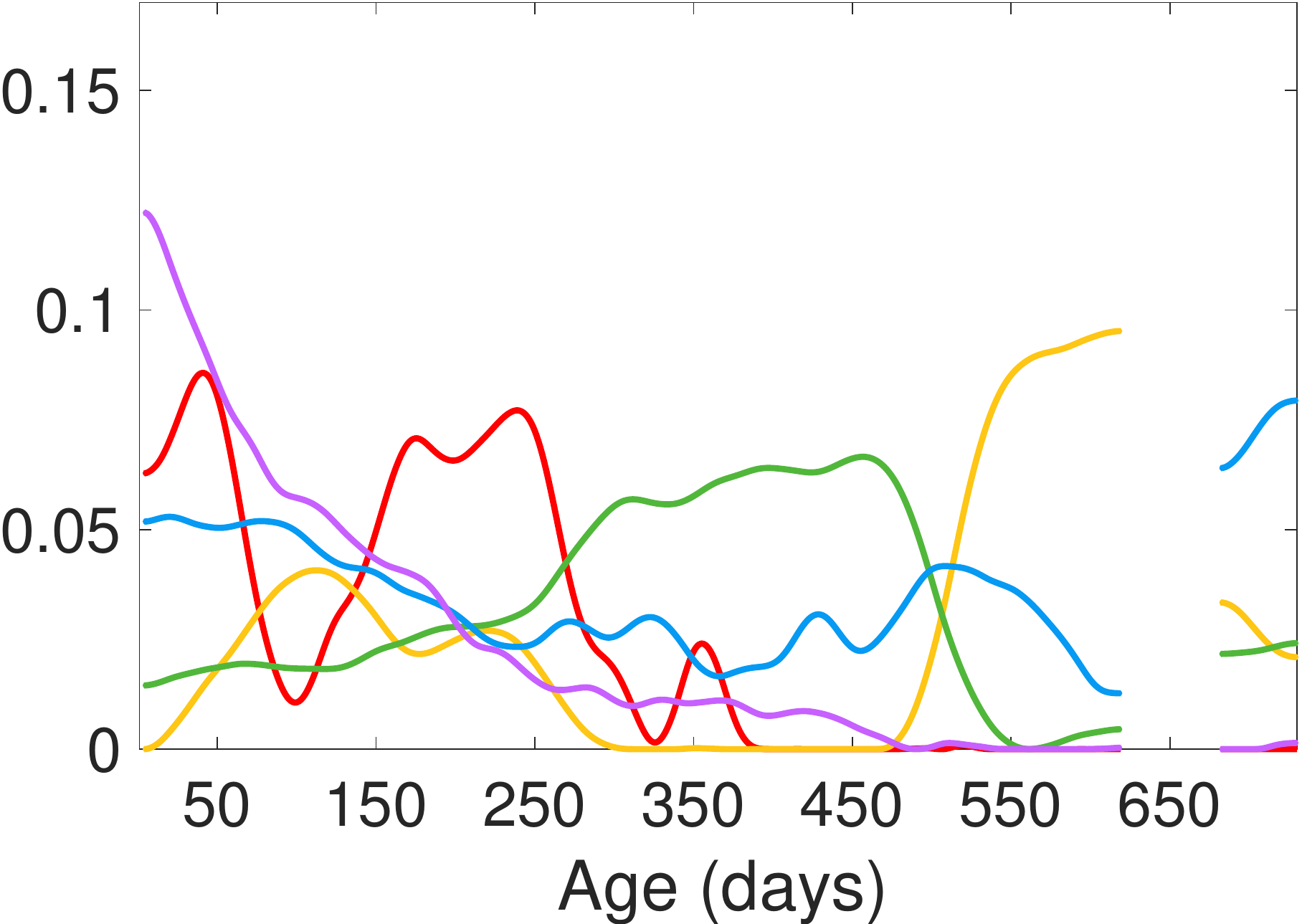} & \includegraphics[scale=0.125]{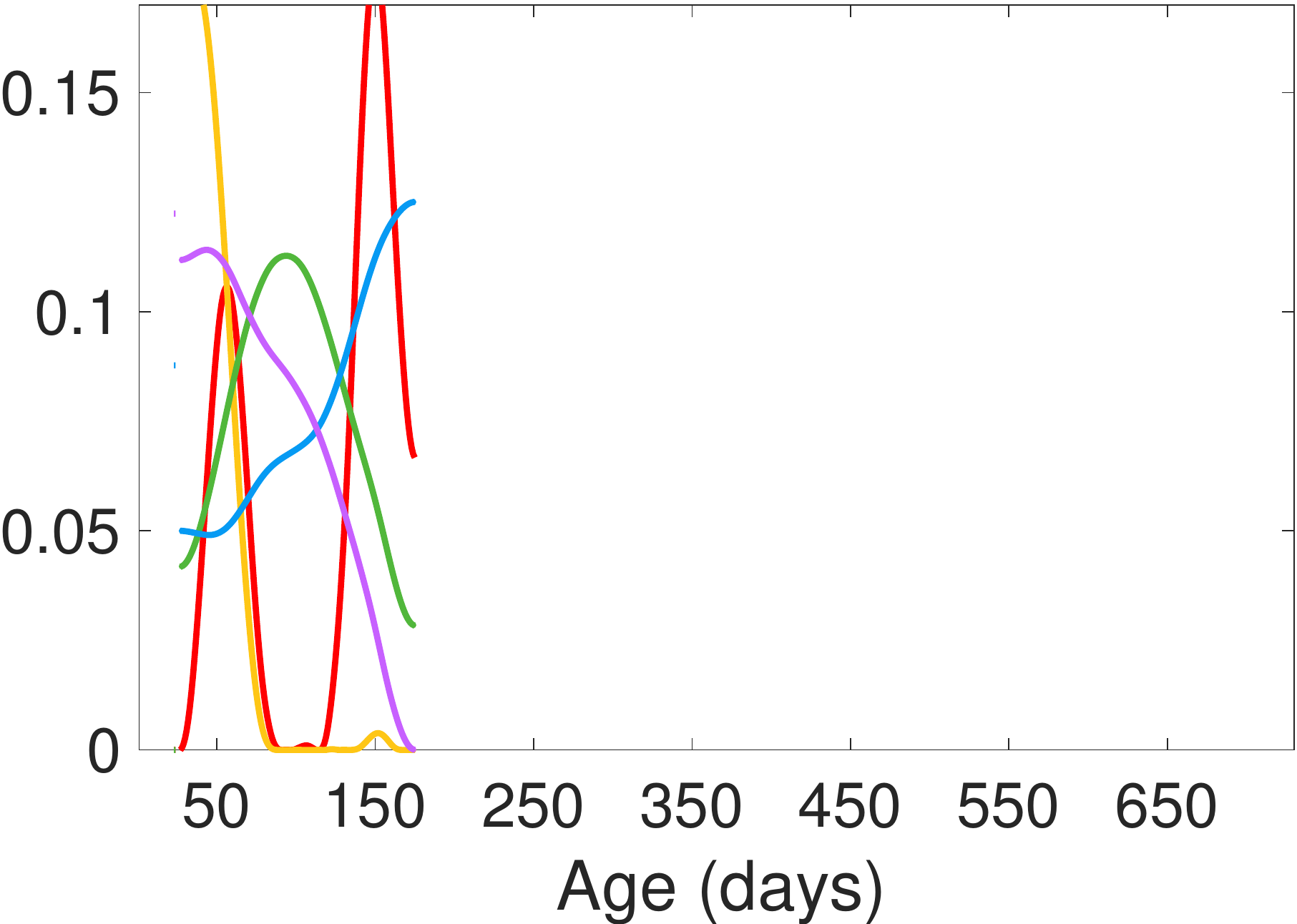} &  \includegraphics[scale=0.125]{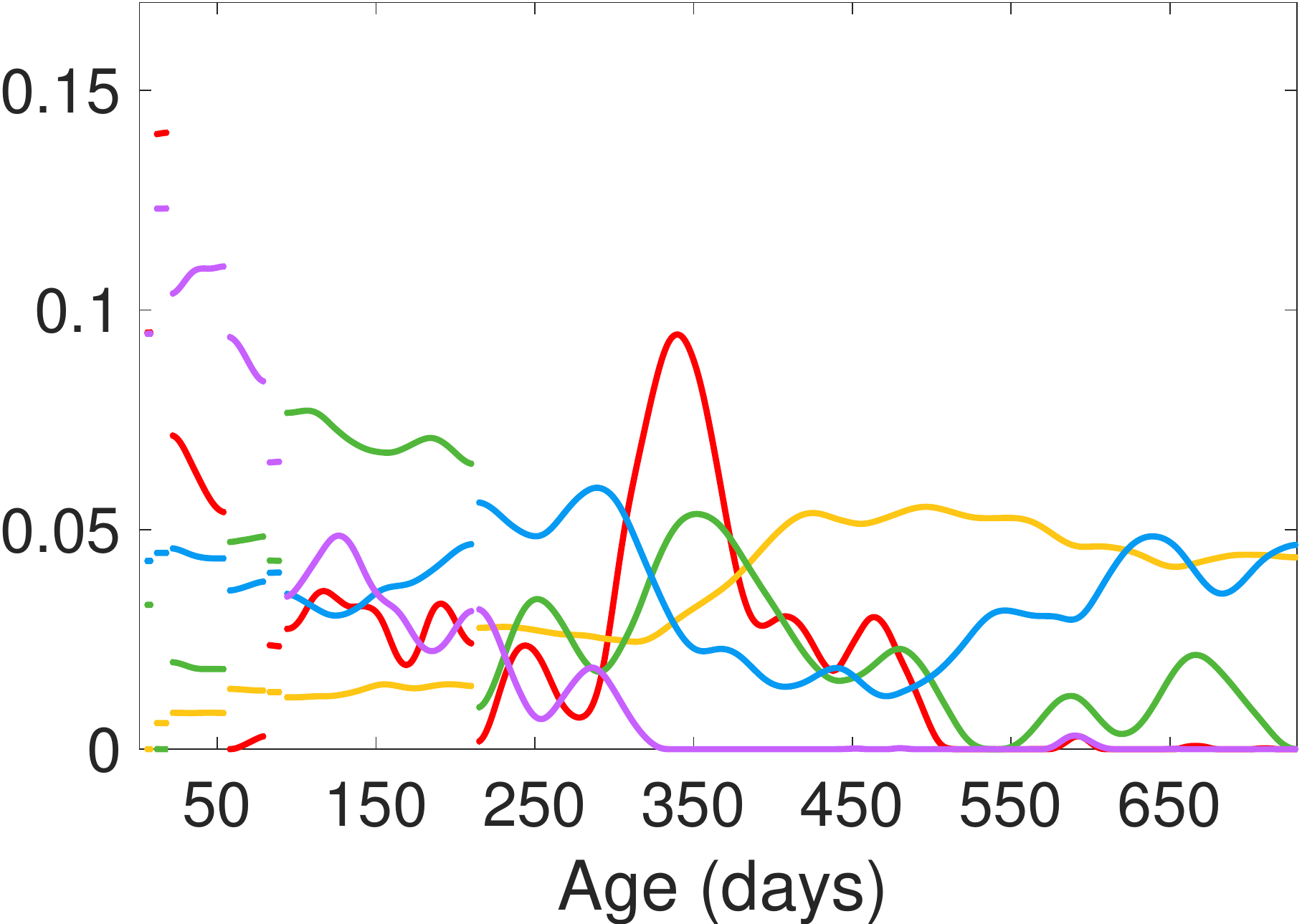} &  \includegraphics[scale=0.125]{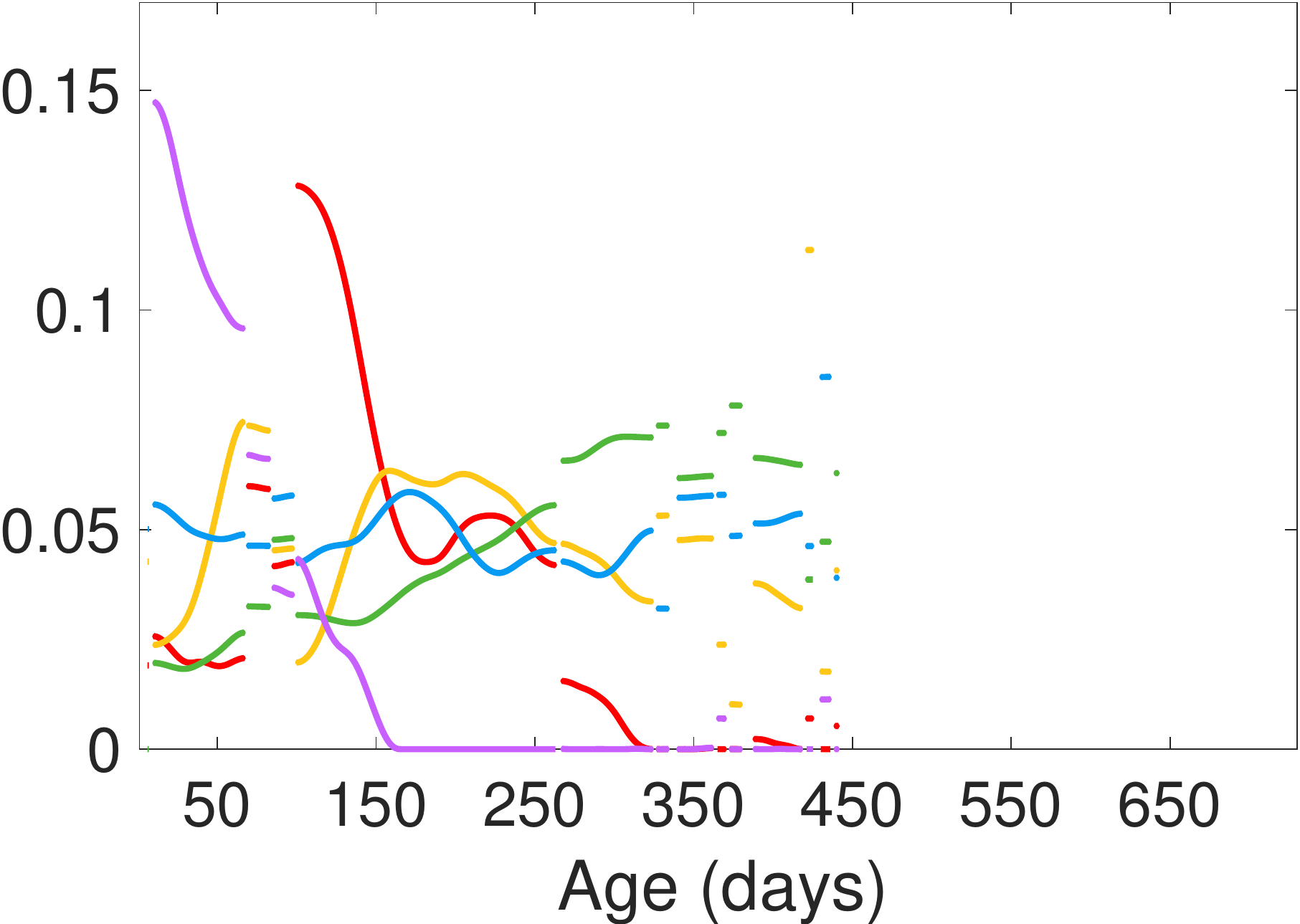}&  \includegraphics[scale=0.125]{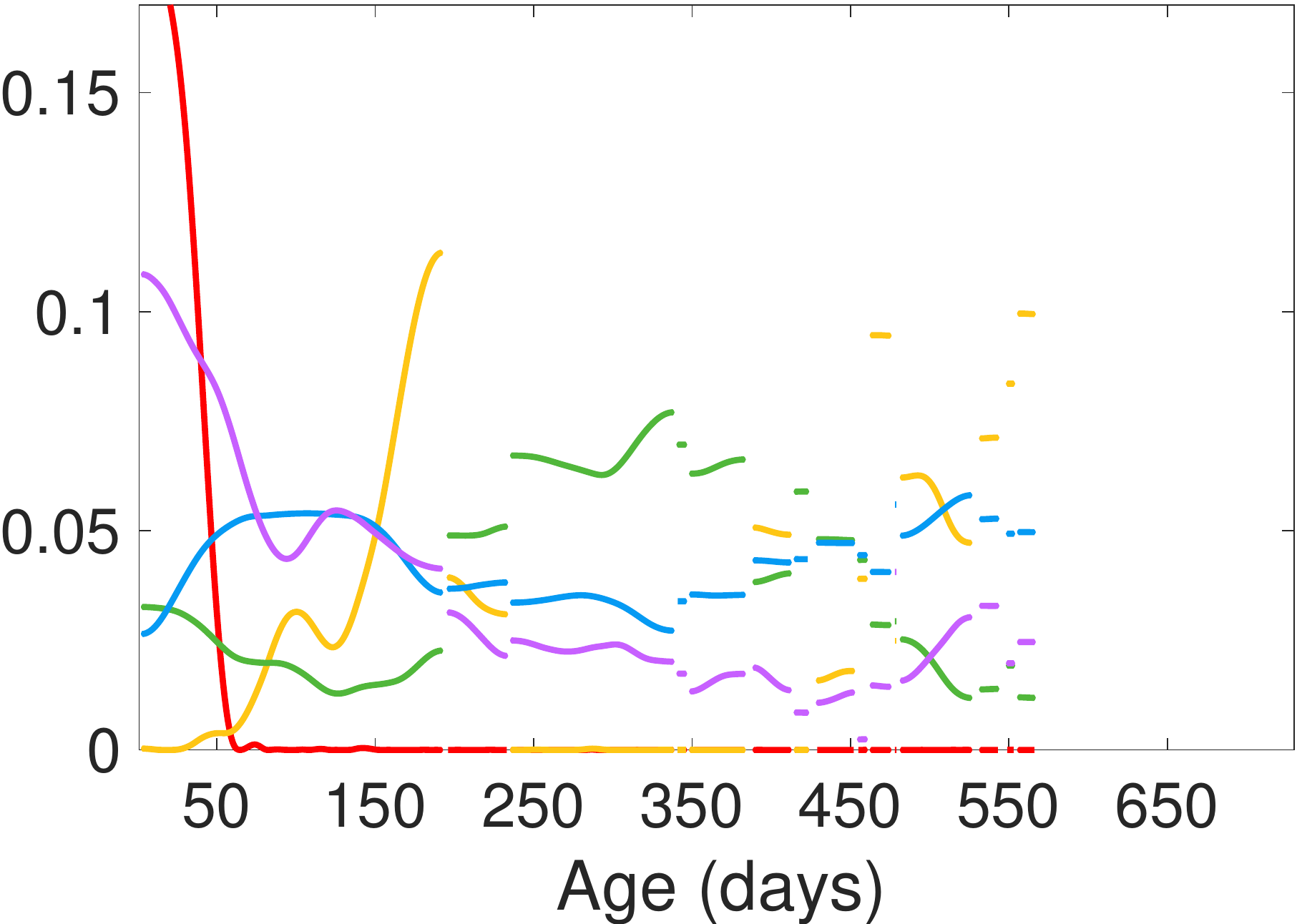} \\

 \end{tabular}
 \caption{Raw data from infants randomly selected from the dataset. Yellow indicates that the infant is awake, and blue that they are asleep. Gray values indicate missing data. Below each data matrix we show the coefficients from the NMF-TS representation.}
 \label{fig:random_samples}
\end{figure}

\begin{figure}[tp]
\begin{center}
\includegraphics[scale=0.35]{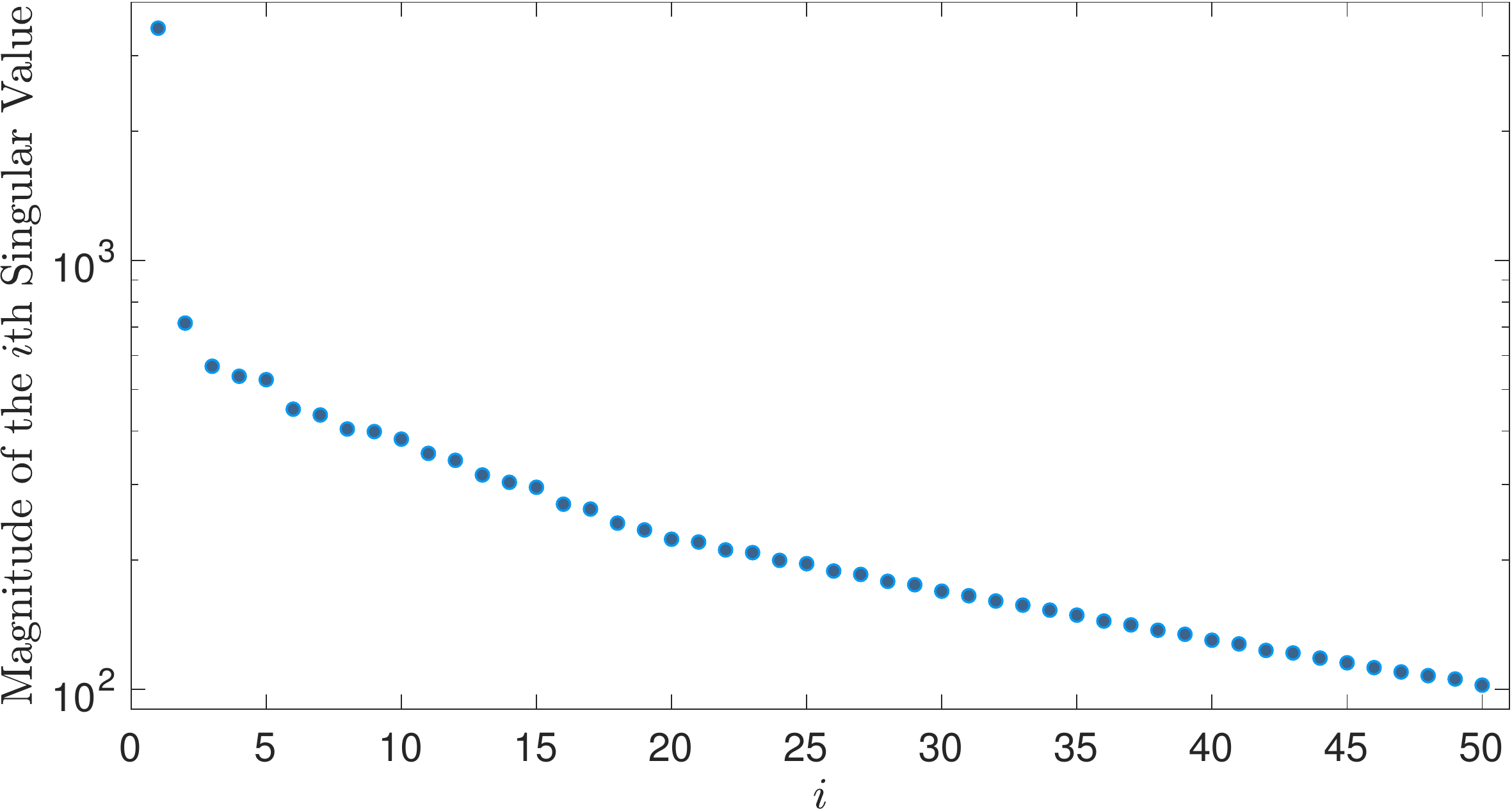}
 \end{center}
 \caption{First 50 singular values obtained by computing the singular-value decomposition of the infant-sleep dataset.}
 \label{fig:svs}
\end{figure}

\begin{figure}[hp]
\hspace{-1.2cm}
\begin{tabular}{ >{\centering\arraybackslash}m{0.07\linewidth} >{\centering\arraybackslash}m{0.2250\linewidth} >{\centering\arraybackslash}m{0.2250\linewidth} >{\centering\arraybackslash}m{0.2250\linewidth} >{\centering\arraybackslash}m{0.2250\linewidth}}
 \vspace{0.35cm} &  Rank 3\vspace{0.35cm} & Rank 4 \vspace{0.2cm} &  Rank 5 \vspace{0.2cm}&  Rank 6 \vspace{0.2cm}\\
PCA & \includegraphics[scale=0.15]{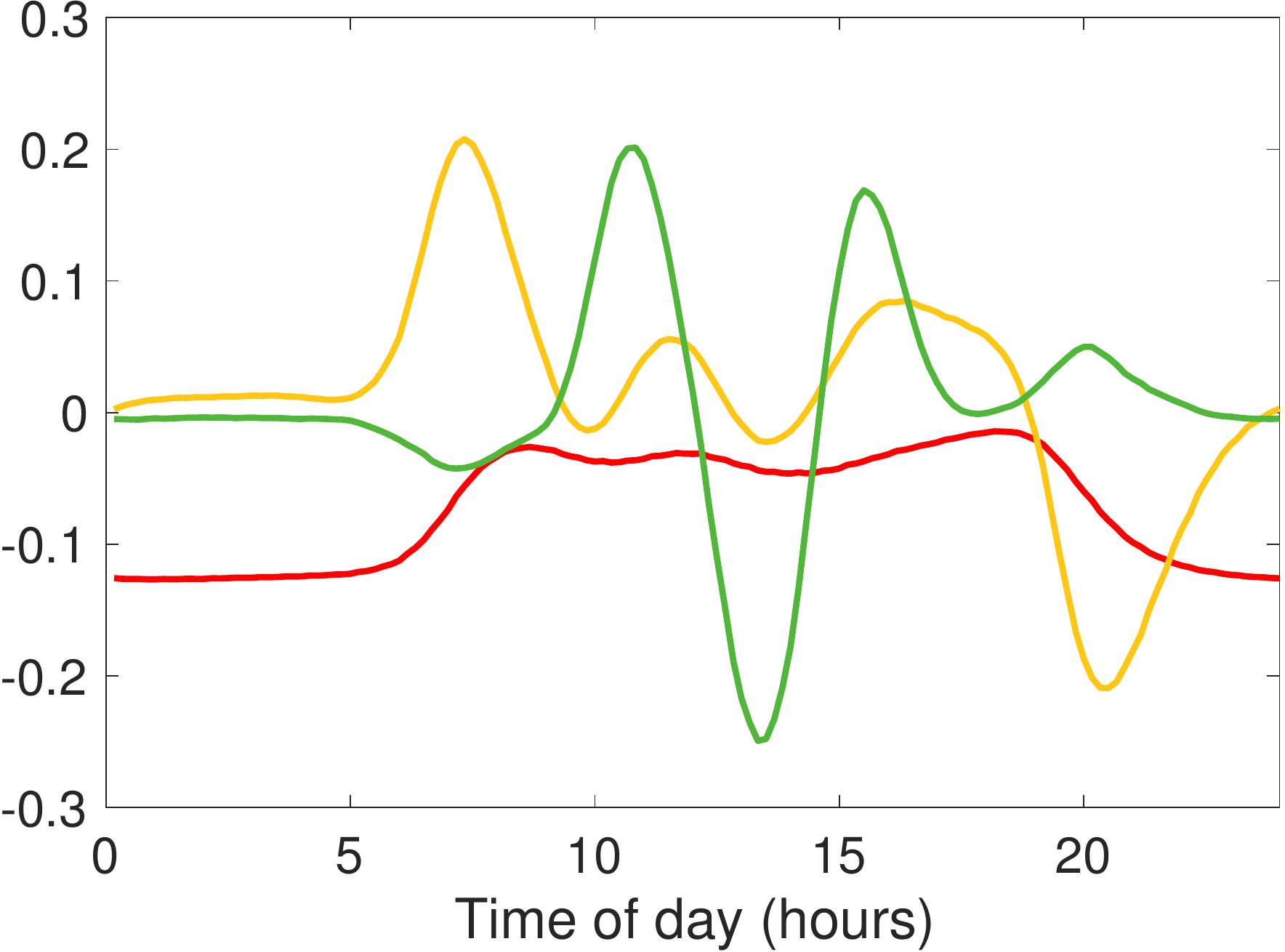} & \includegraphics[scale=0.15]{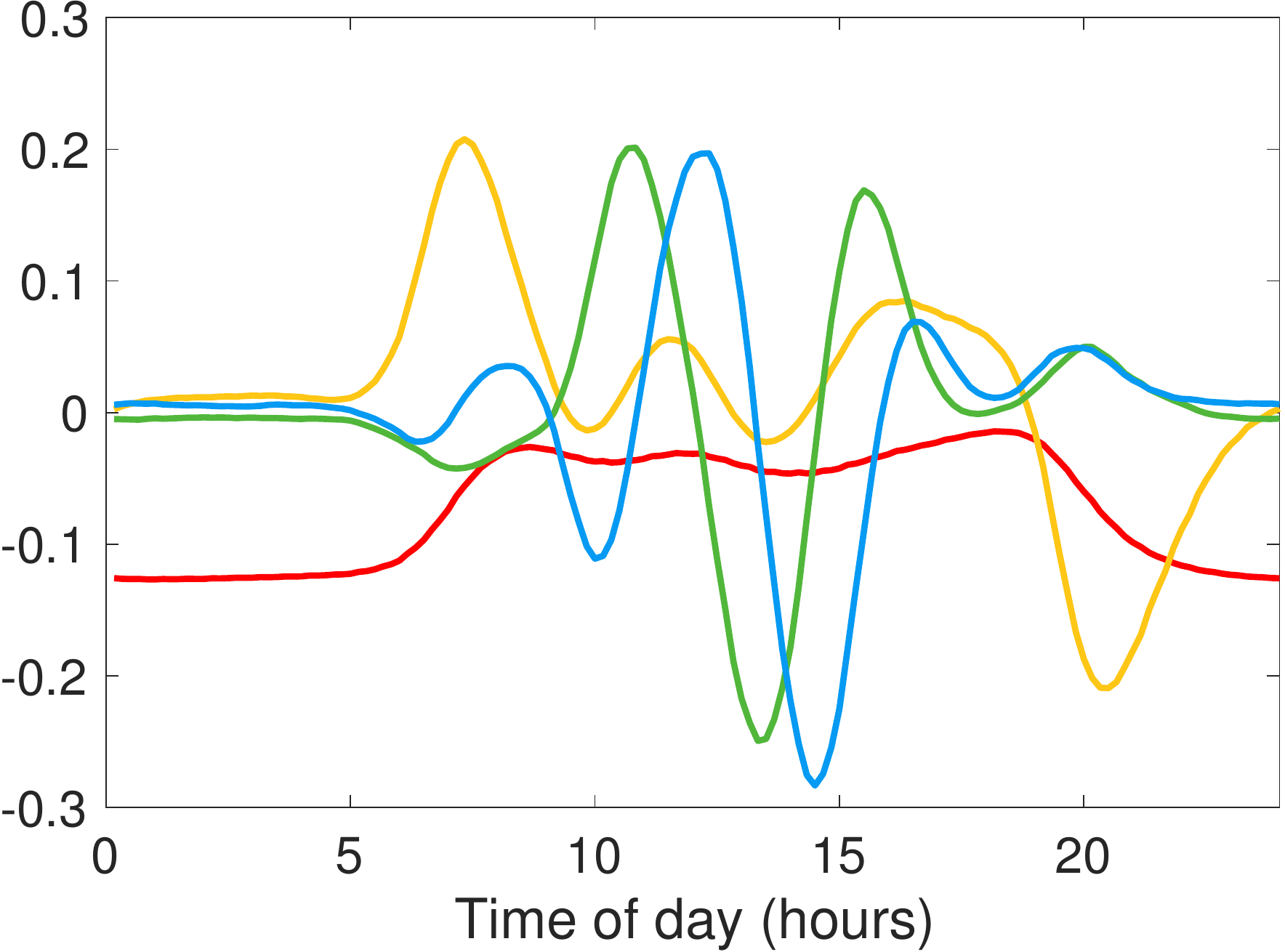} &  \includegraphics[scale=0.15]{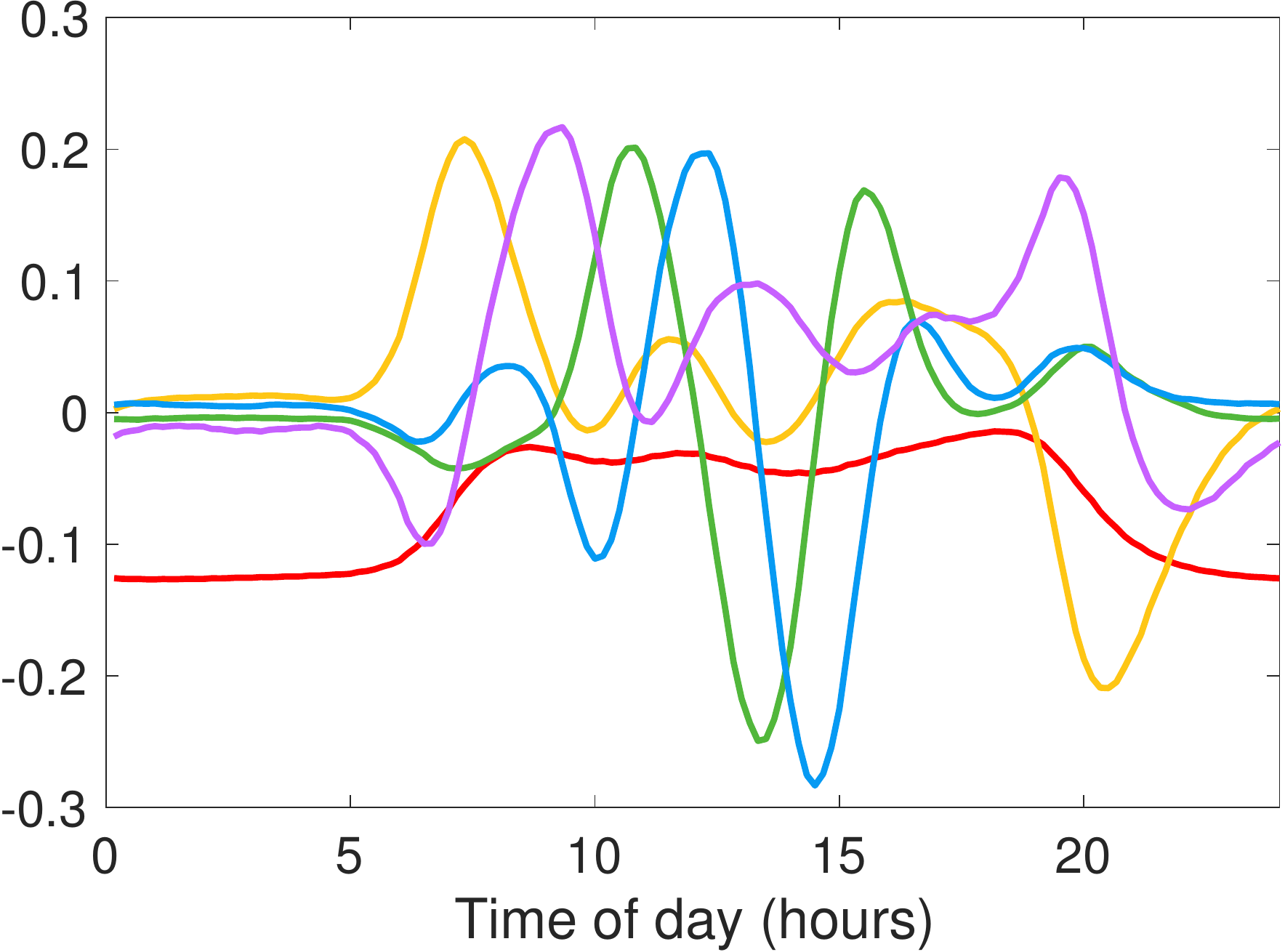} &  \includegraphics[scale=0.15]{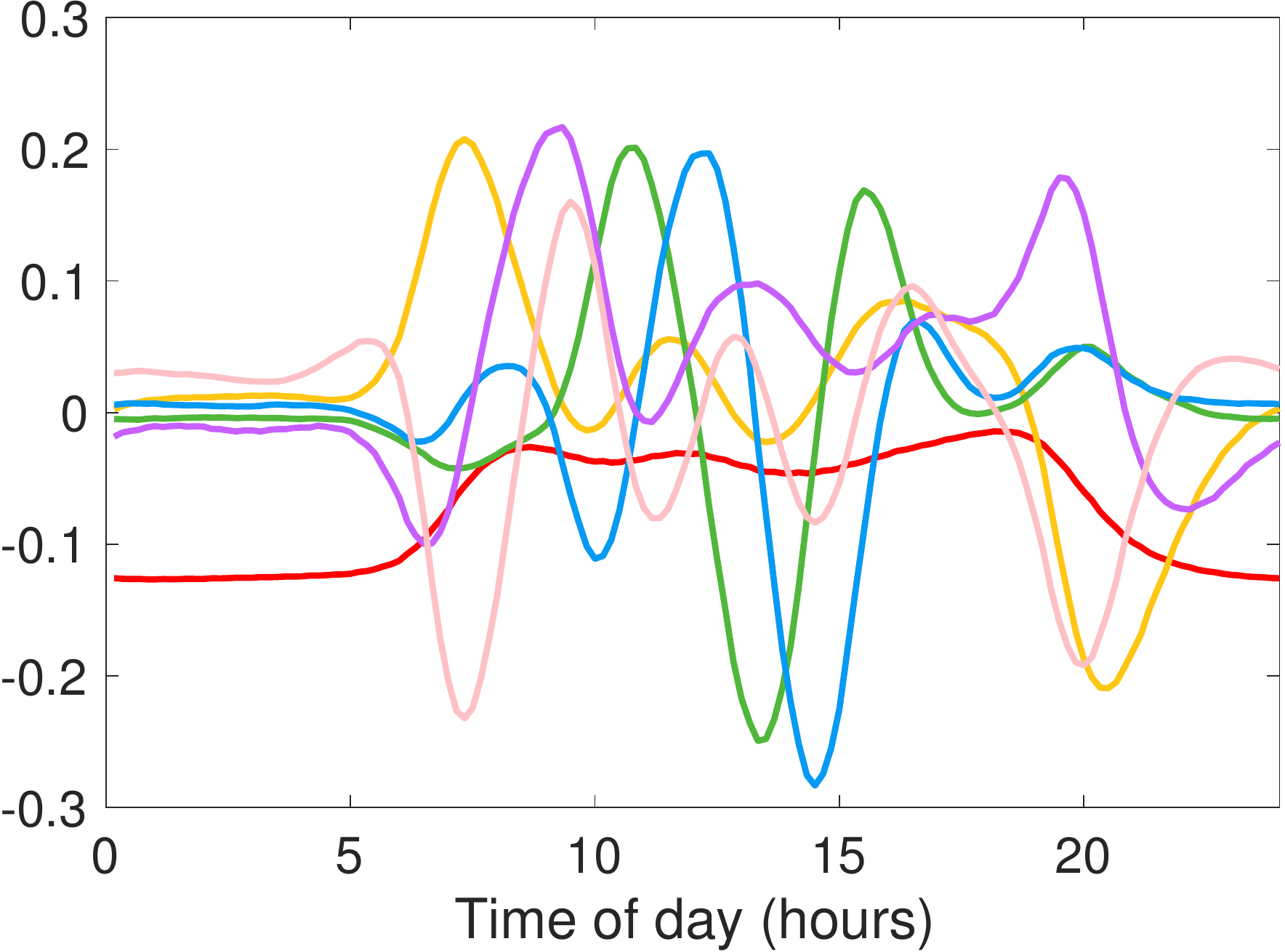}\\
NMF & \includegraphics[scale=0.15]{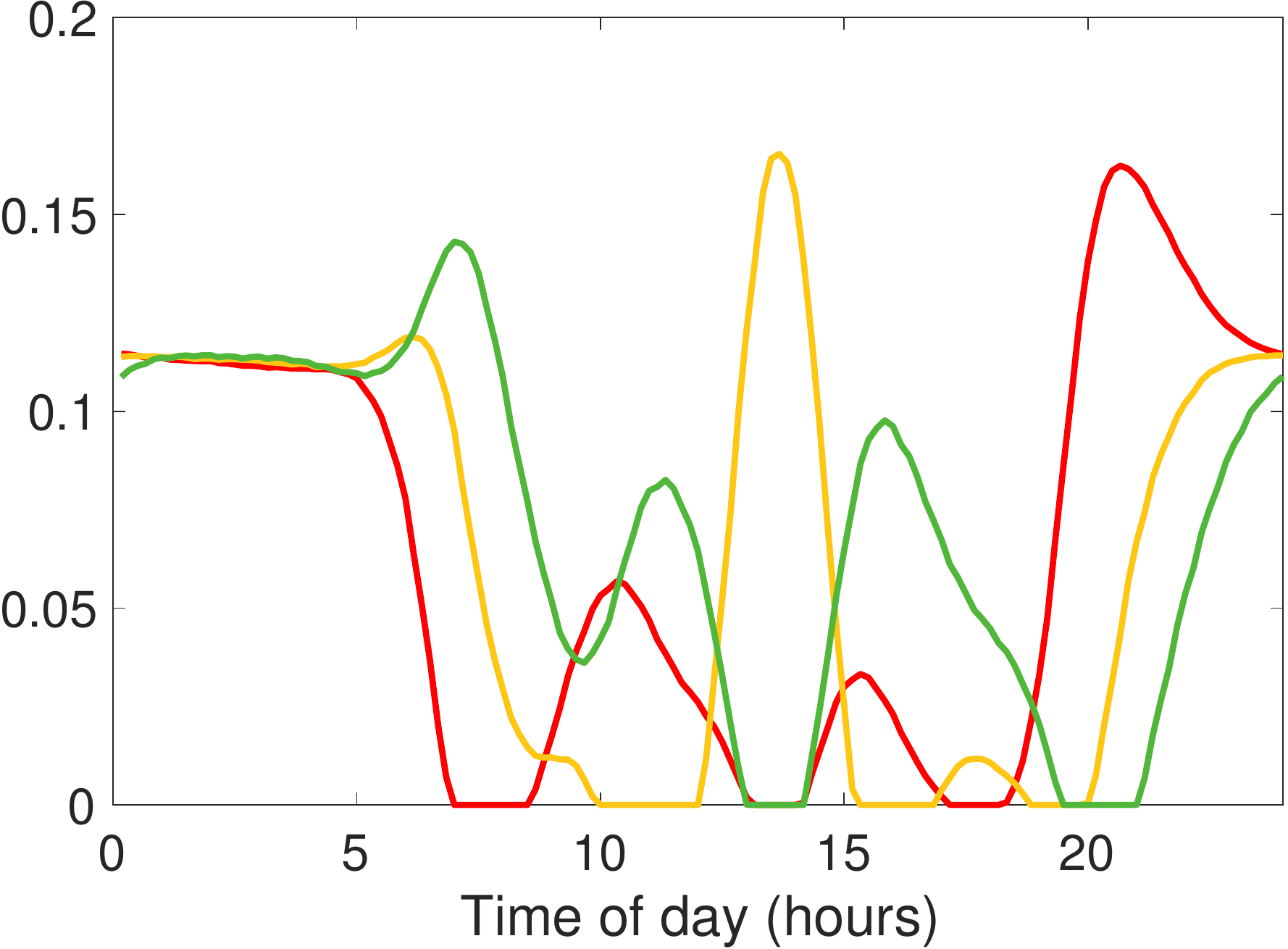} & \includegraphics[scale=0.15]{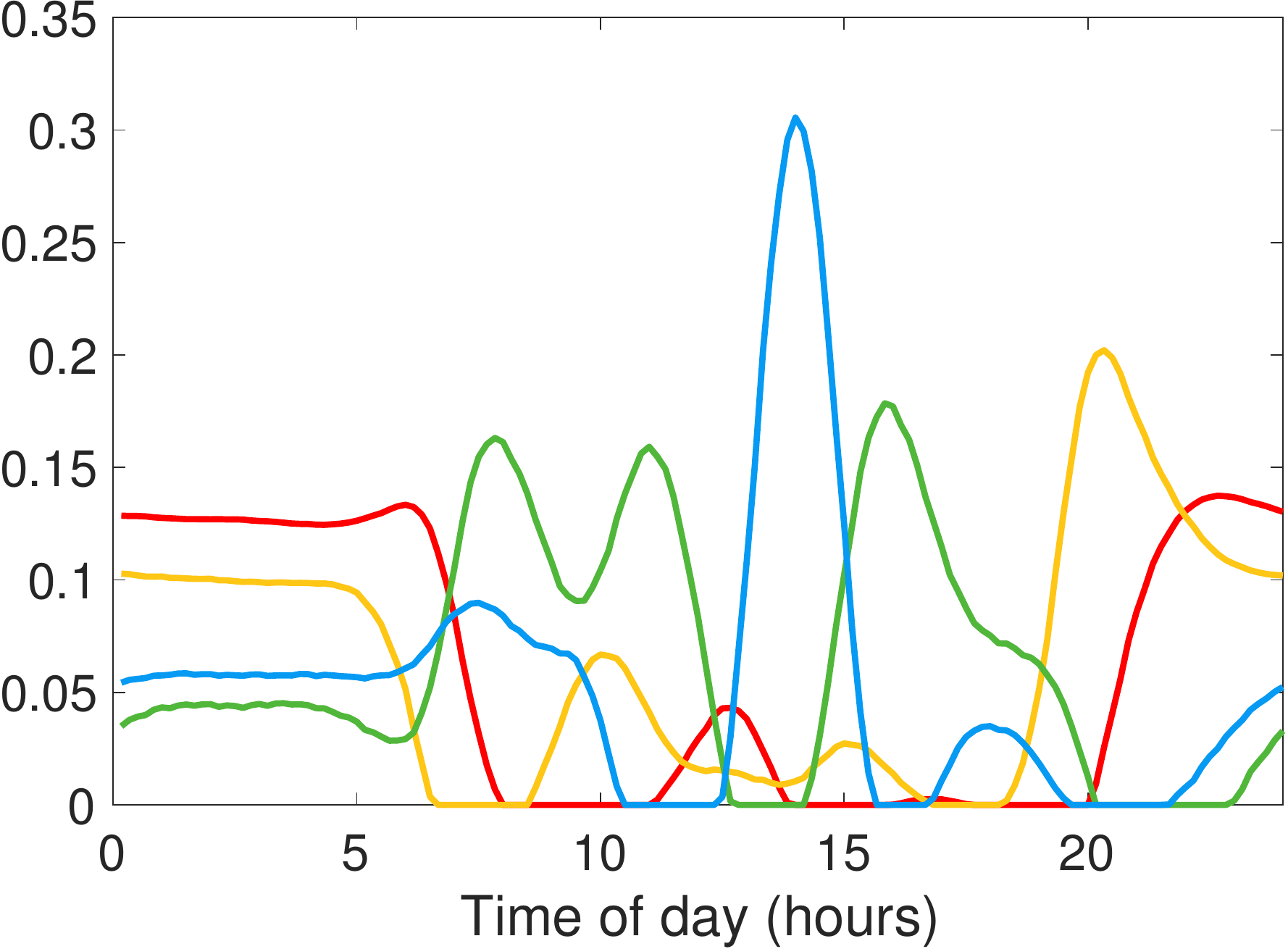} &  \includegraphics[scale=0.15]{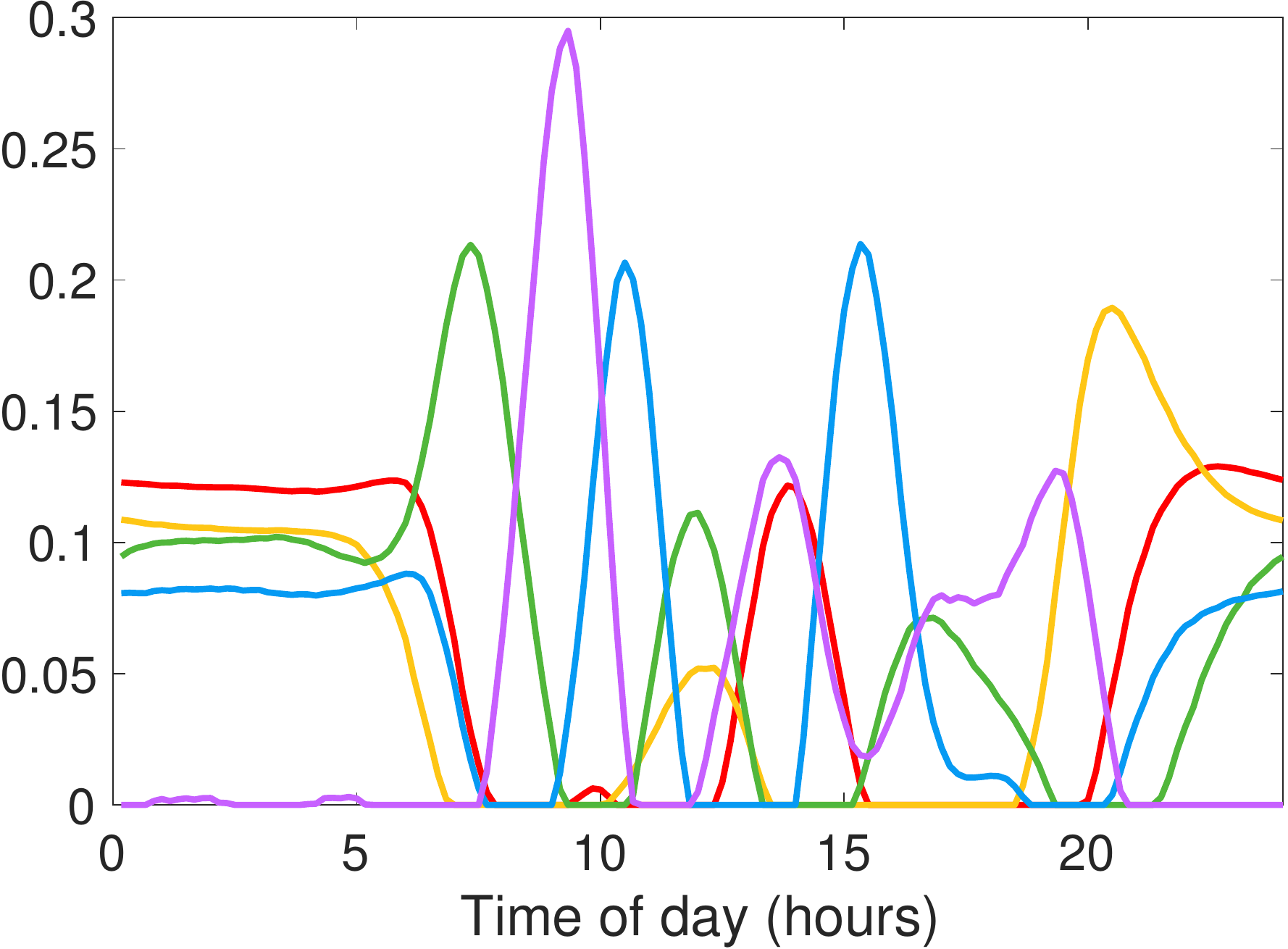} & \includegraphics[scale=0.15]{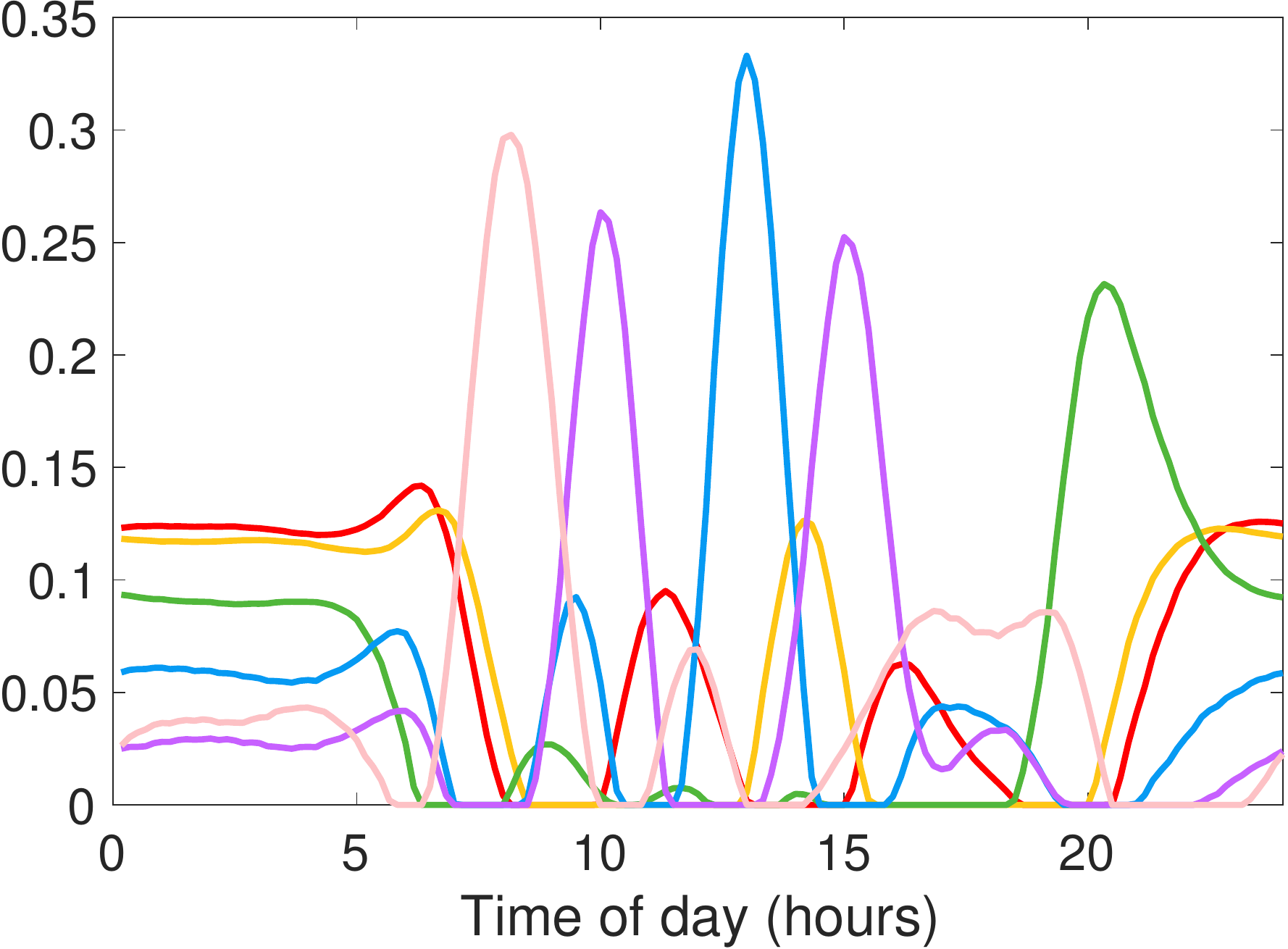}\\
NMF-TS & \includegraphics[scale=0.15]{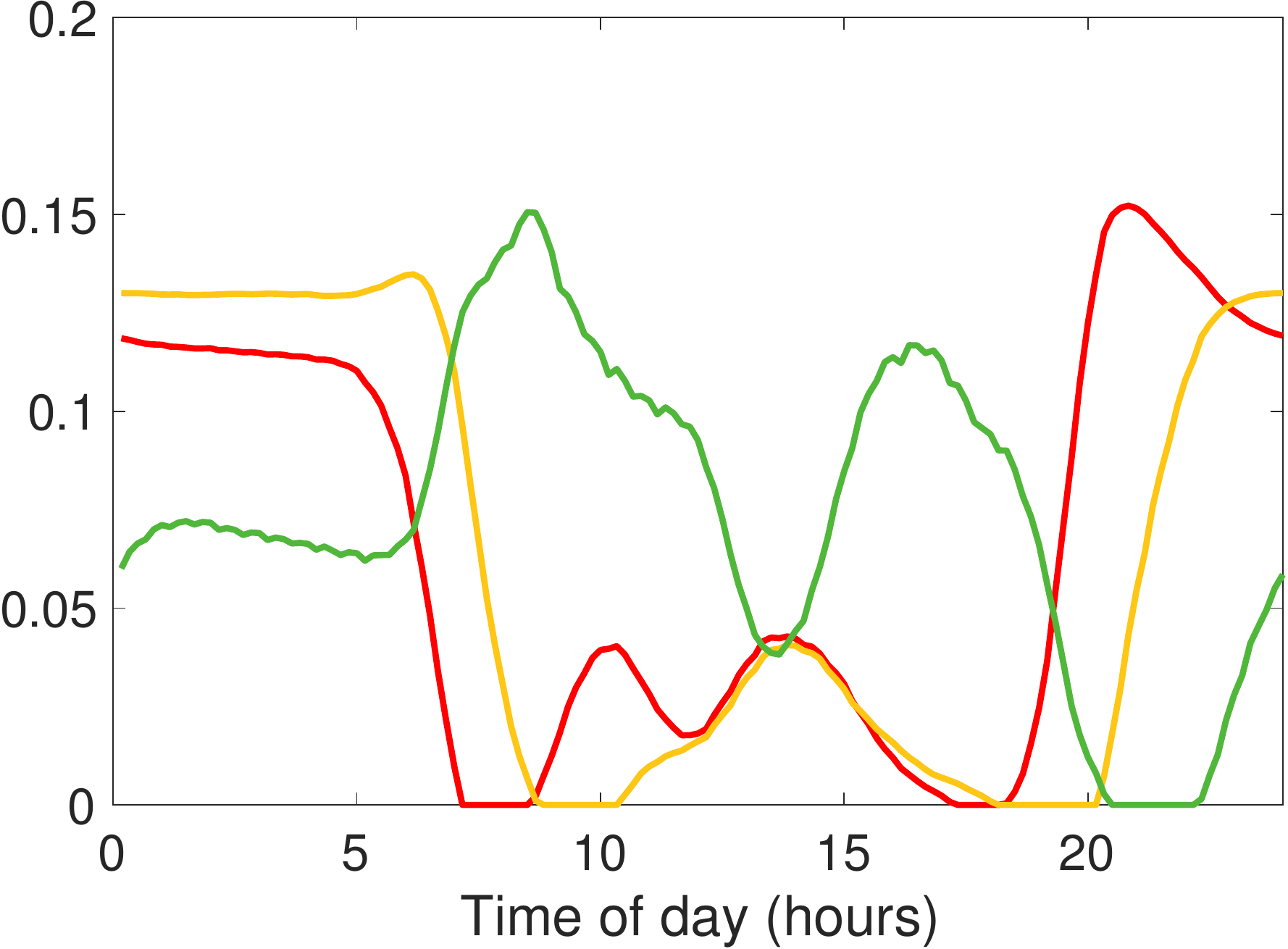} & \includegraphics[scale=0.15]{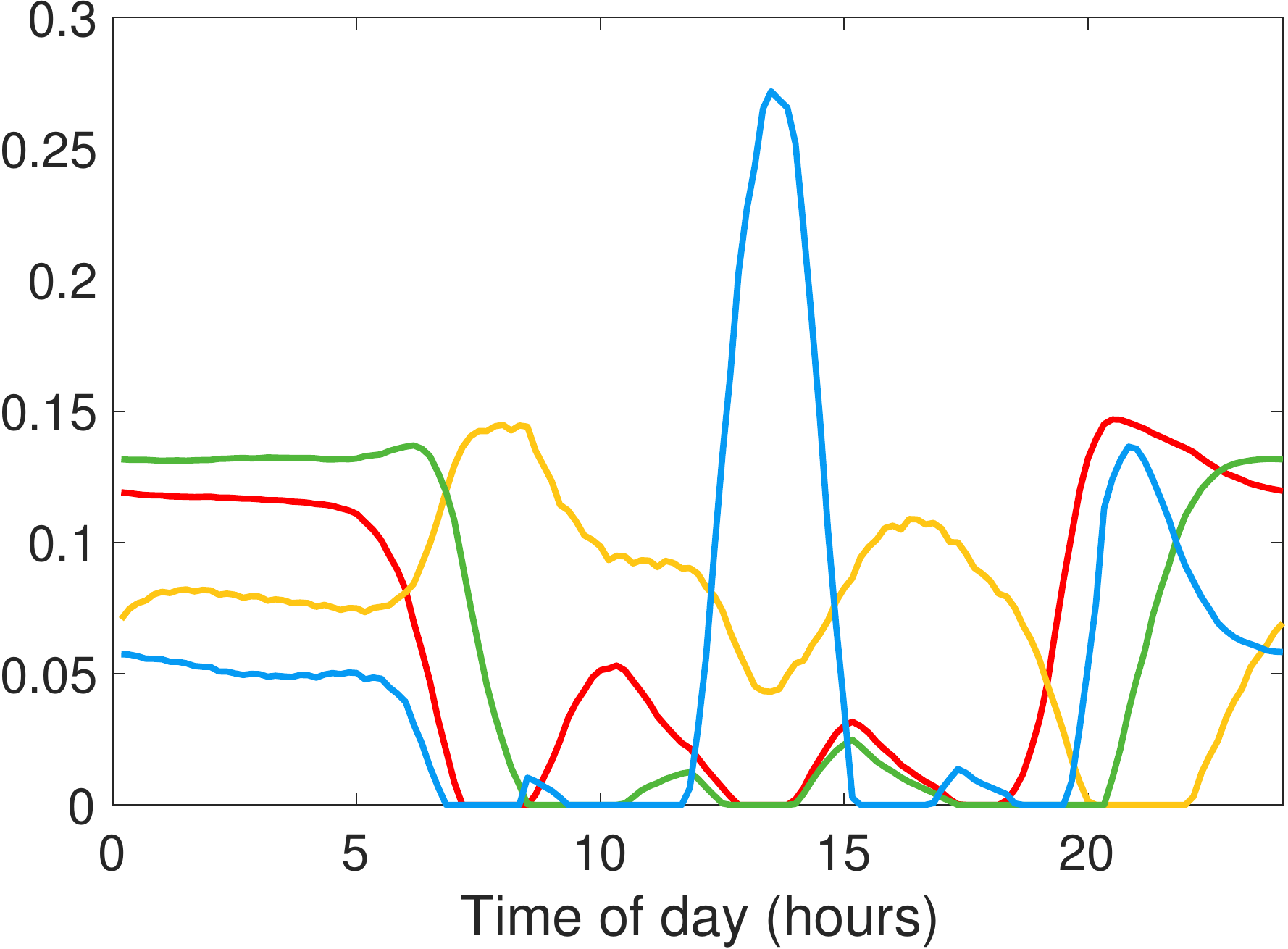} &  \includegraphics[scale=0.15]{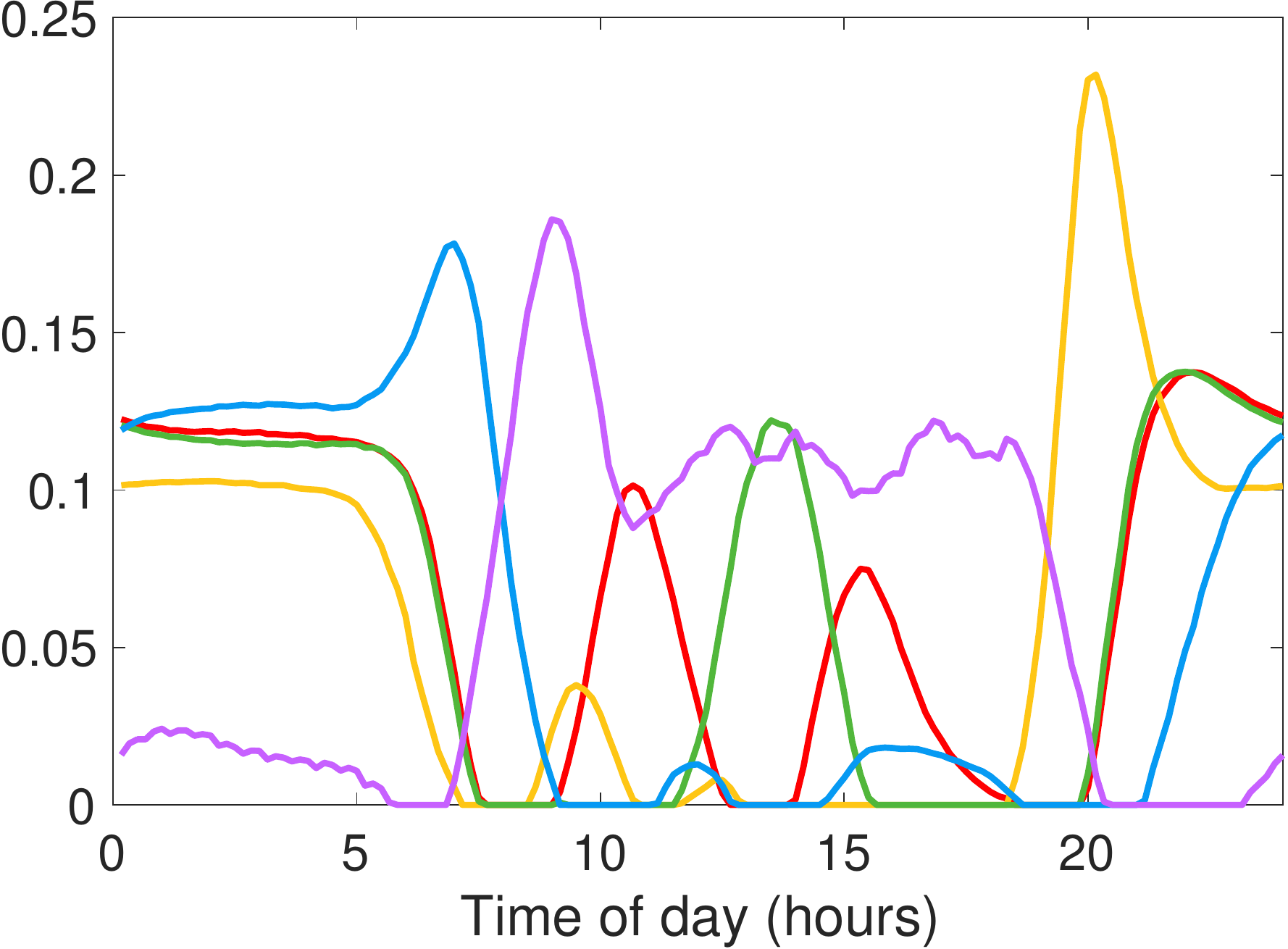} & \includegraphics[scale=0.15]{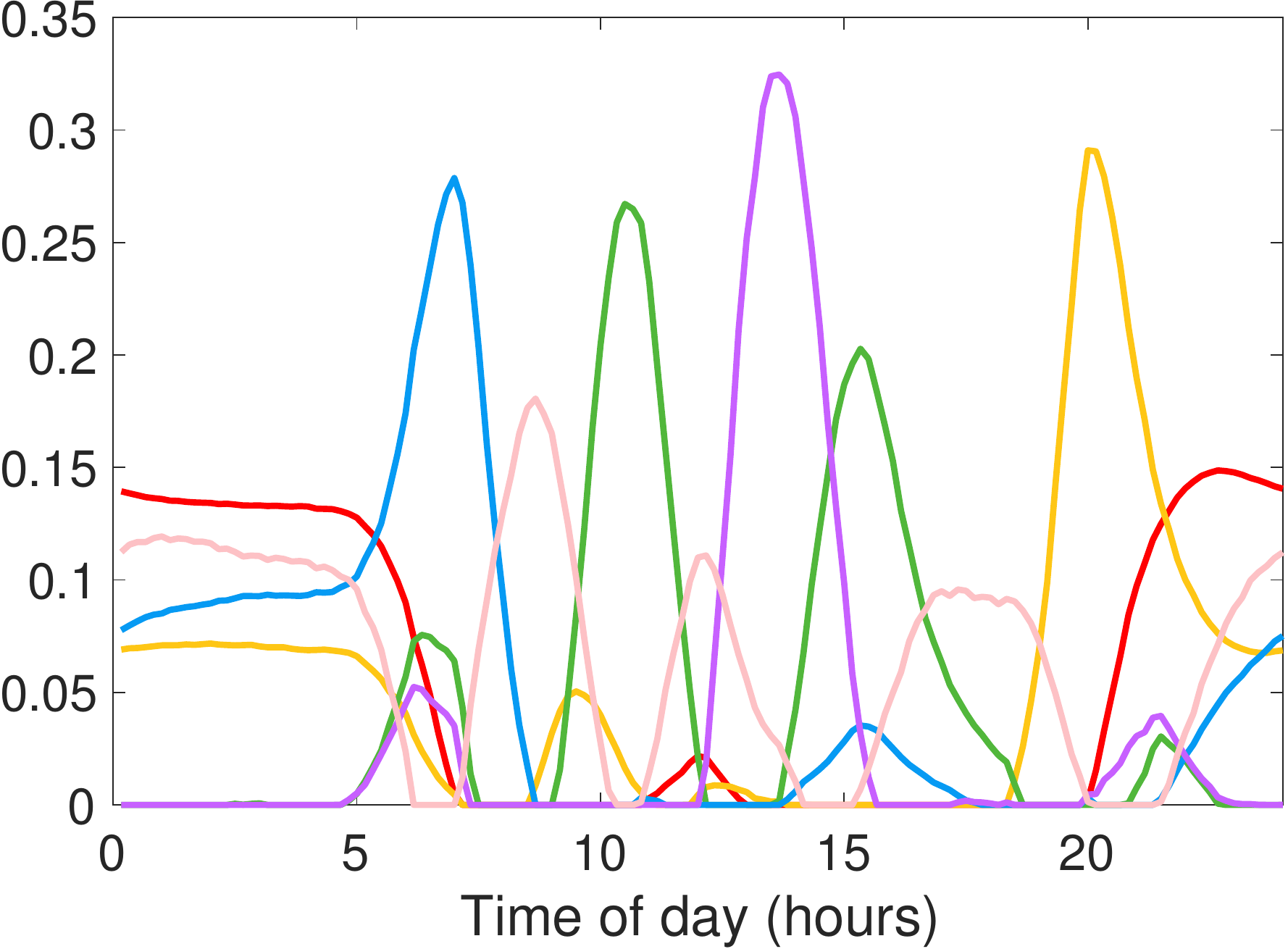}
 \end{tabular}
 \caption{Basis functions extracted from the whole dataset using PCA (top), NMF (center) and NMF-TS (bottom) for different values of the rank $r$. For NMF-TS we set $\lambda := 10^5$.}
 \label{fig:rank}
\end{figure}

\begin{figure}[hp]
\hspace{-1.2cm}
\begin{tabular}{ >{\centering\arraybackslash}m{0.09\linewidth} >{\centering\arraybackslash}m{0.22\linewidth} >{\centering\arraybackslash}m{0.22\linewidth} >{\centering\arraybackslash}m{0.22\linewidth} >{\centering\arraybackslash}m{0.22\linewidth}}
 \vspace{0.38cm} & $F_1$, \ldots, $F_5$ \vspace{0.2cm} & $C_1^{[n_1]}$, \ldots, $C_5^{[n_1]}$ \vspace{0.2cm} & $C_1^{[n_2]}$, \ldots, $C_5^{[n_2]}$ \vspace{0.2cm} & $C_1^{[n_3]}$, \ldots, $C_5^{[n_3]}$\vspace{0.2cm}\\
{\footnotesize NMF} & \includegraphics[scale=0.15]{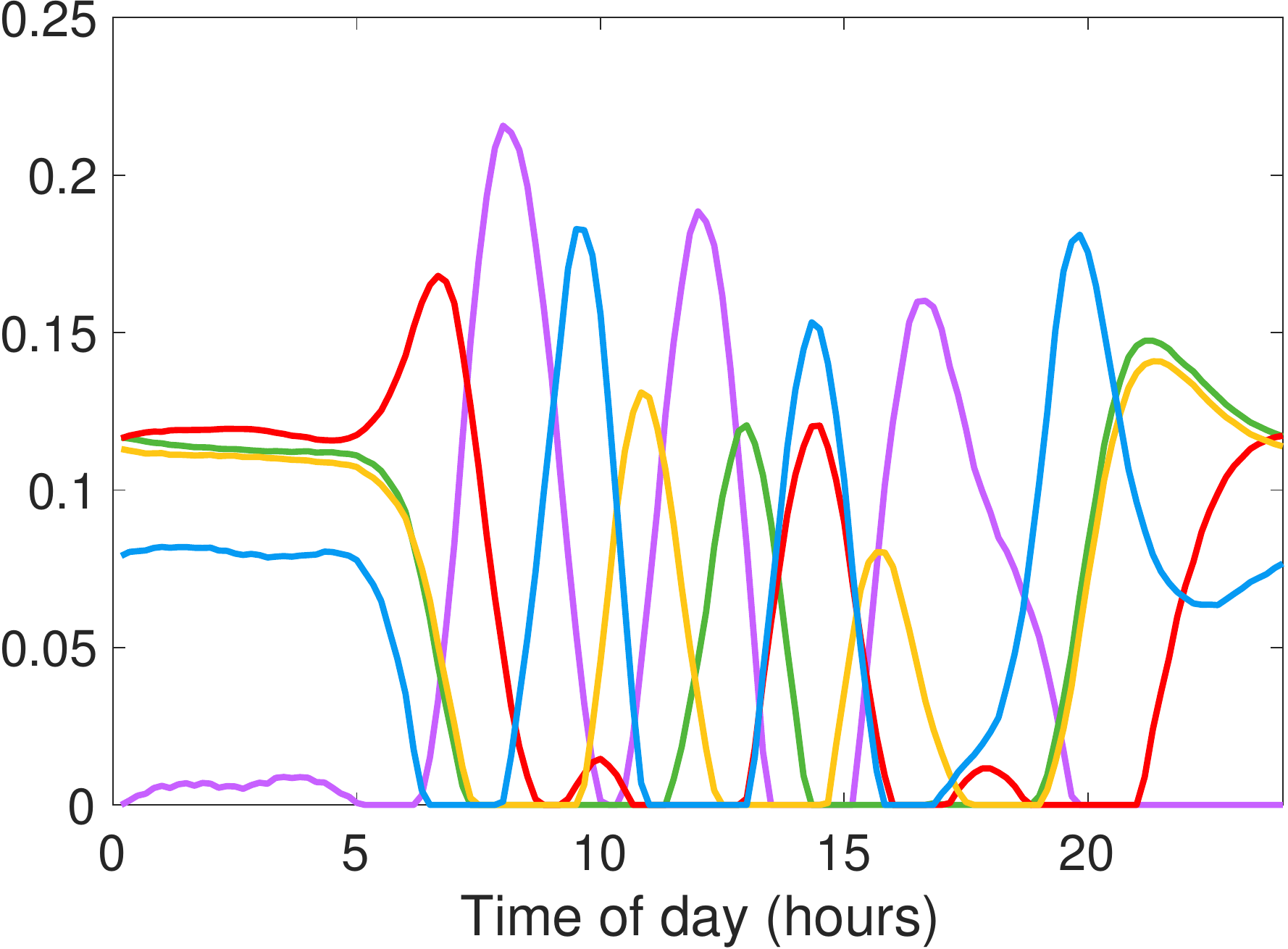} & \includegraphics[scale=0.15]{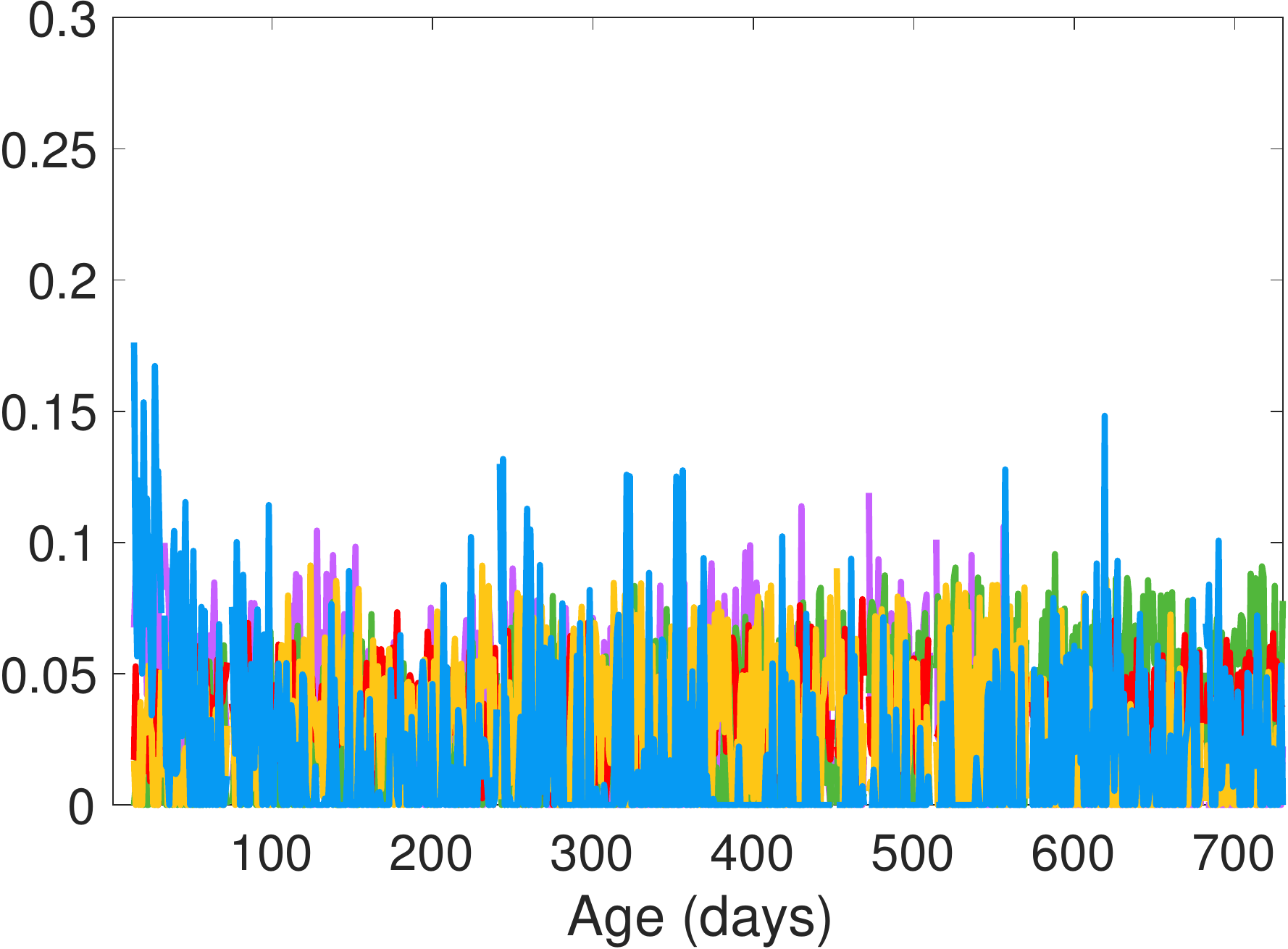} &  \includegraphics[scale=0.15]{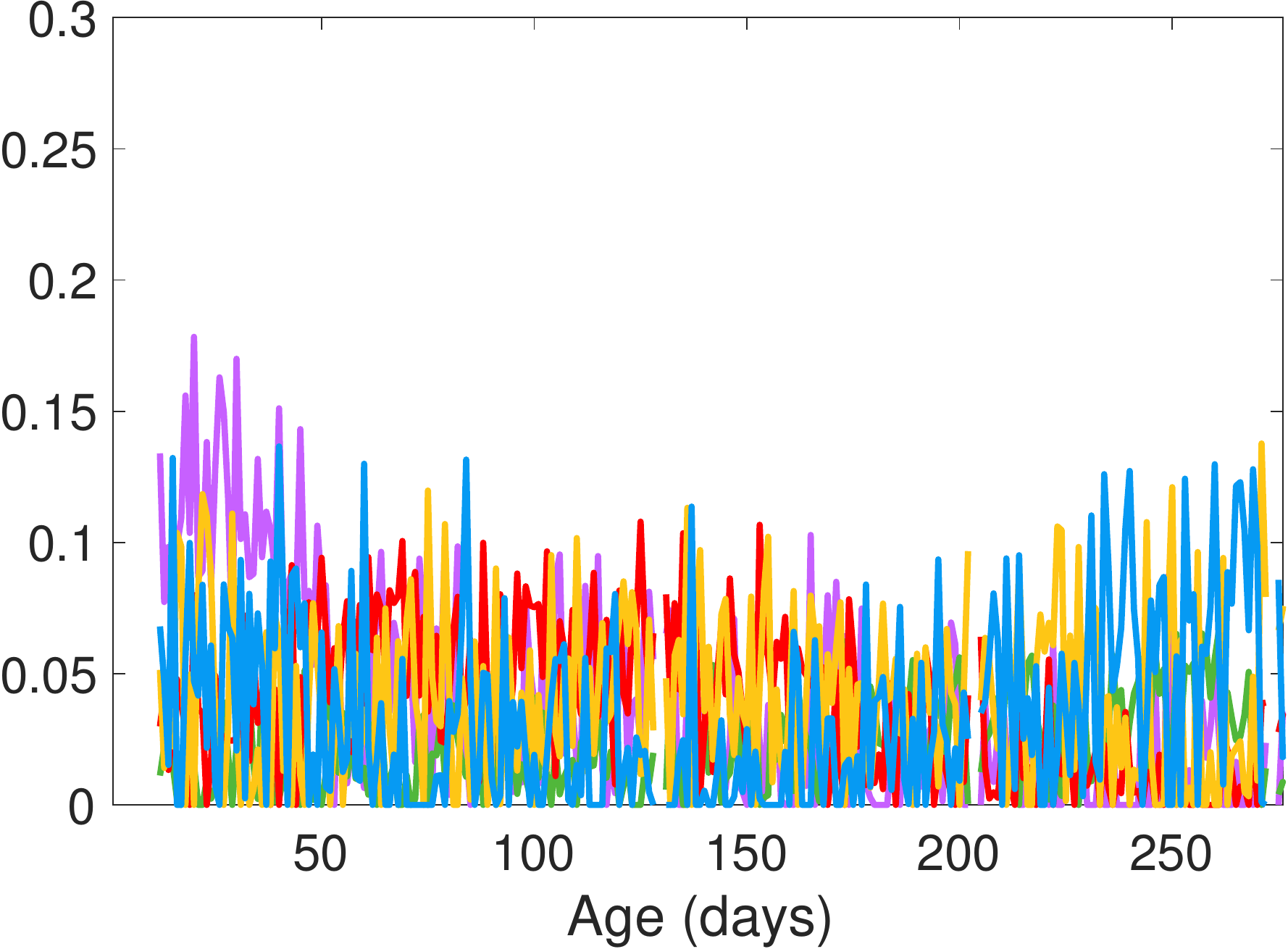} &  \includegraphics[scale=0.15]{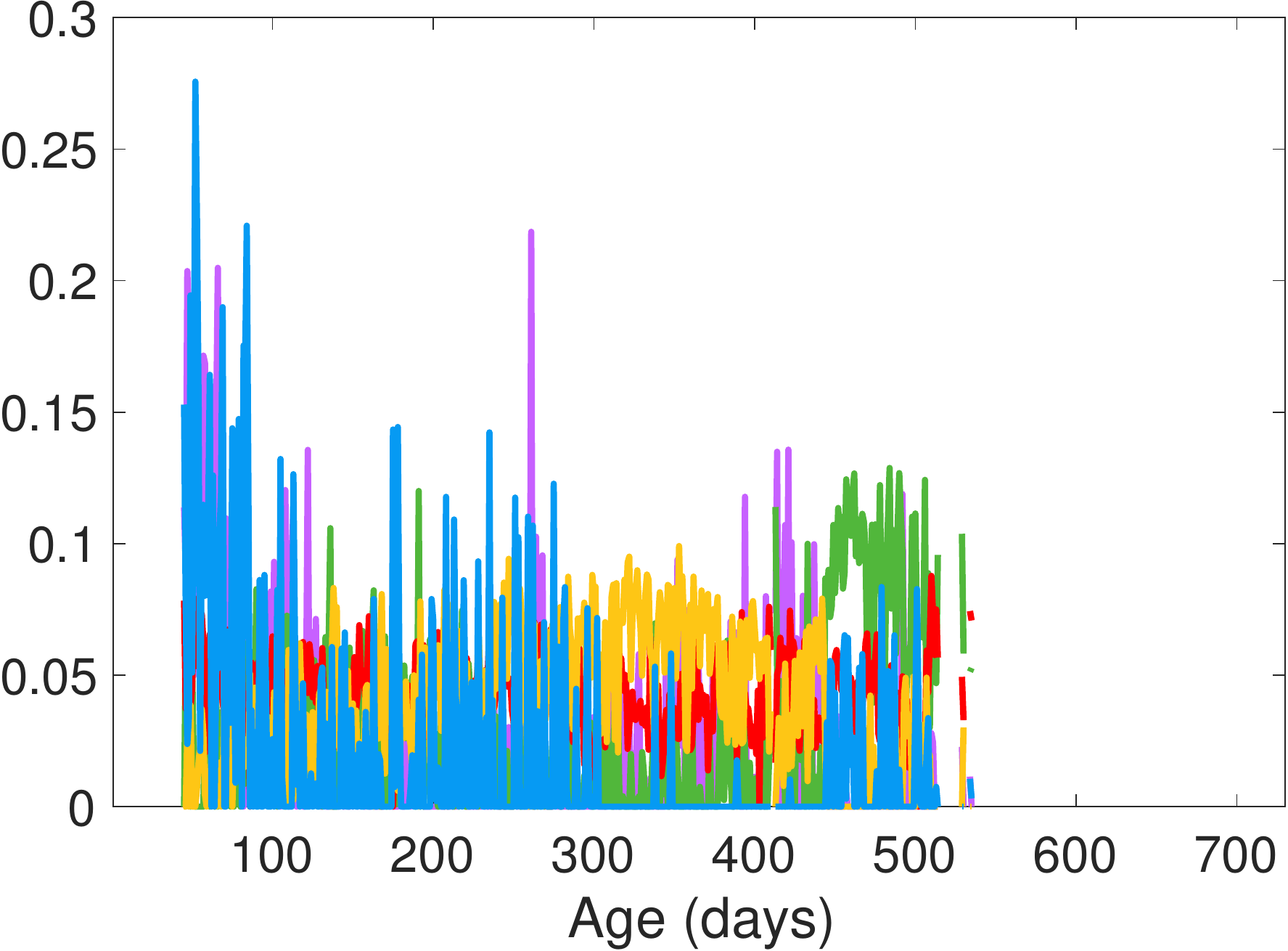}\\
{\footnotesize NMF-TS $\lambda=10$} & \includegraphics[scale=0.15]{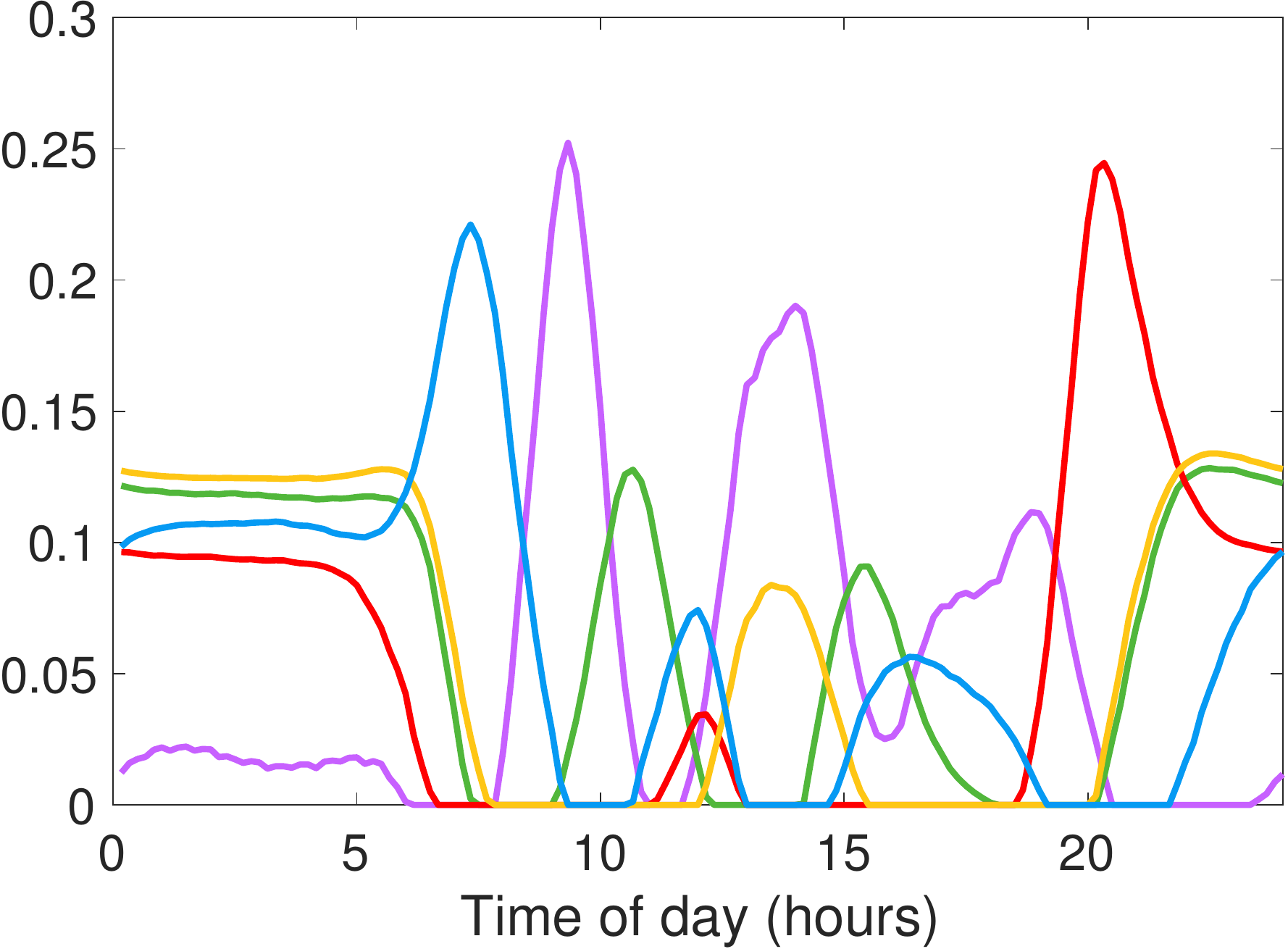} & \includegraphics[scale=0.15]{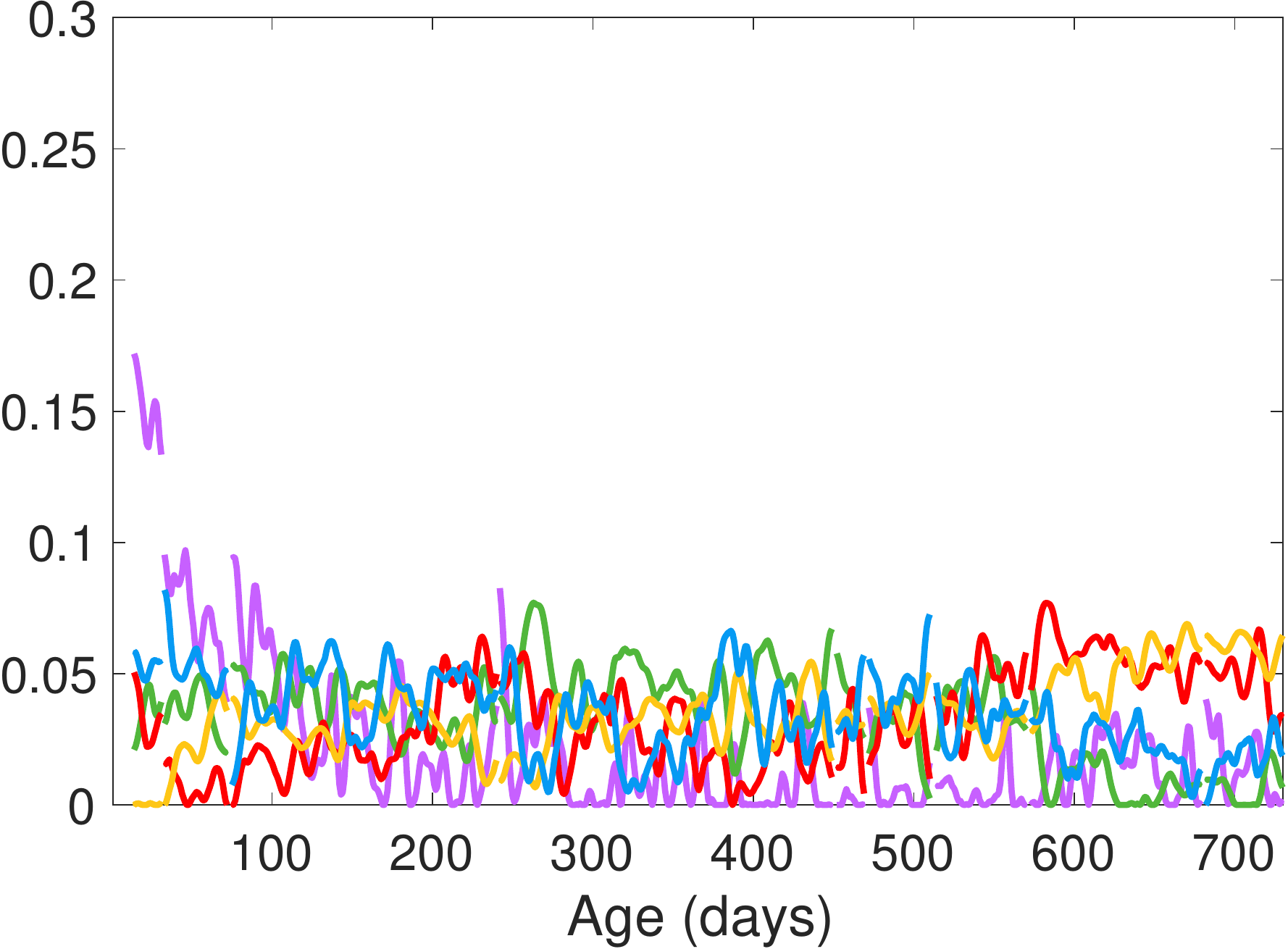} &  \includegraphics[scale=0.15]{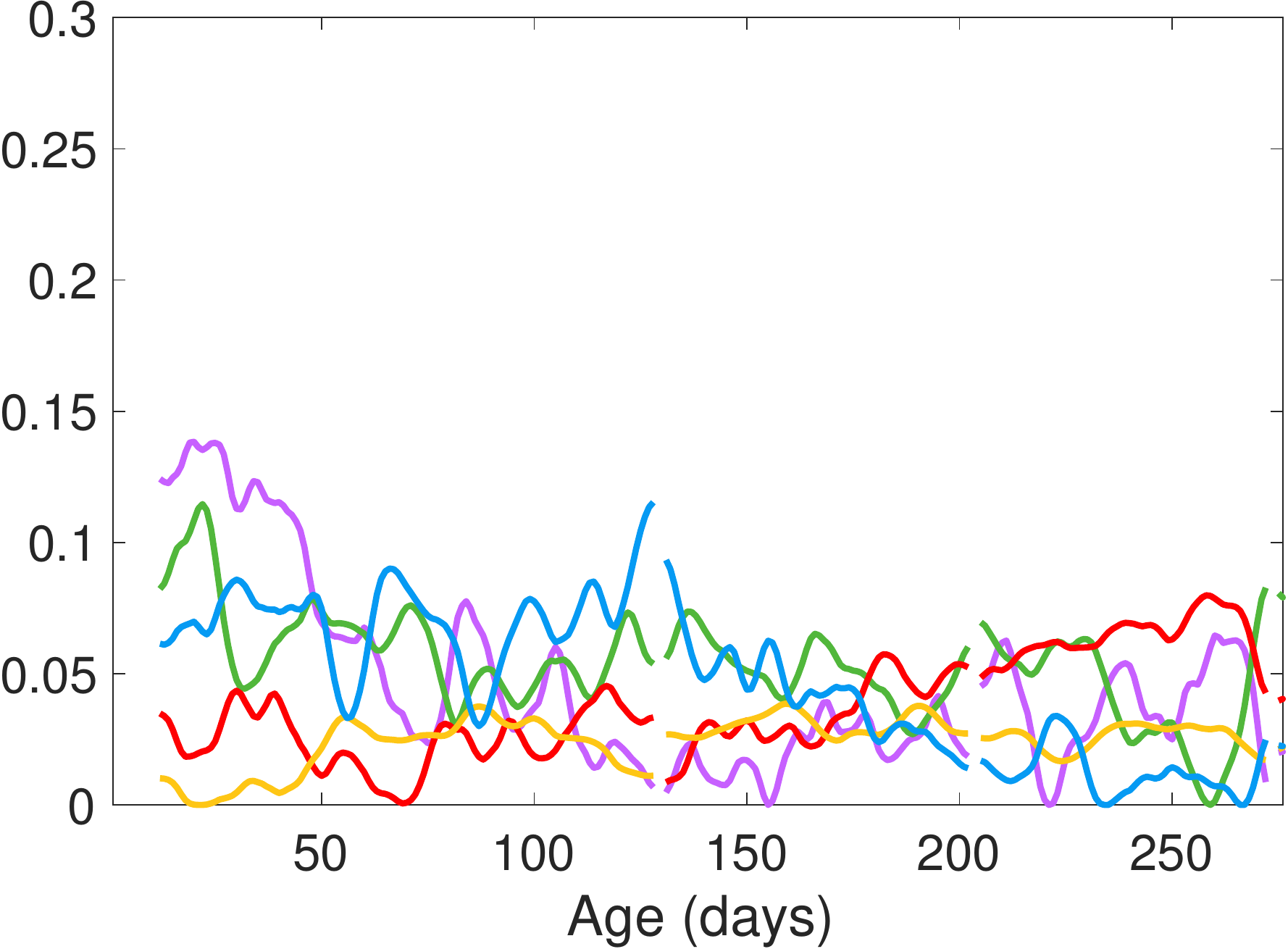} &  \includegraphics[scale=0.15]{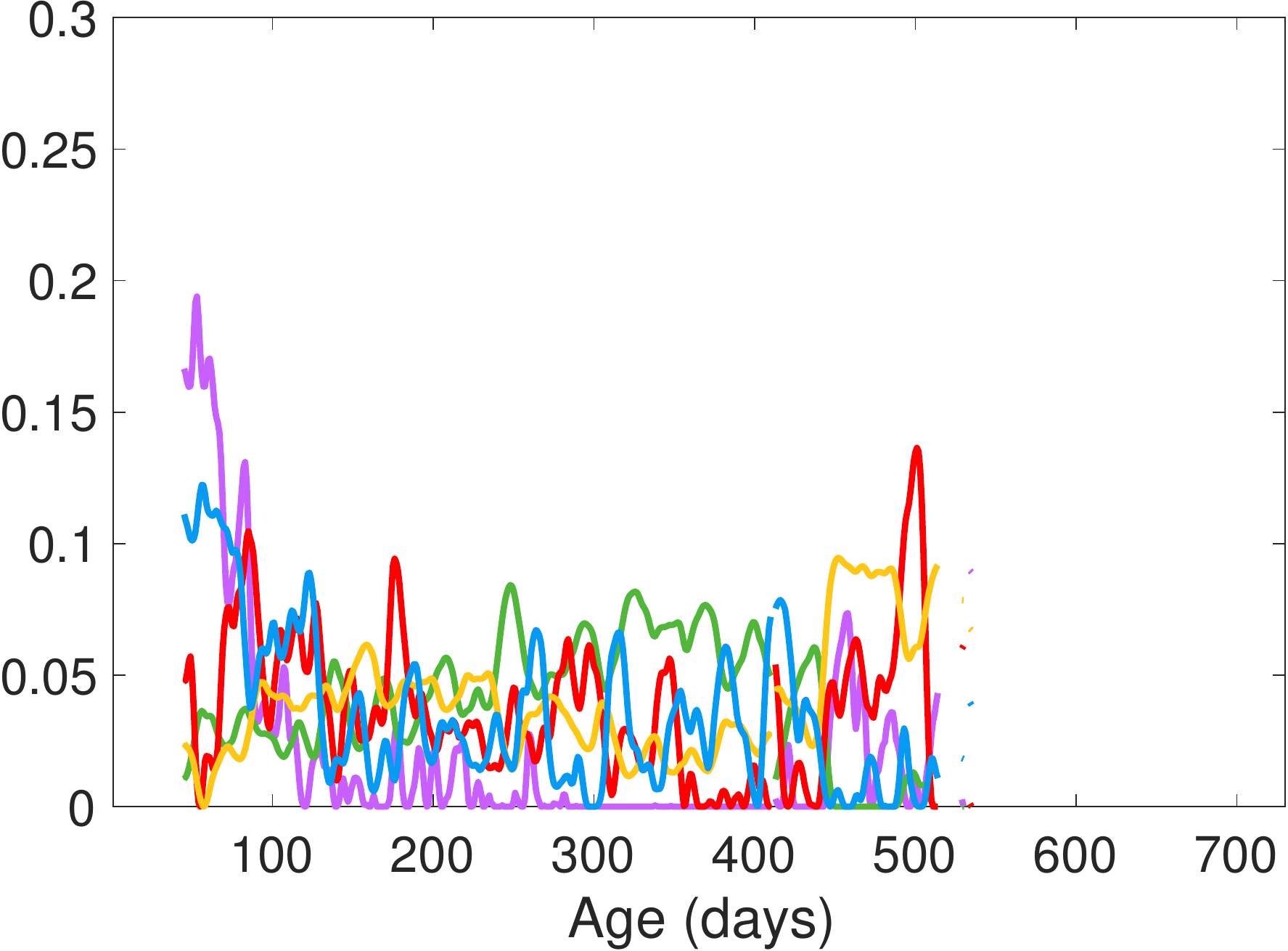}\\
{\footnotesize NMF-TS $\lambda=10^3$} & \includegraphics[scale=0.15]{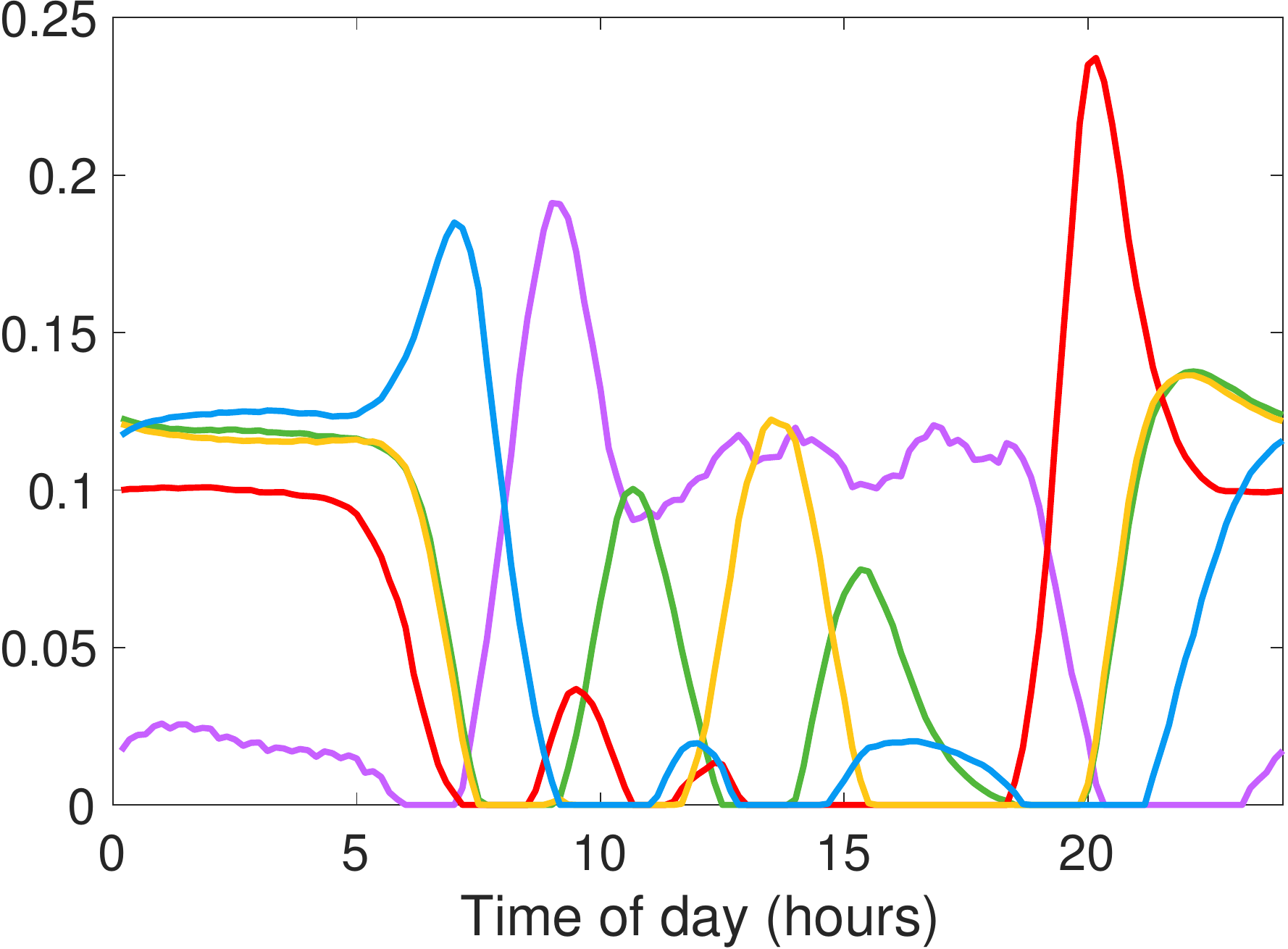} & \includegraphics[scale=0.15]{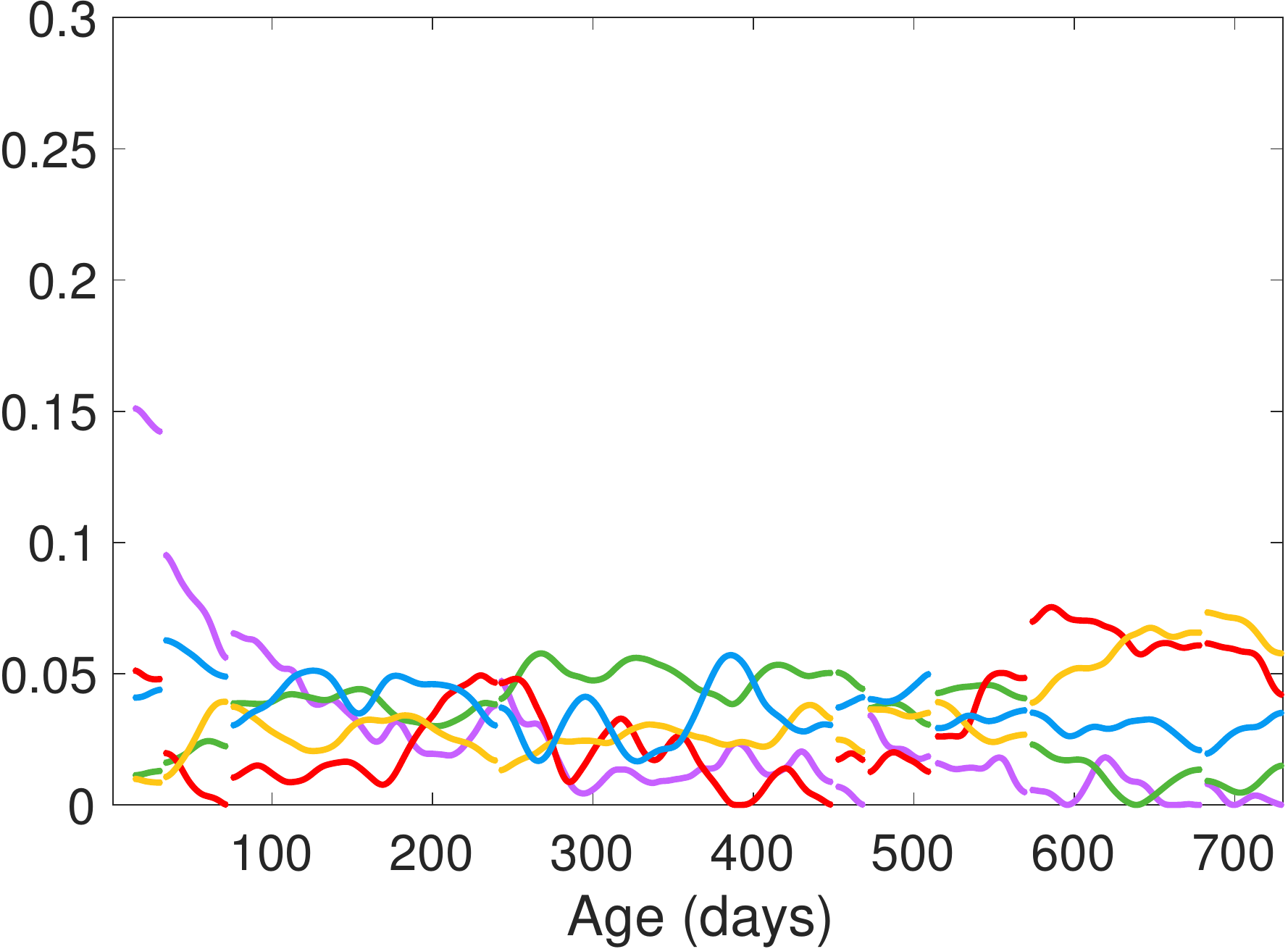} &  \includegraphics[scale=0.15]{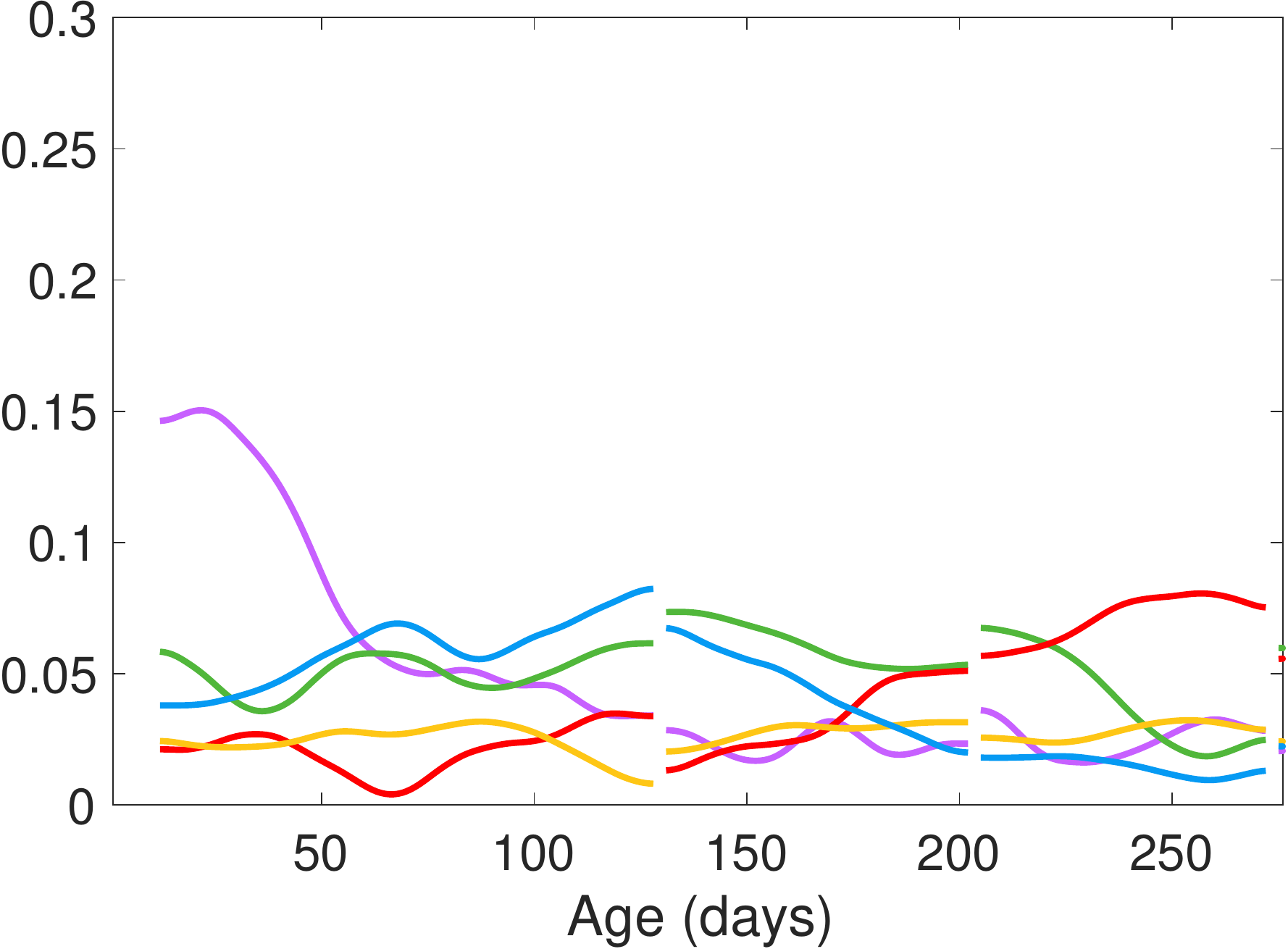} &  \includegraphics[scale=0.15]{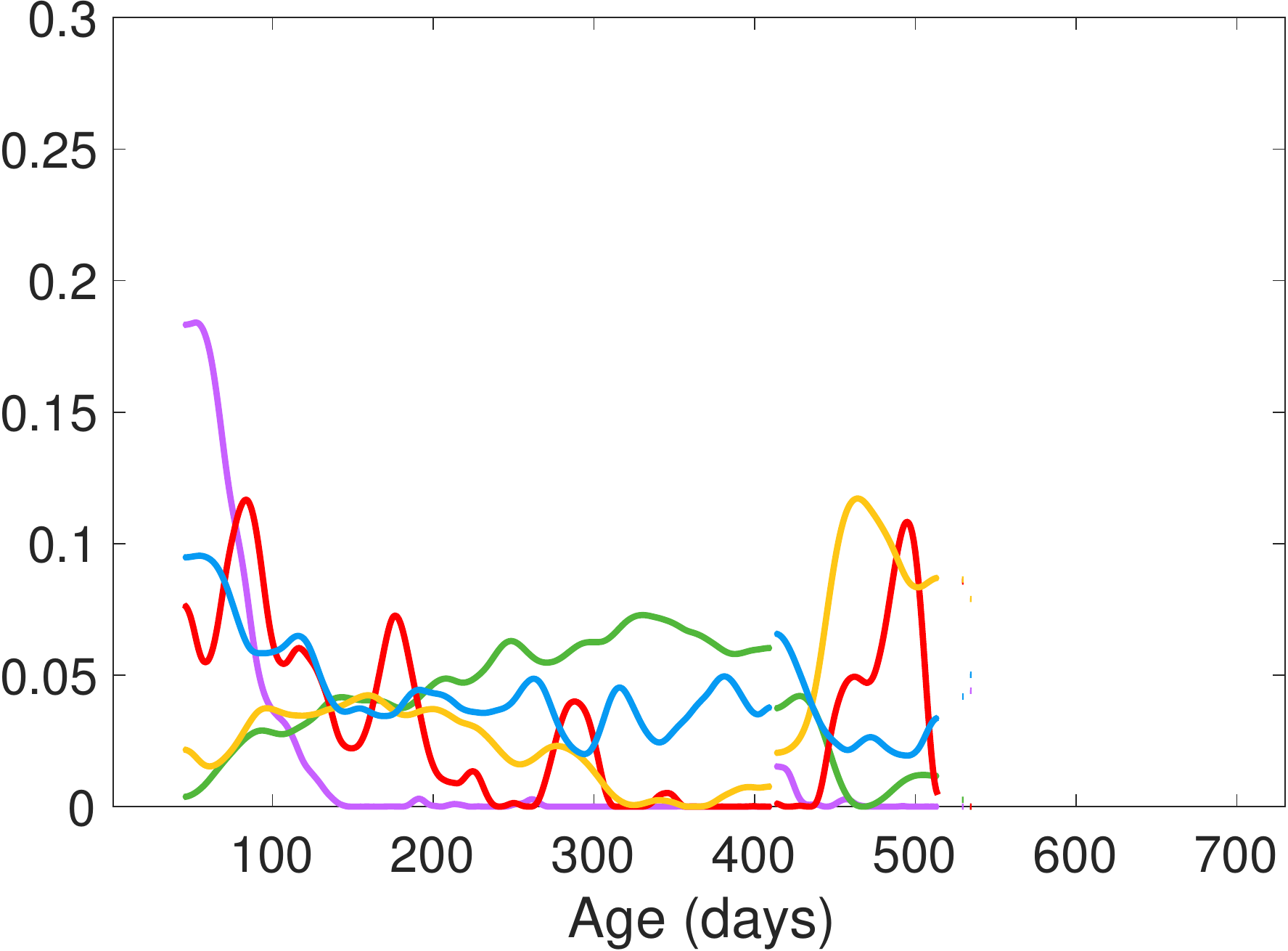}\\
{\footnotesize NMF-TS $\lambda=10^5$} & \includegraphics[scale=0.15]{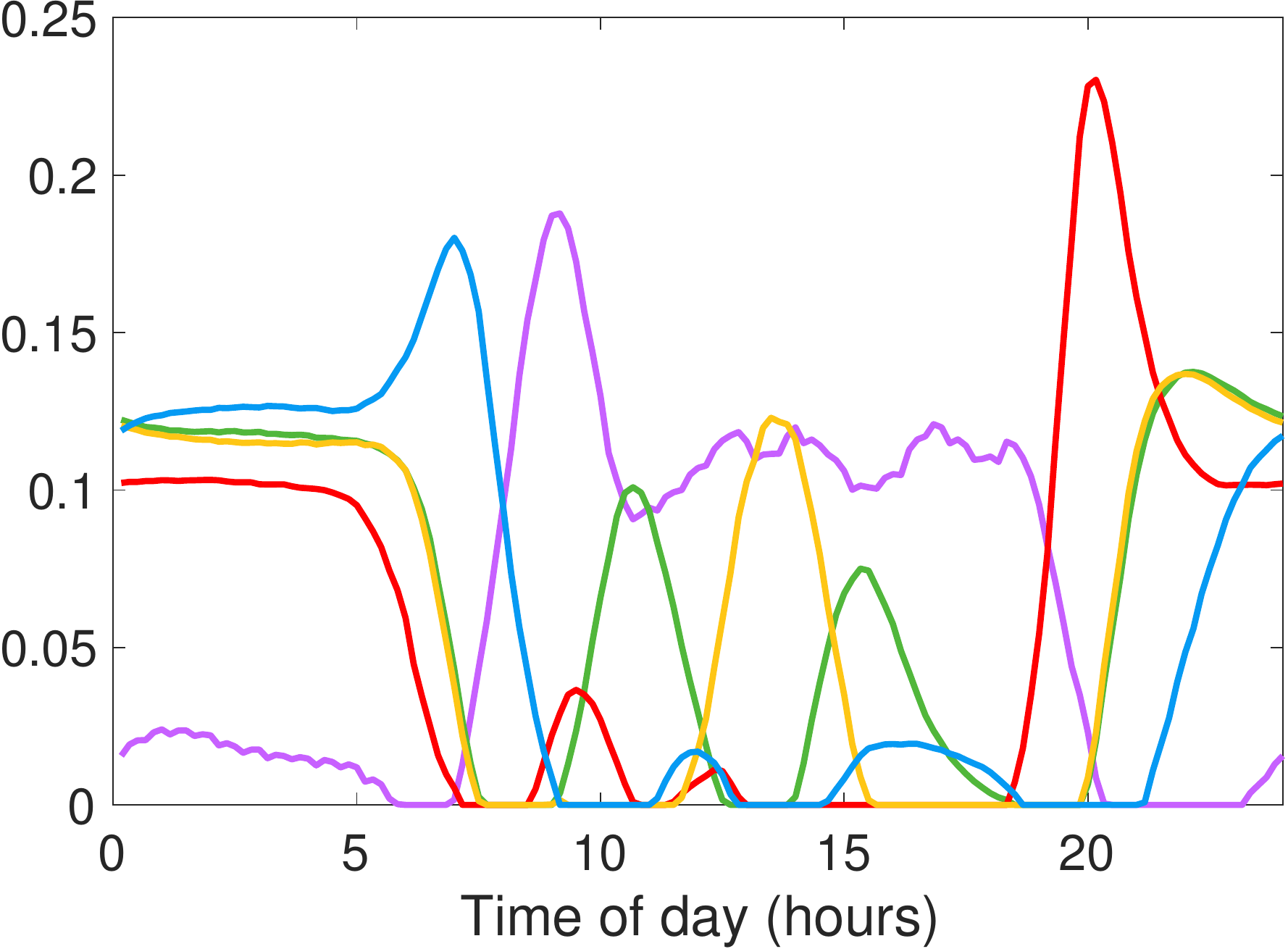} & \includegraphics[scale=0.15]{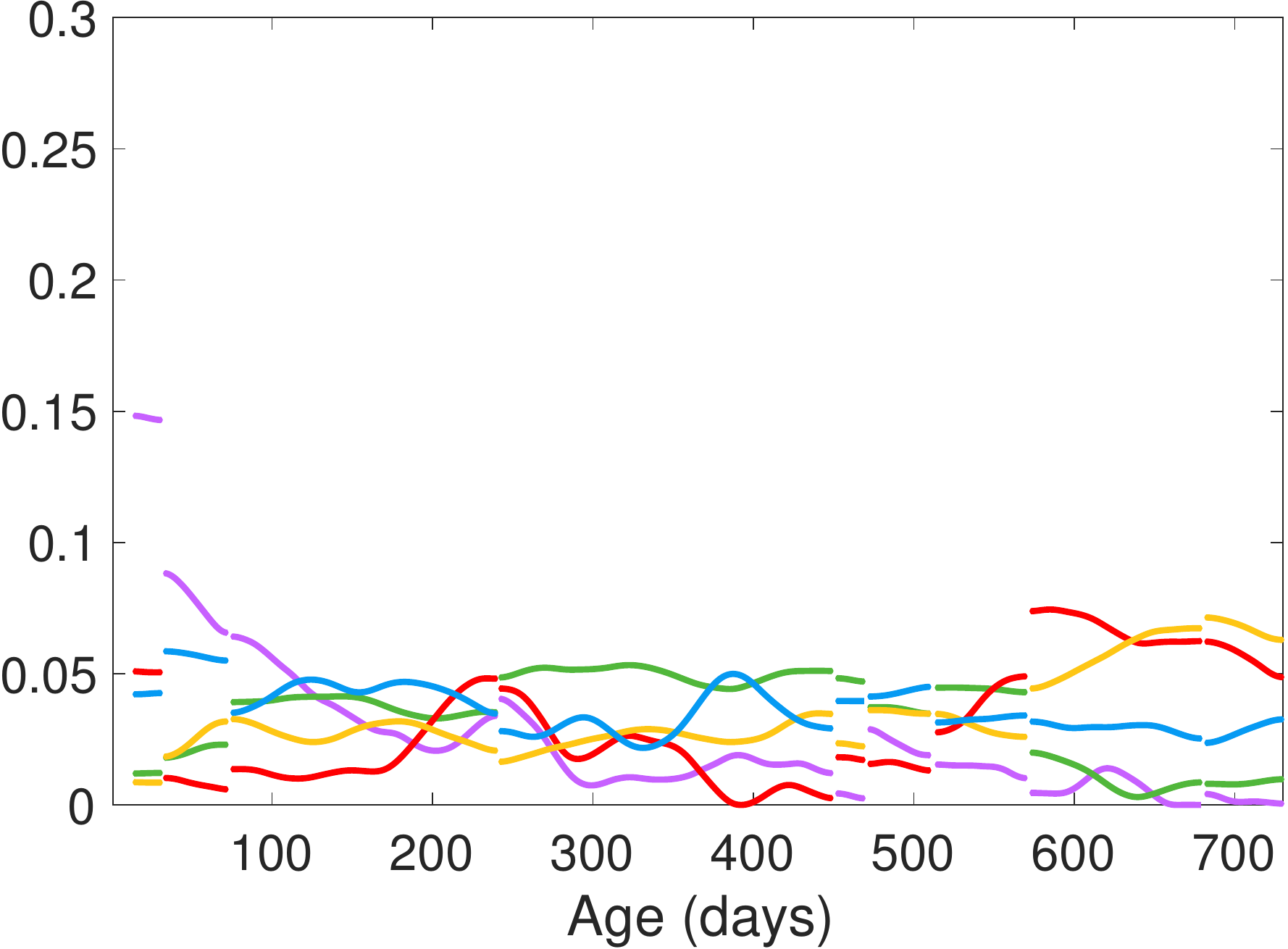} &  \includegraphics[scale=0.15]{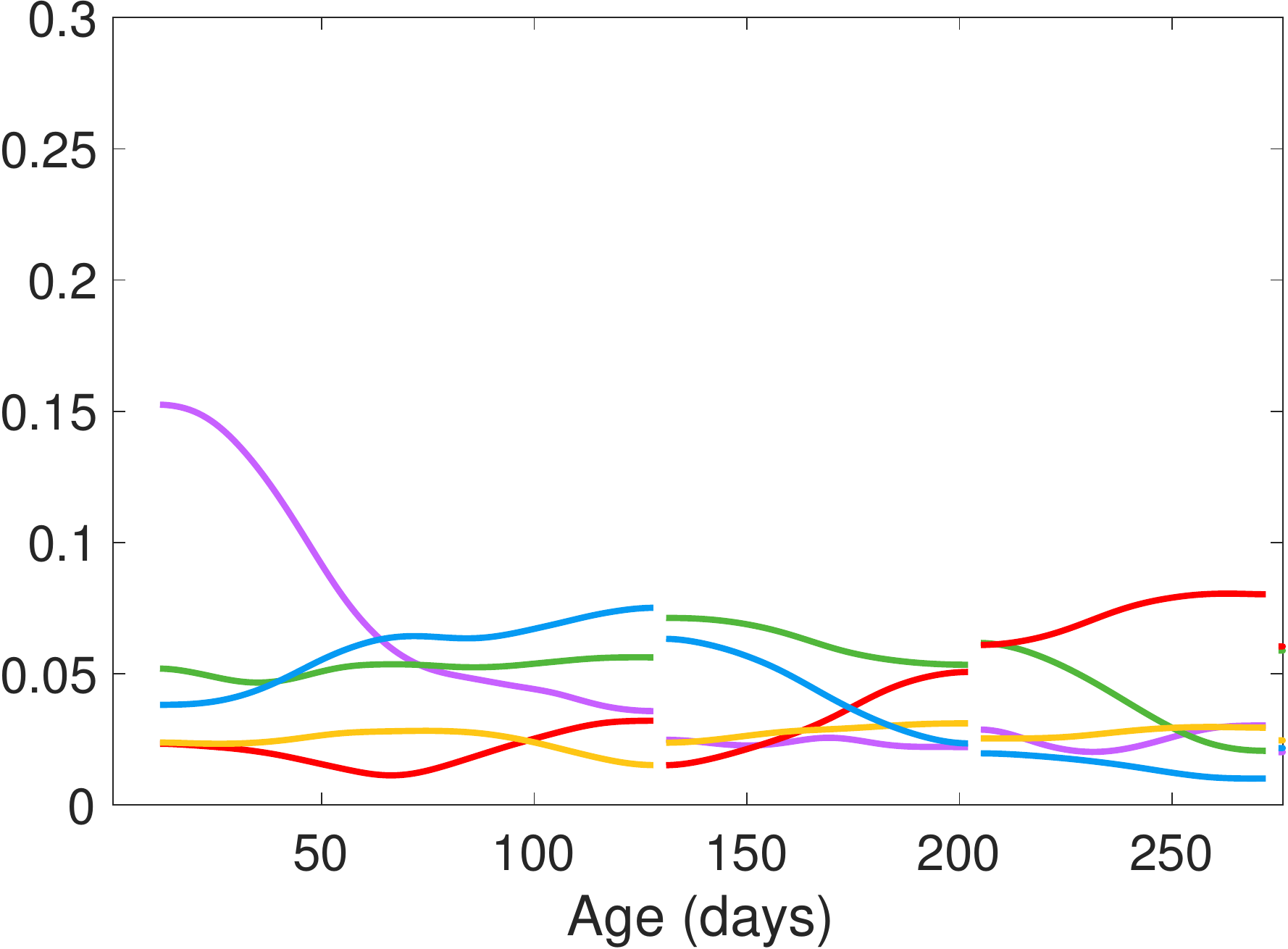} &  \includegraphics[scale=0.15]{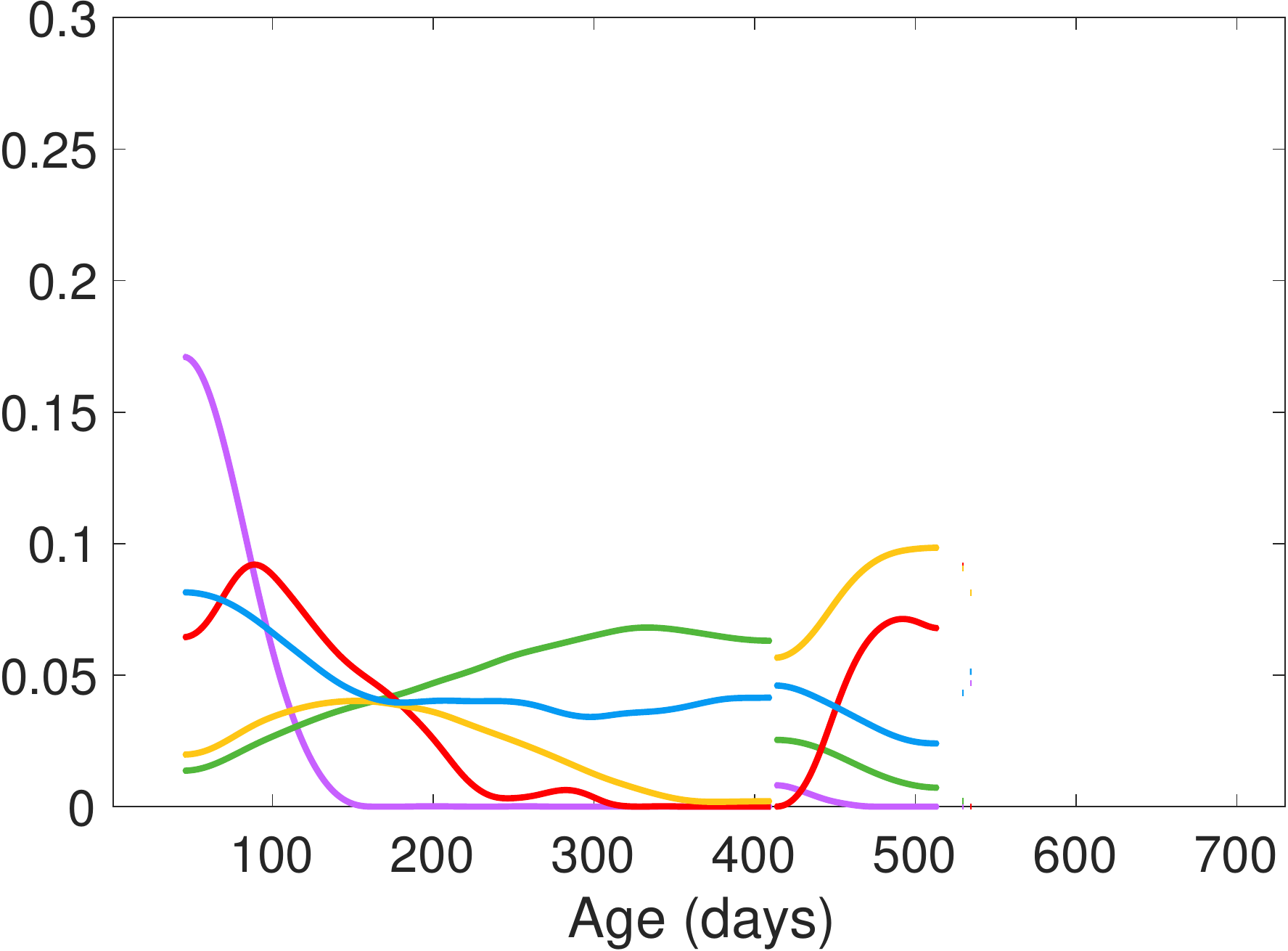}\\
{\footnotesize NMF-TS $\lambda=10^6$} & \includegraphics[scale=0.15]{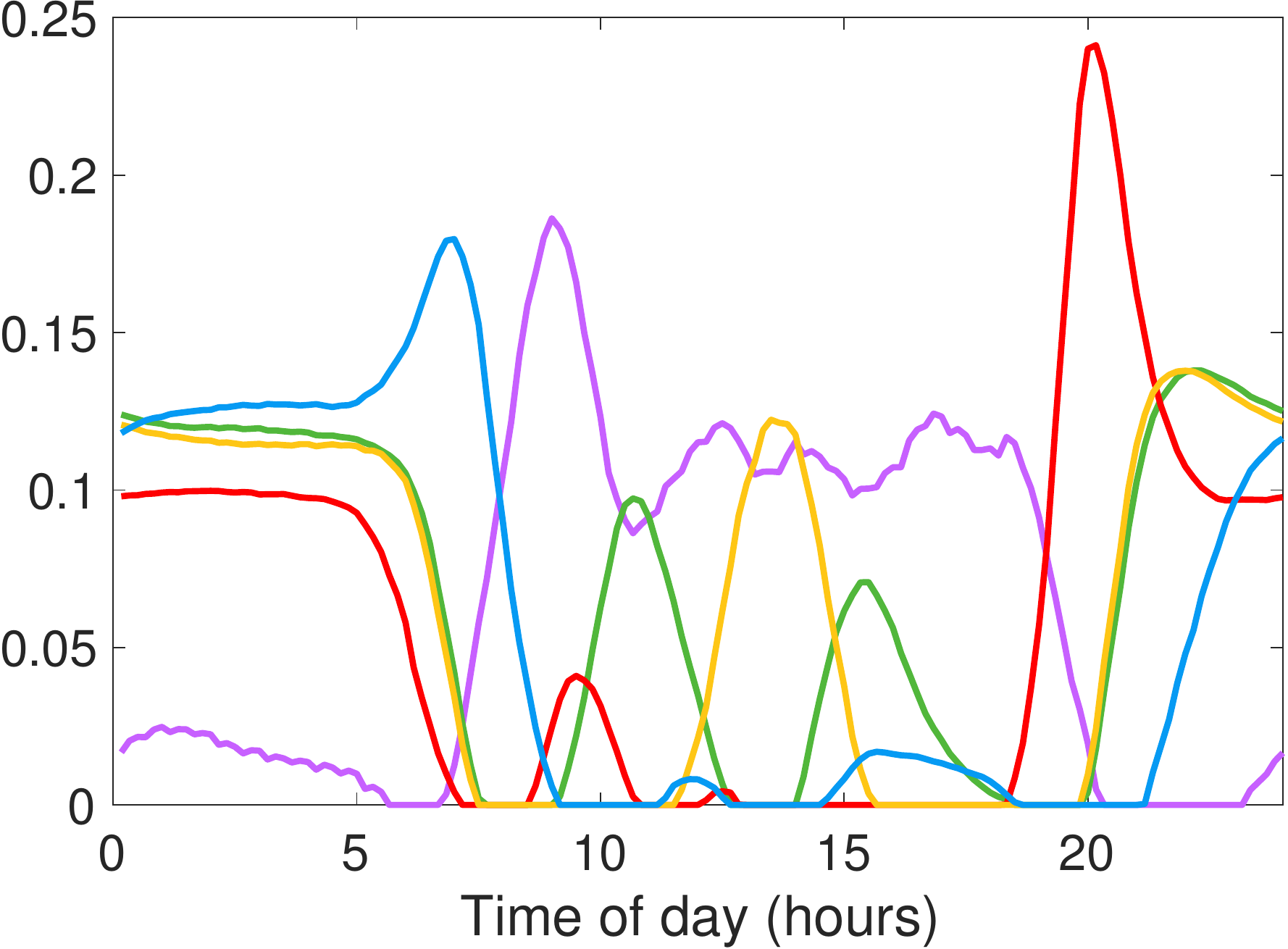} & \includegraphics[scale=0.15]{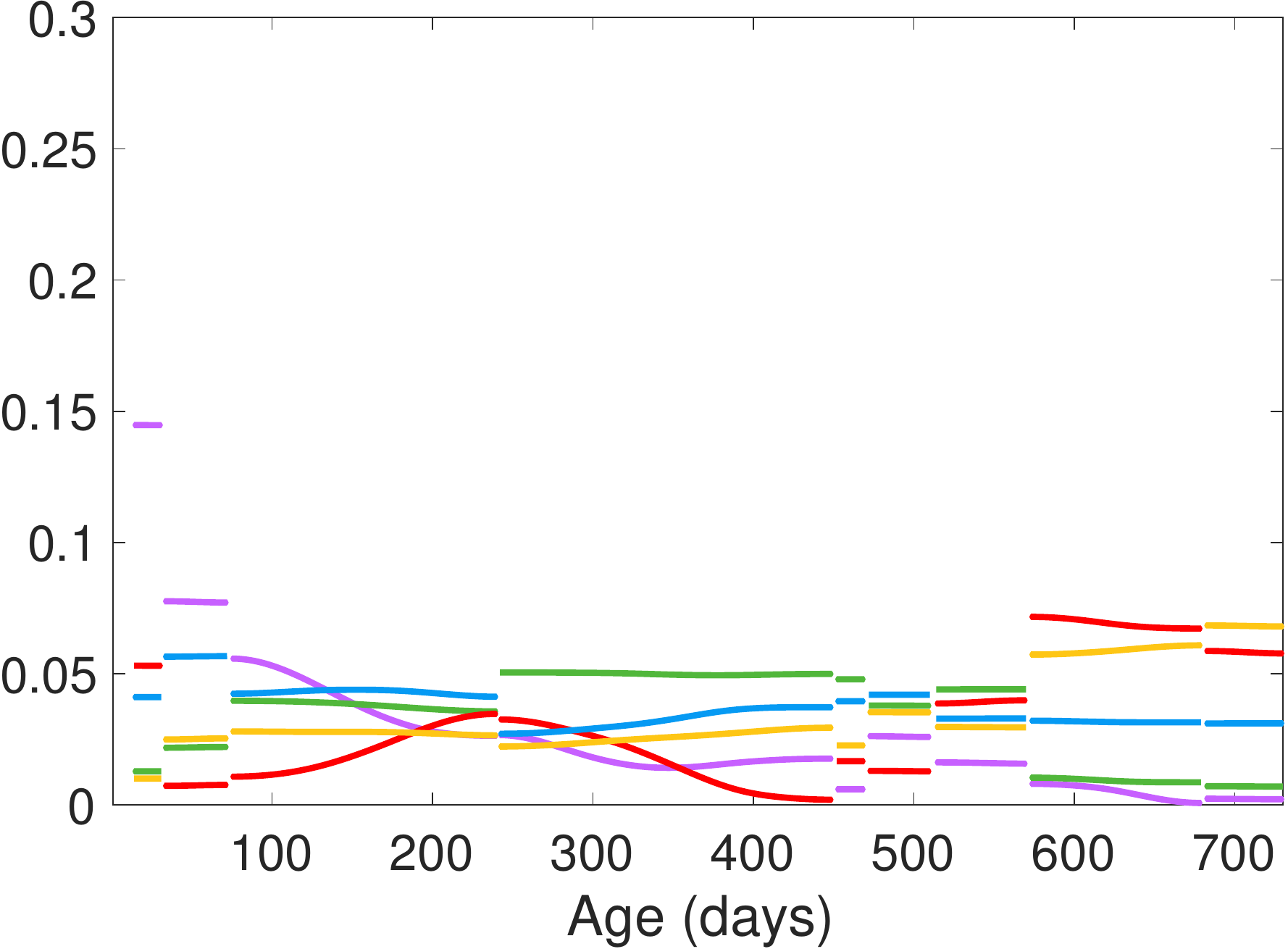} &  \includegraphics[scale=0.15]{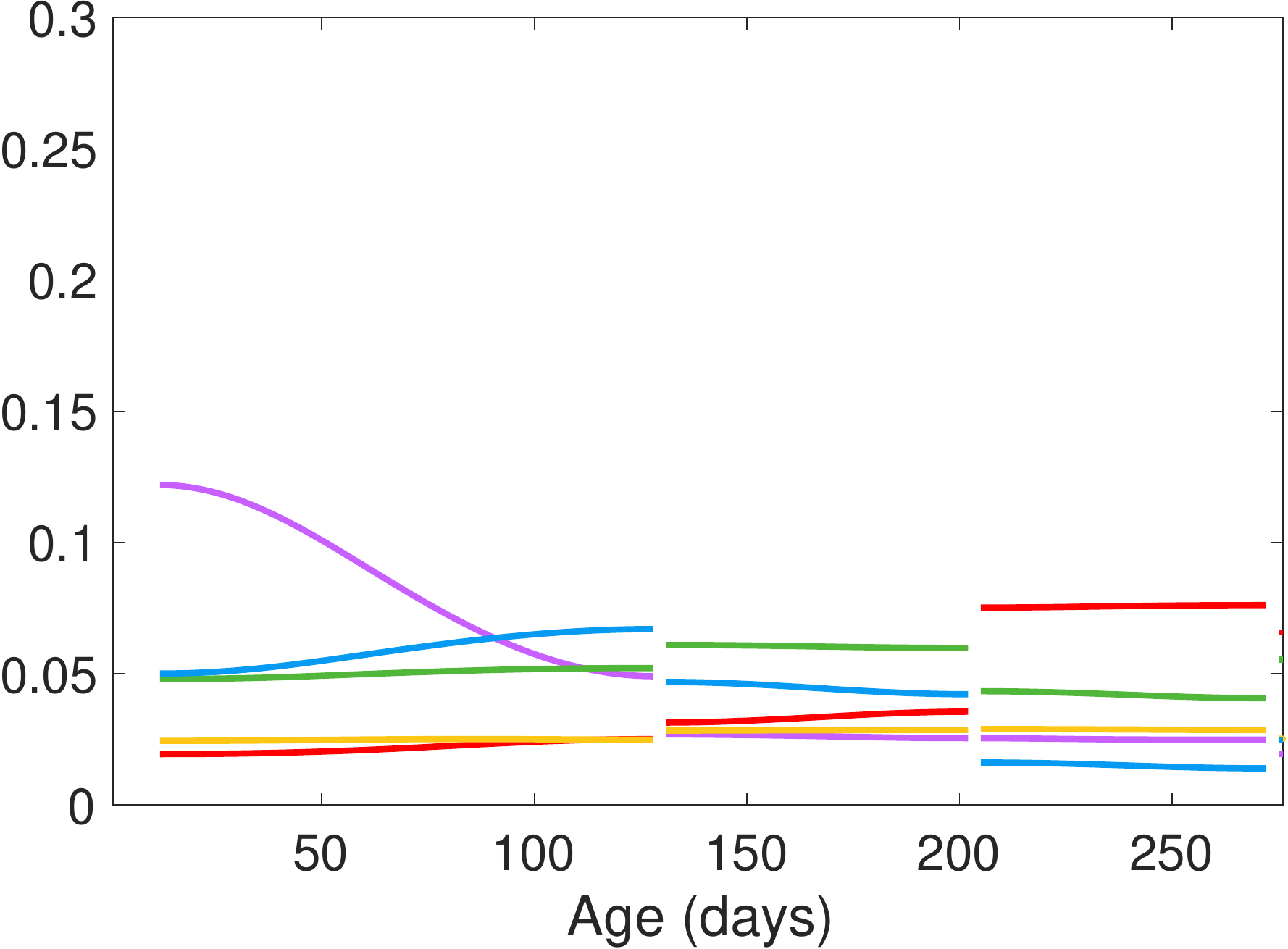} &  \includegraphics[scale=0.15]{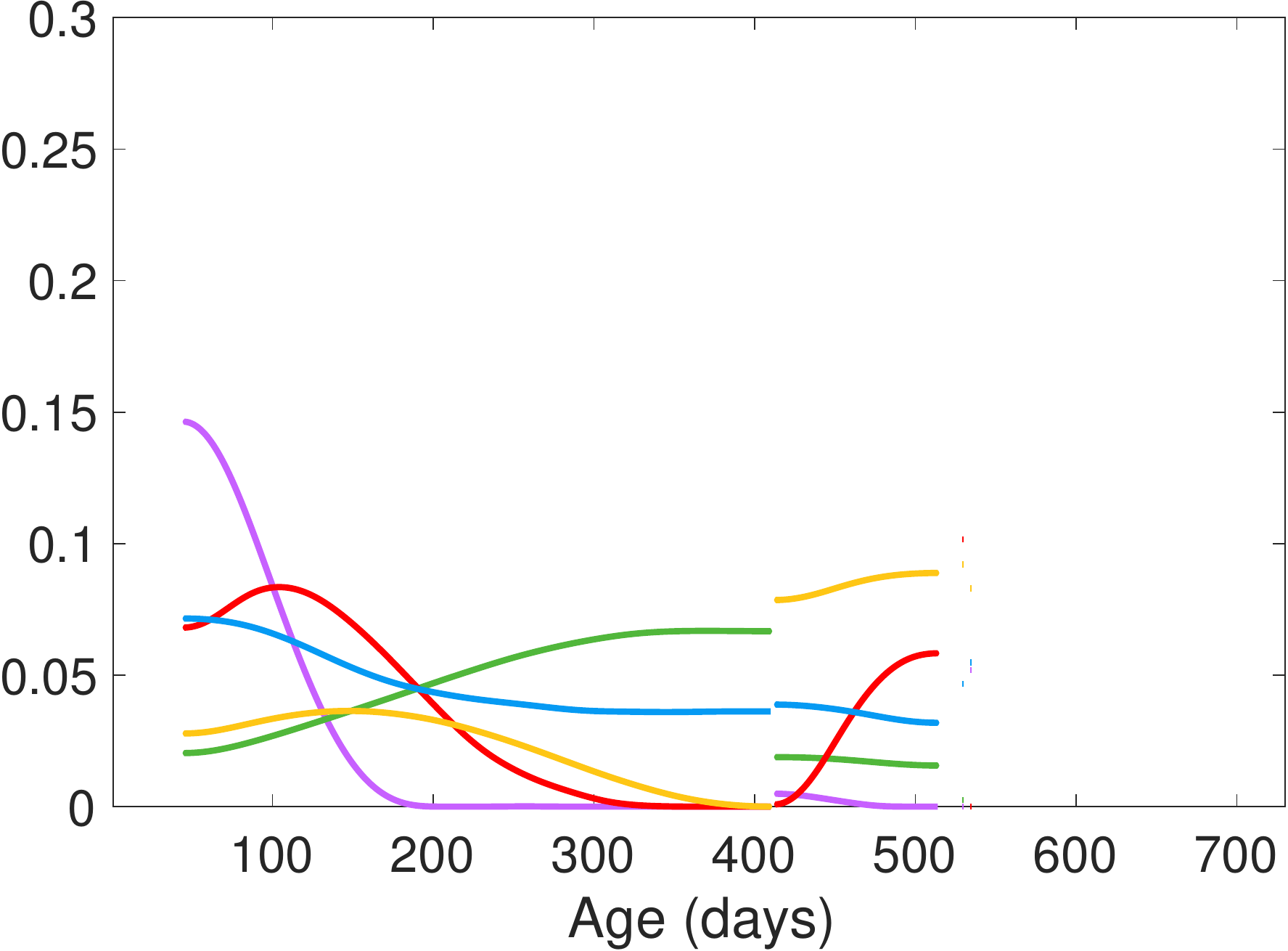}
 \end{tabular}
 \caption{Effect of varying the parameter $\lambda$ on the basis functions $F_1$, \ldots, $F_r$ and the coefficients $C^{[n]}_{1}$, \ldots, $C^{[n]}_{r}$ for three different time series $n_1$, $n_2$, $n_3$. The basis functions change only slightly once $\lambda$ is large enough. The coefficients become smoother as we increase $\lambda$.}
 \label{fig:smoothness}
\end{figure}

\begin{figure}[t]
\hspace{-1.8cm}
{\small
\begin{tabular}{ >{\centering\arraybackslash}m{0.1\linewidth}  >{\centering\arraybackslash}m{0.17\linewidth} >{\centering\arraybackslash}m{0.17\linewidth} >{\centering\arraybackslash}m{0.17\linewidth} >{\centering\arraybackslash}m{0.17\linewidth} >{\centering\arraybackslash}m{0.17\linewidth}}
Cluster \vspace{0.2cm} & Mean \vspace{0.2cm} & Nearest \vspace{0.2cm} &  1st quartile \vspace{0.2cm}&  Median \vspace{0.2cm} &  3rd quartile \vspace{0.2cm}\\
1 & \includegraphics[scale=0.15]{figs/clustering_results/raw_data/cluster_center_1} & \includegraphics[scale=0.15]{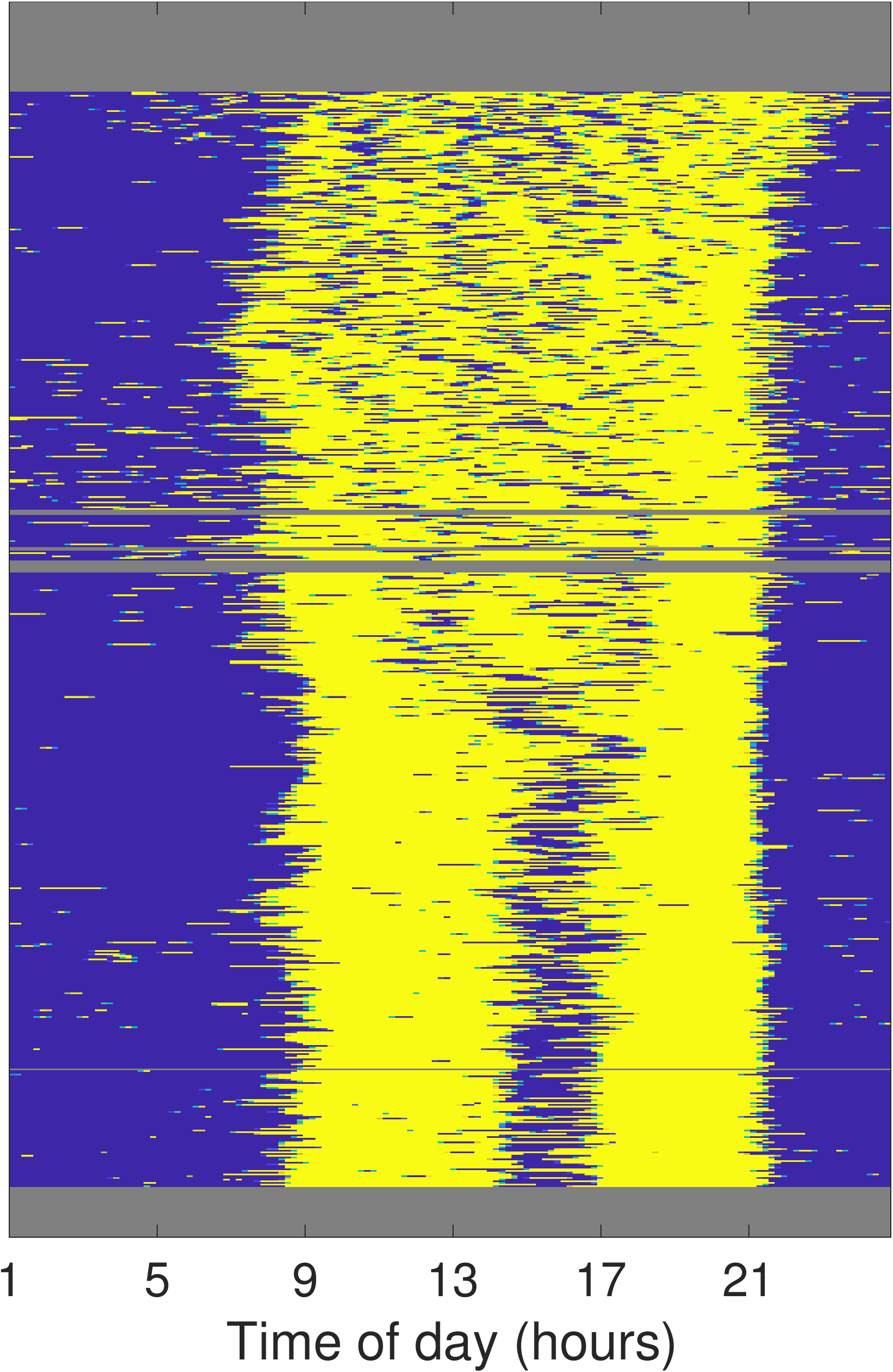}  &  \includegraphics[scale=0.15]{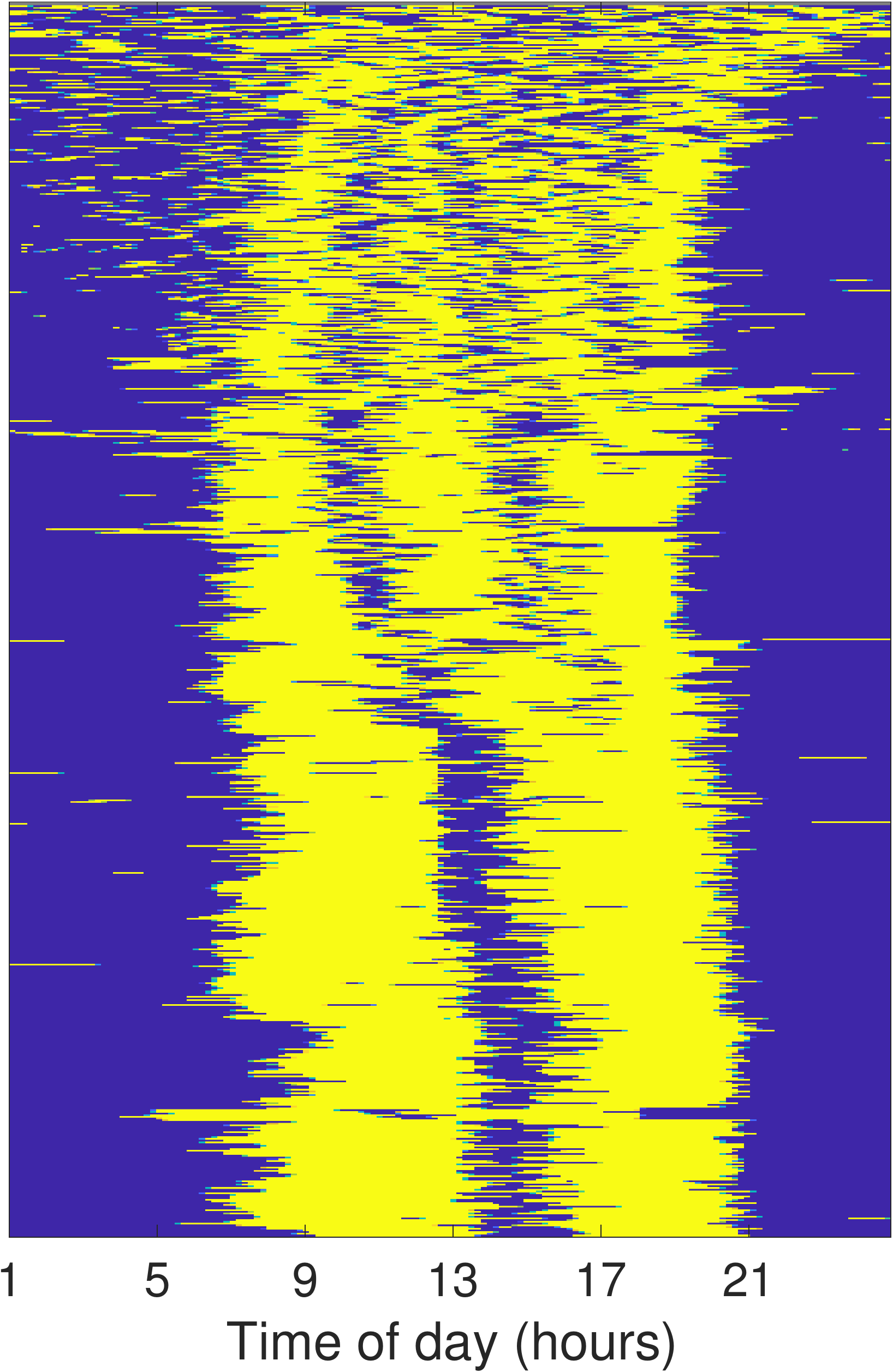}&  \includegraphics[scale=0.15]{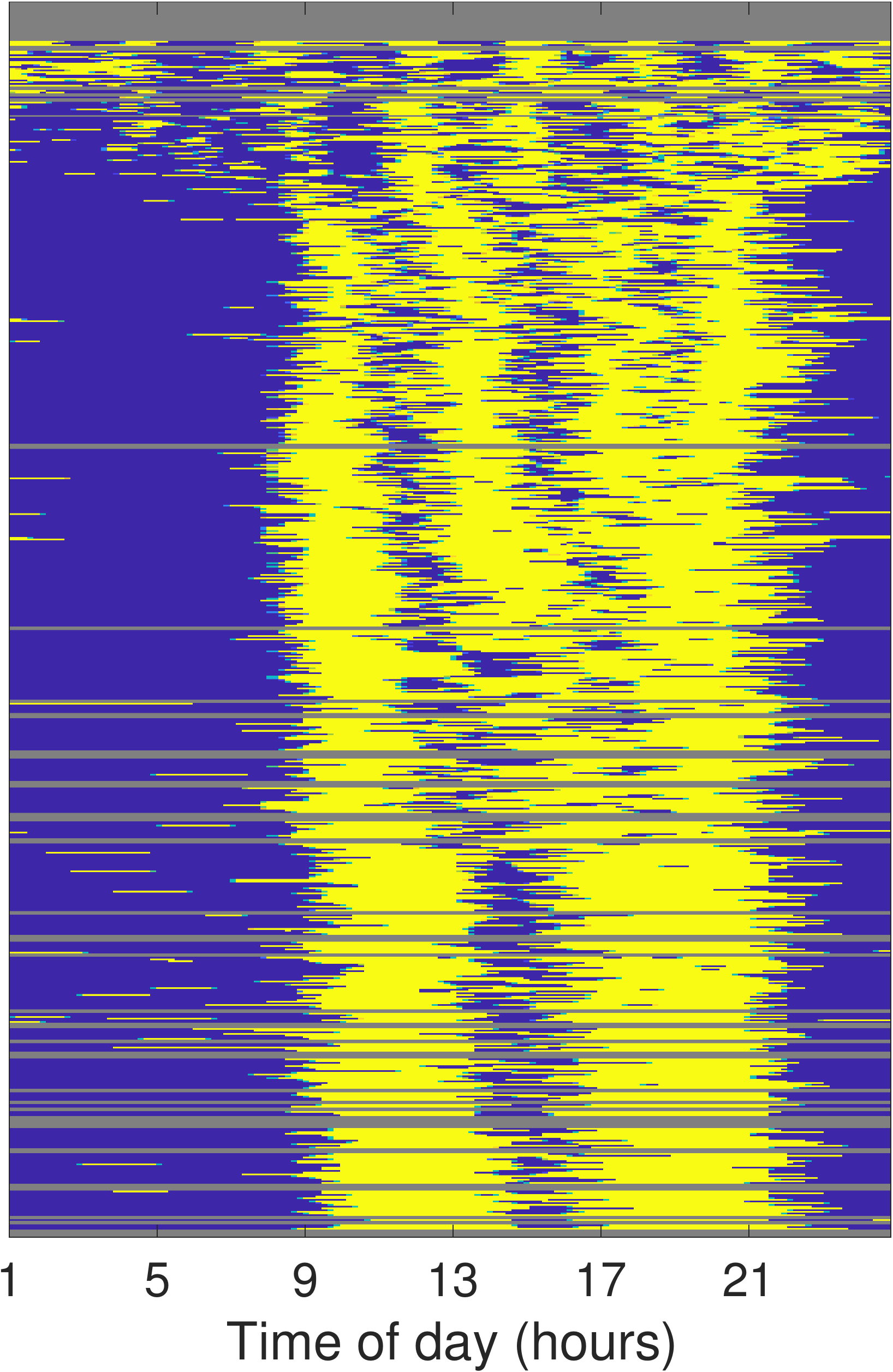} &  \includegraphics[scale=0.15]{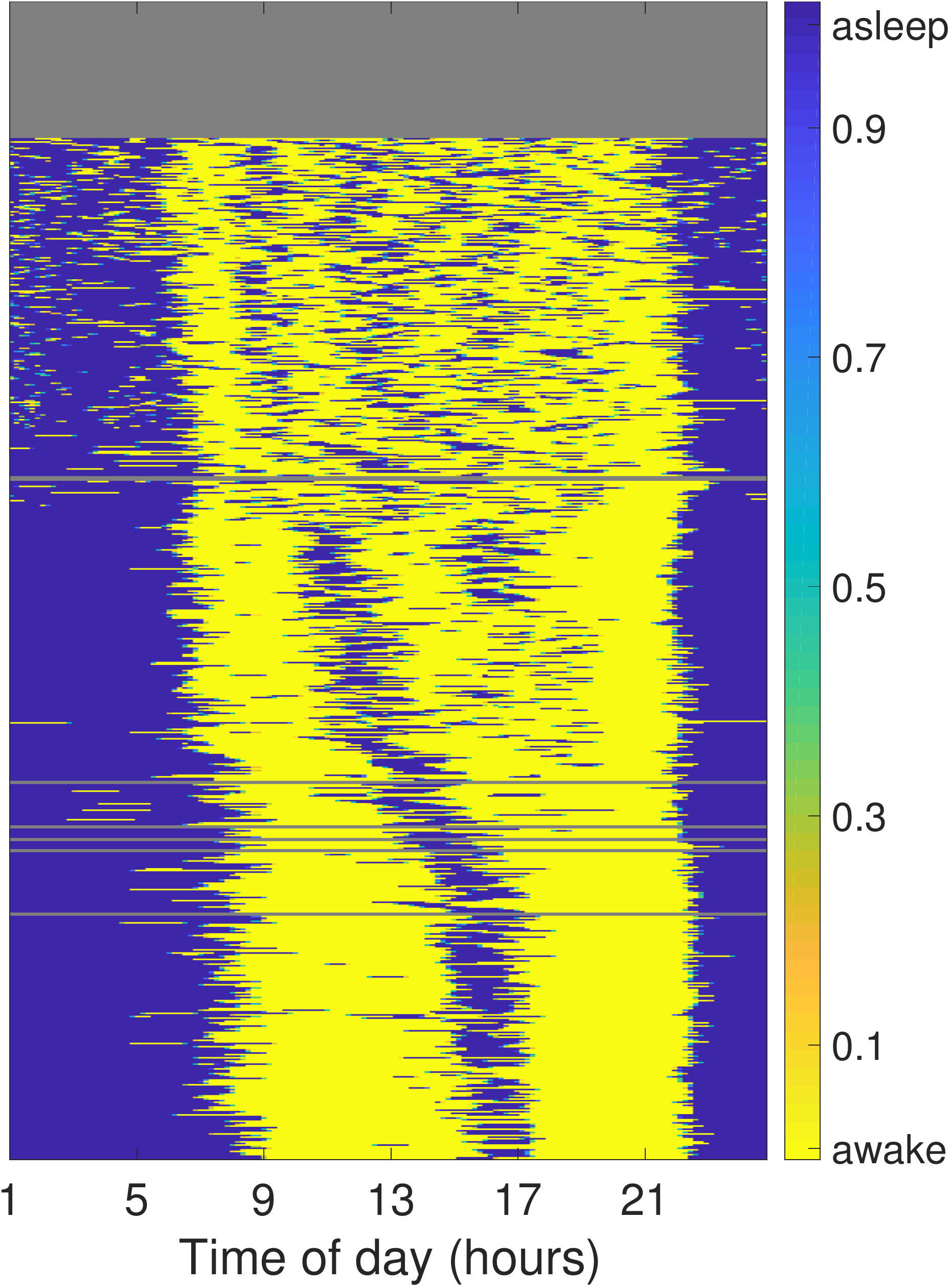}\\
2 & \includegraphics[scale=0.15]{figs/clustering_results/raw_data/cluster_center_2} & \includegraphics[scale=0.15]{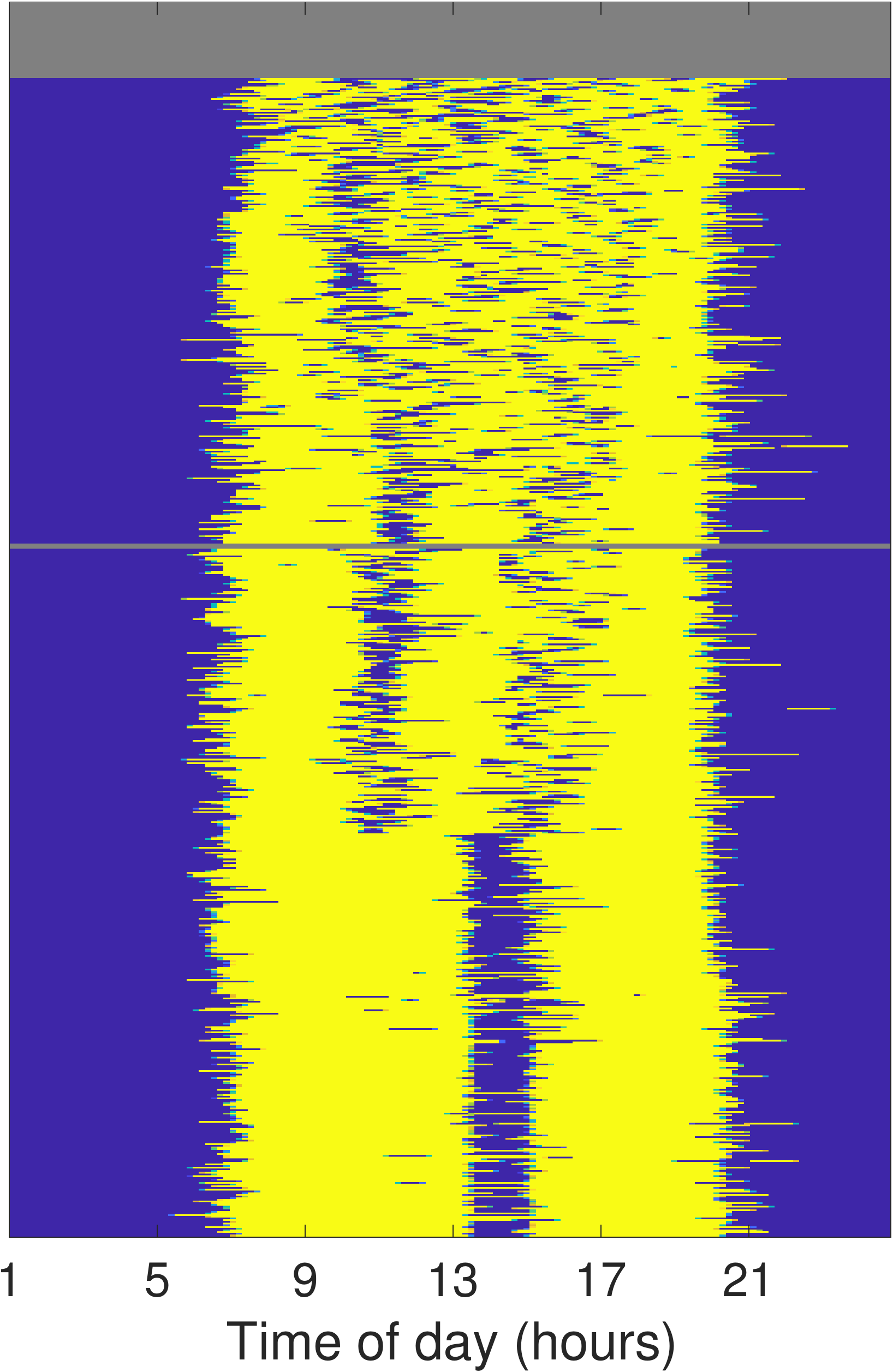} &  \includegraphics[scale=0.15]{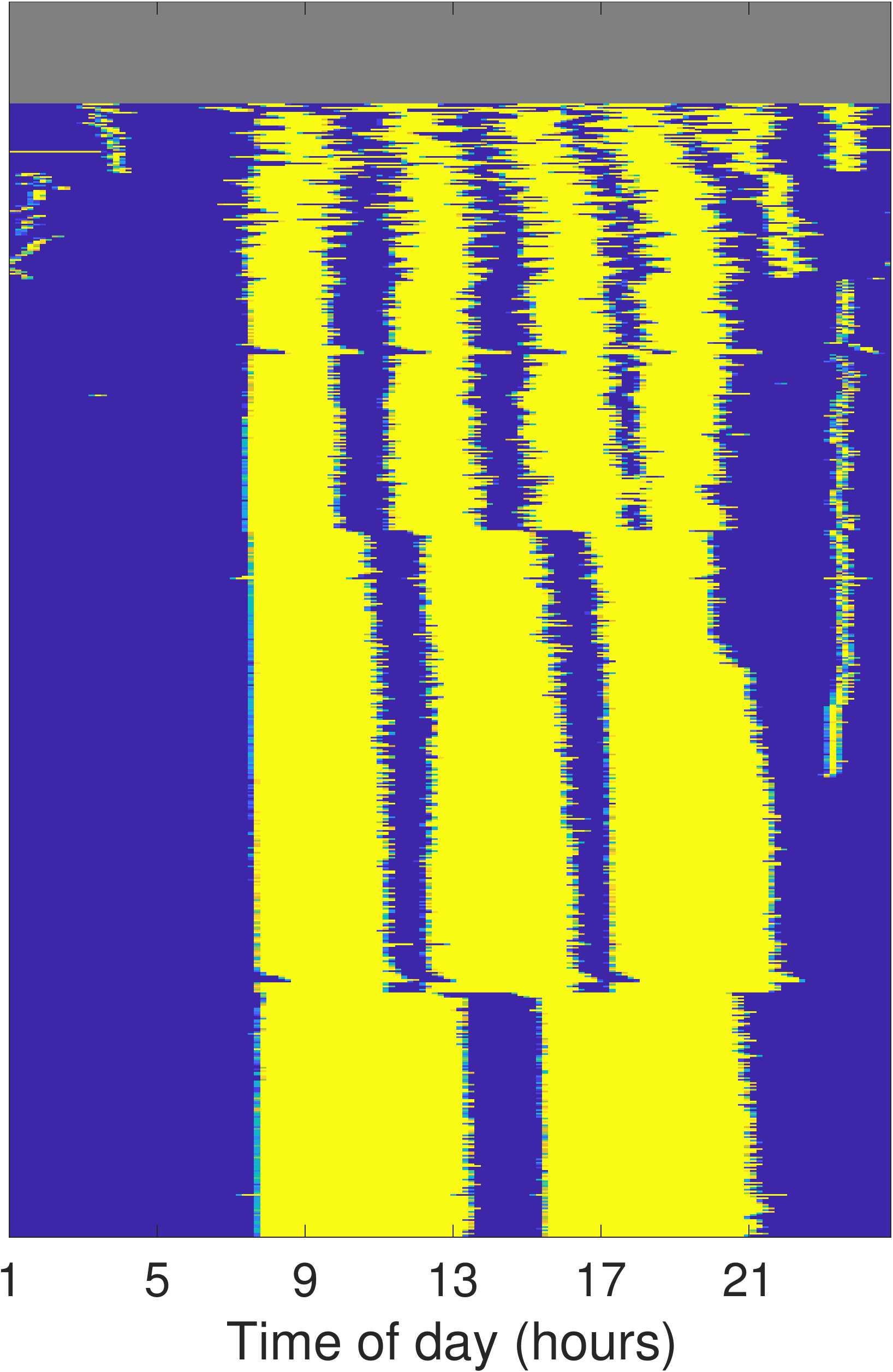}&  \includegraphics[scale=0.15]{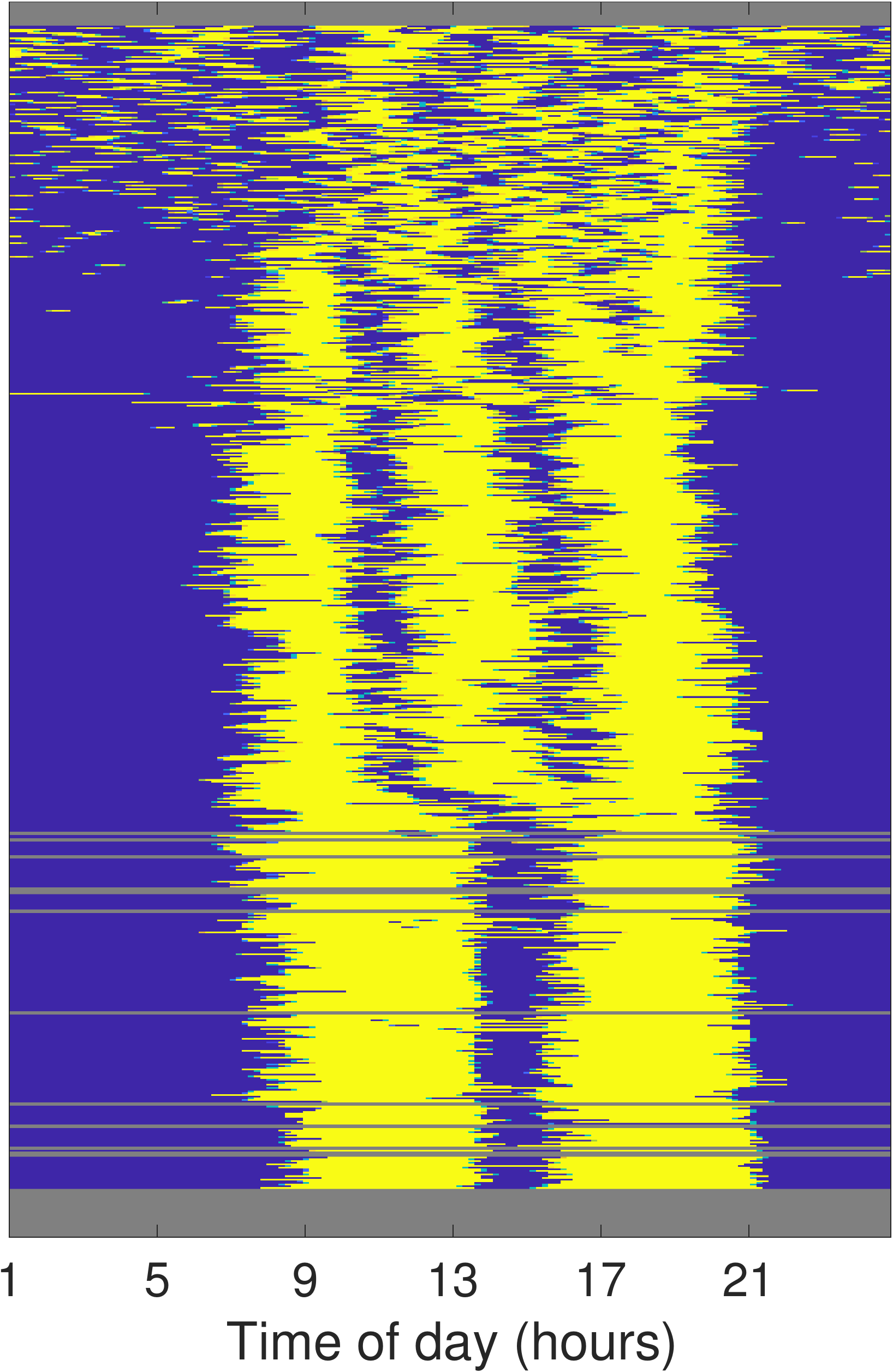} &  \includegraphics[scale=0.15]{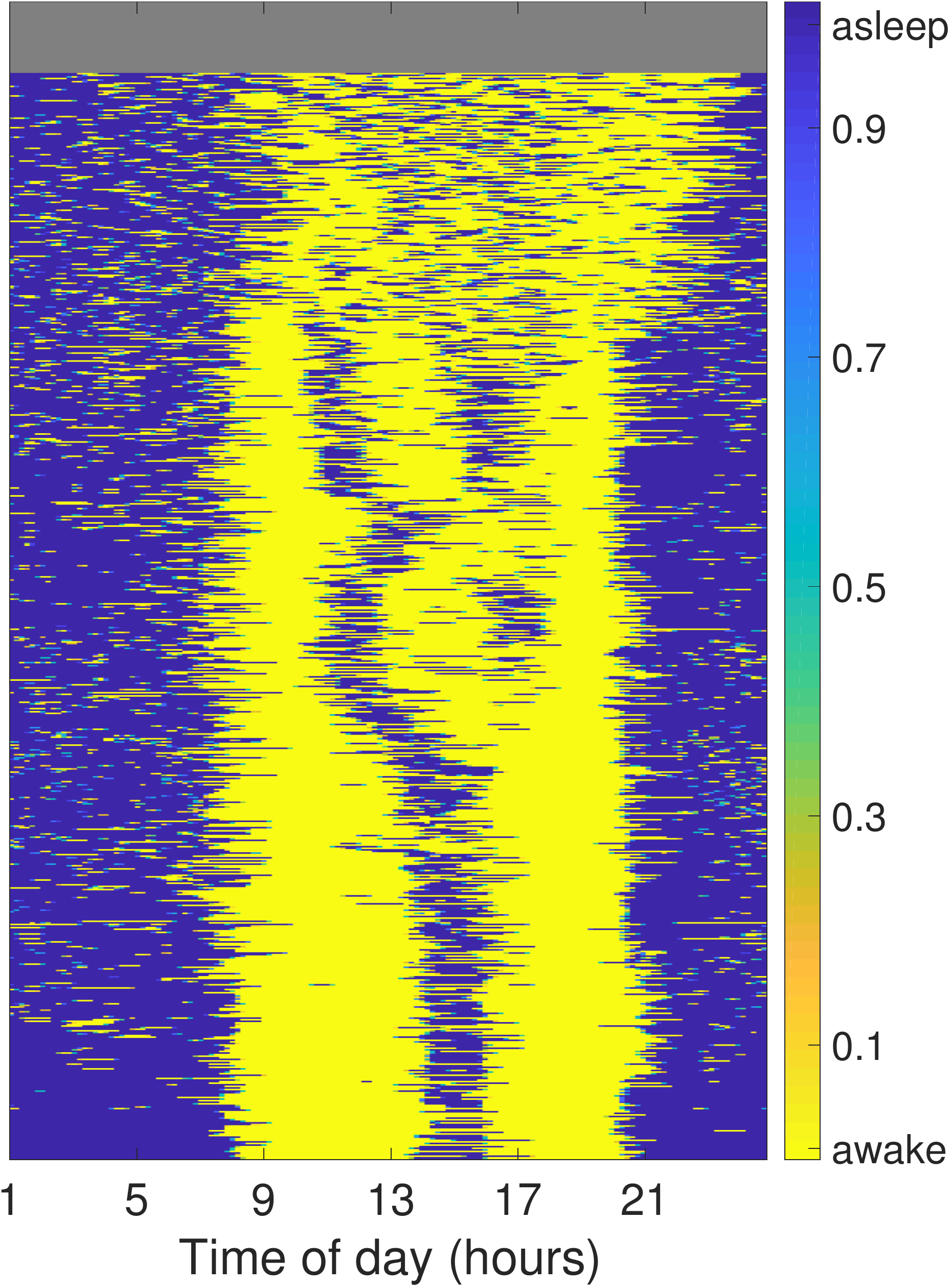}\\
 \end{tabular}
 }
 \caption{Results of applying Lloyd's algorithm to perform k-means clustering with $k :=2$ to the infant-sleep data. For the two clusters we show the cluster mean (1st column), the closest infant to the mean (2nd column), and the first quartile, median and third quartile infants in terms of distance to the mean out of infants with at least 80\% available data (3rd, 4th and 5th columns respectively). The first cluster contains 308 infants. The second cluster contains 393. Gray values indicate missing data.}
 \label{fig:rawdata_results}
\end{figure}

\begin{figure}[tp]
{\small
\hspace{-1.9cm}
\begin{tabular}{ >{\centering\arraybackslash}m{0.08\linewidth} >{\centering\arraybackslash}m{0.195\linewidth} >{\centering\arraybackslash}m{0.17\linewidth} >{\centering\arraybackslash}m{0.17\linewidth} >{\centering\arraybackslash}m{0.17\linewidth} >{\centering\arraybackslash}m{0.17\linewidth}}
 Cluster \vspace{0.2cm} & Mean \vspace{0.2cm} & Nearest \vspace{0.2cm} &  1st quartile \vspace{0.2cm}&  Median \vspace{0.2cm} &  3rd quartile\vspace{0.2cm}\\
 1 & \includegraphics[scale=0.15]{figs/clustering_results/125/cluster_low_rank_center_1} & \includegraphics[scale=0.15]{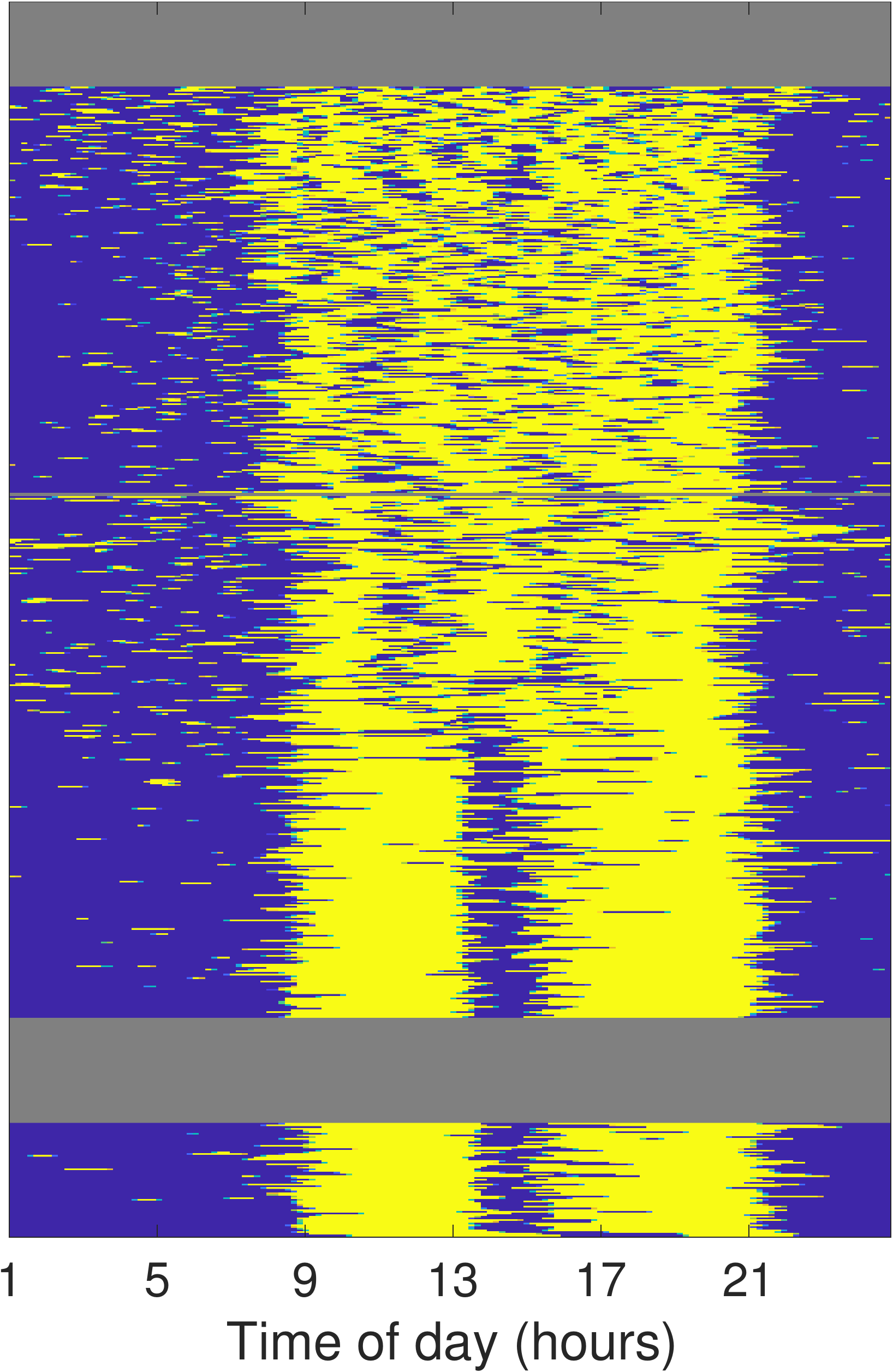} &  \includegraphics[scale=0.15]{figs/clustering_results/125/cluster_1_baby_25_raw_data}&  \includegraphics[scale=0.15]{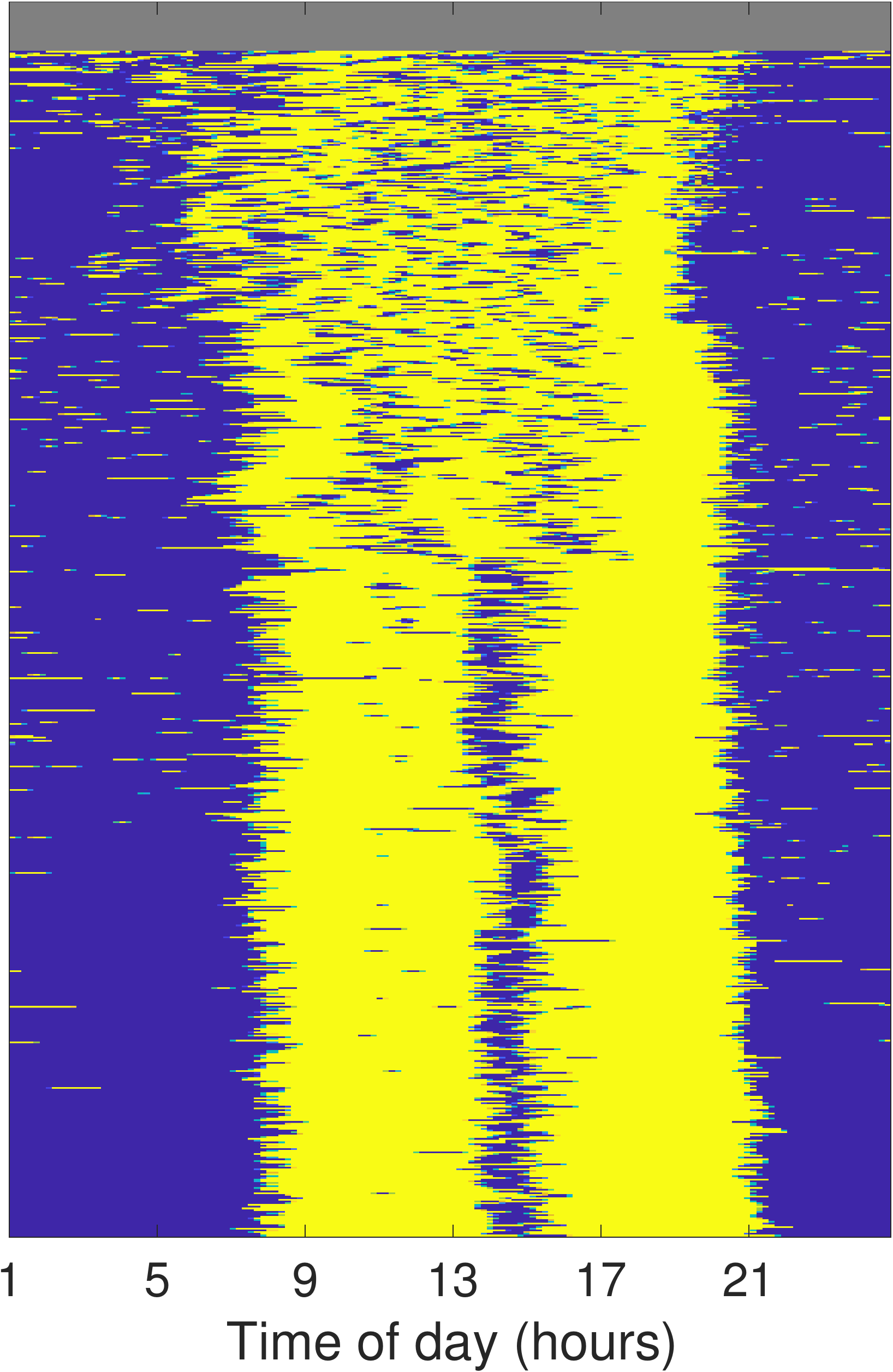} &  \includegraphics[scale=0.15]{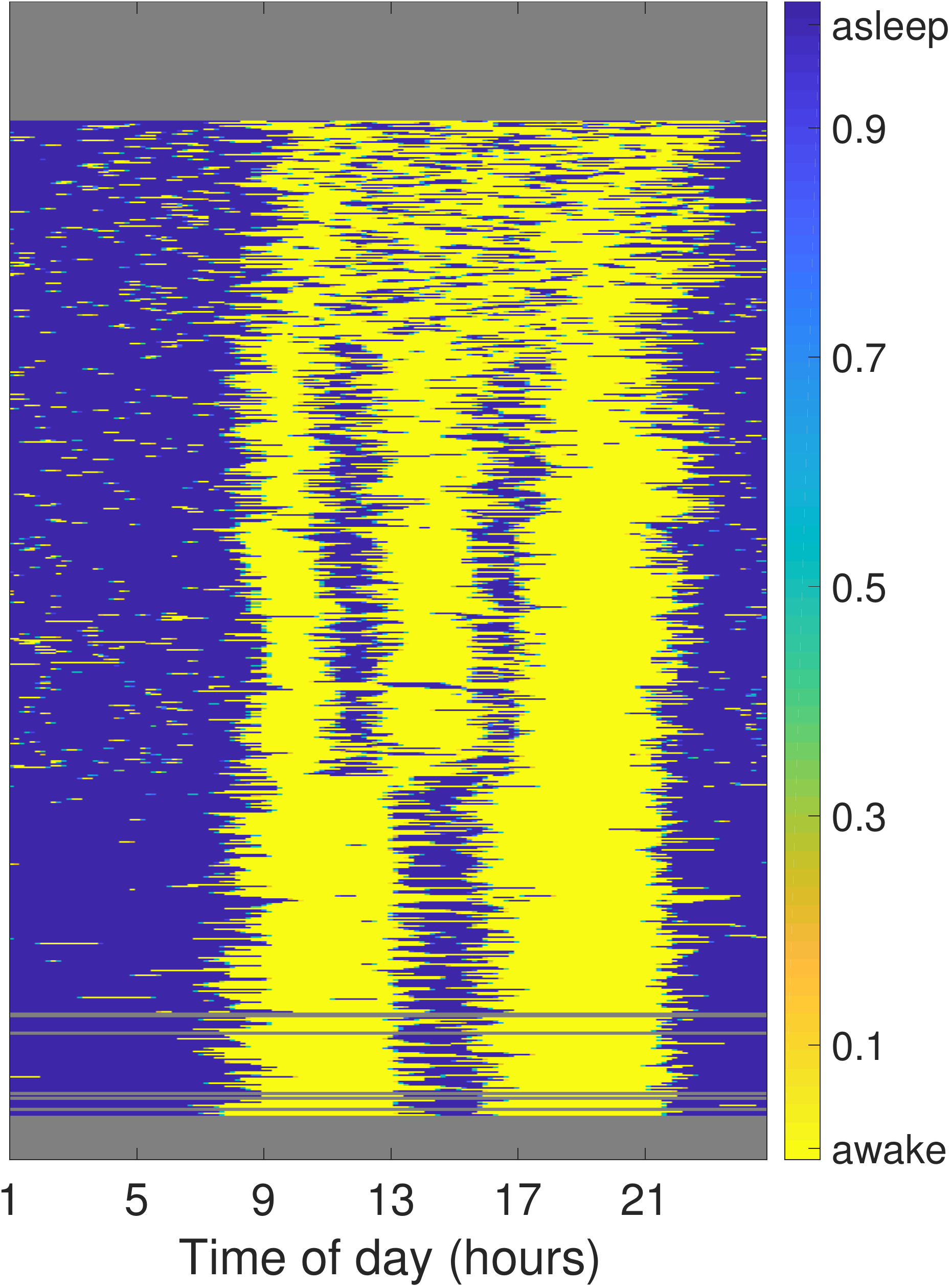}\\
 & \includegraphics[scale=0.15]{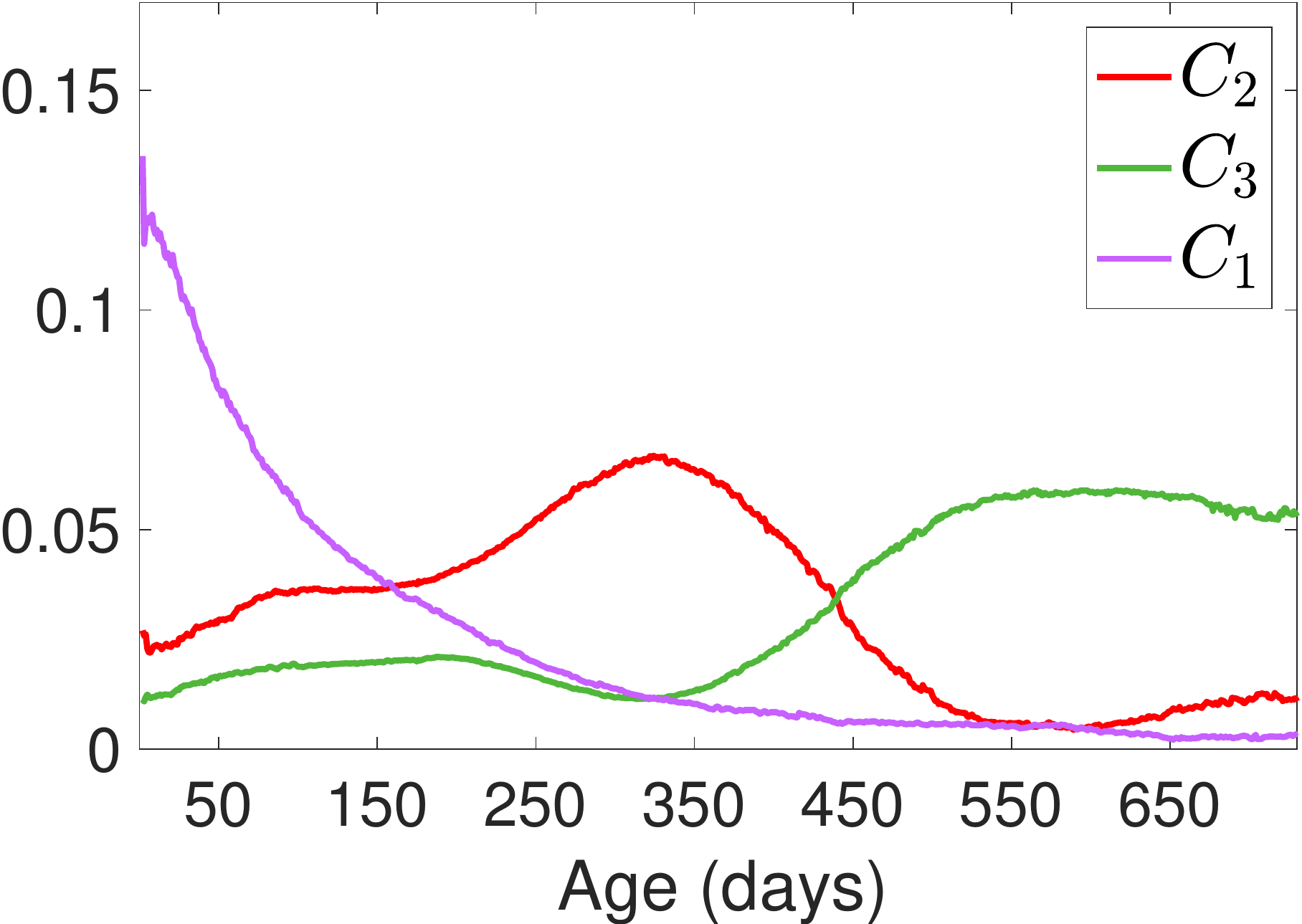} & \includegraphics[scale=0.15]{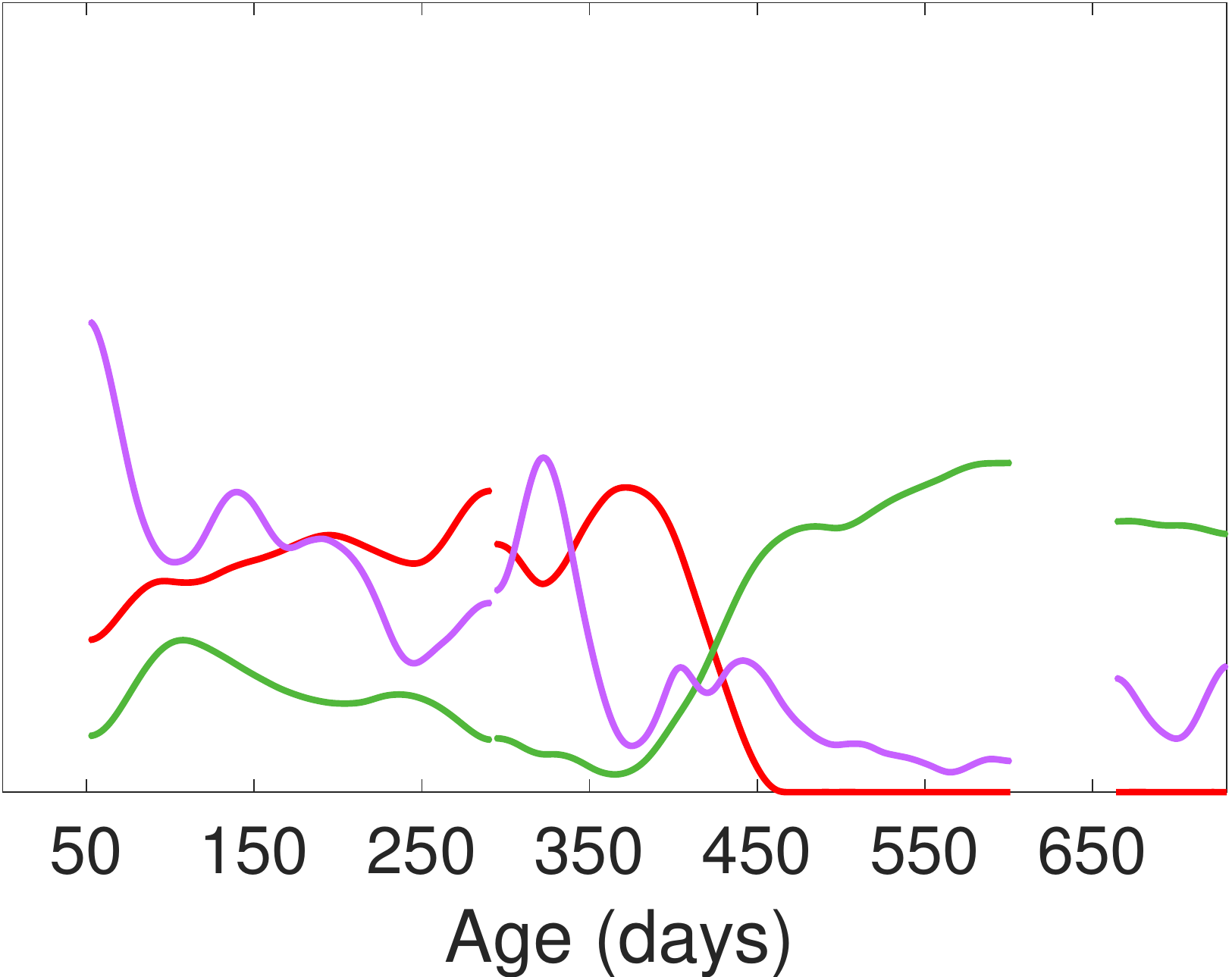}  &  \includegraphics[scale=0.15]{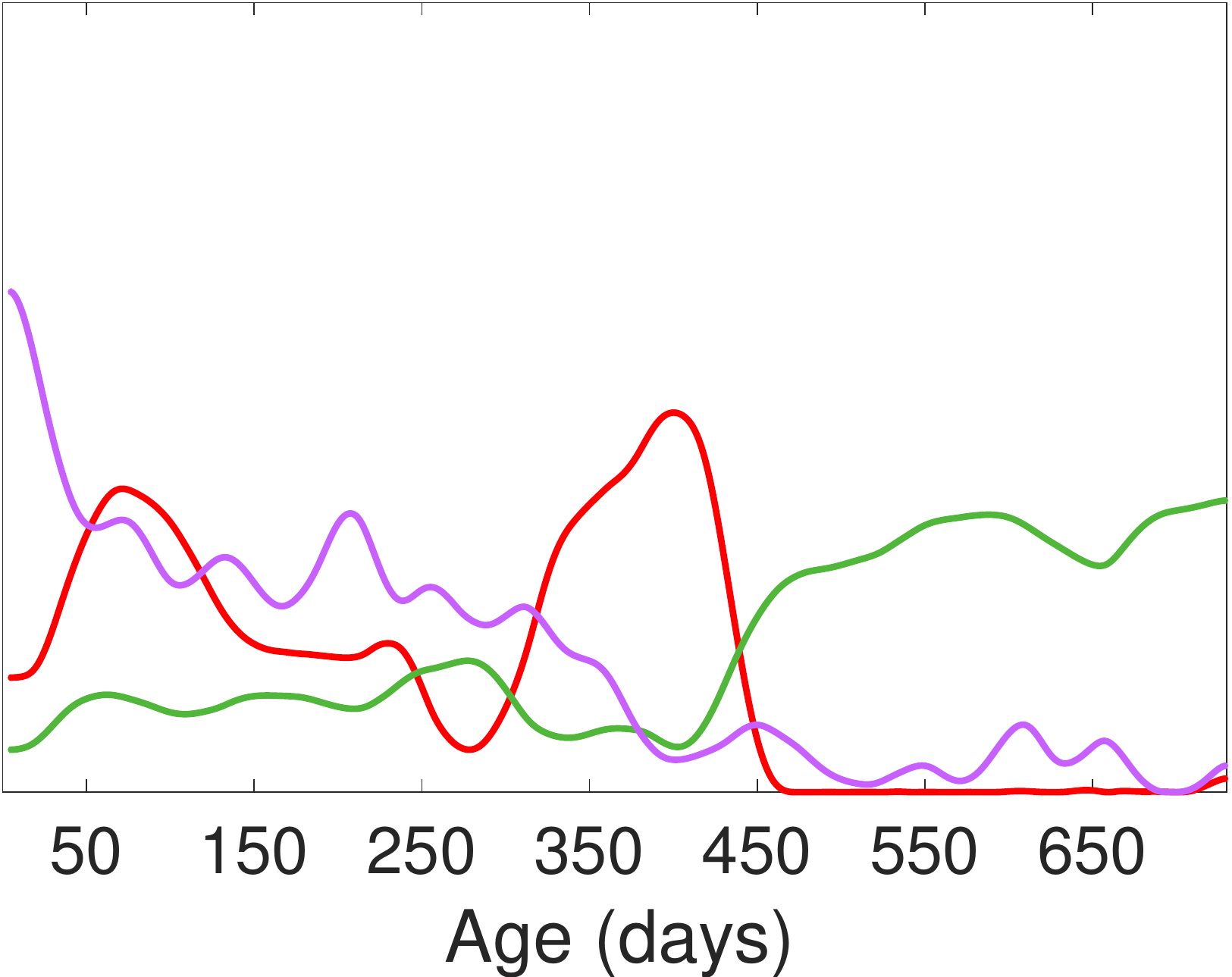}&  \includegraphics[scale=0.15]{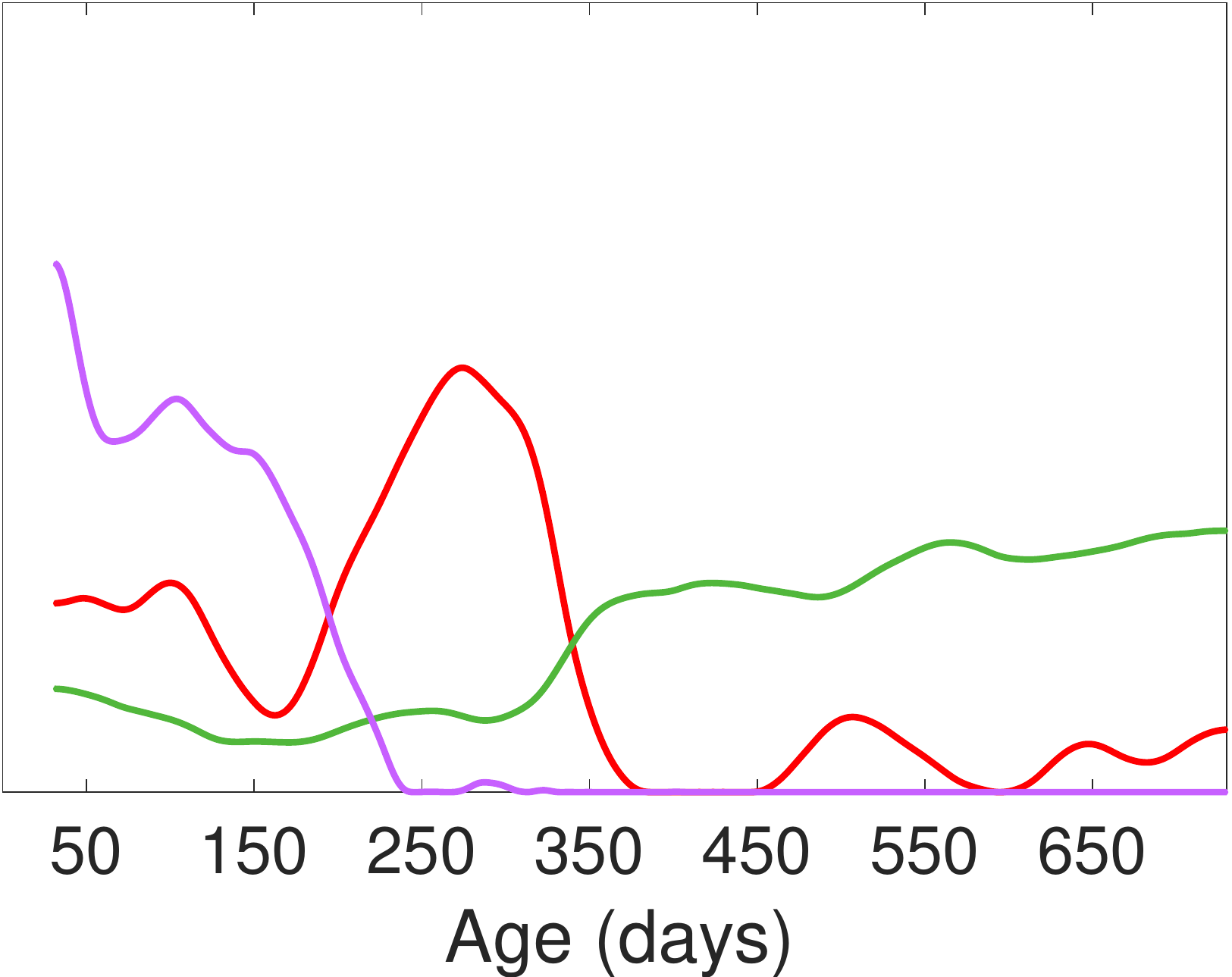} &  \includegraphics[scale=0.15]{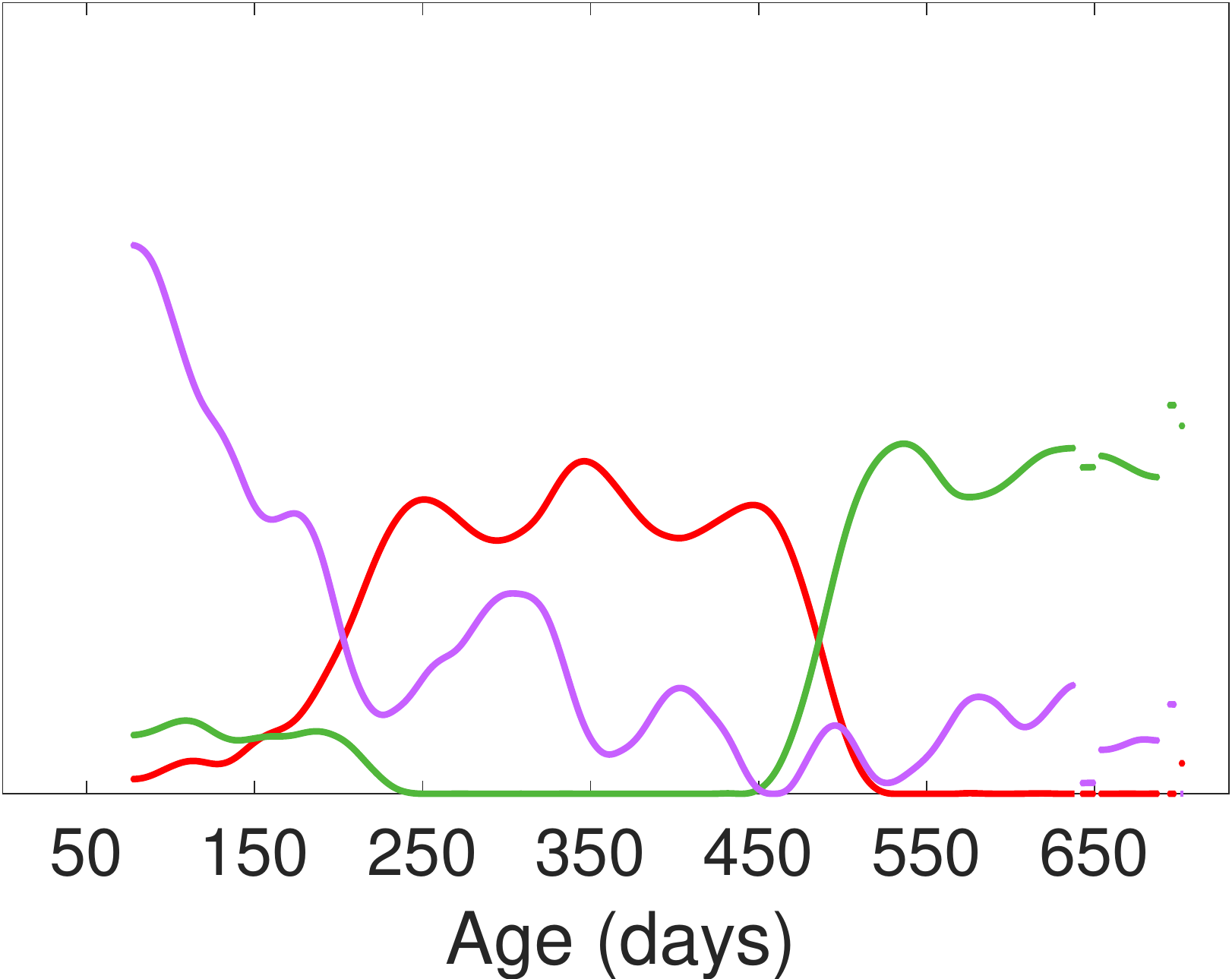}\\
2 & \includegraphics[scale=0.15]{figs/clustering_results/125/cluster_low_rank_center_2} & \includegraphics[scale=0.15]{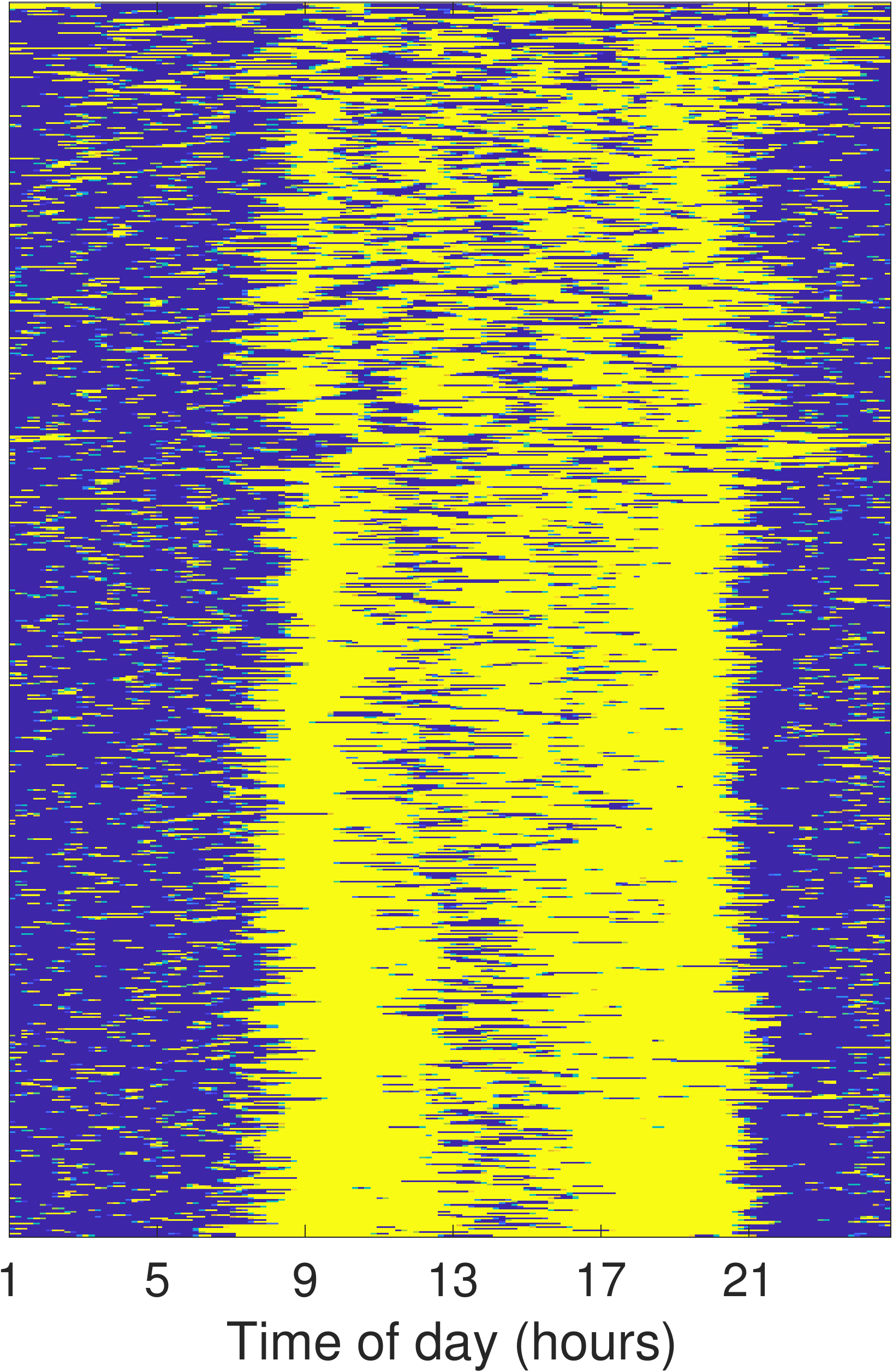}  &  \includegraphics[scale=0.15]{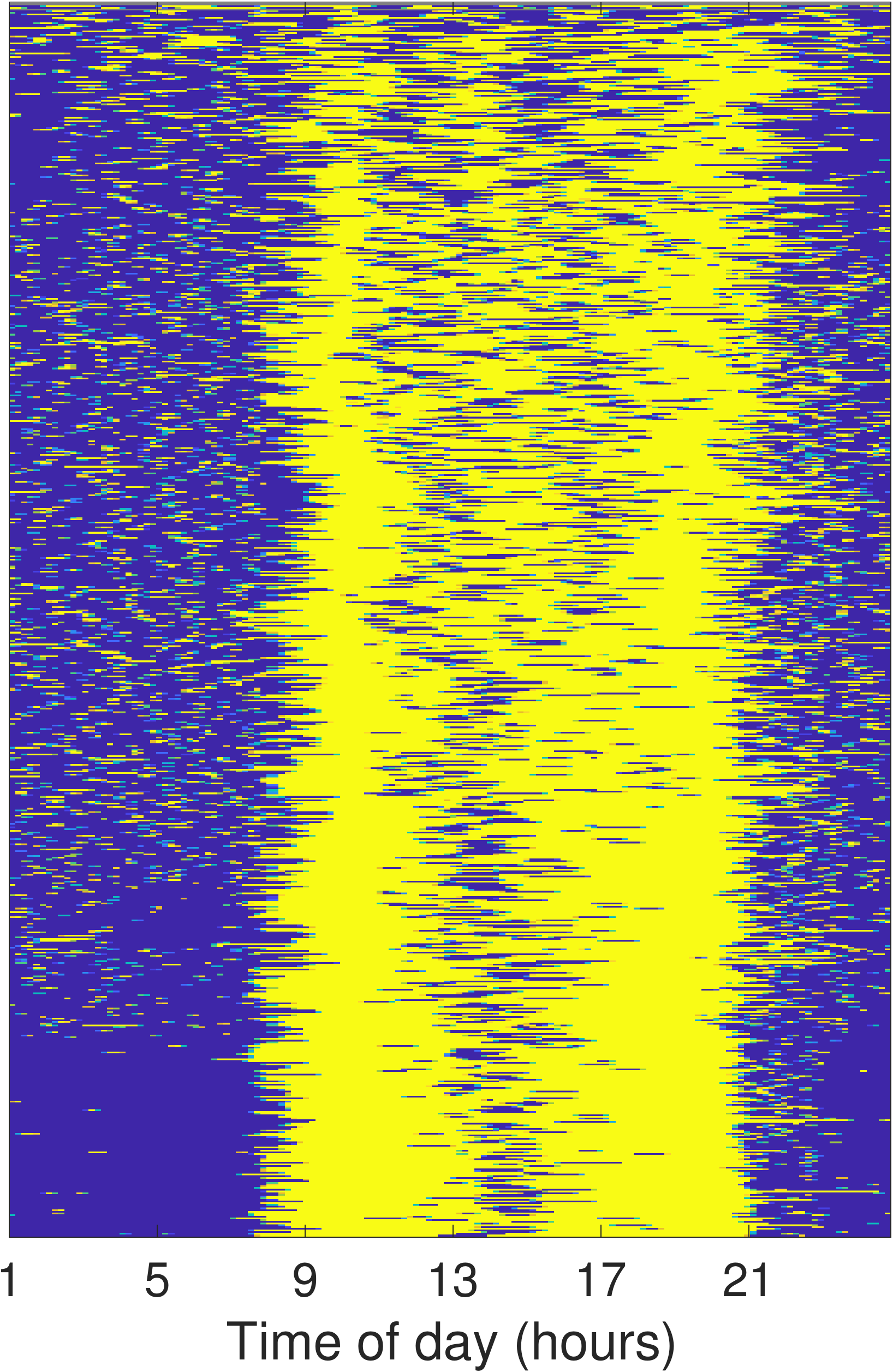}&  \includegraphics[scale=0.15]{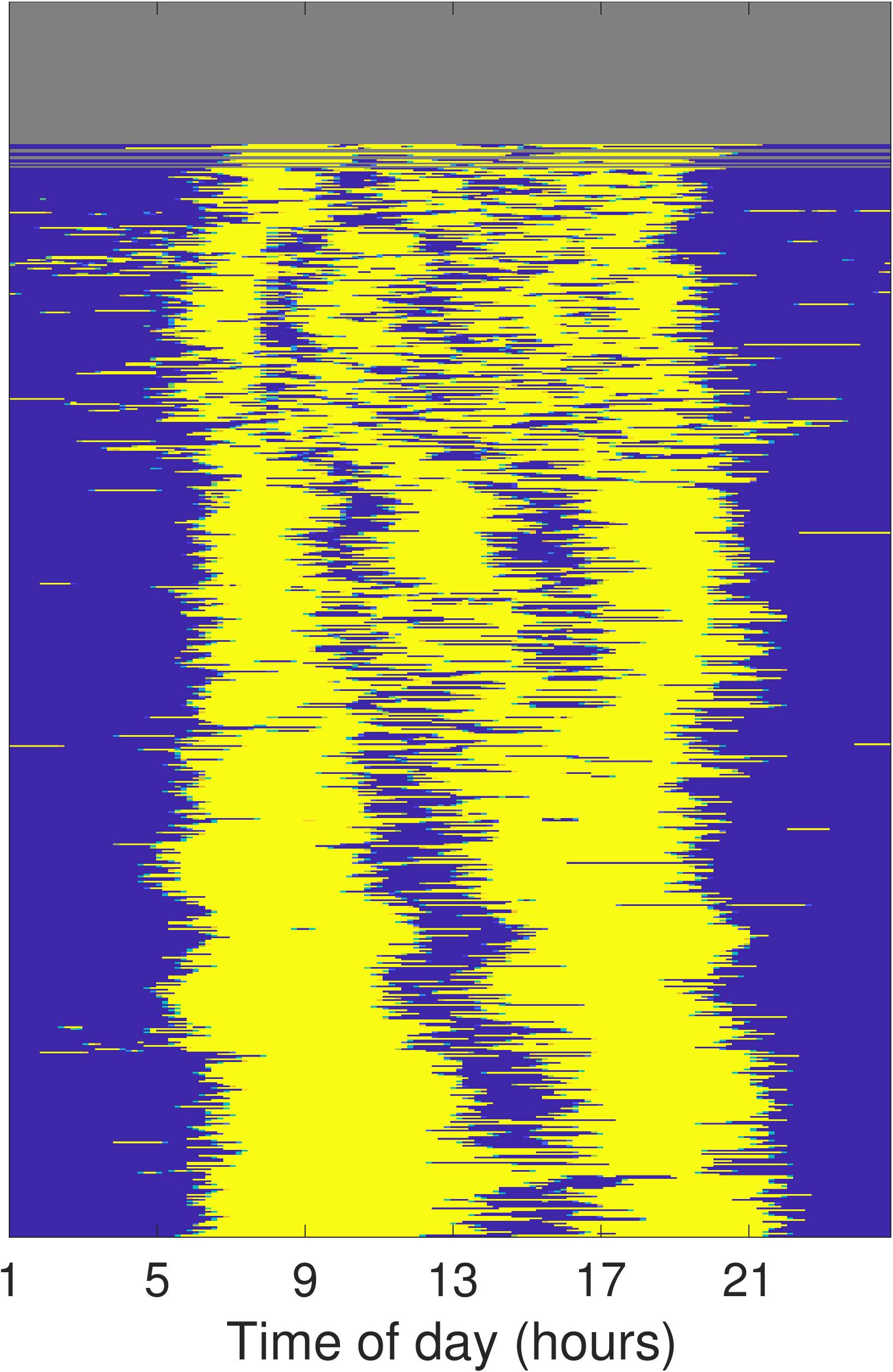} &  \includegraphics[scale=0.15]{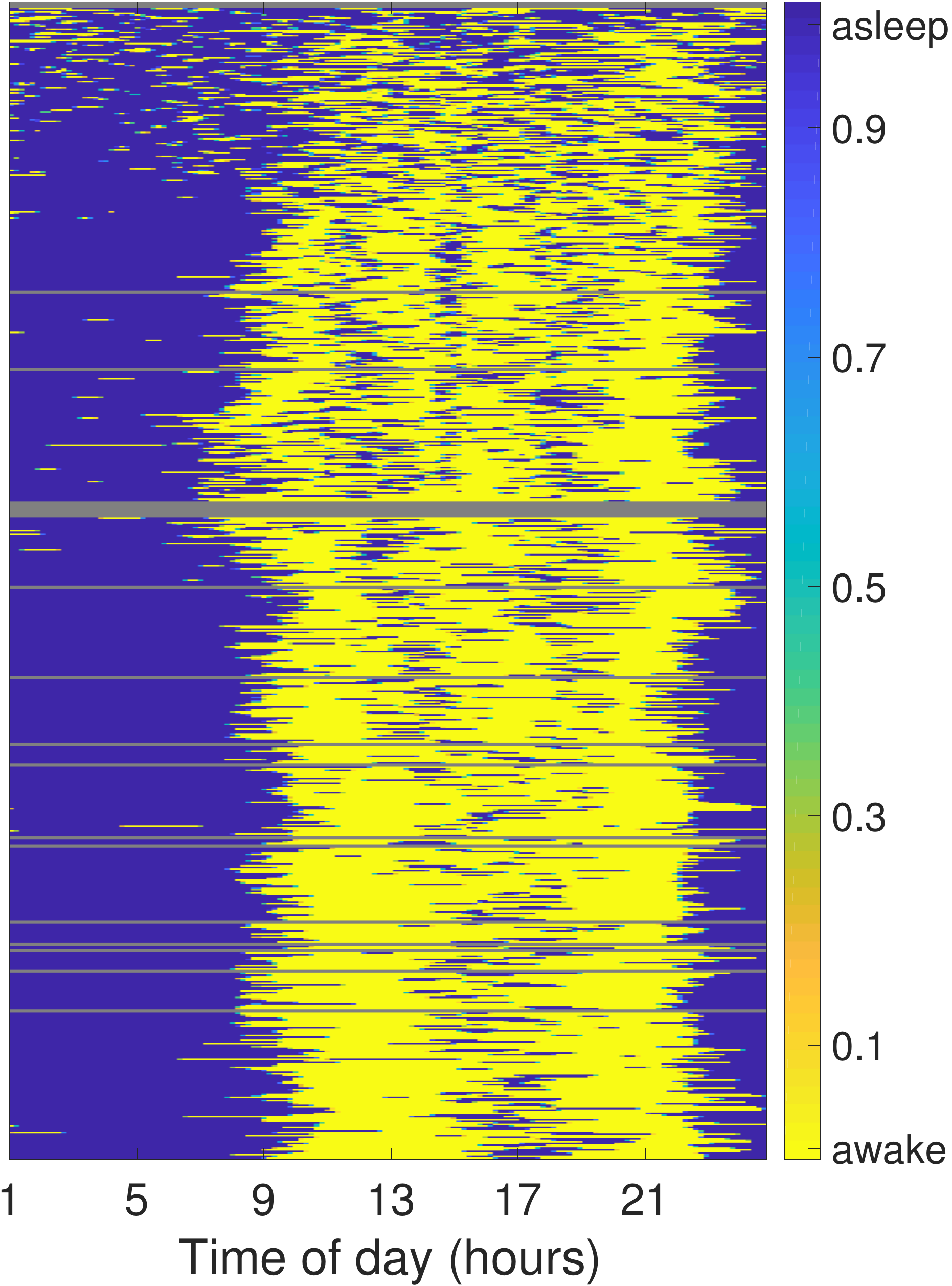}\\
 & \includegraphics[scale=0.15]{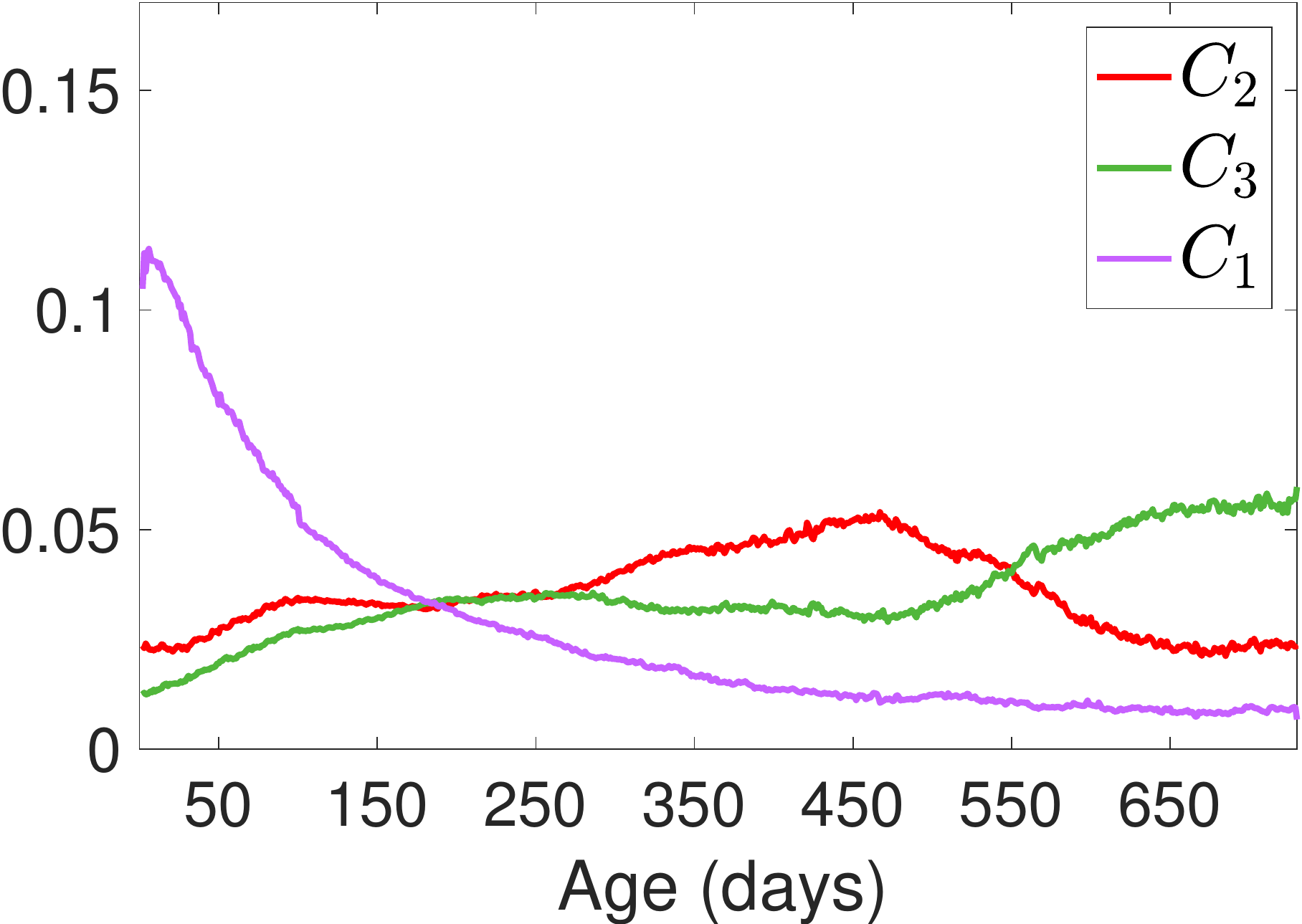} & \includegraphics[scale=0.15]{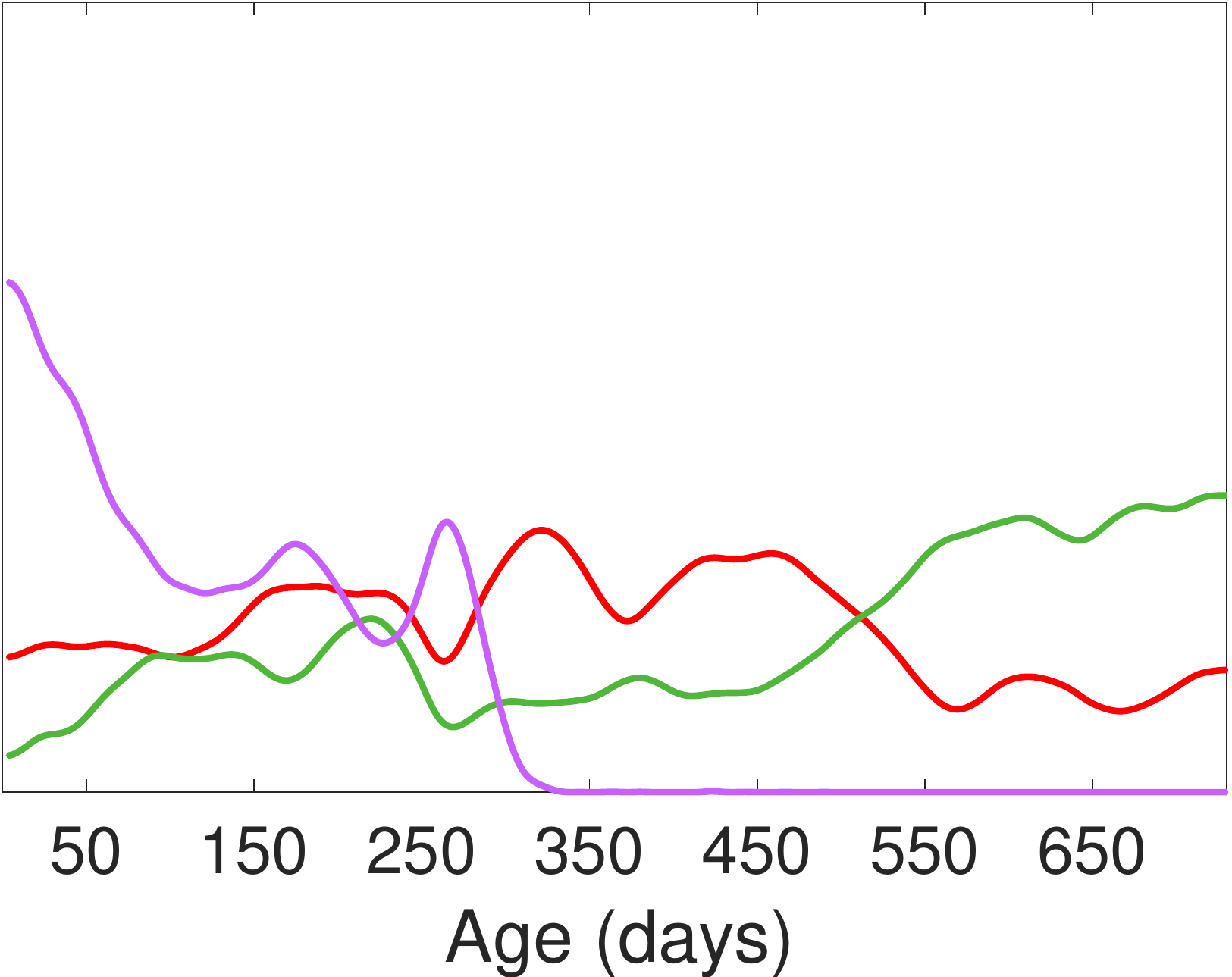} &  \includegraphics[scale=0.15]{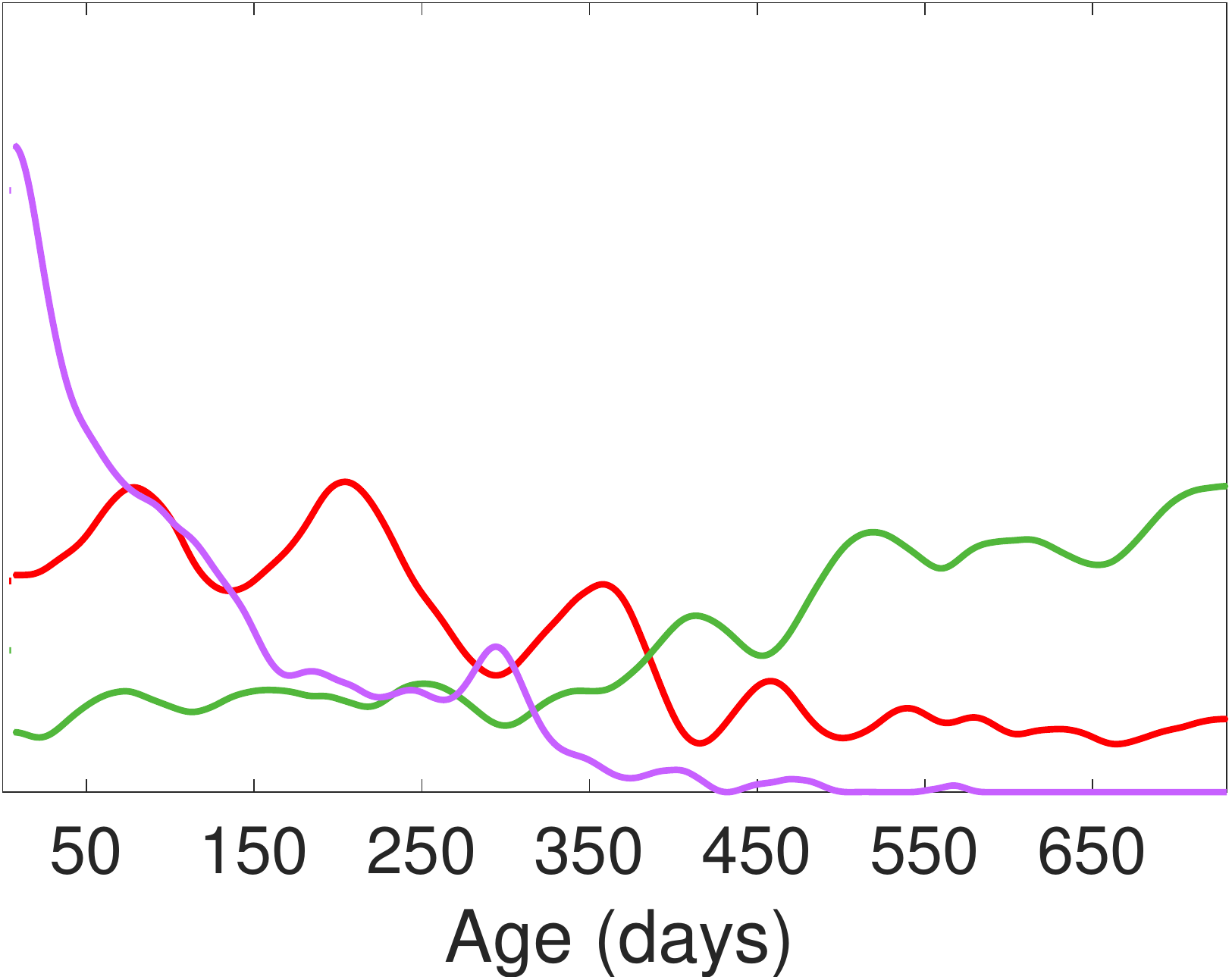}&  \includegraphics[scale=0.15]{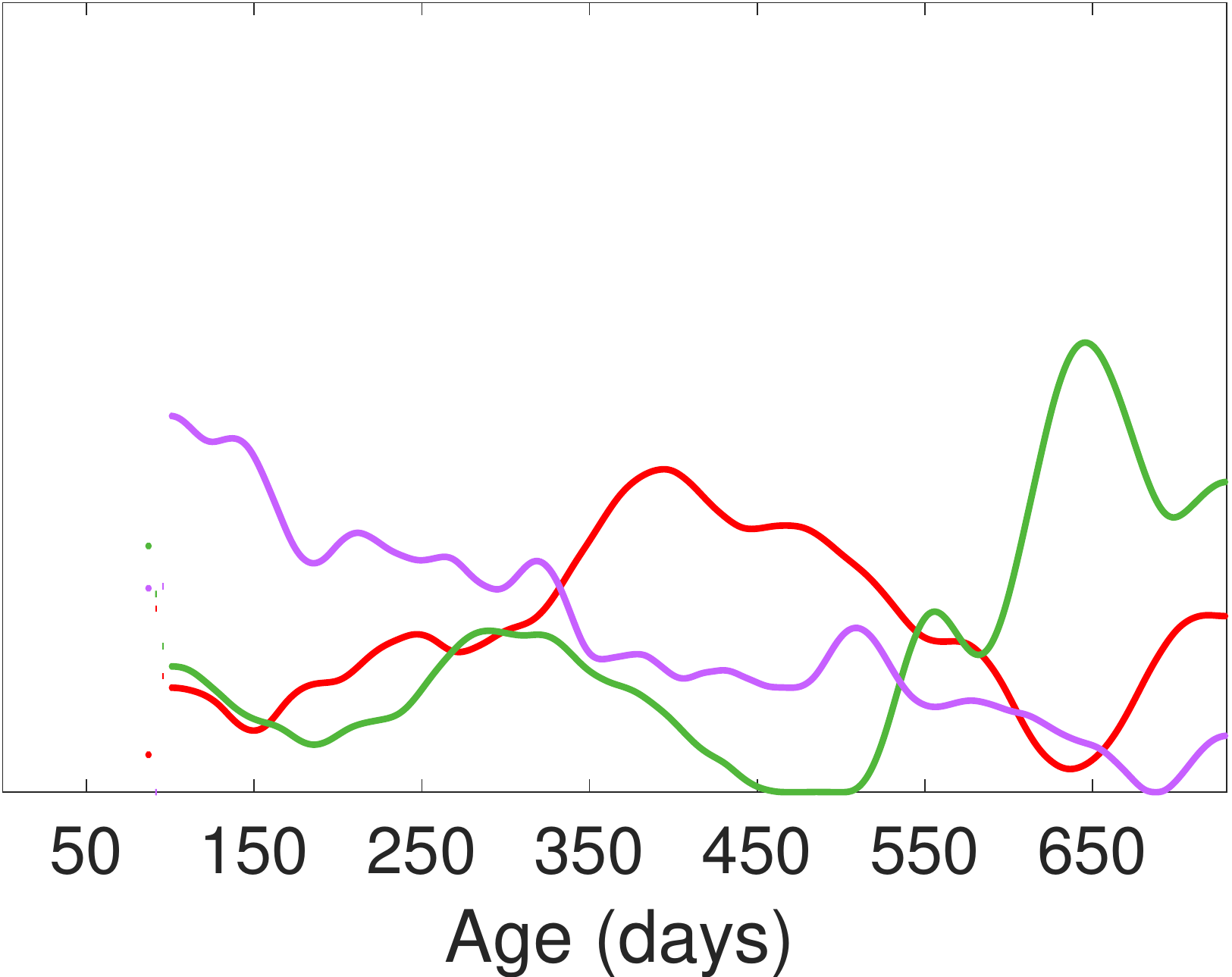} &  \includegraphics[scale=0.15]{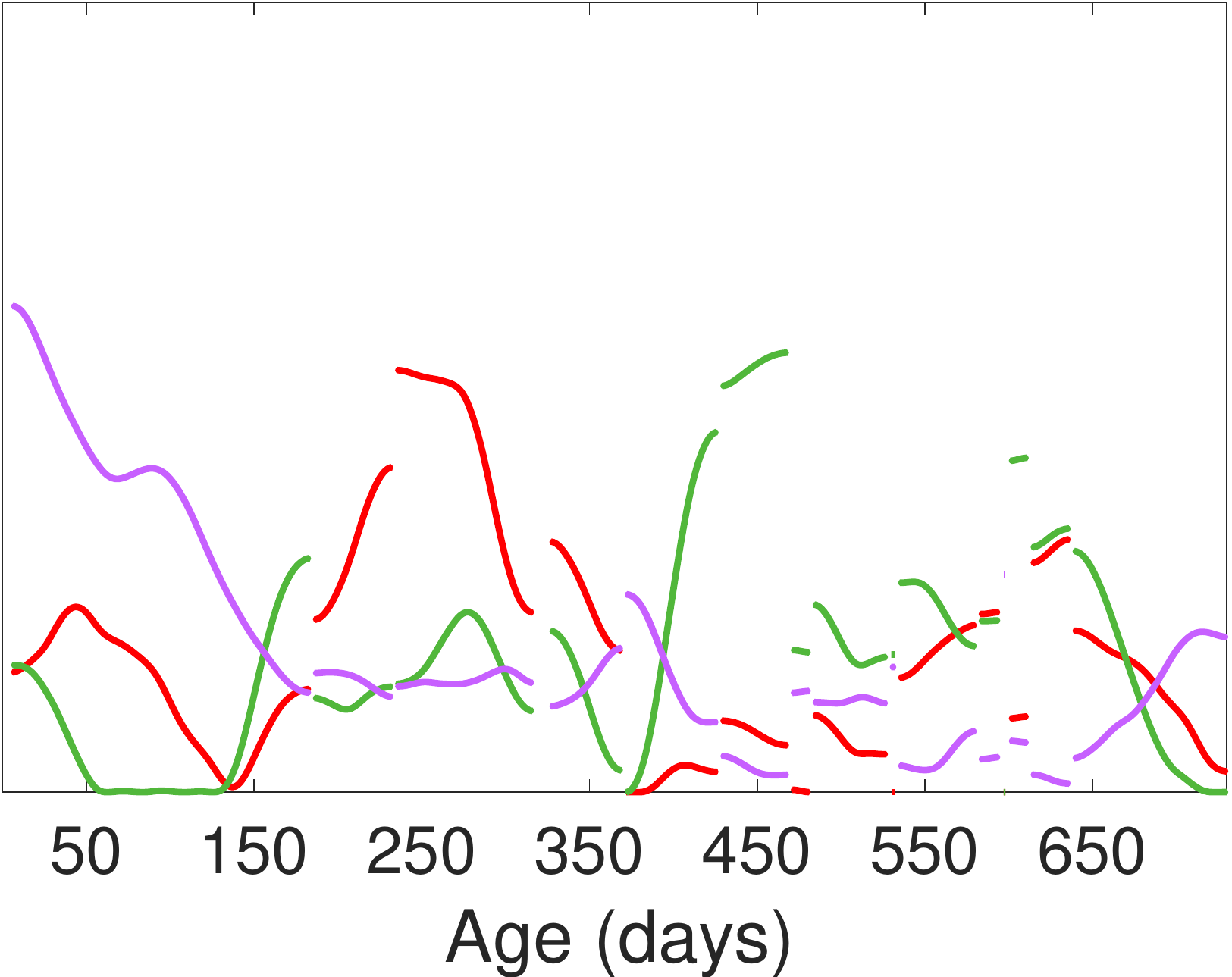}
 \end{tabular}
 }
 \caption{Results of applying Lloyd's algorithm to perform k-means clustering with $k :=2$ based on the metric $d_{\op{NMF-TS}}$ defined in Eq.~\eqref{eq:metric}. For the two clusters we show the low-rank cluster mean (1st column), the closest infant to the mean (2nd column), and the first quartile, median and third quartile infants in terms of distance to the mean out of infants with at least 80\% available data (3rd, 4th and 5th columns respectively). Below each image we also show the coefficients corresponding to the three sleeping patterns included in the metric. The first cluster contains 359 infants. The second cluster contains 342. Gray values indicate missing data.}
 \label{fig:rnmf_results}
\end{figure}

\begin{figure}[tp]
\hspace{-0.5cm}
\begin{tabular}{ >{\centering\arraybackslash}m{0.18\linewidth} >{\centering\arraybackslash}m{0.18\linewidth} >{\centering\arraybackslash}m{0.18\linewidth} >{\centering\arraybackslash}m{0.18\linewidth} >{\centering\arraybackslash}m{0.18\linewidth}}
 Raw Data \vspace{0.2cm} &  Mean\vspace{0.2cm} & KR (raw data) \vspace{0.2cm} &  KR (NMF-TS) \vspace{0.2cm}  &  Rank-5 GT \vspace{0.2cm}\\
 \includegraphics[scale=0.15]{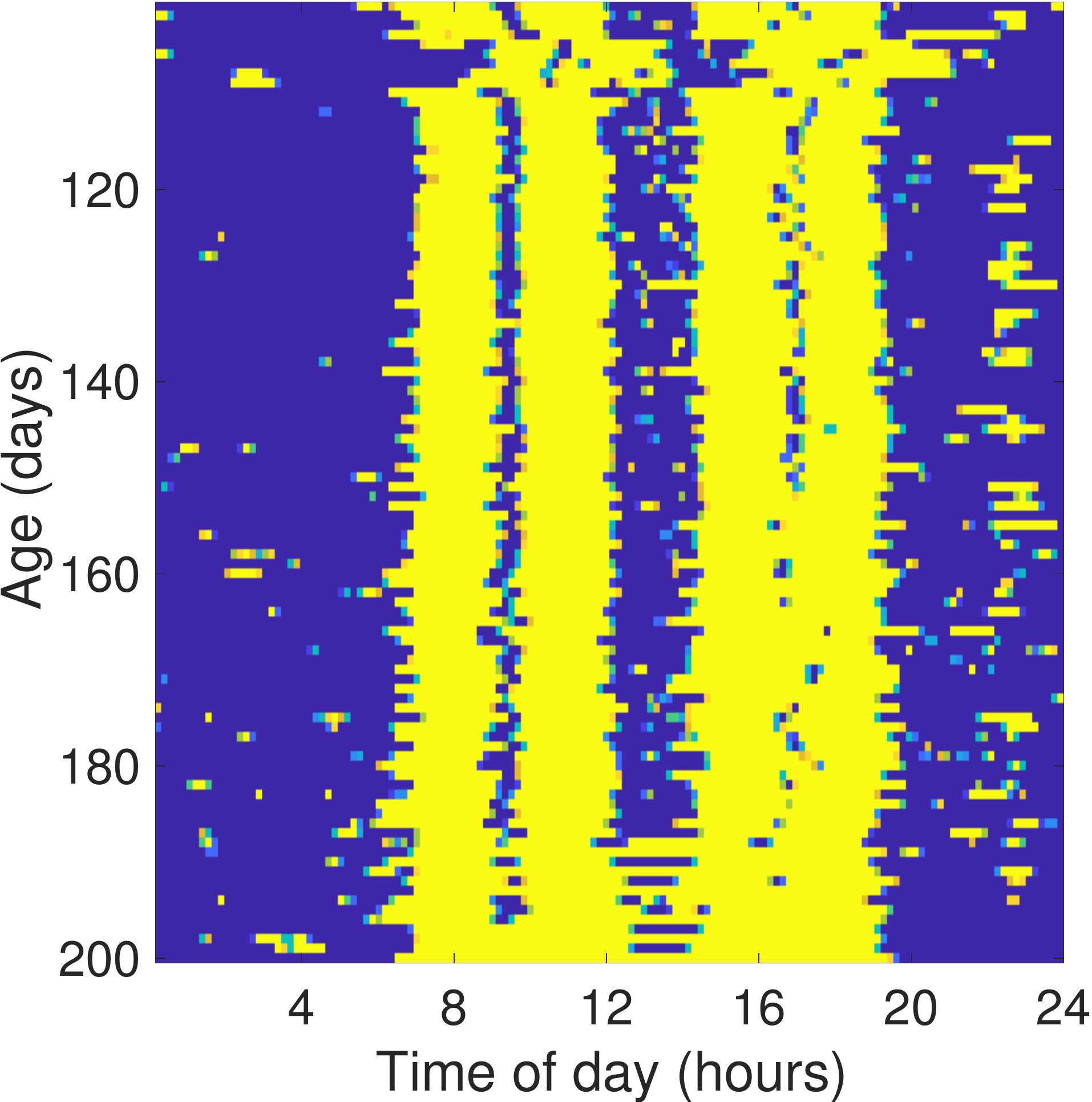} & \includegraphics[scale=0.15]{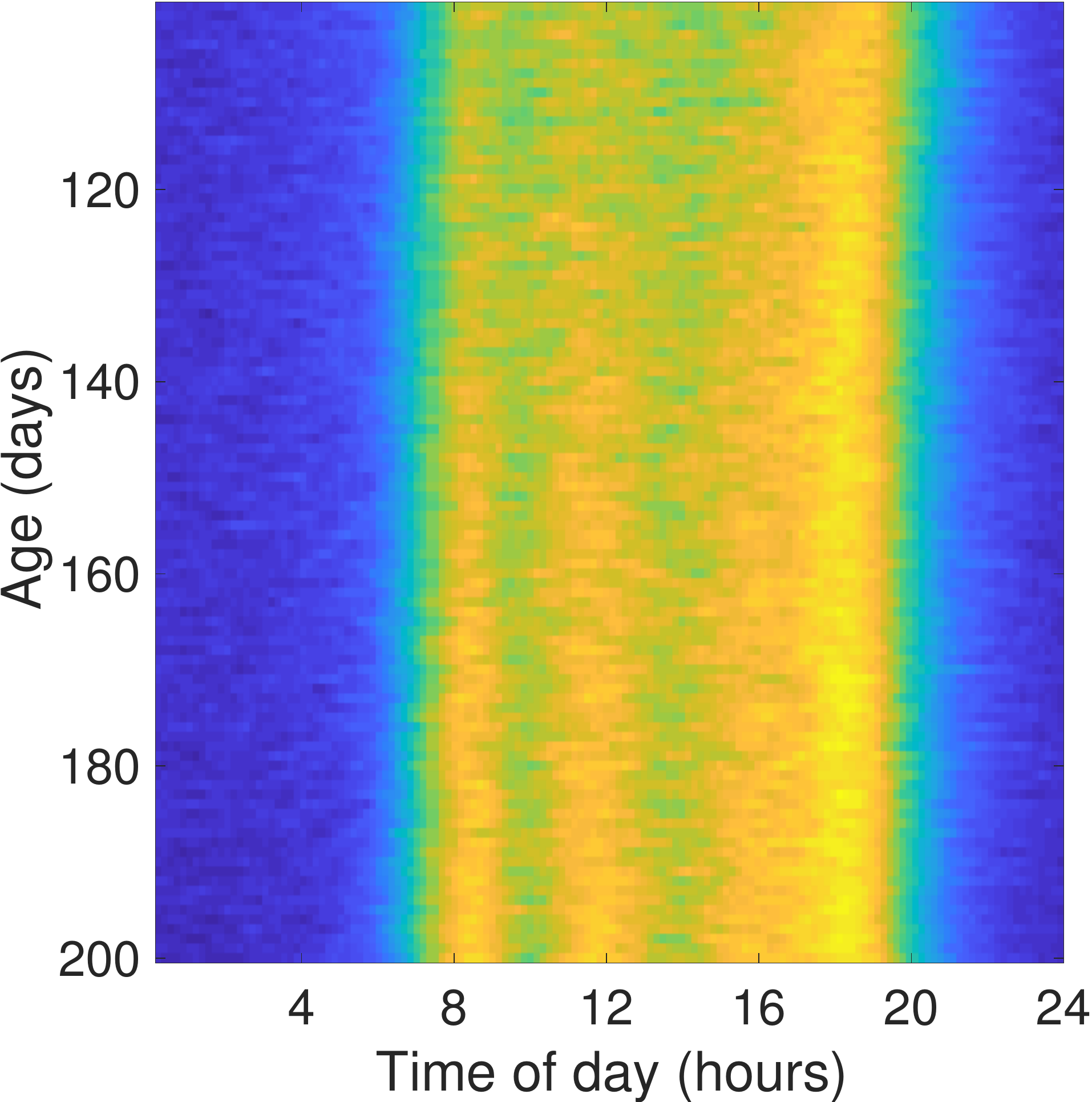} &  \includegraphics[scale=0.15]{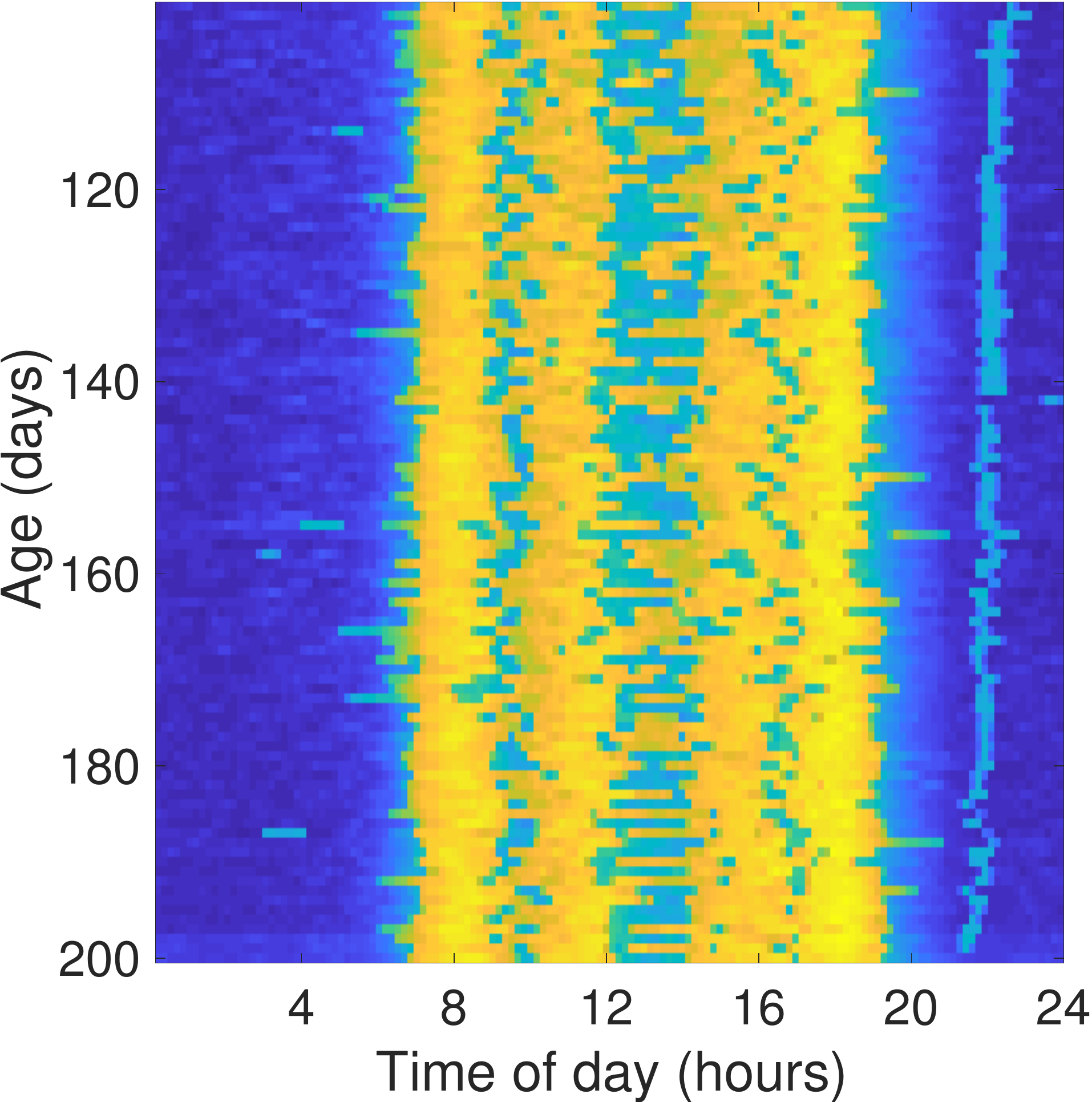}&  \includegraphics[scale=0.15]{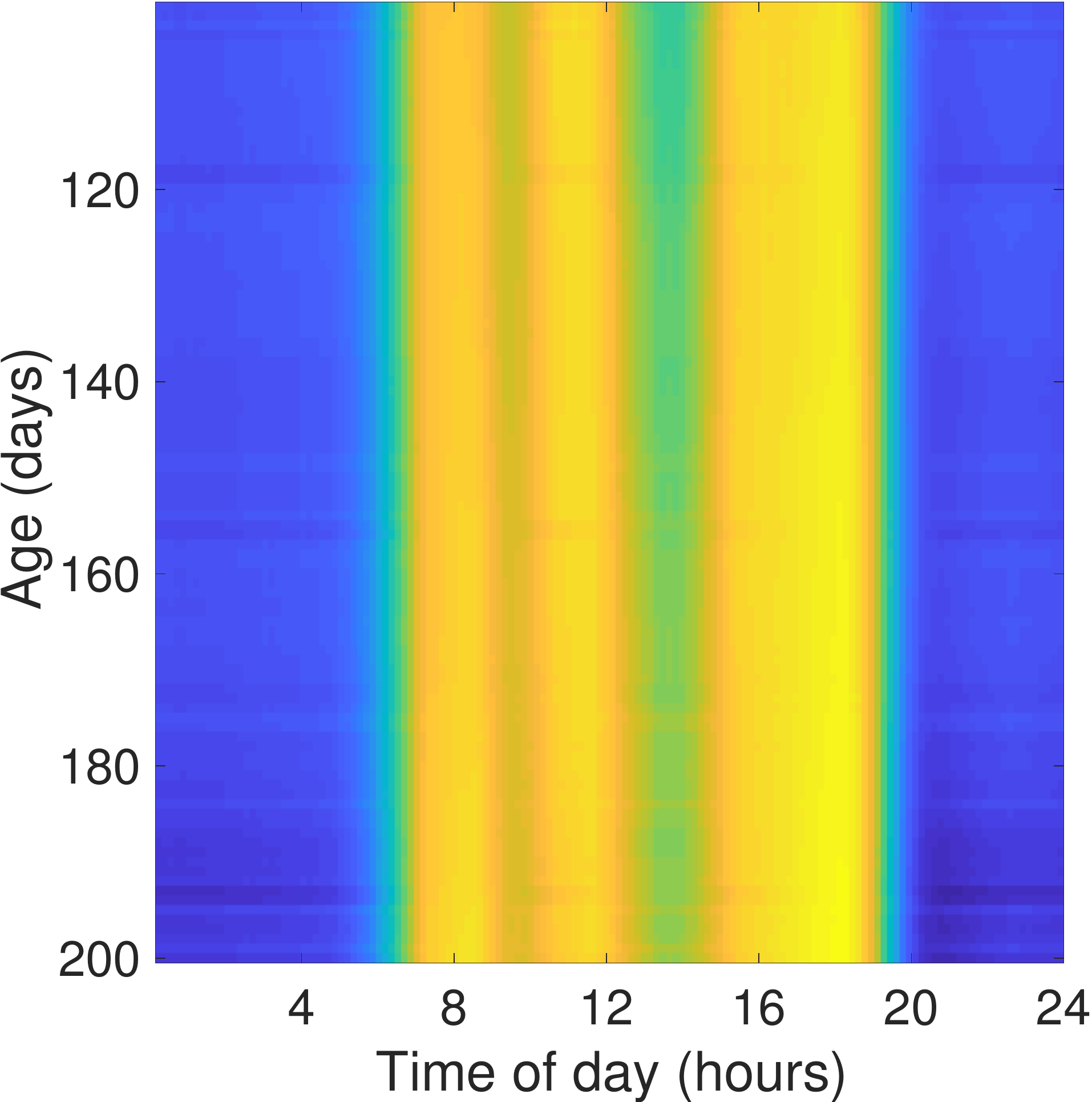} &  \includegraphics[scale=0.15]{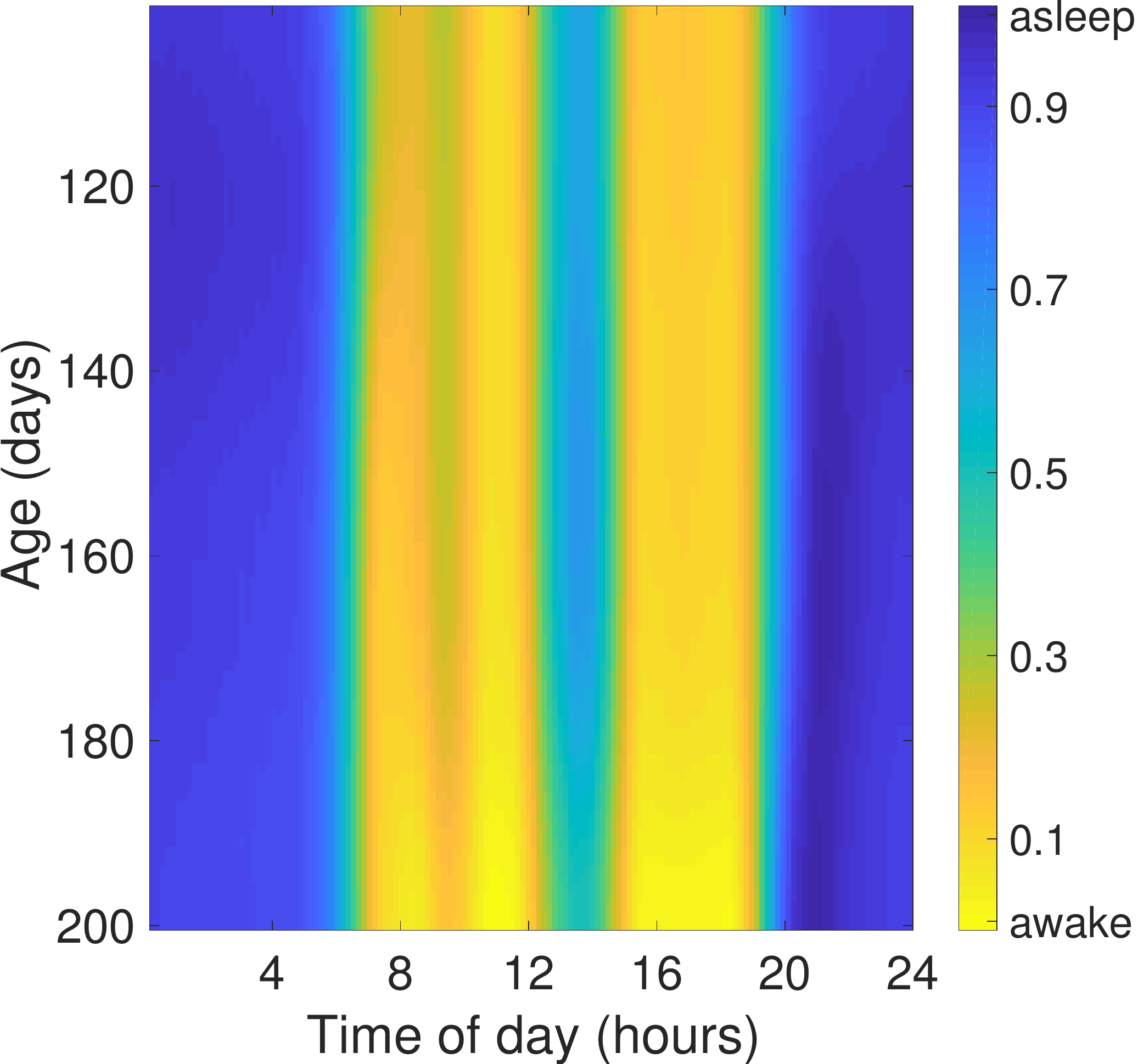}\\
\includegraphics[scale=0.15]{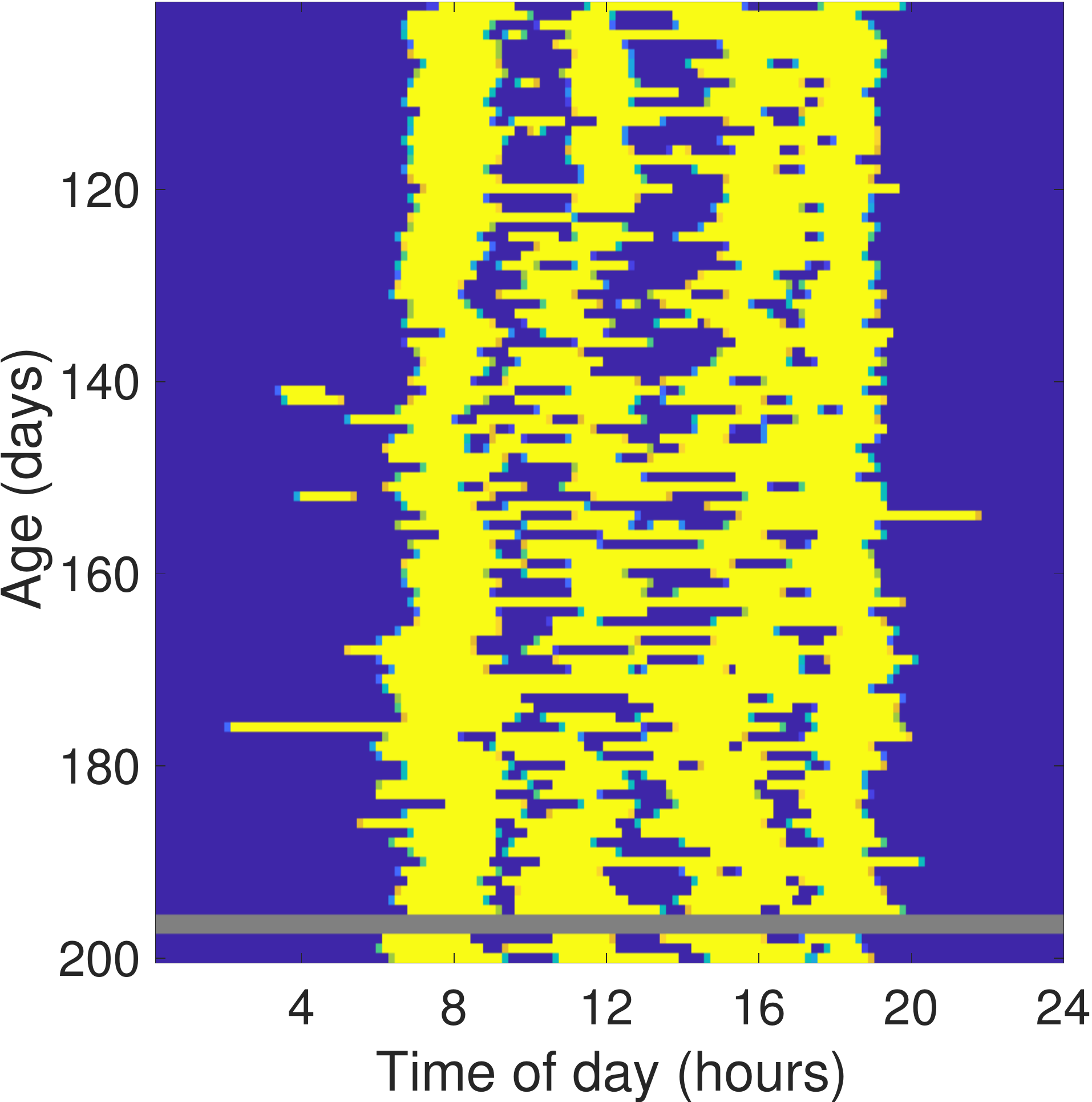} & \includegraphics[scale=0.15]{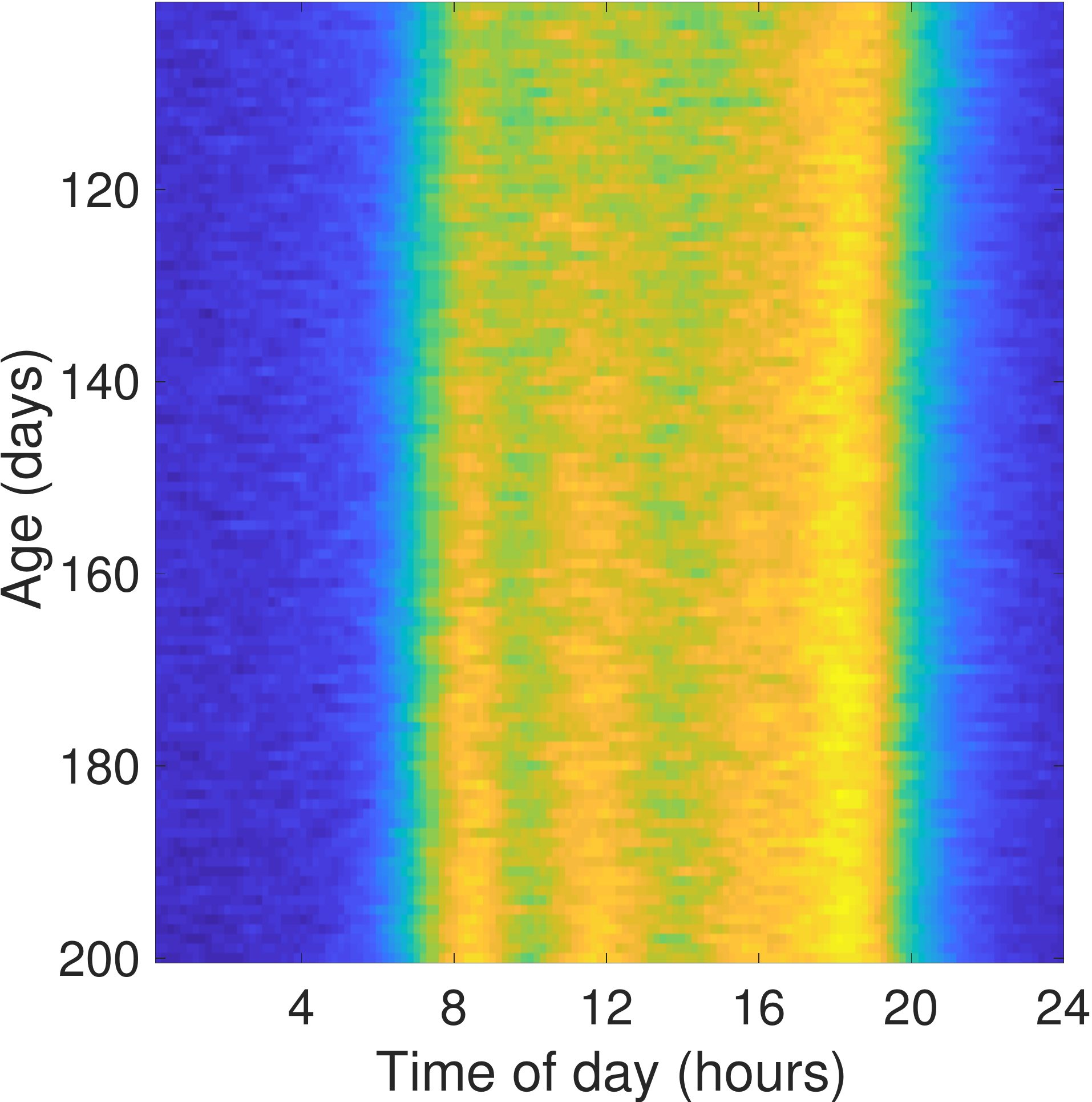}  &  \includegraphics[scale=0.15]{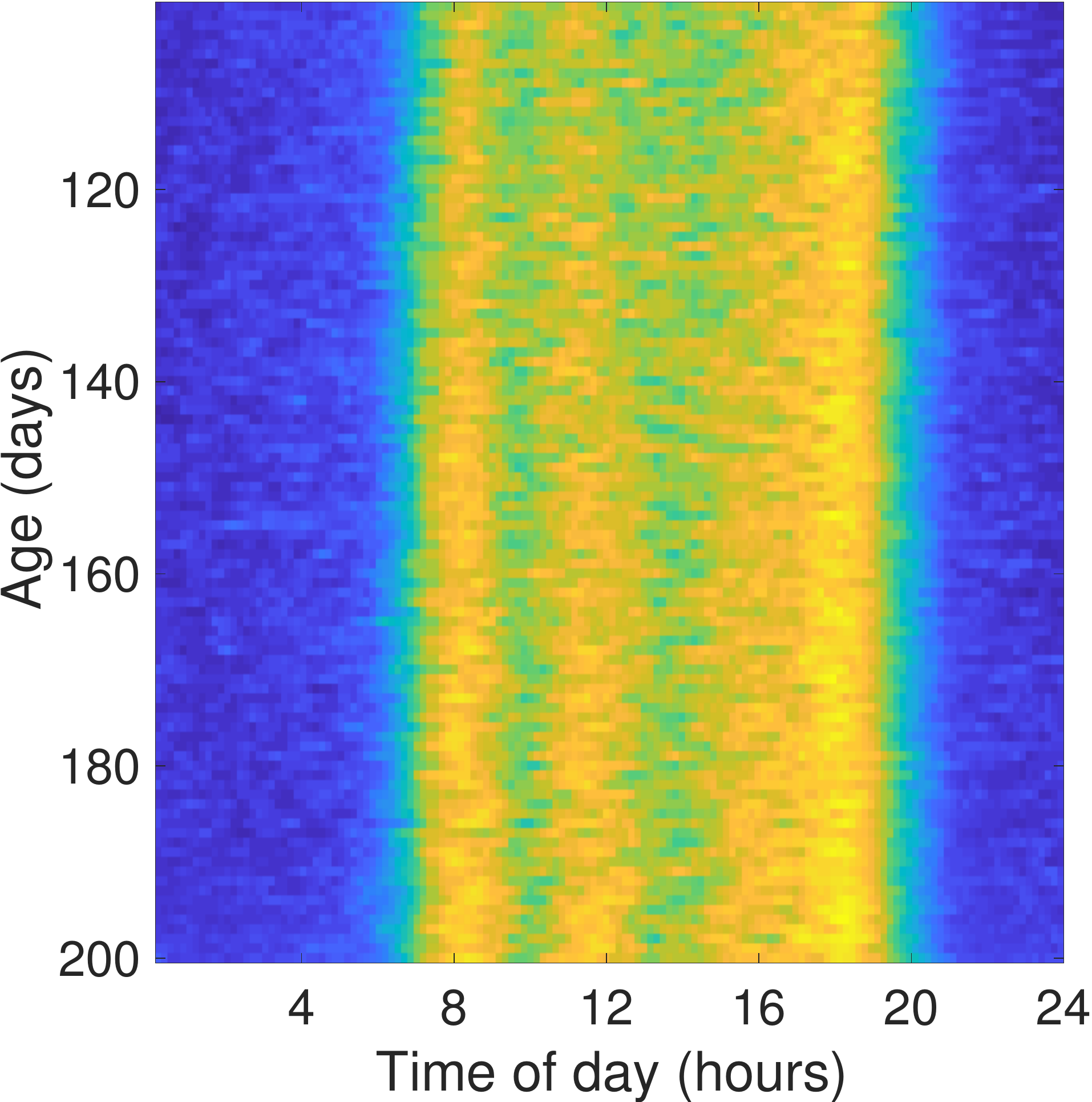}&  \includegraphics[scale=0.15]{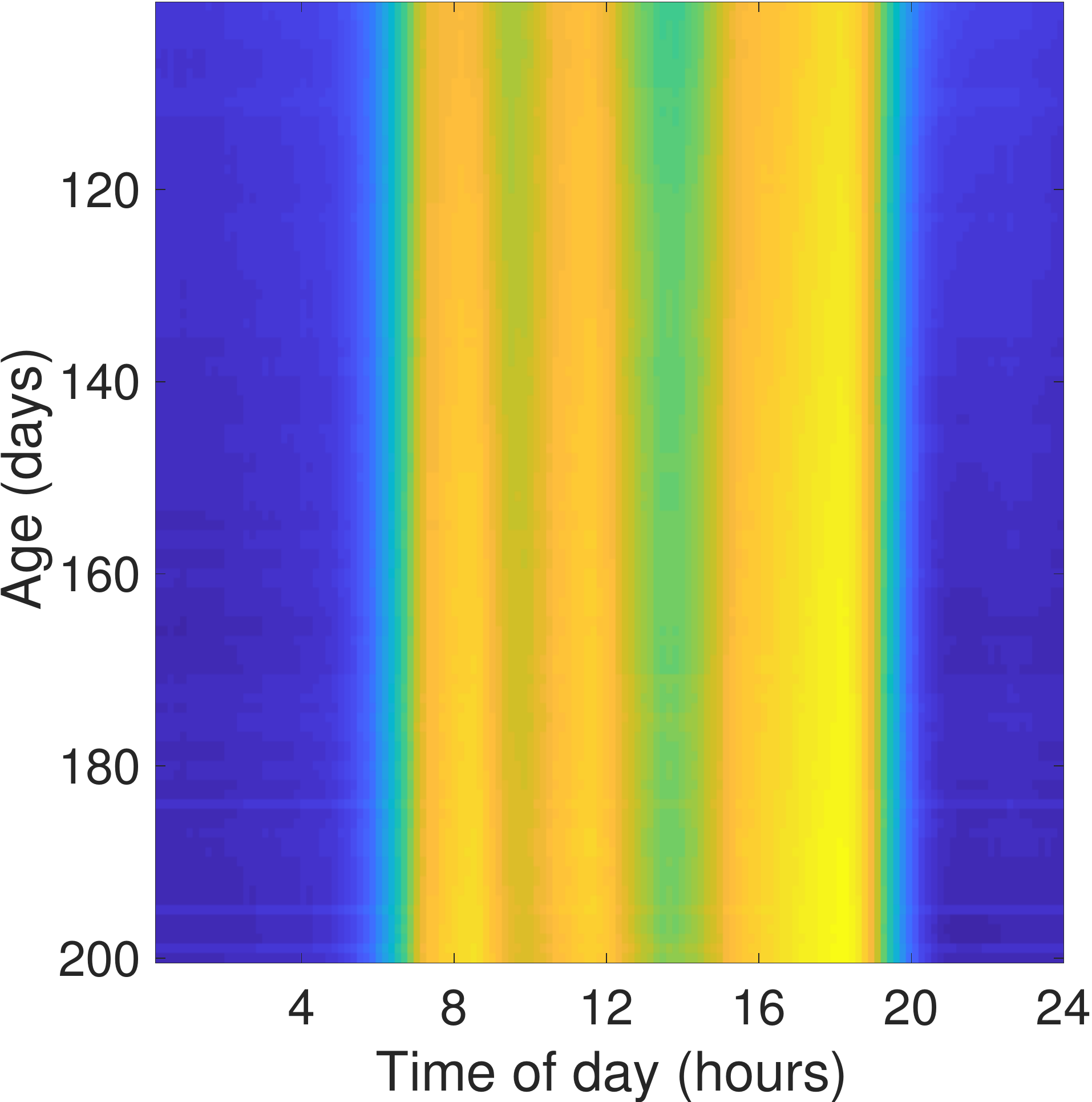} &  \includegraphics[scale=0.15]{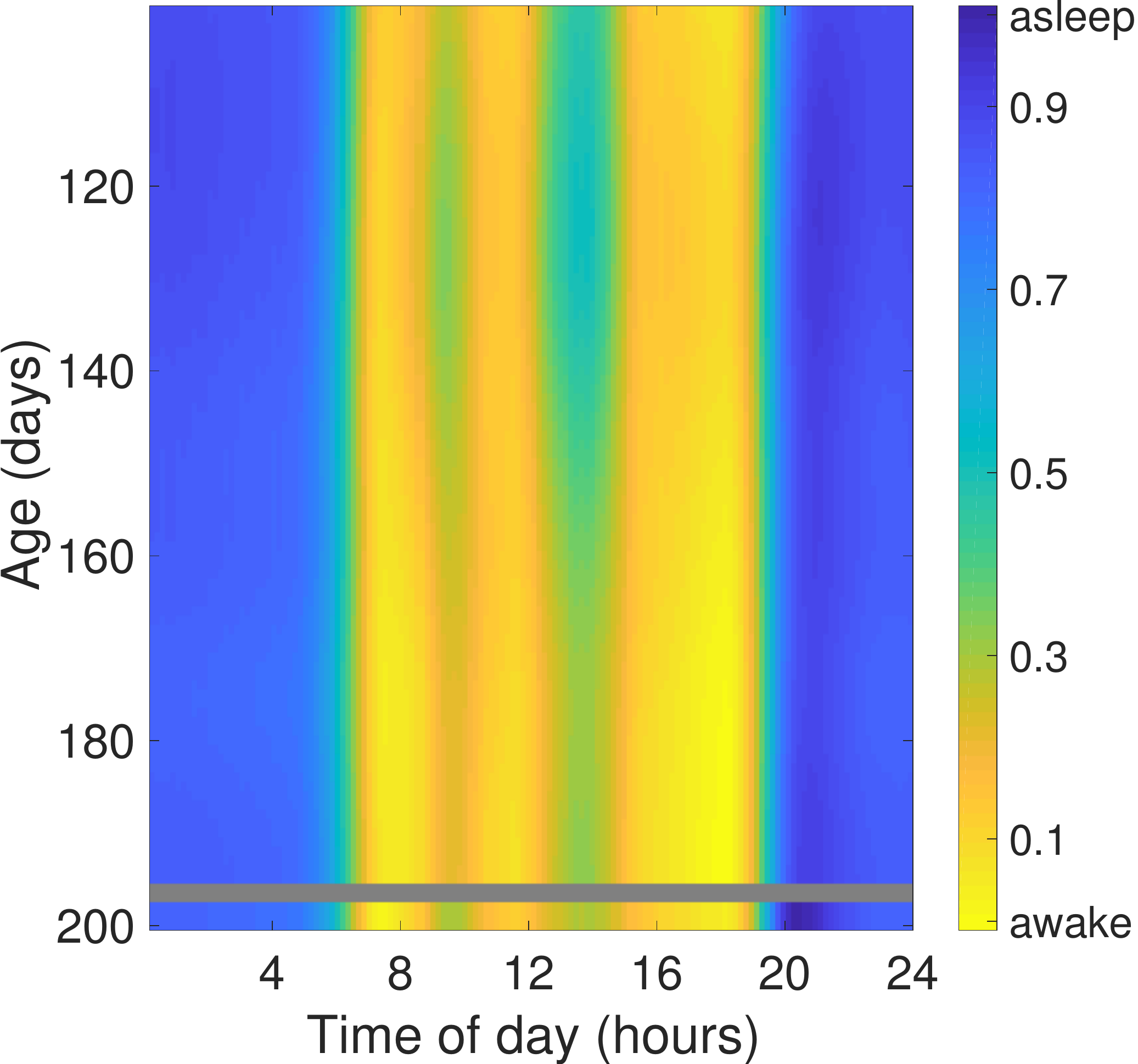}\\
\includegraphics[scale=0.15]{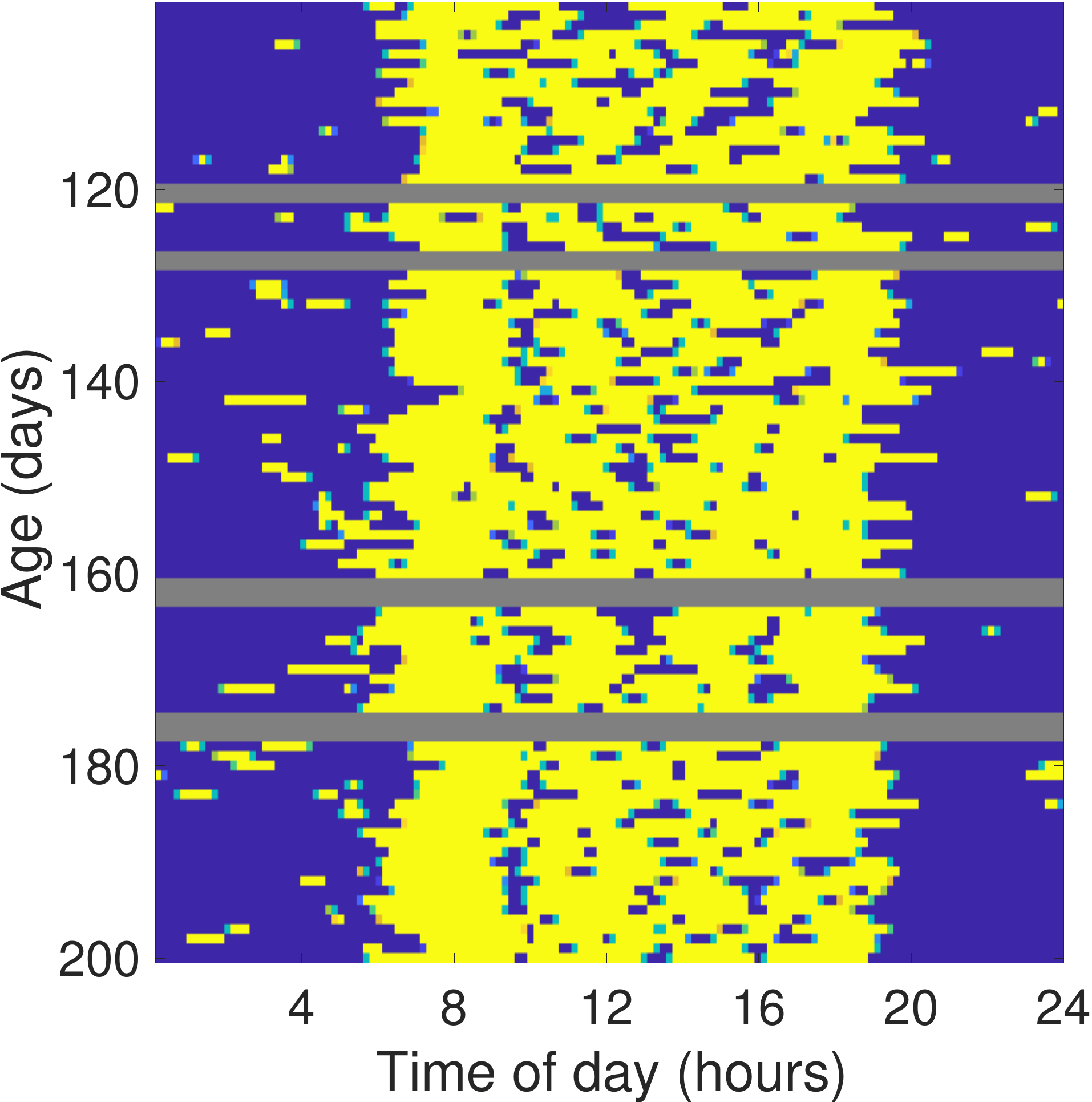} & \includegraphics[scale=0.15]{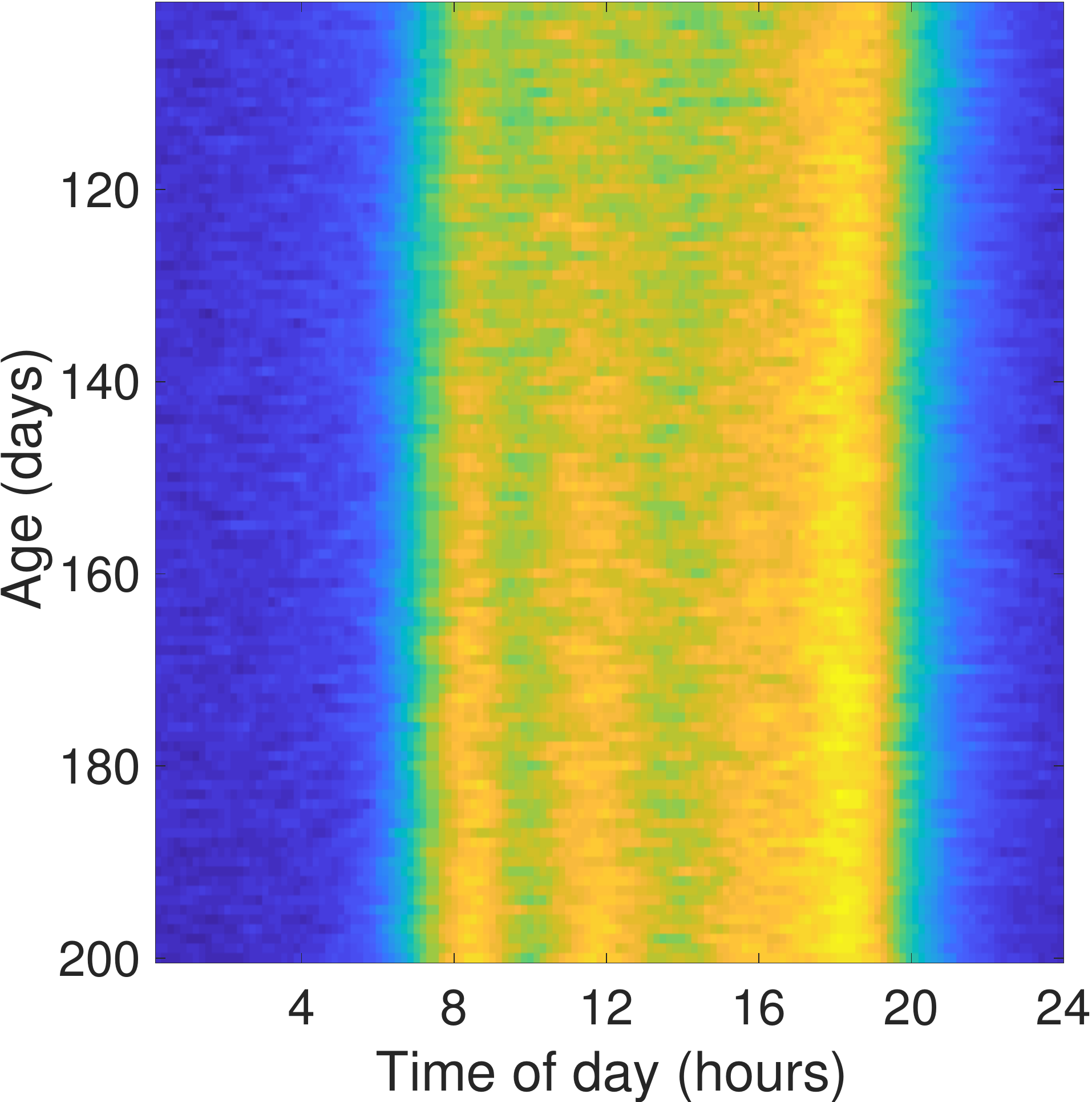}  &  \includegraphics[scale=0.15]{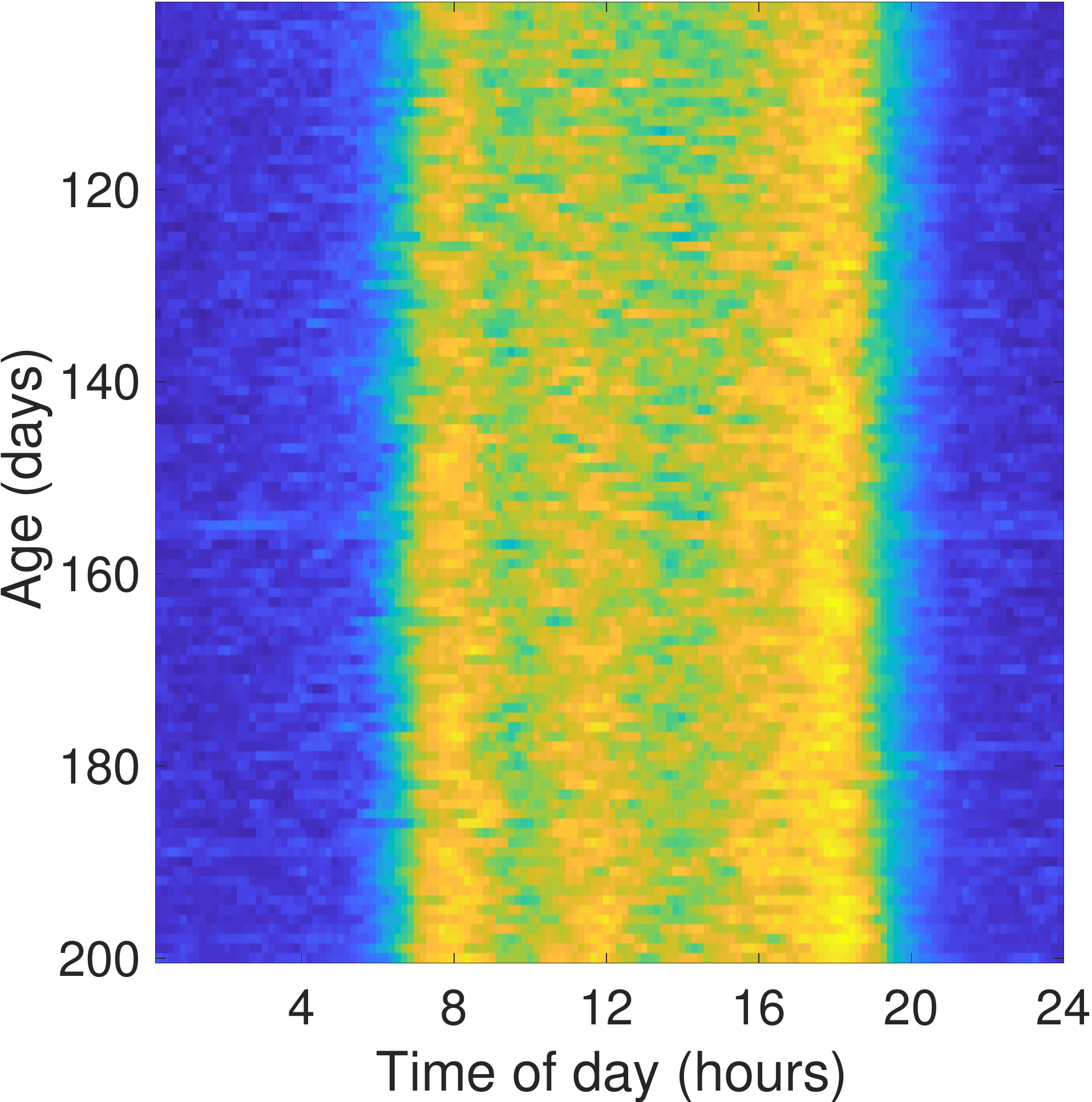}&  \includegraphics[scale=0.15]{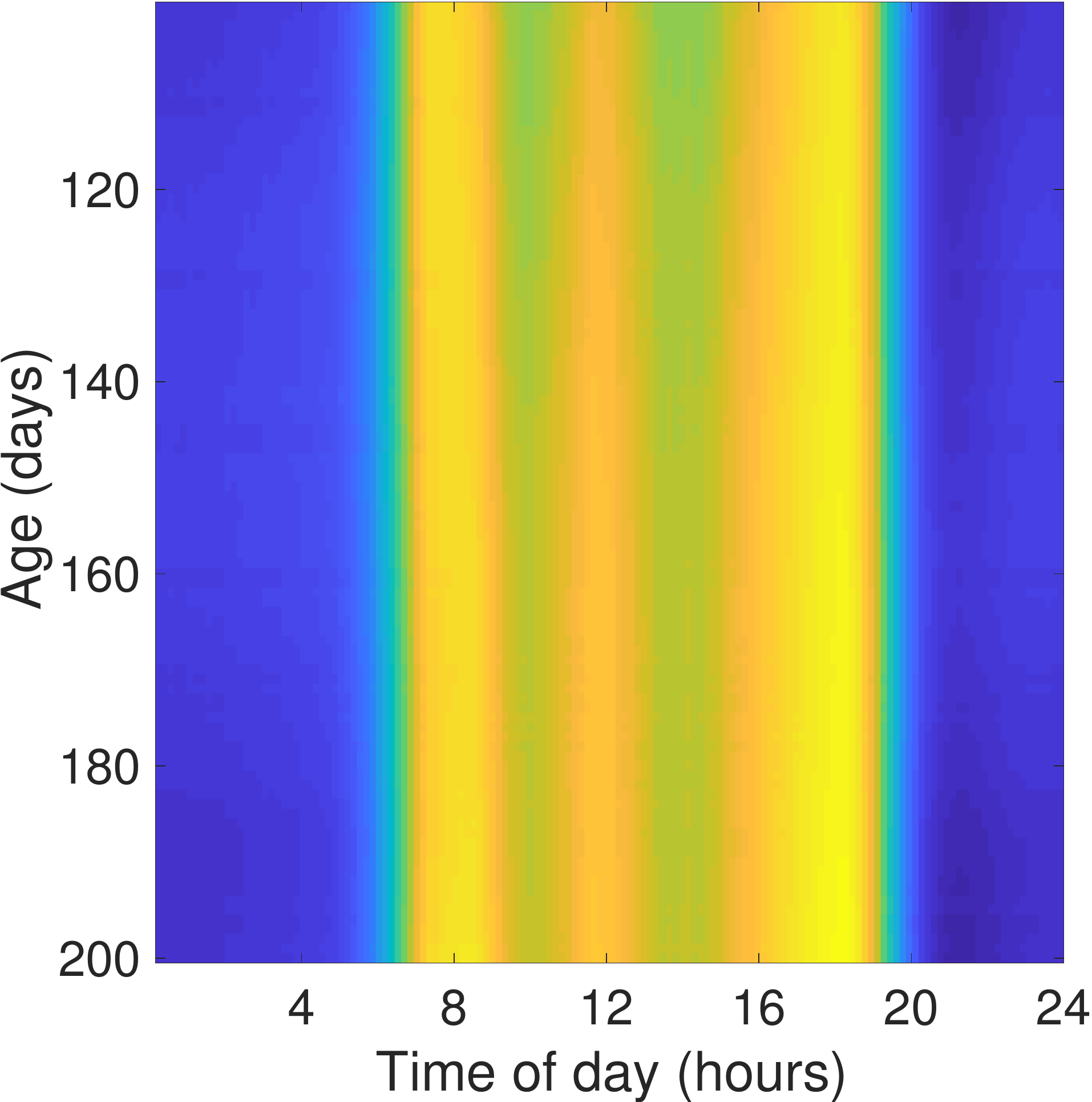} &  \includegraphics[scale=0.15]{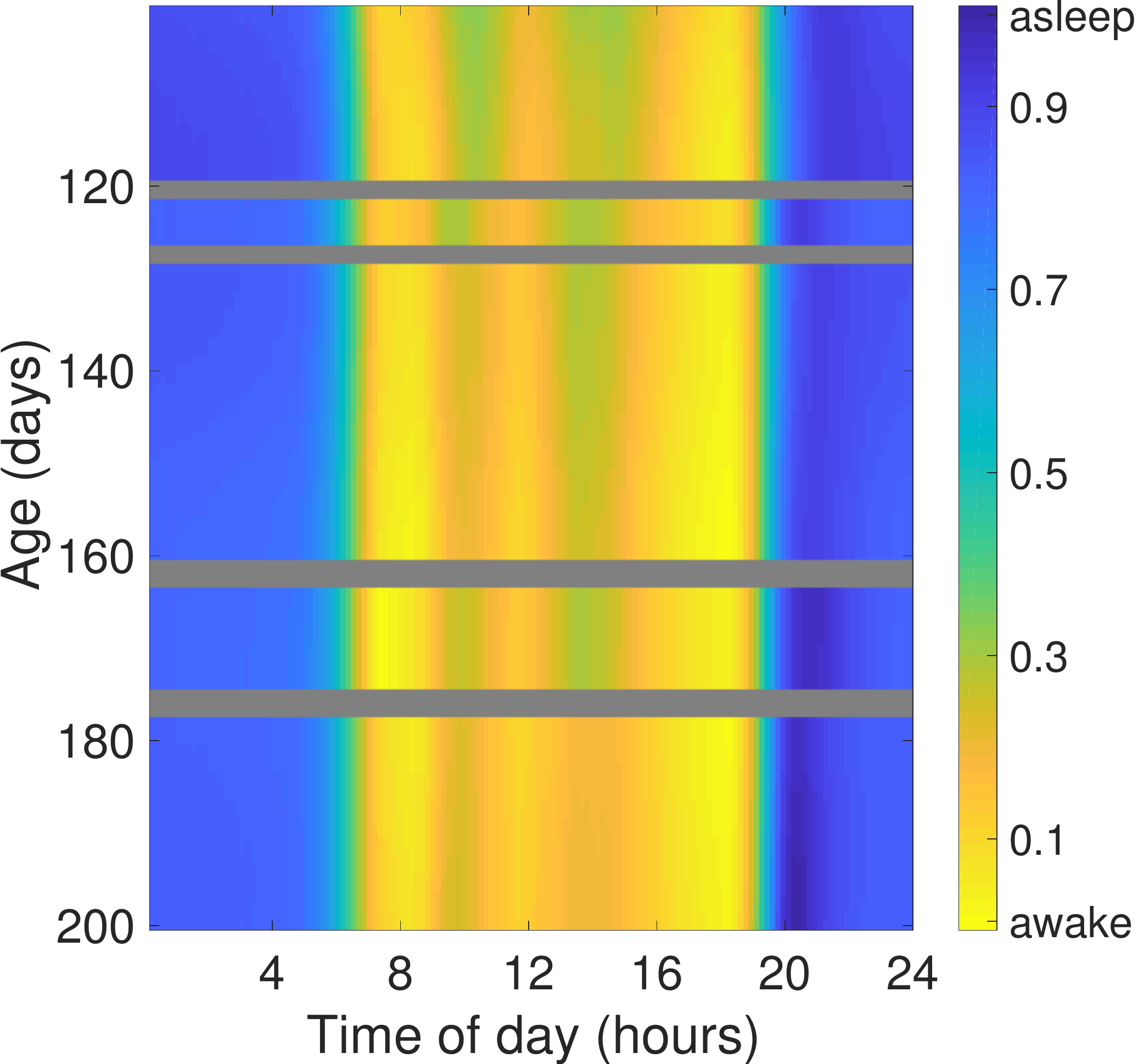}\\
 \end{tabular}
 \caption{Examples of sleep forecasts from days 100 to 200 of age, computed from the data of days 0 to 100. For each example, we show the future data, the mean of the training set, the kernel-regression (KR) using the raw training data, the kernel-regression prediction based on NMF-TS coefficients, and the rank-5 representation of the future data.}
 \label{fig:pred_results_1}
\end{figure}

\begin{figure}[tp]
\hspace{-0.5cm}
\begin{tabular}{  >{\centering\arraybackslash}m{0.18\linewidth} >{\centering\arraybackslash}m{0.18\linewidth} >{\centering\arraybackslash}m{0.18\linewidth} >{\centering\arraybackslash}m{0.18\linewidth} >{\centering\arraybackslash}m{0.18\linewidth}}
 Raw Data \vspace{0.2cm} &  Mean\vspace{0.2cm} & KR (raw data) \vspace{0.2cm} &  KR (NMF-TS) \vspace{0.2cm}  &  Rank-5 GT \vspace{0.2cm}\\
\includegraphics[scale=0.15]{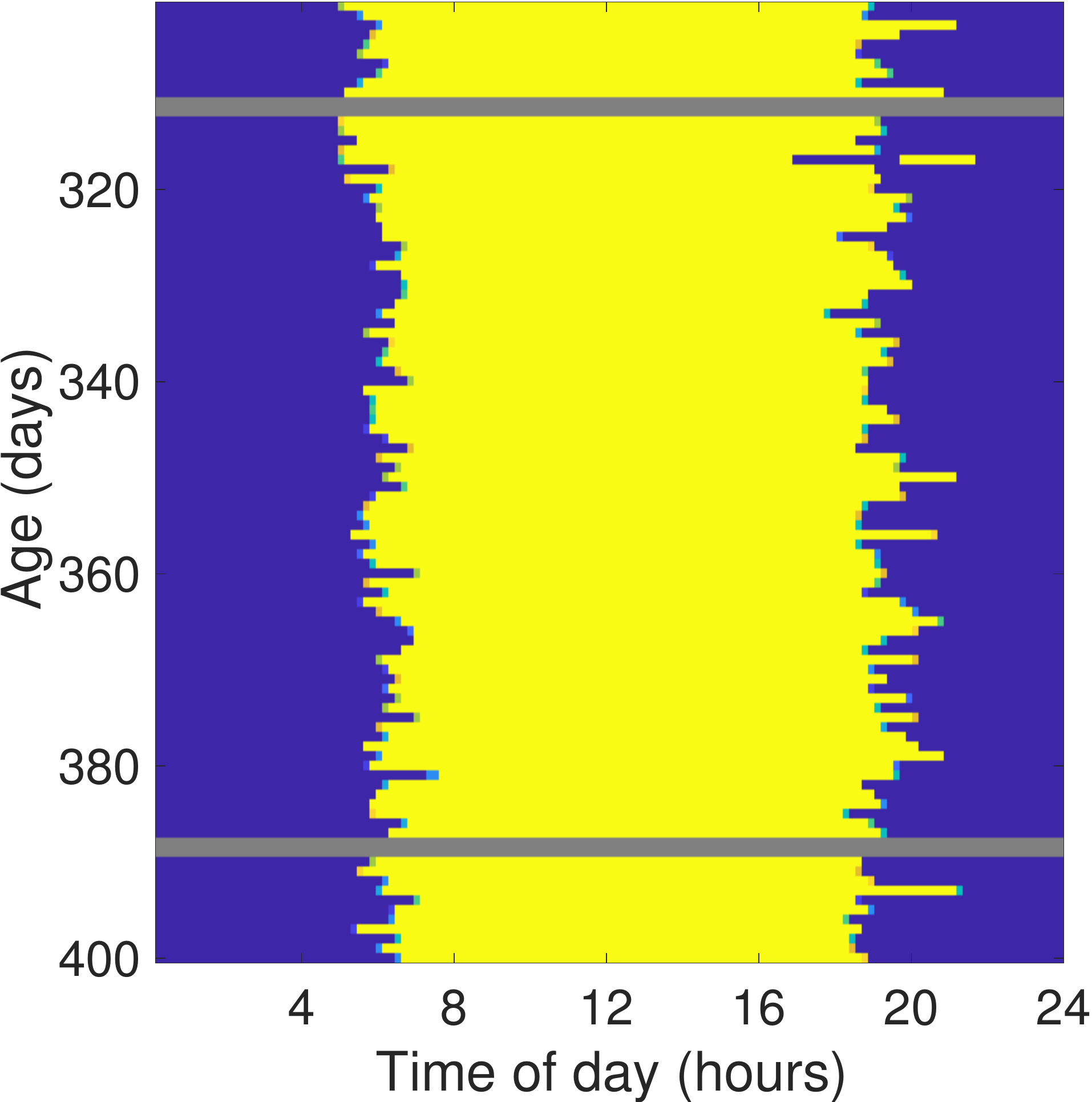} & \includegraphics[scale=0.15]{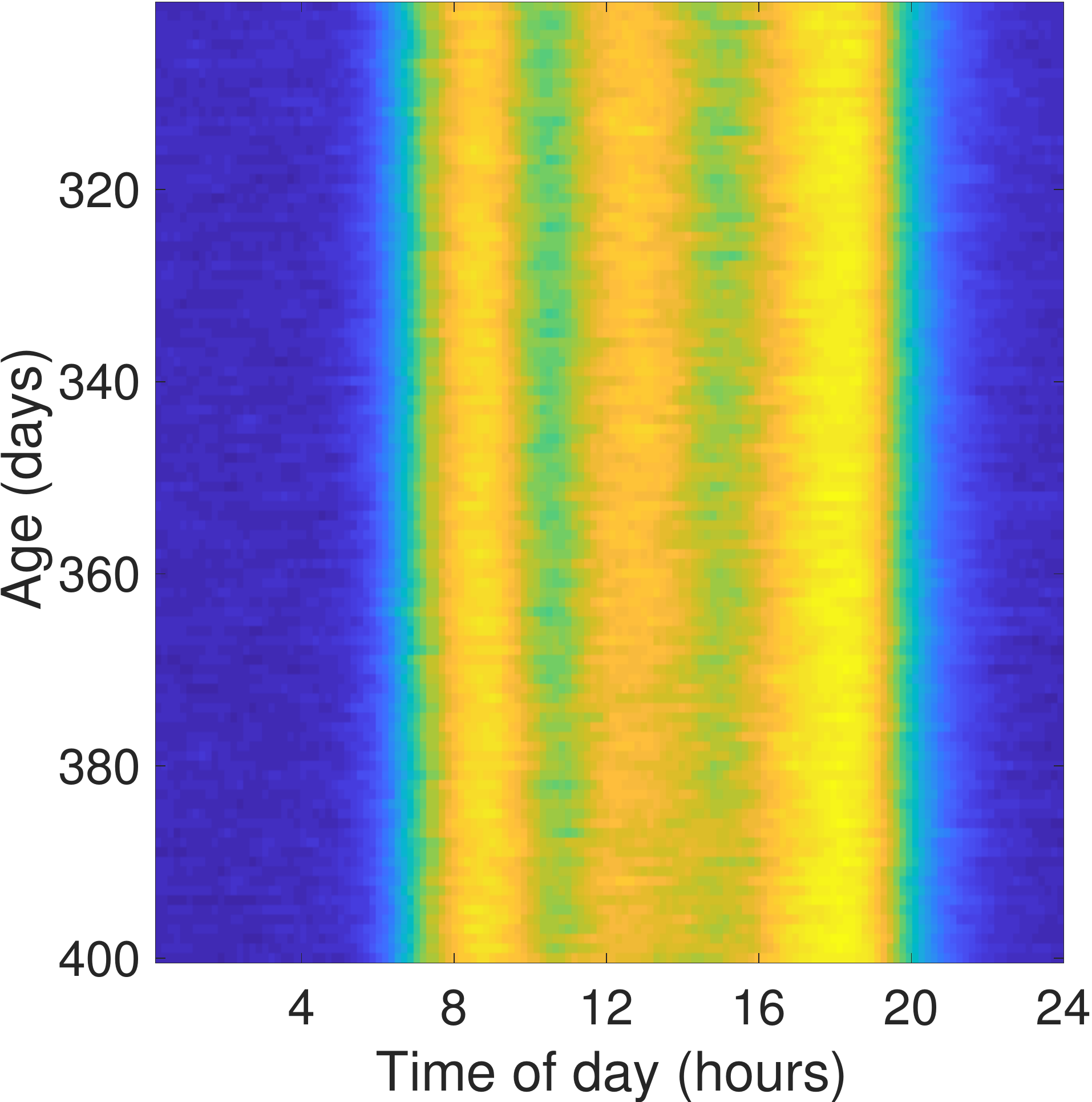} &  \includegraphics[scale=0.15]{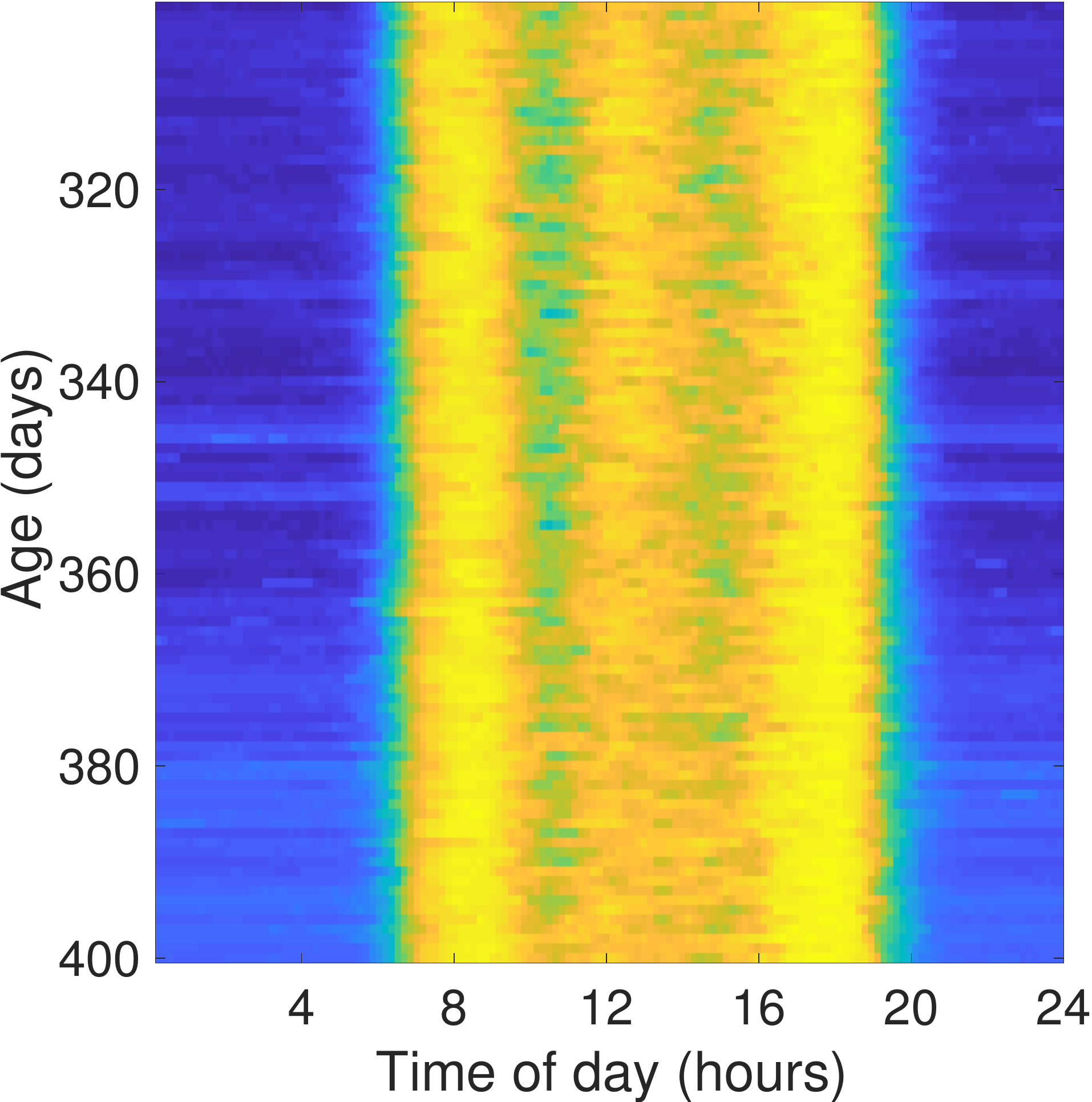}&  \includegraphics[scale=0.15]{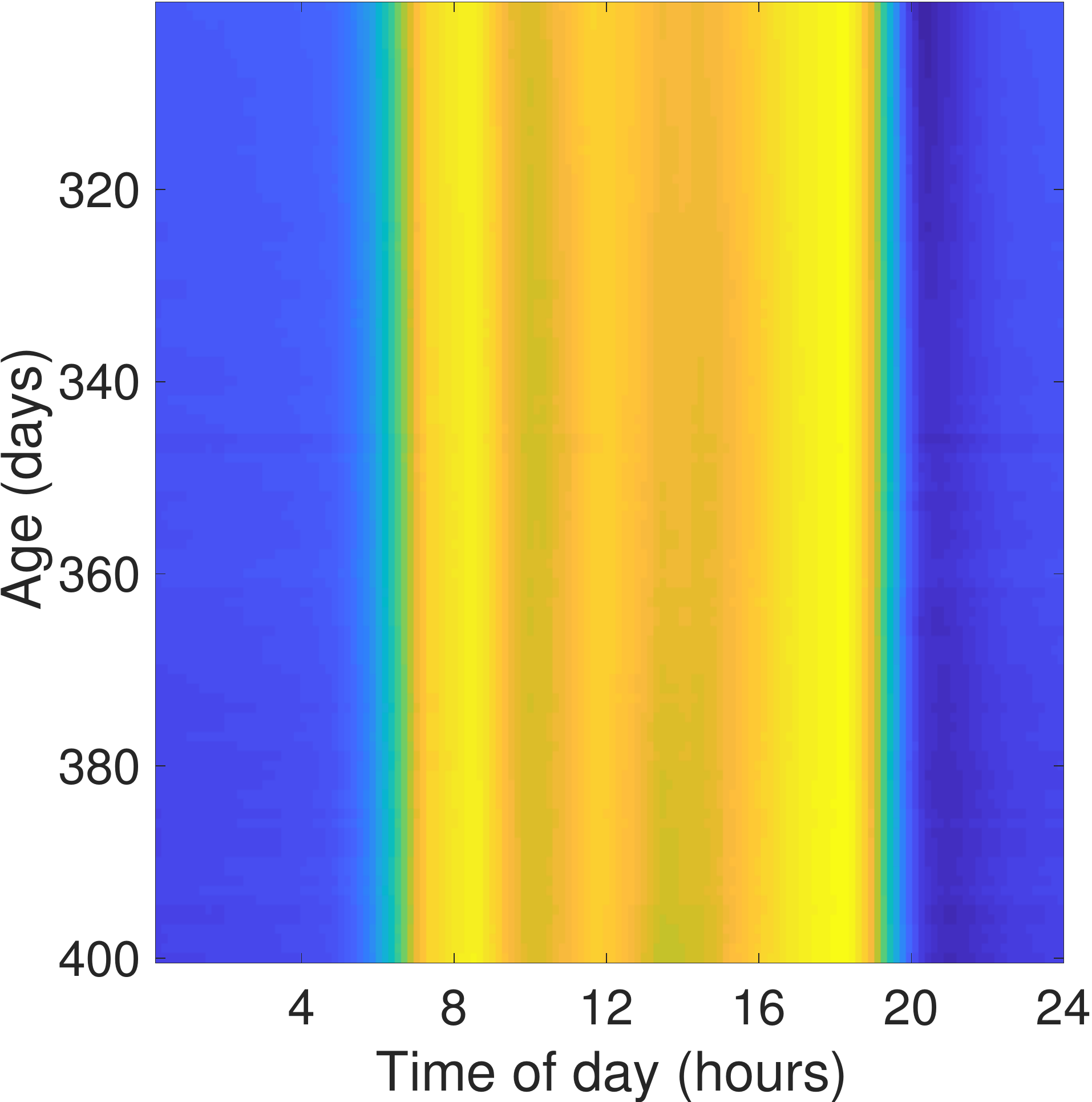} &  \includegraphics[scale=0.15]{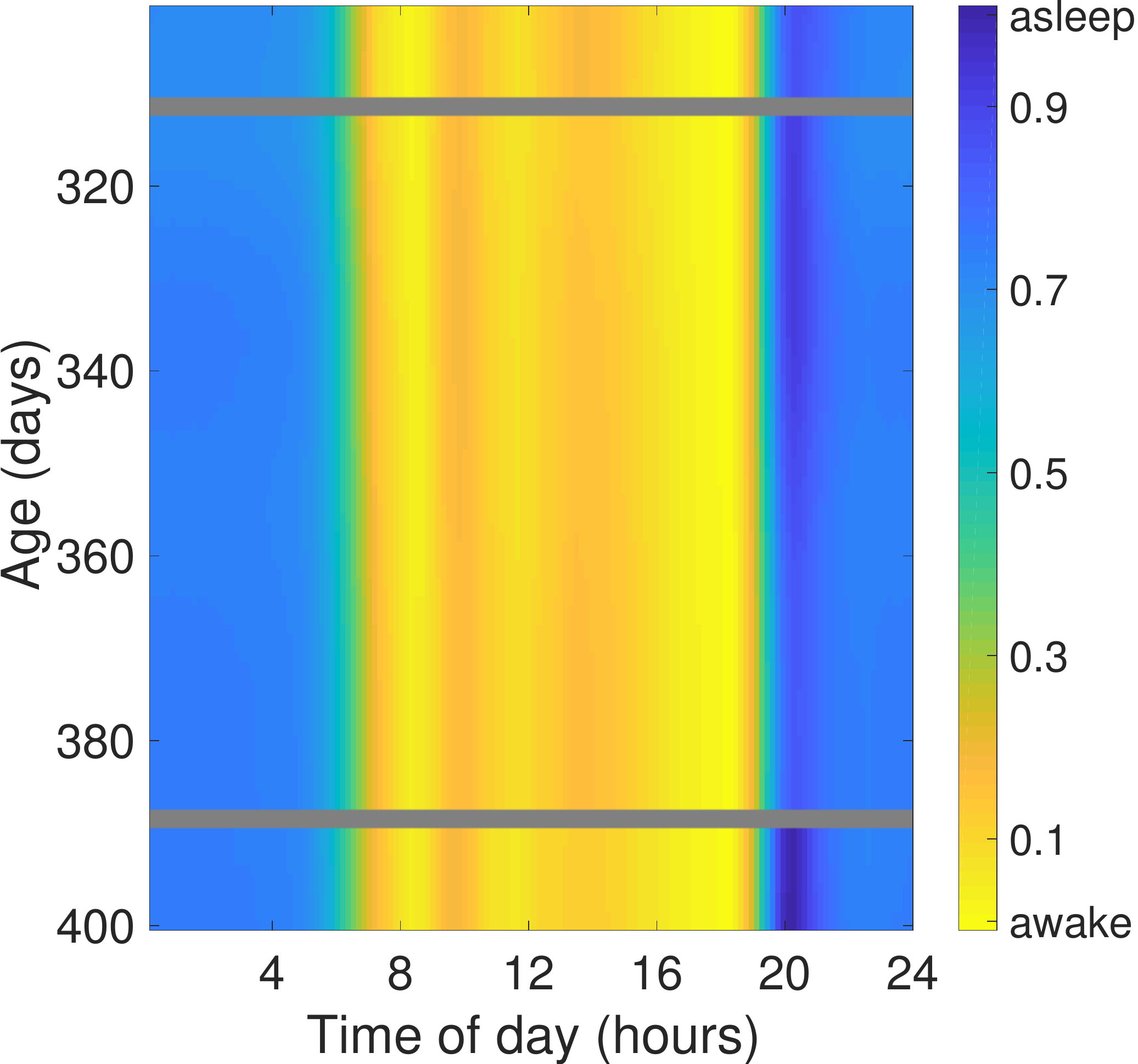}\\
\includegraphics[scale=0.15]{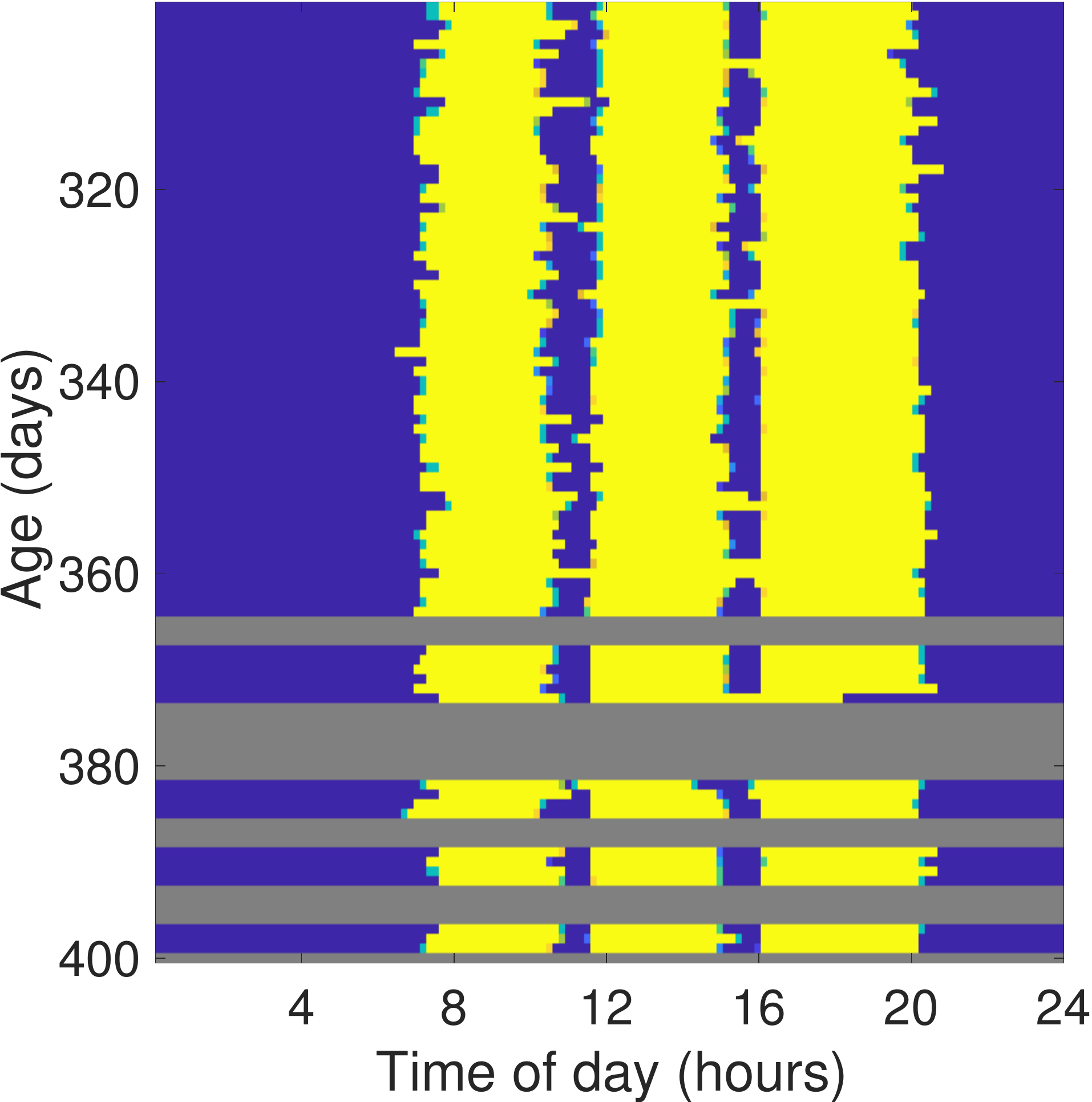} & \includegraphics[scale=0.15]{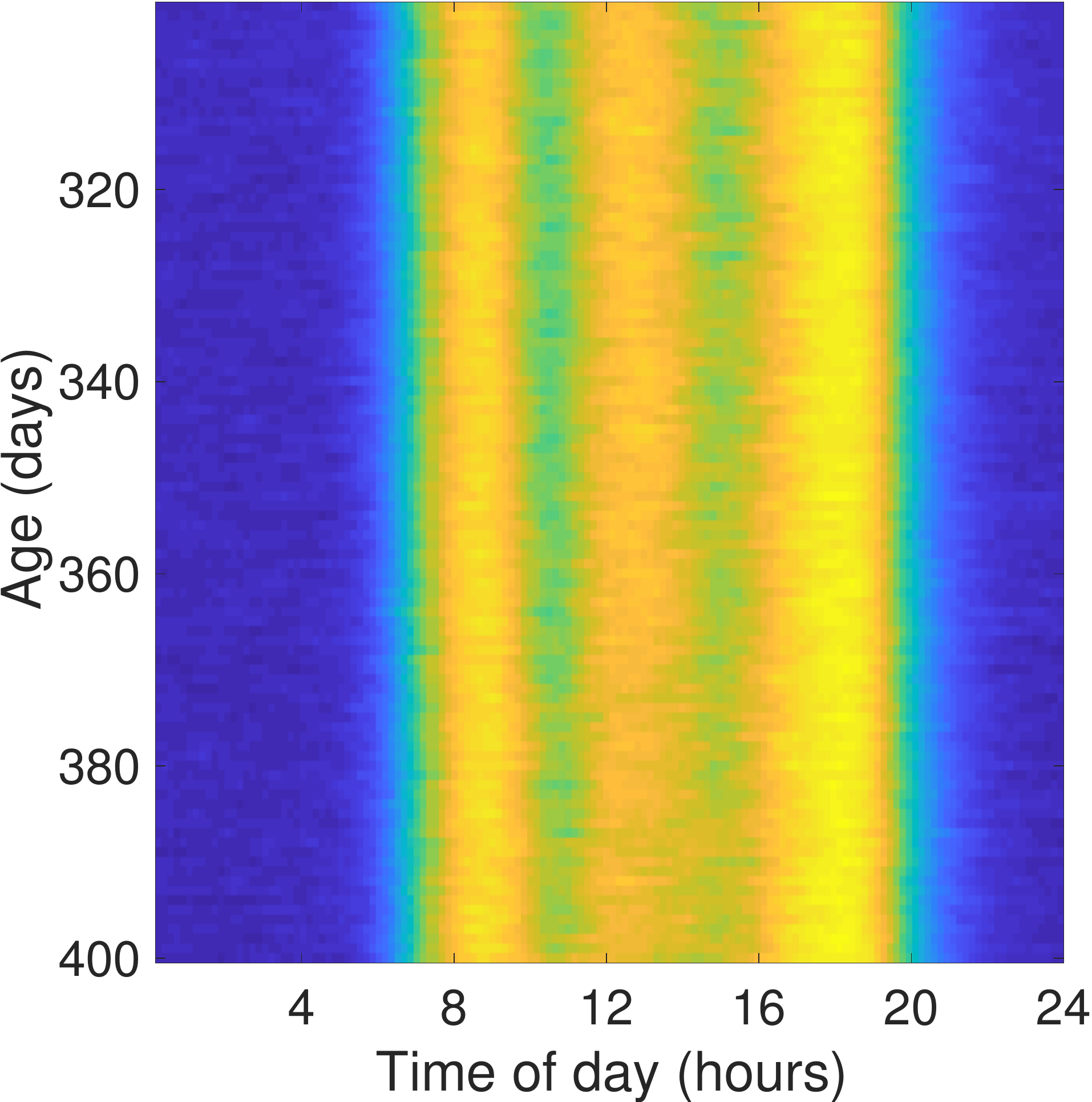}  &  \includegraphics[scale=0.15]{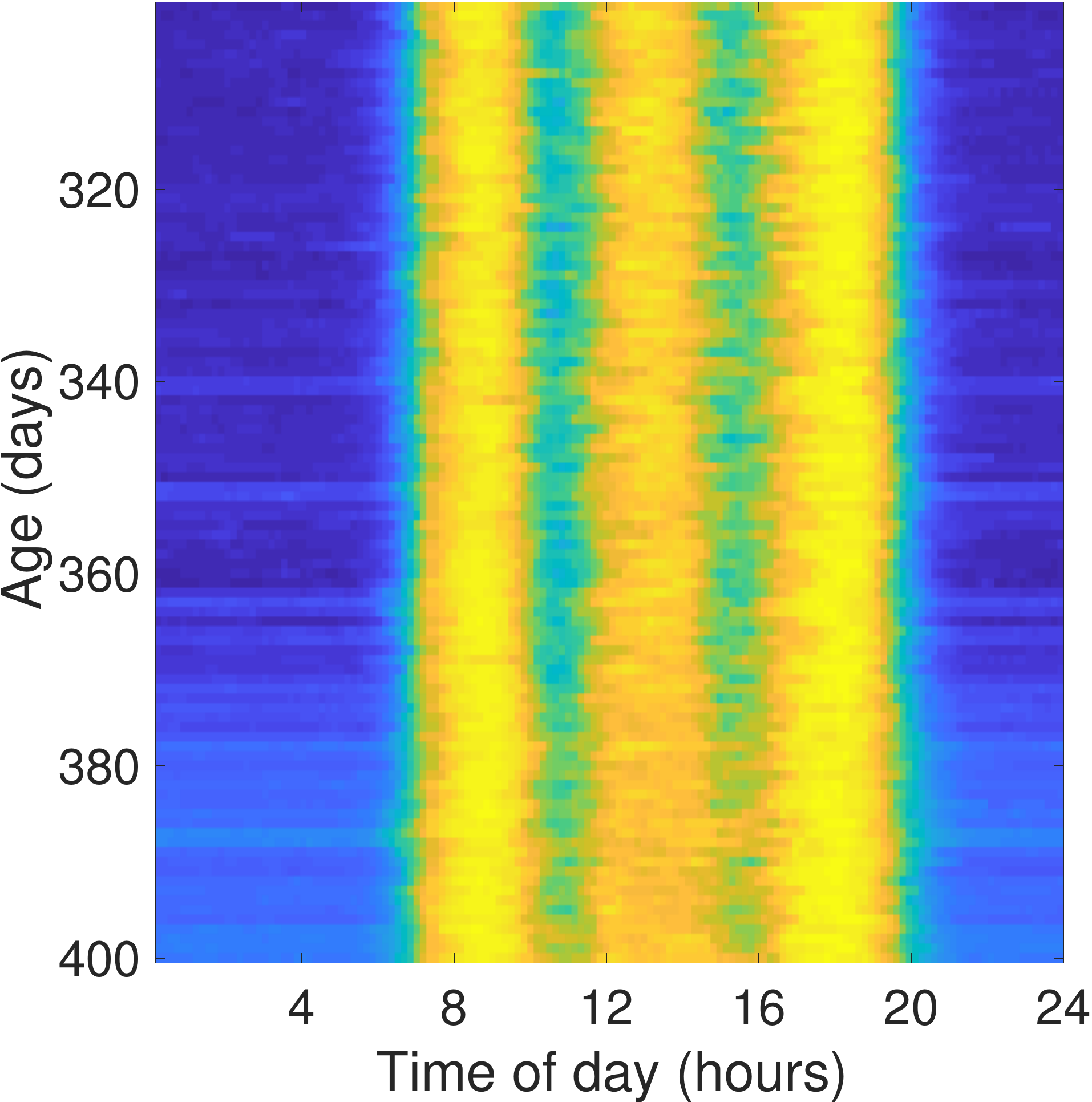}&  \includegraphics[scale=0.15]{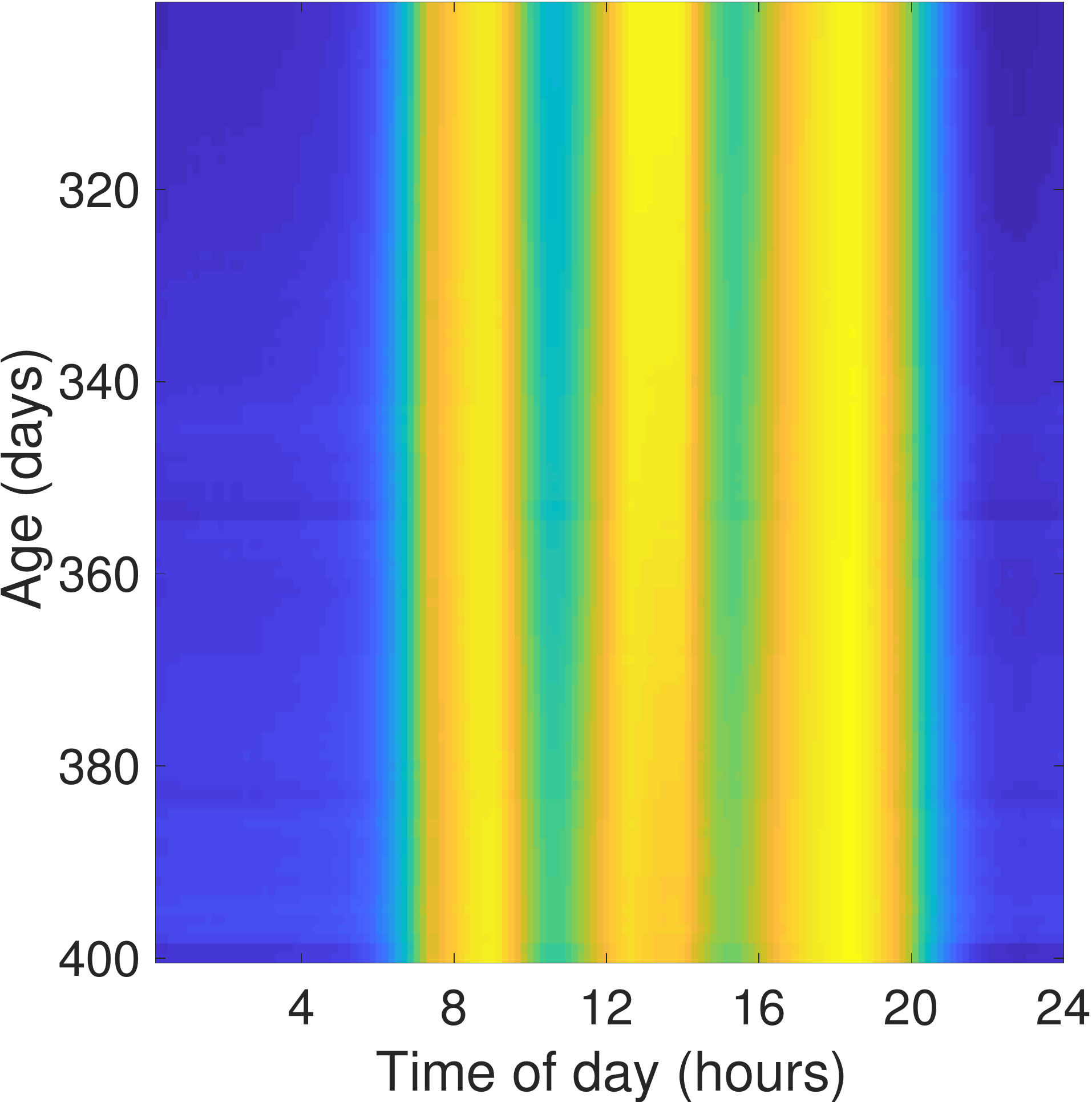} &  \includegraphics[scale=0.15]{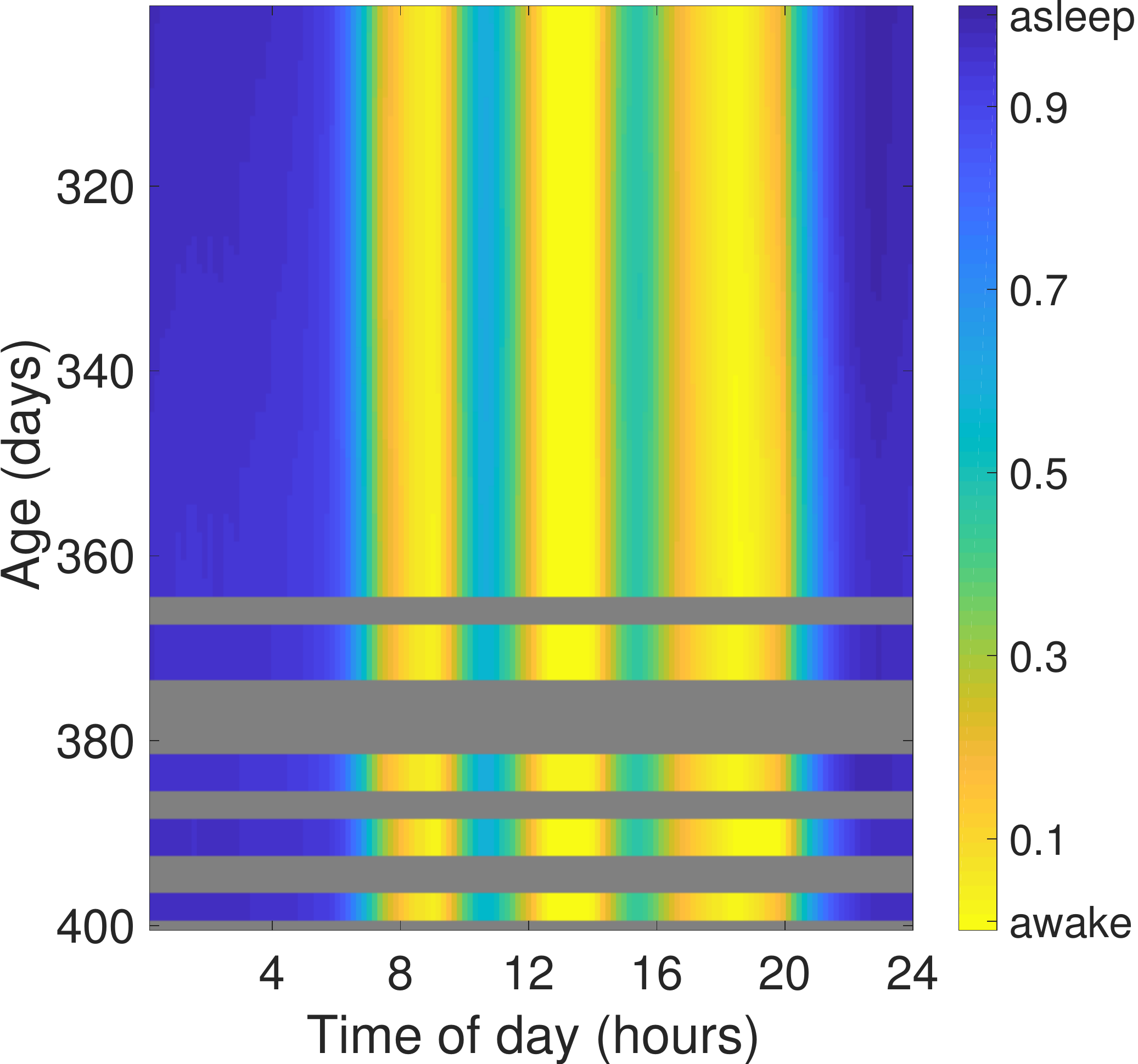}\\
\includegraphics[scale=0.15]{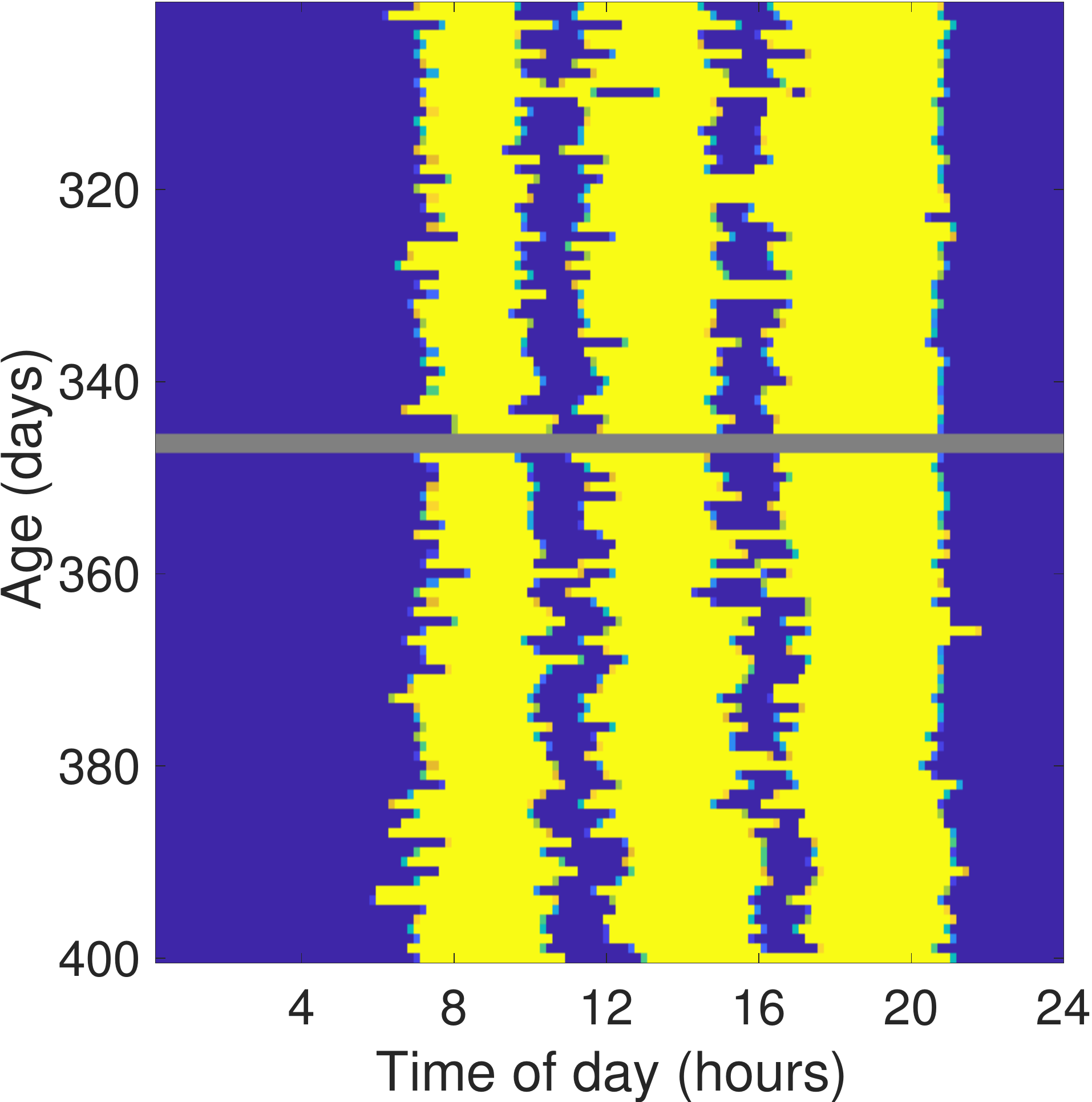} & \includegraphics[scale=0.15]{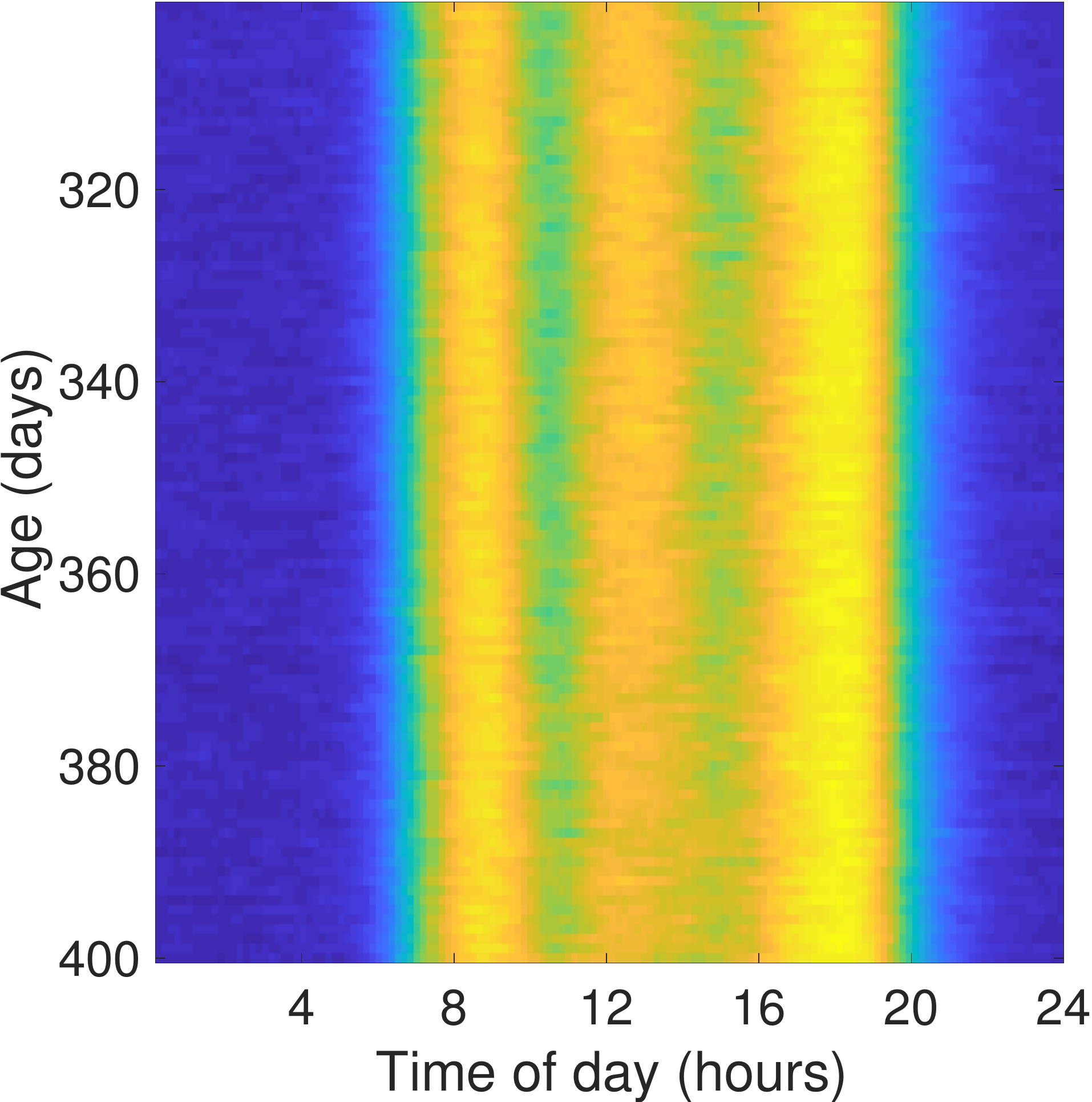}  &  \includegraphics[scale=0.15]{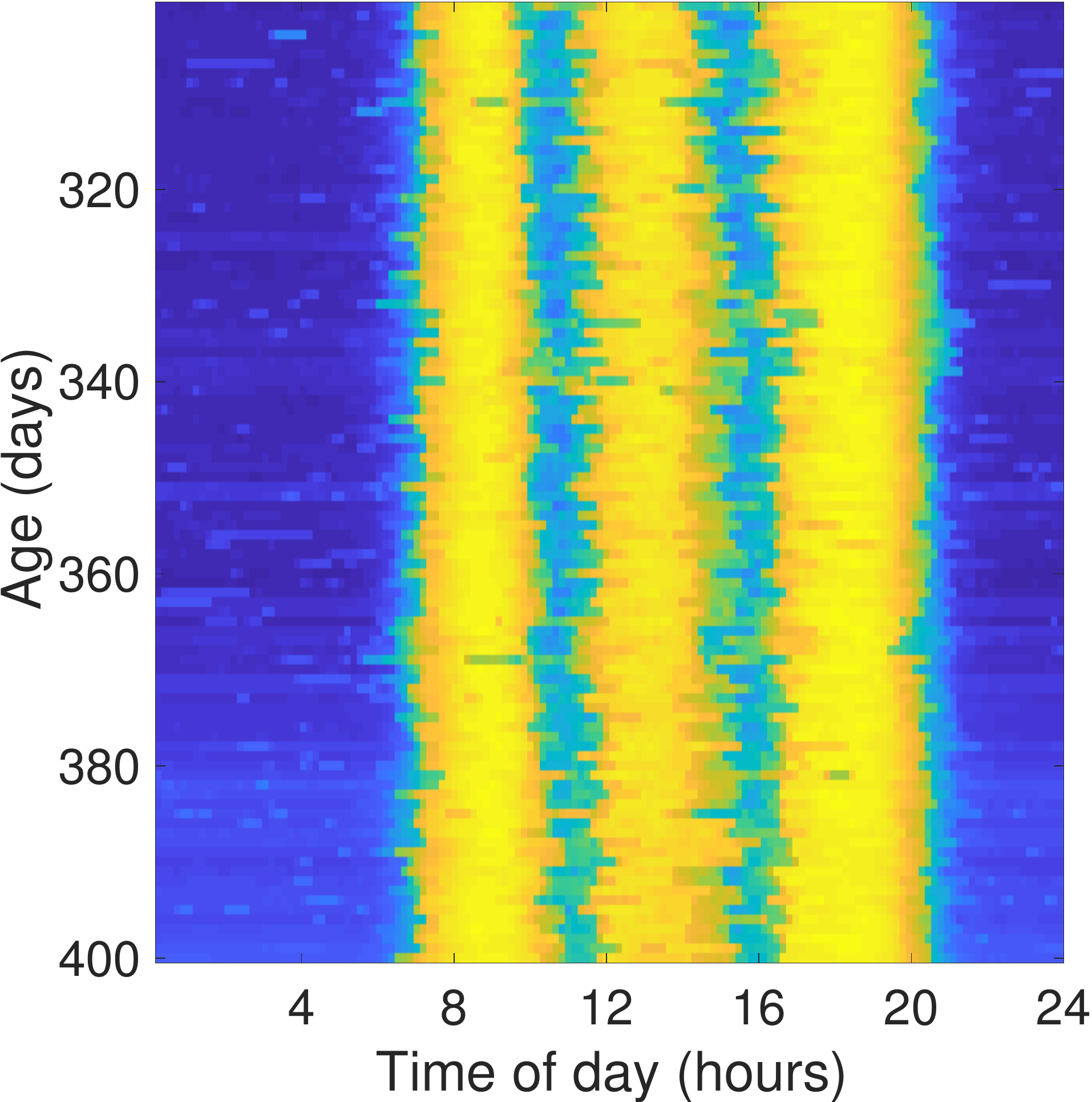}&  \includegraphics[scale=0.15]{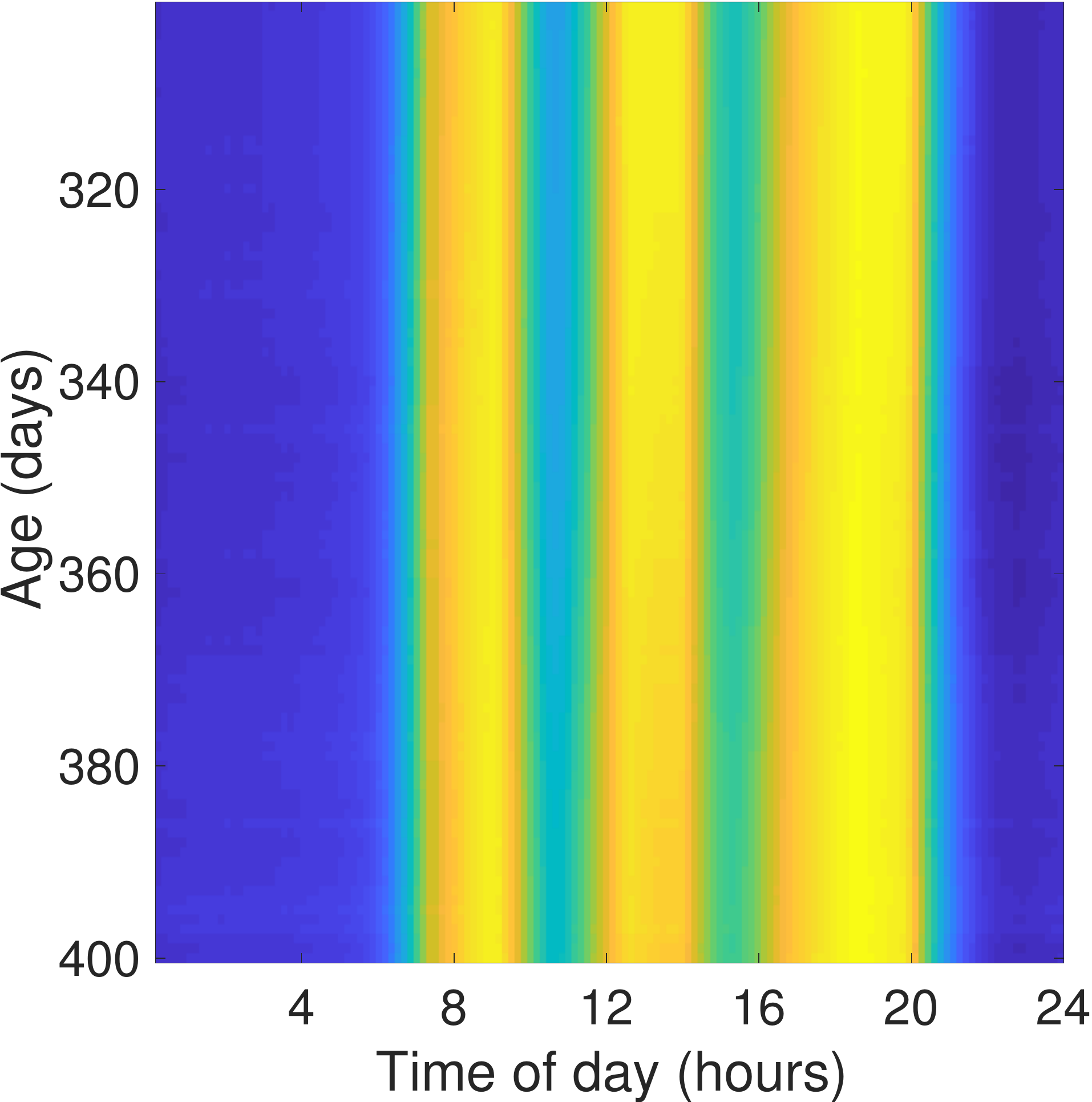} &  \includegraphics[scale=0.15]{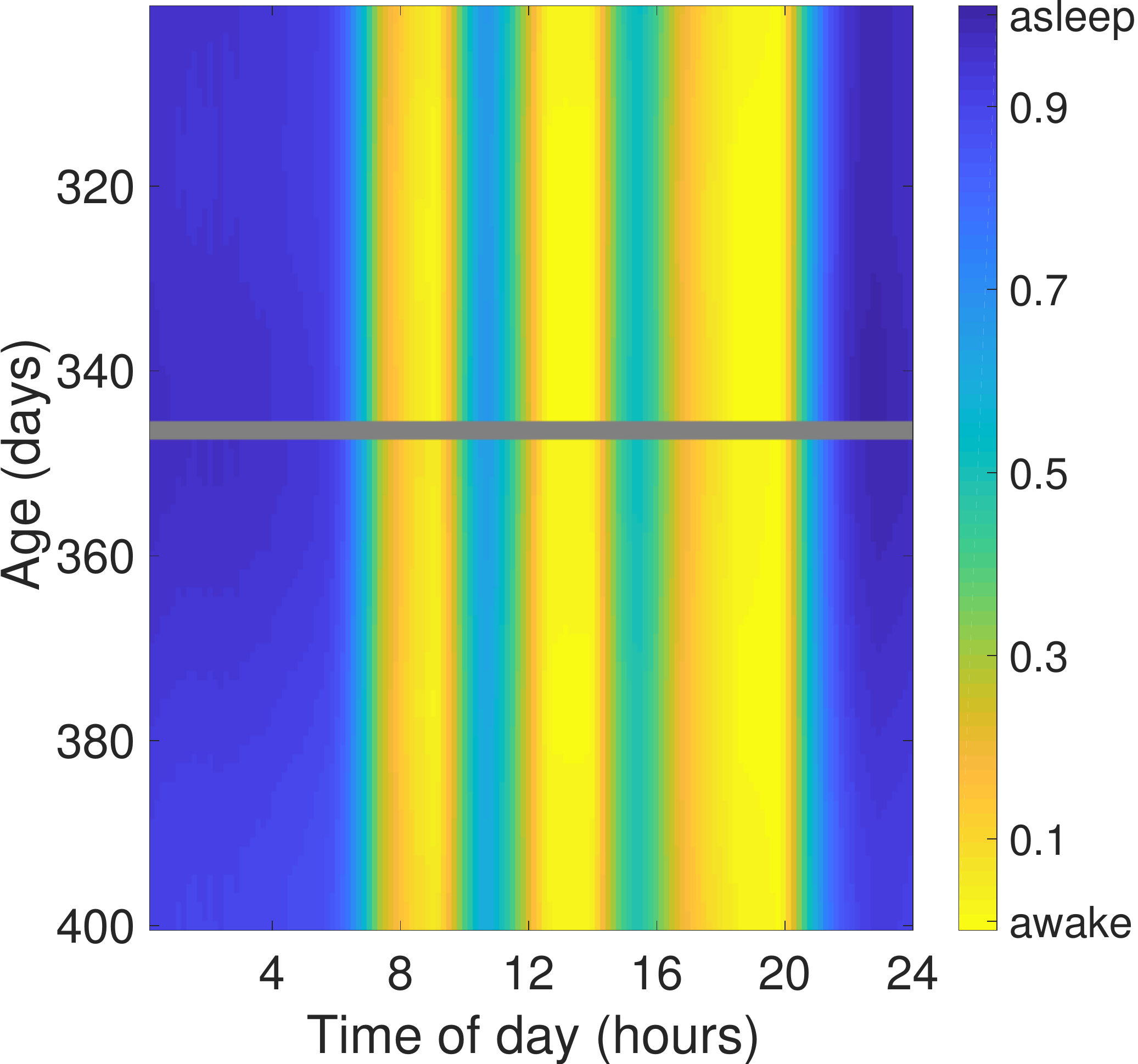}\\
 \end{tabular}
 \caption{Examples of sleep forecasts from days 300 to 400 of age, computed from the data of days 200 to 300. For each example, we show the future data, the mean of the training set, the kernel-regression (KR) using the raw training data, the kernel-regression prediction based on NMF-TS coefficients, and the rank-5 representation of the future data.}
 \label{fig:pred_results_2}
\end{figure}

\begin{figure}[tp]
\hspace{-0.5cm}
\begin{tabular}{  >{\centering\arraybackslash}m{0.18\linewidth} >{\centering\arraybackslash}m{0.18\linewidth} >{\centering\arraybackslash}m{0.18\linewidth} >{\centering\arraybackslash}m{0.18\linewidth} >{\centering\arraybackslash}m{0.18\linewidth}}
 Raw Data \vspace{0.2cm} &  Mean\vspace{0.2cm} & KR (raw data) \vspace{0.2cm} &  KR (NMF-TS) \vspace{0.2cm}  &  Rank-5 GT \vspace{0.2cm}\\
\includegraphics[scale=0.15]{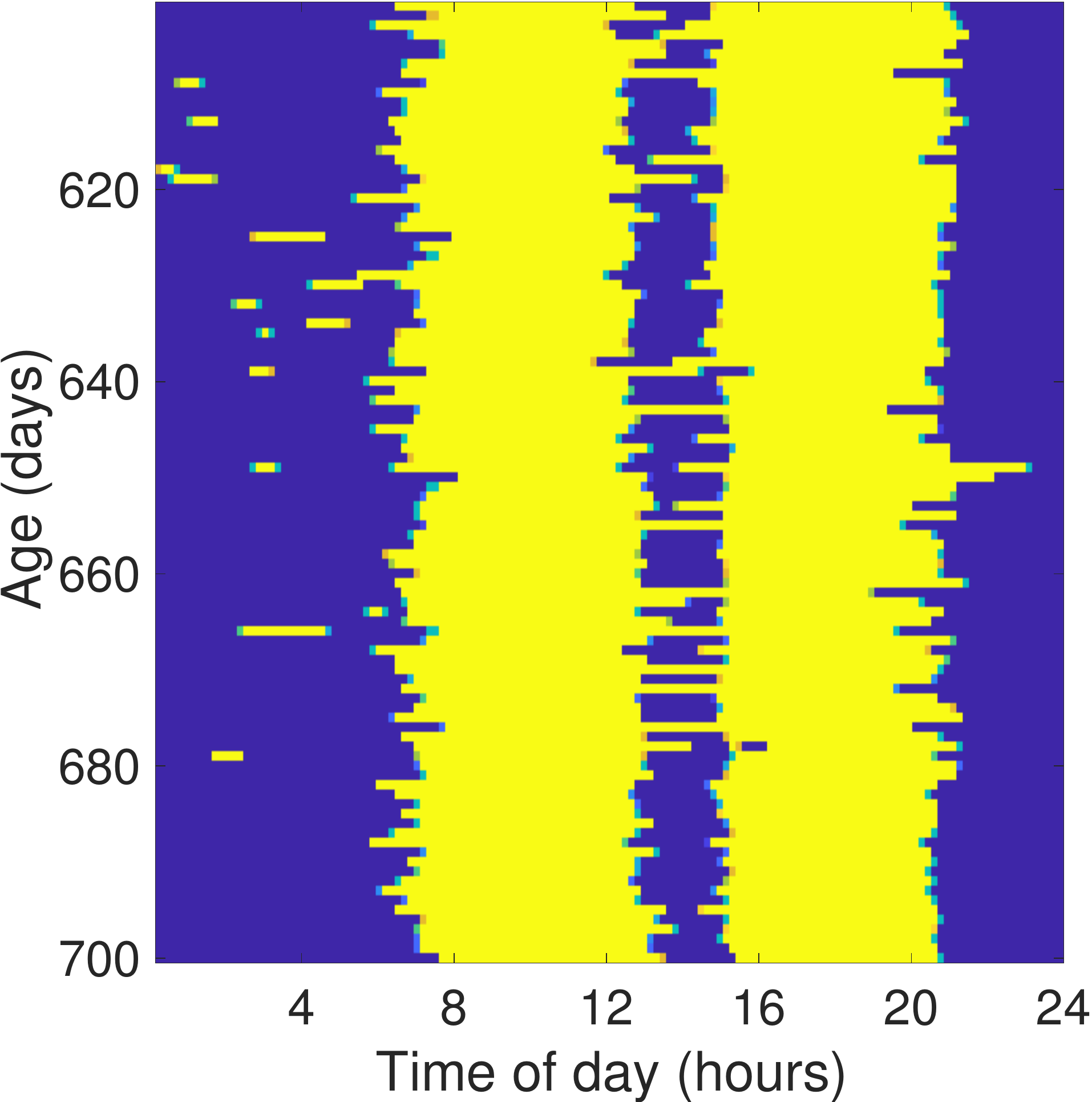} & \includegraphics[scale=0.15]{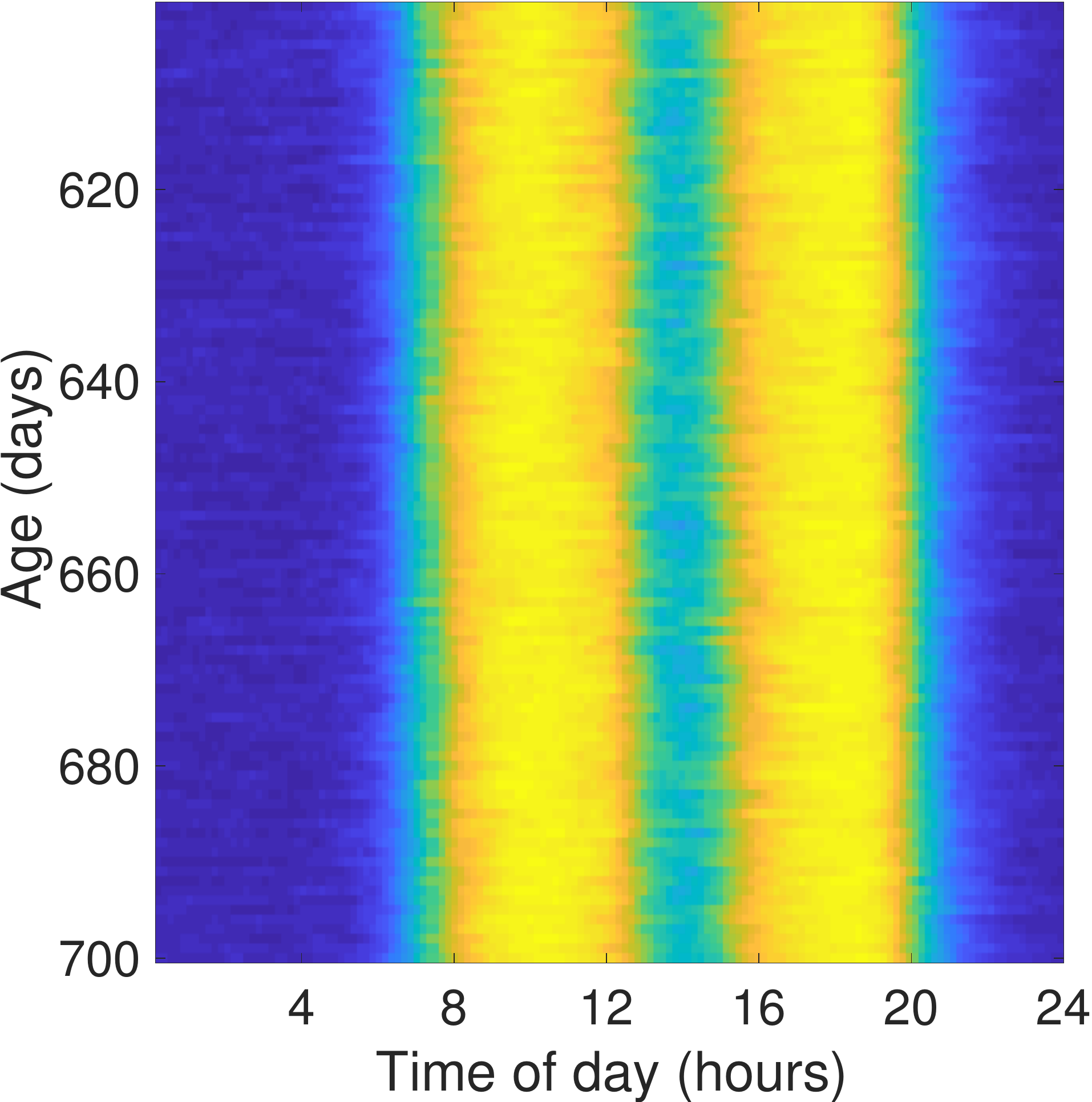} &  \includegraphics[scale=0.15]{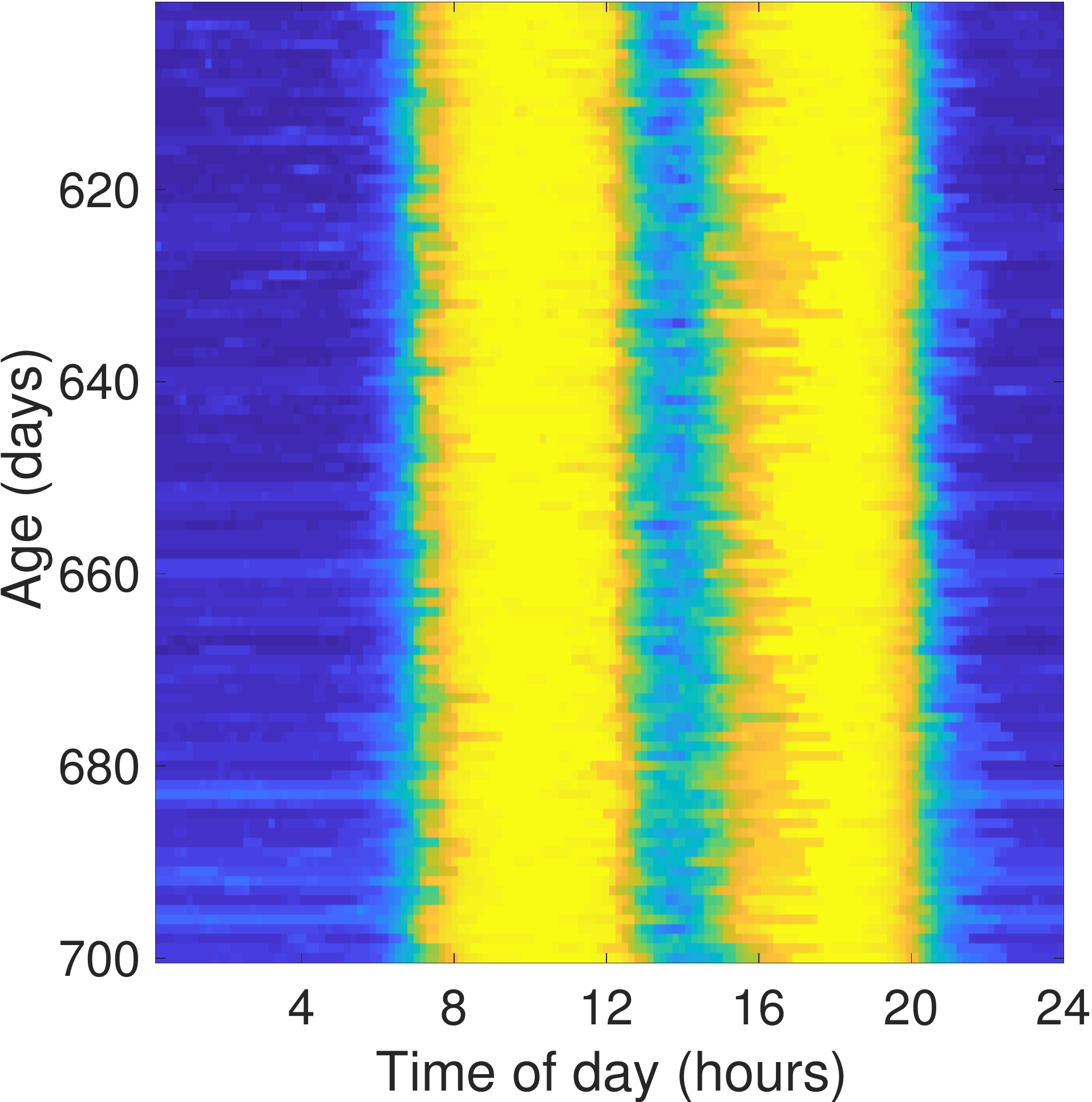}&  \includegraphics[scale=0.15]{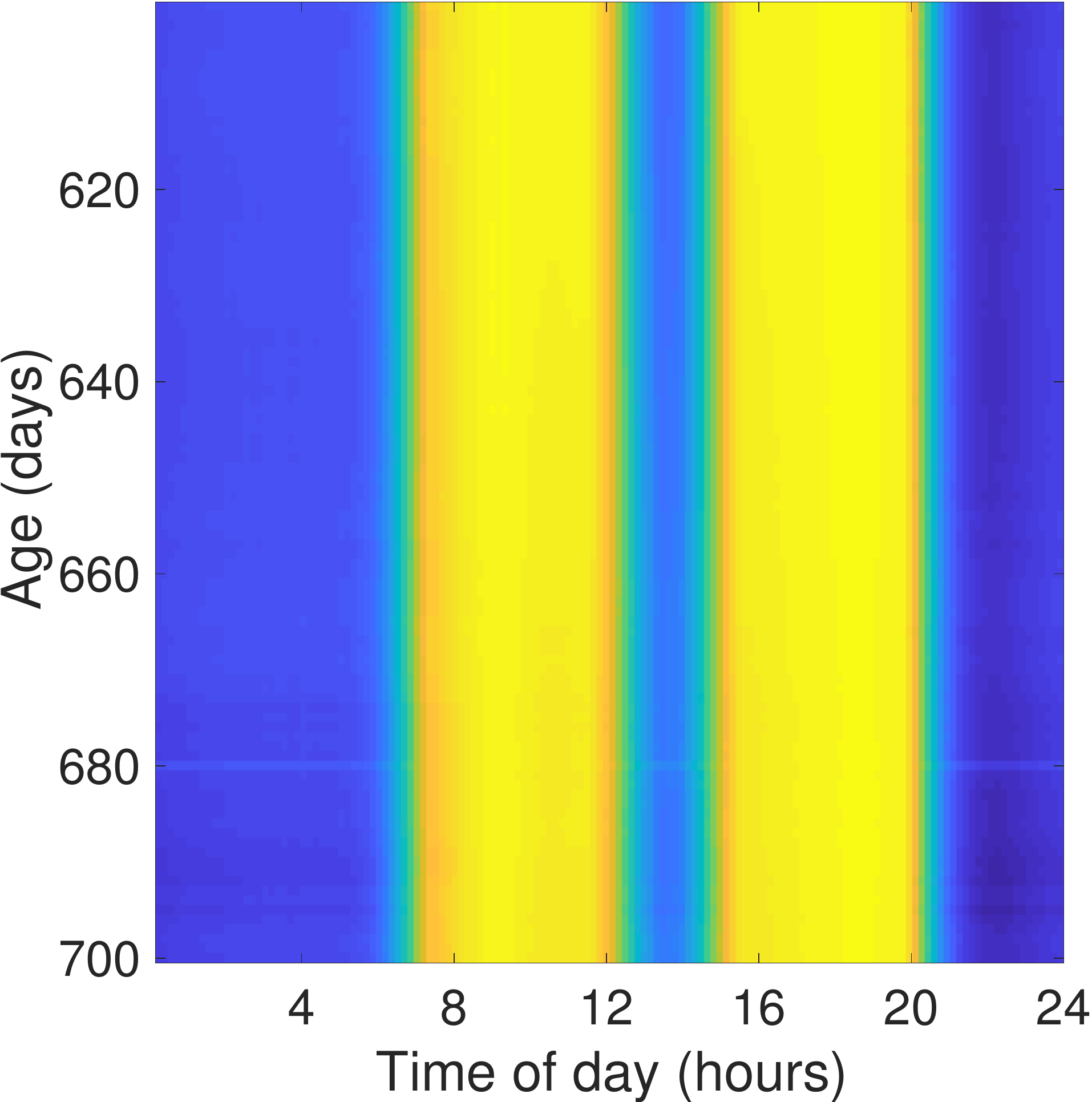} &  \includegraphics[scale=0.15]{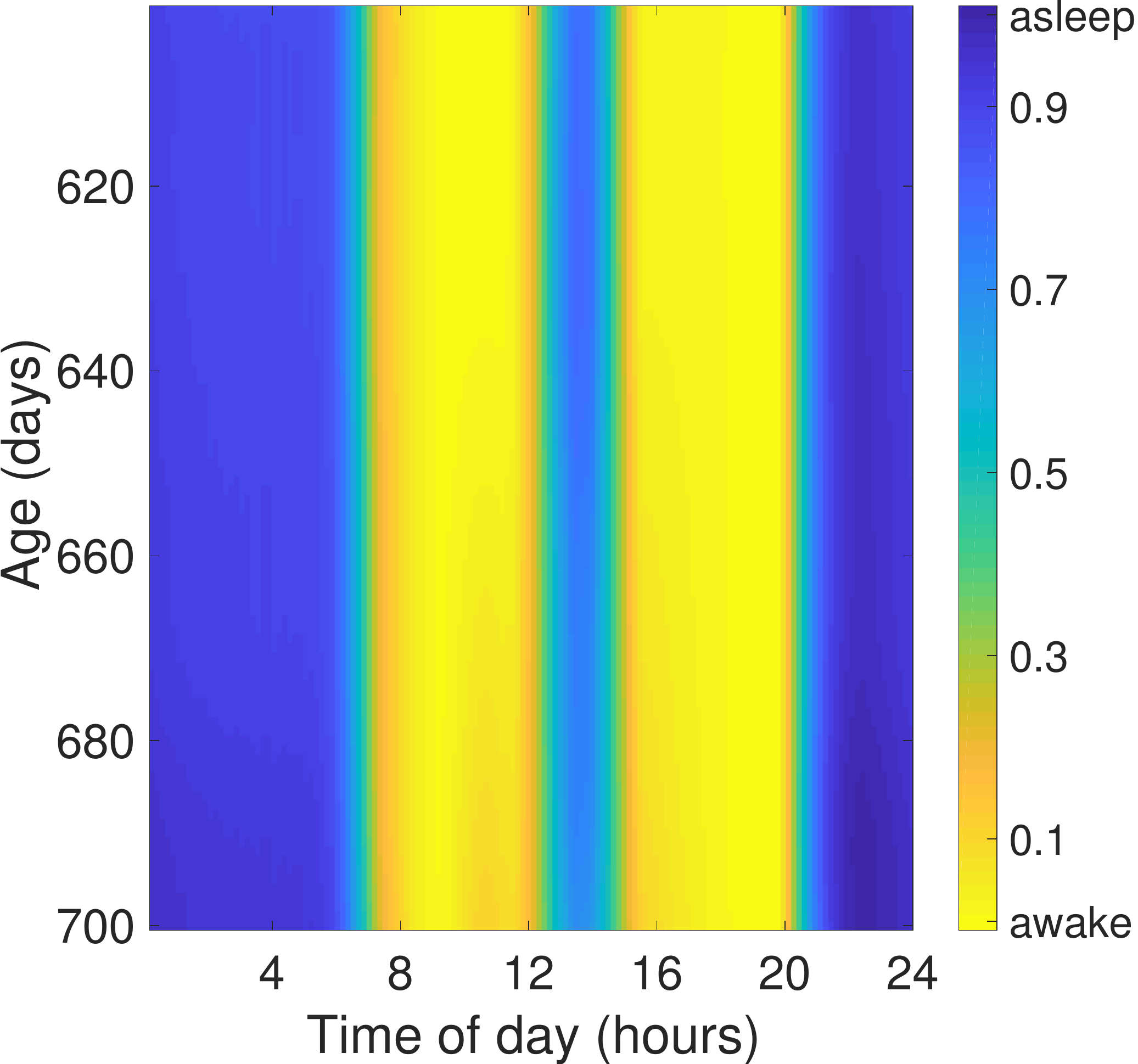}\\
\includegraphics[scale=0.15]{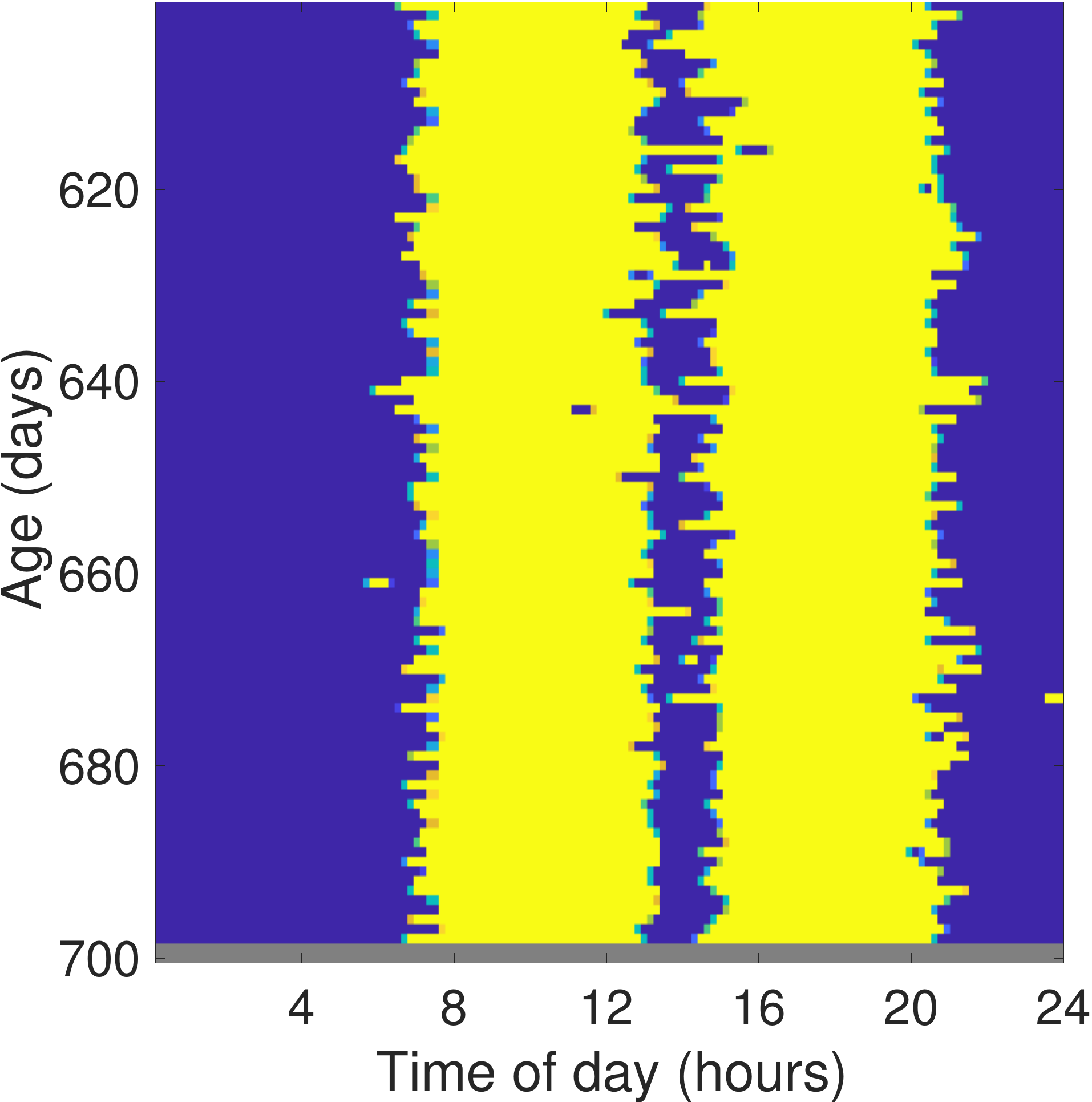} & \includegraphics[scale=0.15]{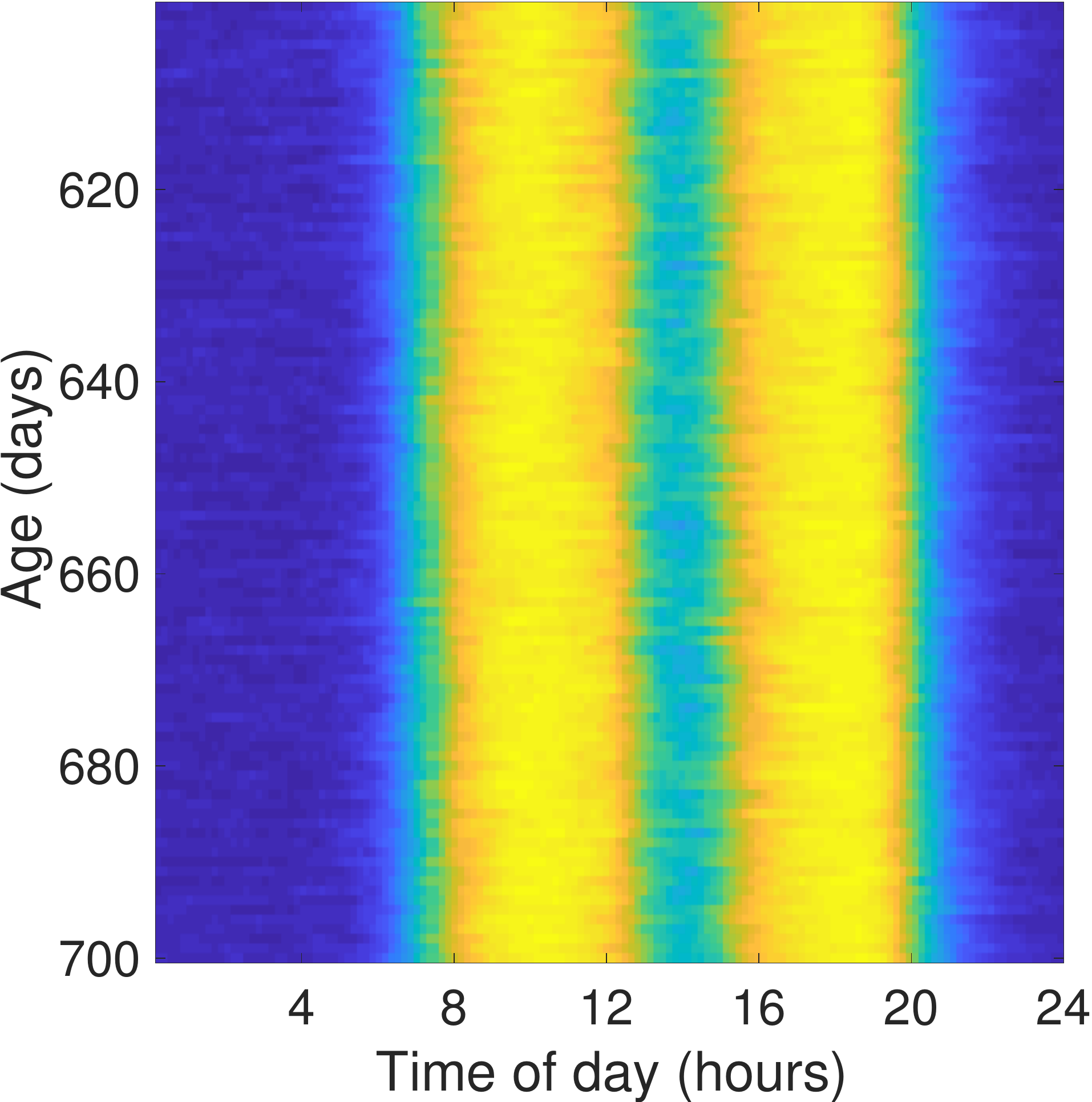}  &  \includegraphics[scale=0.15]{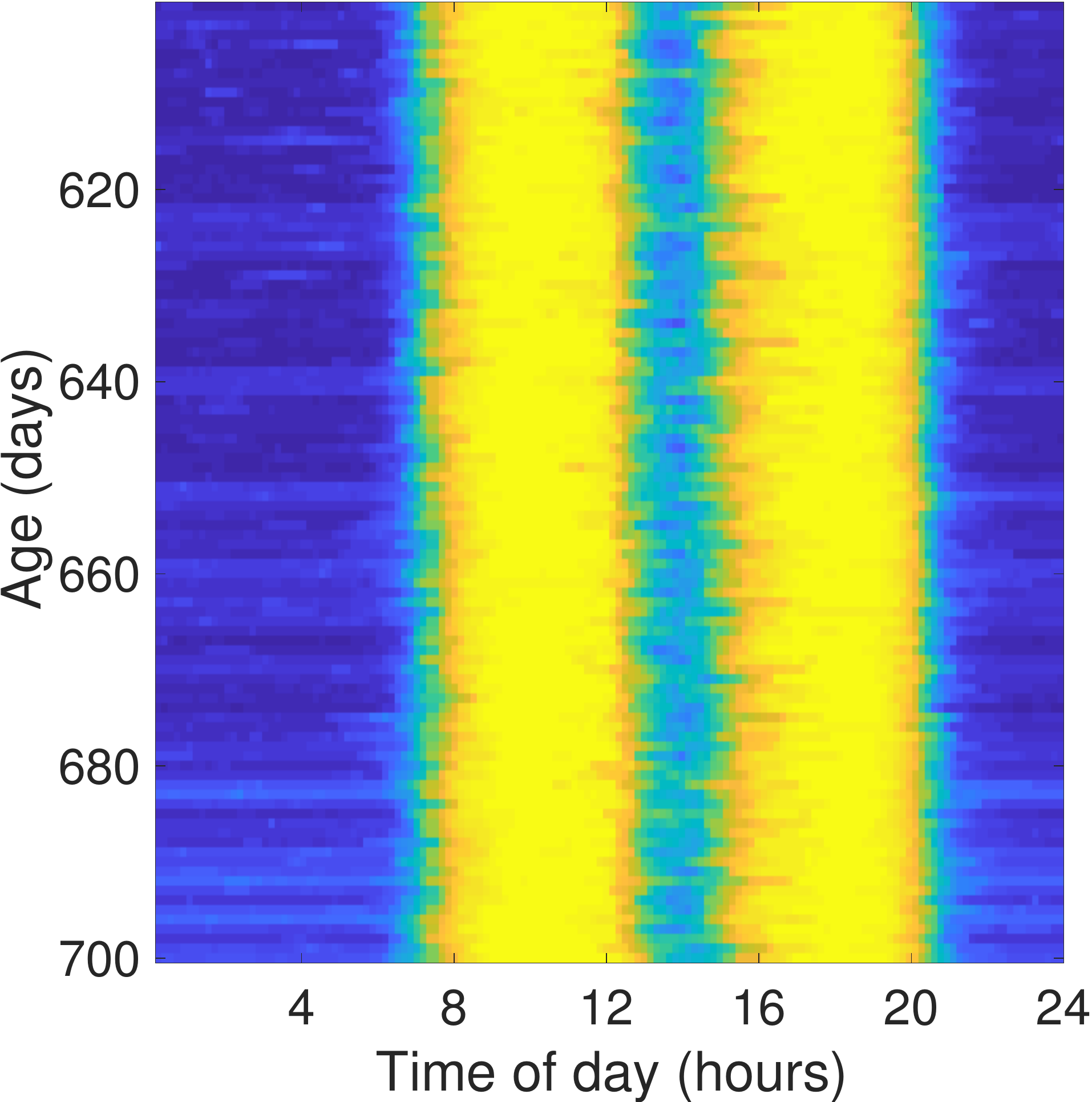}&  \includegraphics[scale=0.15]{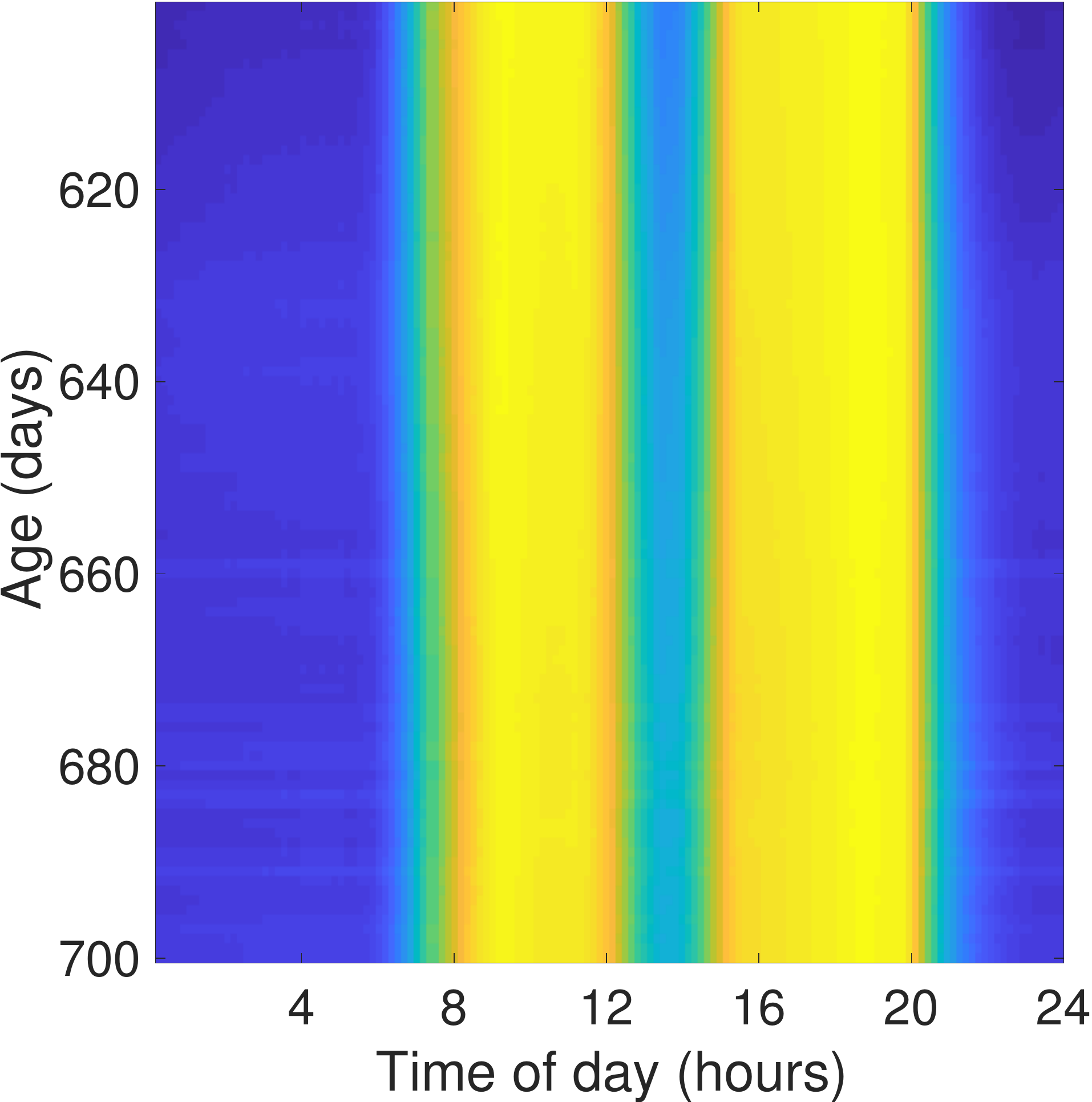} &  \includegraphics[scale=0.15]{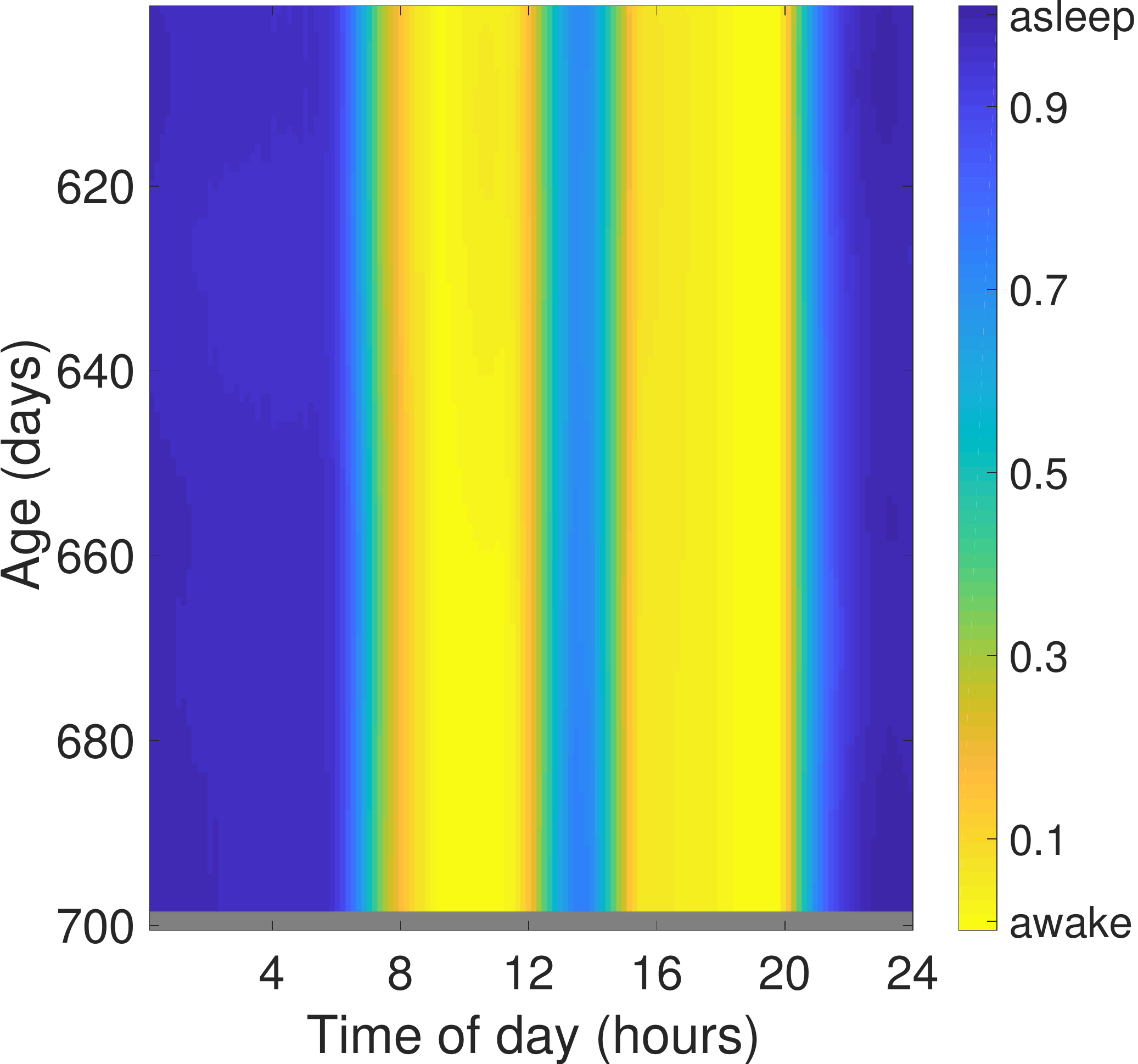}\\
\includegraphics[scale=0.15]{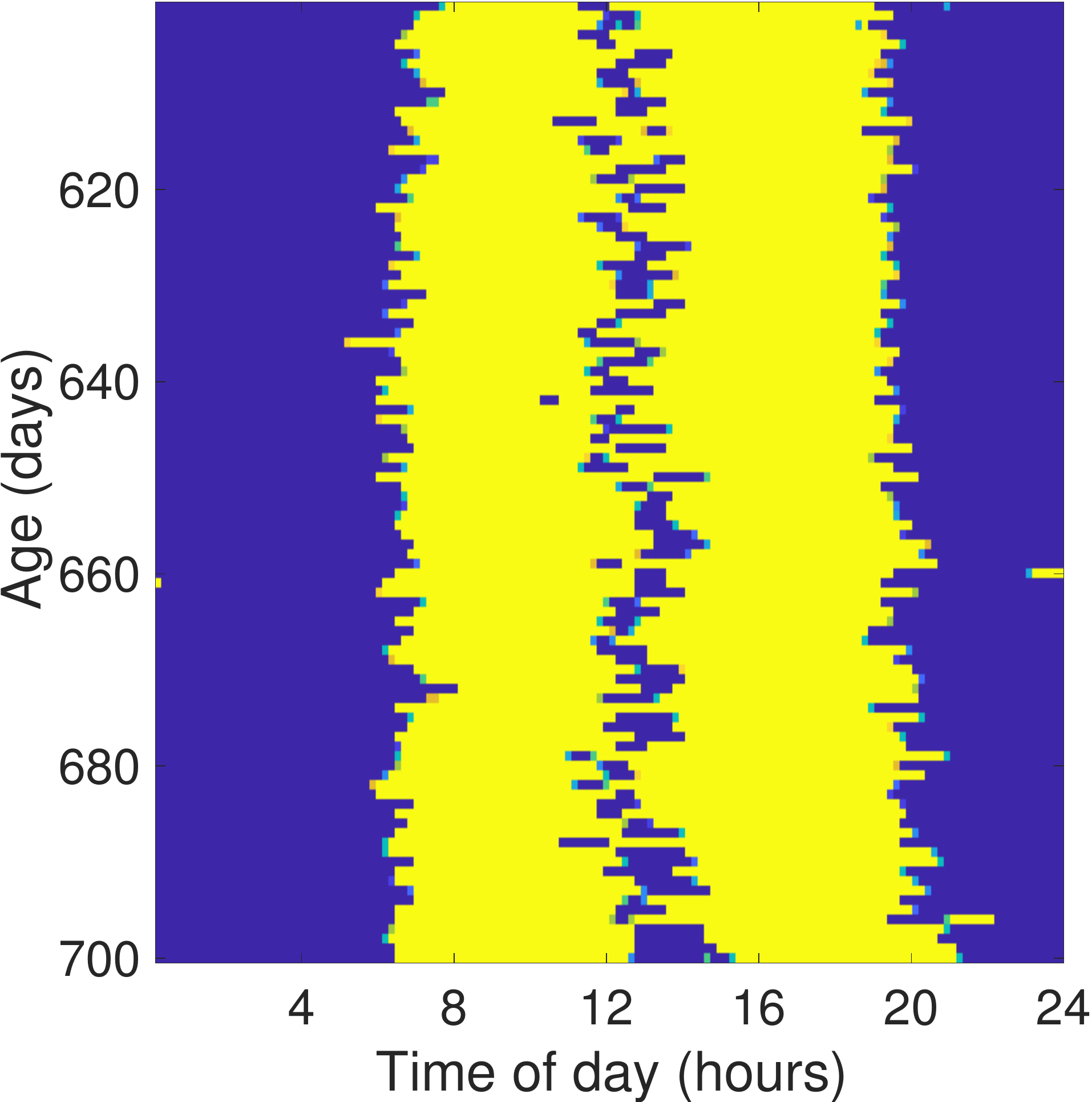} & \includegraphics[scale=0.15]{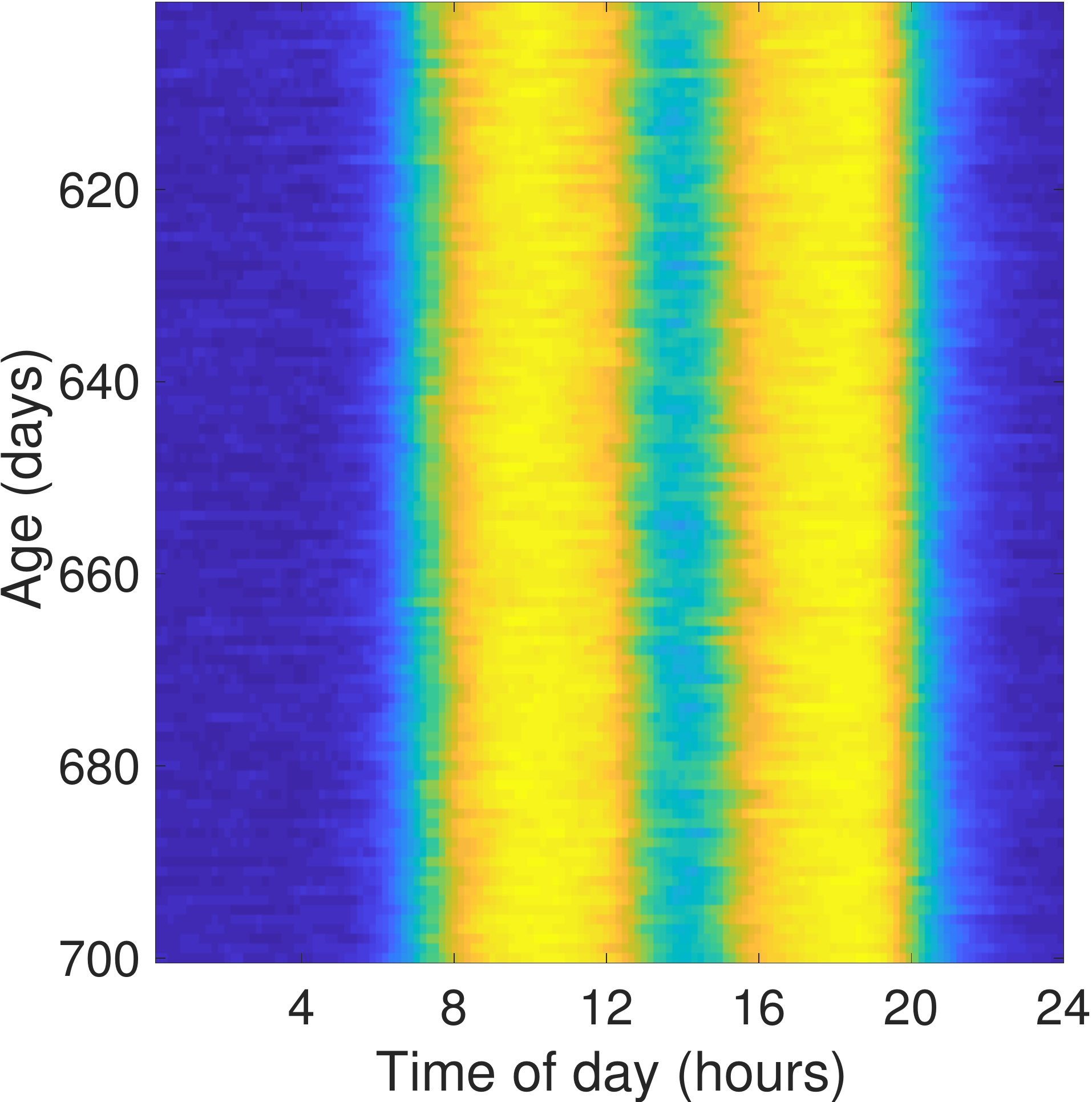}  &  \includegraphics[scale=0.15]{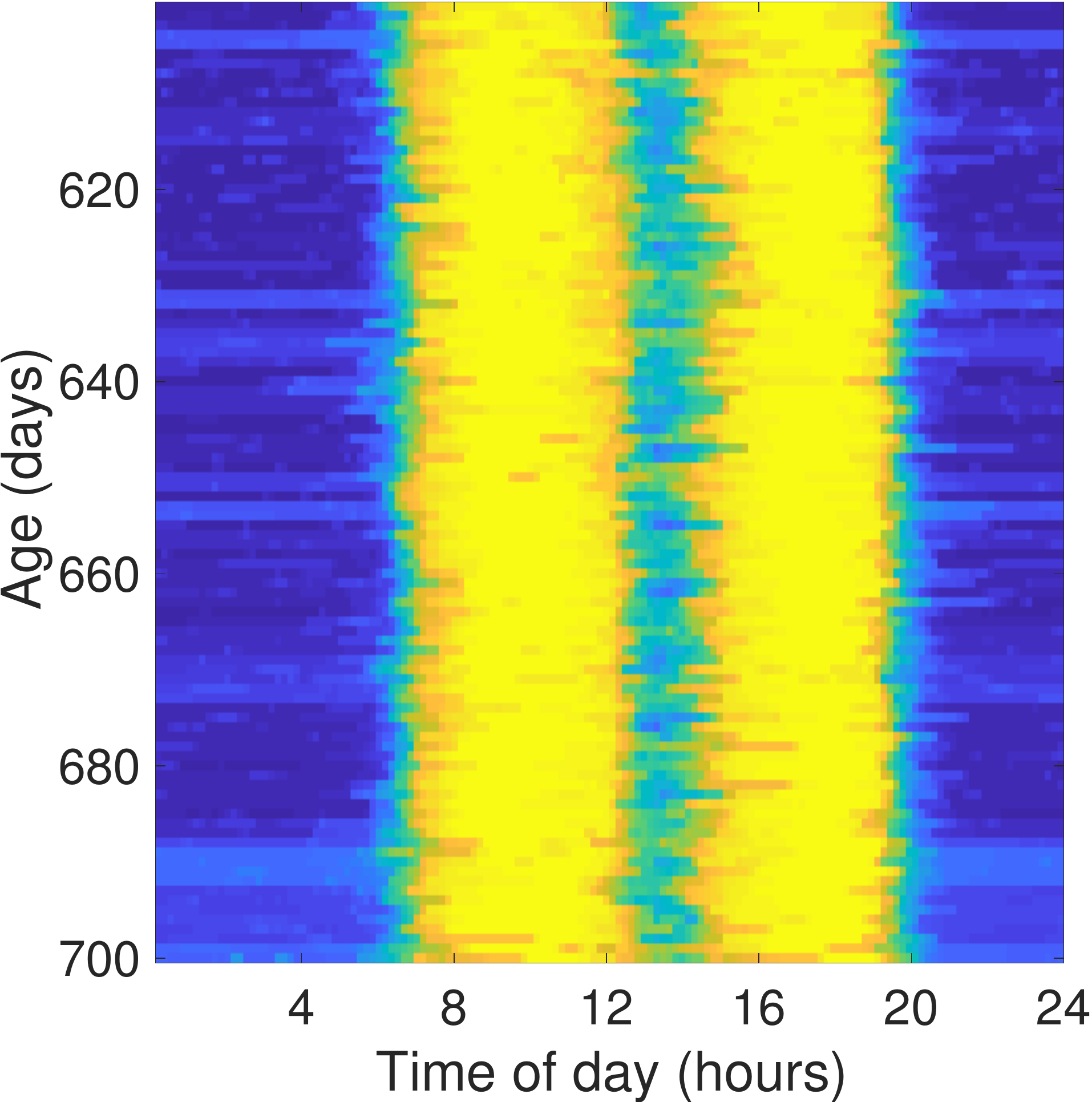}&  \includegraphics[scale=0.15]{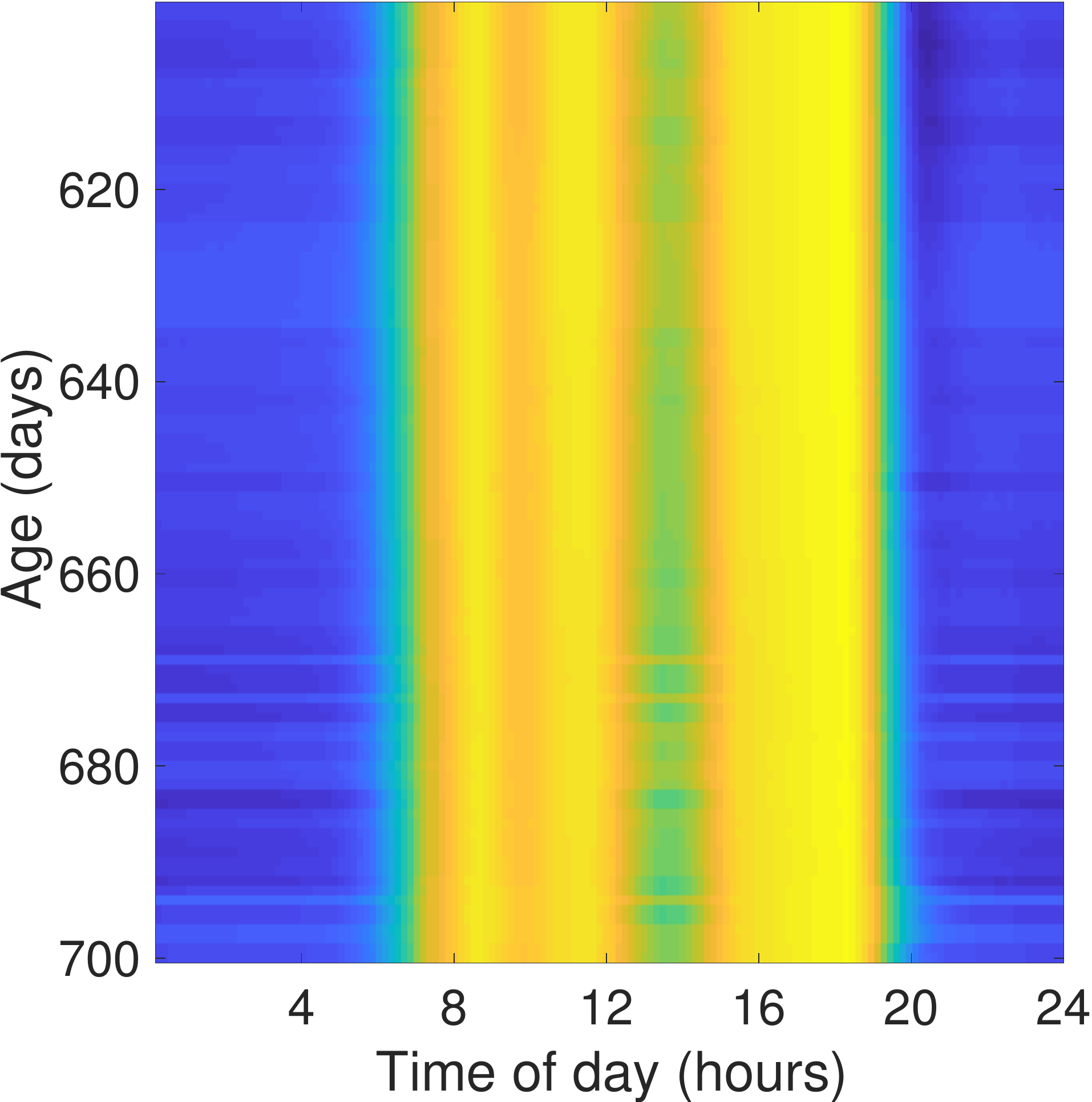} &  \includegraphics[scale=0.15]{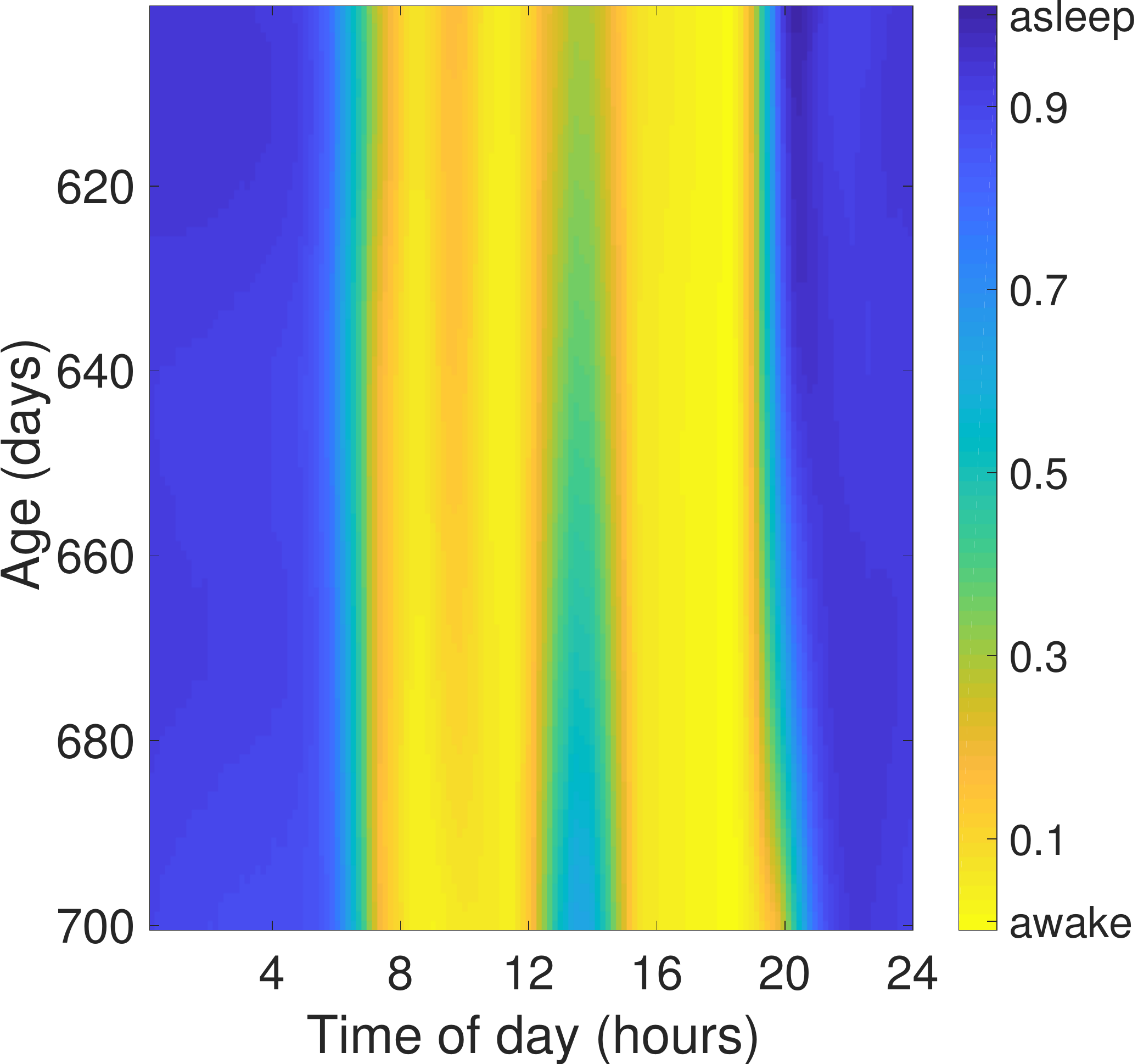}\\
 \end{tabular}
 \caption{Examples of sleep forecasts from days 600 to 700 of age, computed from the data of days 500 to 600. For each example, we show the future data, the mean of the training set, the kernel-regression (KR) using the raw training data, the kernel-regression prediction based on NMF-TS coefficients, and the rank-5 representation of the future data.}
 \label{fig:pred_results_3}
\end{figure}

In this section we provide additional figures to illustrate our methods and analysis. Figure~\ref{fig:low_rank_approximation} shows Low-rank approximation obtained from different methods. Figure~\ref{fig:daily_sleep_patterns} shows basis functions extracted from the infant-sleep dataset using PCA, NMF and NMF-TS. Figure~\ref{fig:age_development_patterns} shows coefficients corresponding to the basis functions  for the infant whose data is shown in Figure~\ref{fig:low_rank_approximation}. Figure~\ref{fig:random_samples} shows raw data from infants randomly selected from the dataset. Each data matrix is shown accompanied by the coefficients of its NMF-TS low-rank representation. Figure~\ref{fig:svs} shows the singular values computed over the whole dataset. The jump between the 5th and 6th singular value suggests using a rank-5 low-rank representation. Figure~\ref{fig:rank} shows the basis functions extracted by PCA (top), NMF (center) and the proposed approach (bottom) for different values of the rank $r$. The value $r=5$, which we use for our analysis, yields the most interpretable model. Figure~\ref{fig:smoothness} illustrates the effect of varying the parameter $\lambda$ on the NMF-TS basis functions and coefficients. The daily-sleep patterns change only slightly, whereas the age-development become smoother as we increase $\lambda$. We set $\lambda=10^5$ for our analysis. Figure~\ref{fig:rawdata_results} and Figure~\ref{fig:rnmf_results} show the results of applying Lloyd's algorithm to perform k-means clustering on the raw infant-sleep data and on the three first NMF-TS coefficients respectively. Figures~\ref{fig:pred_results_1}, \ref{fig:pred_results_2} and \ref{fig:pred_results_3} show examples of forecasts computed as described in Section~\ref{sec:prediction}. The low-rank prediction obtained by applying kernel regression on the NMF-TS coefficients is visually similar to the true rank-5 representation of the future data.


\end{document}

%% file: macros.tex
\newcommand{\MAT}[1]{\begin{bmatrix} #1 \end{bmatrix}}

\newcommand{\abs}[1]{\left| #1 \right|}
\newcommand{\keys}[1]{\left\{ #1 \right\}}

\newcommand{\brac}[1]{\left( #1 \right) }
\newcommand{\ml}[1]{\mathcal{ #1 } }
\newcommand{\op}[1]{ \operatorname{#1} }

\newcommand{\normTwo}[1]{\left\| #1 \right\| _{2}}

\newcommand{\R}{\mathbb{R}}